\pdfoutput=1

\documentclass[11pt]{article}

\usepackage[table,dvipsnames]{xcolor}

\usepackage[pagebackref,breaklinks,colorlinks,citecolor=mydarkblue]{hyperref}
\usepackage{colortbl}
\usepackage[most]{tcolorbox}

\usepackage[preprint]{acl}

\usepackage{times}
\usepackage{latexsym}

\usepackage[T1]{fontenc}

\usepackage[utf8]{inputenc}

\usepackage{microtype}

\usepackage{inconsolata}

\usepackage{graphicx}

\usepackage{siunitx}
\usepackage[utf8]{inputenc} 
\usepackage[T1]{fontenc}    
\usepackage{url}            
\usepackage{booktabs}       
\usepackage{amsfonts}       
\usepackage{nicefrac}       
\usepackage{microtype}      
\usepackage{paralist}
\usepackage{xspace}
\usepackage{graphicx}
\usepackage{subcaption}
\usepackage{float} 
\usepackage{multirow}
\usepackage{adjustbox} 
\usepackage{caption}
\usepackage{array}
\usepackage{courier}
\usepackage{arydshln}

\usepackage[utf8]{inputenc} 
\usepackage[T1]{fontenc}    
\usepackage{url}            
\usepackage{booktabs}       
\usepackage{amsfonts}       
\usepackage{nicefrac}       
\usepackage{microtype}      
\usepackage{booktabs}
\usepackage{makecell}  
\usepackage{paralist}

\usepackage{soul}
\usepackage{calc}  
\usepackage{array} 
\usepackage{graphicx}
\usepackage{subcaption}
\usepackage{amssymb}  
\usepackage{pifont}   
\usepackage{multirow} 
\usepackage{mdframed}
\tcbuselibrary{listingsutf8}
\usepackage{listings}  
\tcbuselibrary{listings} 
\usepackage{geometry}  
\tcbuselibrary{listingsutf8}
\usepackage{tabularx}

\definecolor{mydarkblue}{rgb}{0,0.53,0.96}

\usepackage{cleveref}
\crefname{figure}{Fig.}{Figs.}
\Crefname{figure}{Fig.}{Figs.}

\crefname{table}{Tab.}{Tabs.}
\Crefname{table}{Tab.}{Tabs.}

\crefname{section}{Sec.}{Secs.}
\Crefname{section}{Sec.}{Secs.}

\crefname{suppsection}{Appendix}{Appendices}
\Crefname{suppsection}{Appendix}{Appendices}

\newcommand{\subsec}[1]{\noindent\textbf{#1}~~}

\definecolor{codegreen}{rgb}{0,0.6,0}
\definecolor{codegray}{rgb}{0.5,0.5,0.5}

\definecolor{backcolour}{RGB}{245,248,250}
\definecolor{emph}{RGB}{166,88,53}
\definecolor{nightblue}{RGB}{9,49,105}
\definecolor{keywords}{RGB}{207,33,46}
\definecolor{lightpurple}{RGB}{130,81,223}

\definecolor{MyLightGray}{rgb}{0.95, 0.95, 0.95}
\definecolor{CLIPBlue}{rgb}{0.192, 0.454, 0.643}

\newcommand{\eg}{e.g.\xspace}
\newcommand{\ie}{i.e.\xspace}
\newcommand{\sq}[1]{\textcolor{#1}{\rule{2ex}{2ex}}}

\lstdefinelanguage{json}{
  keywords={true, false, null},
  sensitive=true,
  comment=[l]{//},
  morecomment=[s]{/*}{*/},
  string=[s]{"}{"},
}

\definecolor{ForestGreen}{RGB}{34, 139, 34}
\definecolor{checkred}{RGB}{178, 34, 34}

\newcommand{\greencheck}{\textcolor{ForestGreen}{\checkmark}}

\newcommand{\datasetsize}{83k}

\newcommand{\totalimagesize}{305k}
\newcommand{\benchmarksize}{328}
\newcommand{\numberOfAction}{15}
\newcommand{\numberOfModels}{46}

\definecolor{lightgray}{RGB}{245, 245, 245}
\definecolor{lightyellow}{RGB}{255, 255, 204} 
\definecolor{verylightblue}{rgb}{0.88, 0.95, 1.0}

\newcommand{\action}[1]{\sethlcolor{lightgray}\hl{\small\sffamily #1}}
\newcommand{\category}[1]{\sethlcolor{lightyellow}\hl{\small\sffamily #1}}

\newcommand{\datasetAllPosts}{98,234}
\newcommand{\datasetAllPostsFinaldataset}{82,976}
\newcommand{\datasetAllEdits}{350,609}
\newcommand{\datasetAllEditsFinaldataset}{305,806}
\newcommand{\datasetAllPushShiftsPosts}{58,624}
\newcommand{\datasetAllRecentPosts}{39,610}

\newcommand{\datasetAllPushShiftsEdits}{90,466}
\newcommand{\datasetAllRecentEdits}{215,340}

\newcommand{\datasetAllSingle}{71,027}
\newcommand{\datasetAllMulti}{11,949}

\newcommand{\InstructPix}{\model{InstructPix2Pix\xspace}}
\newcommand{\SeedEdit}{\model{SeedEdit\xspace}}
\newcommand{\Magic}{\model{MagicQuill\xspace}}
\newcommand{\CosXL}{\model{CosXL\xspace}}
\newcommand{\Ledits}{\model{LEDITS\xspace}}
\newcommand{\CodeFormer}{\model{CodeFormer\xspace}}
\newcommand{\FineEraser}{\model{Finegrain-Object-Eraser\xspace}}
\newcommand{\FineCutter}{\model{Finegrain-Object-Cutter\xspace}}
\newcommand{\BriaEraser}{\model{BRIA-Eraser-API\xspace}}
\newcommand{\BriaRecolor}{\model{BRIA-2.2-ControlNet-Recoloring\xspace}}
\newcommand{\BRIAInpainting}{\model{BRIA-2.3-Inpainting\xspace}}
\newcommand{\BRIAFillAPI}{\model{BRIA-Generative-Fill-API\xspace}}
\newcommand{\RemoveObject}{\model{remove-photo-object\xspace}}
\newcommand{\ReplaceAny}{\model{ReplaceAnything\xspace}}
\newcommand{\Colorize}{\model{text-guided-image-colorization\xspace}}
\newcommand{\PhotoRestore}{\model{old\_photo\_restoration\xspace}}
\newcommand{\ICLight}{\model{IC-Light\xspace}}
\newcommand{\FluxInpaint}{\model{FLUX.1-dev-Inpainting-Model-Beta-GPU\xspace}}
\newcommand{\FluxIP}{\model{flux-IP-adapter\xspace}}
\newcommand{\FluxFillDev}{\model{FLUX.1-Fill-dev\xspace}}

\newcommand{\FluxInpaintDev}{\model{FLUX.1-inpaint-dev\xspace}}
\newcommand{\FluxFillOut}{\model{flux-fill-outpaint\xspace}}
\newcommand{\StableDiff}{\model{stable-diffusion-xl-inpainting\xspace}}
\newcommand{\LeditsPlus}{\model{leditsplusplus\xspace}}
\newcommand{\DiffFast}{\model{diffusers-fast-inpaint\xspace}}
\newcommand{\TurboEdit}{\model{turbo\_edit\xspace}}
\newcommand{\BSharp}{\model{B2BMGMT\_Sharpening\xspace}}
\newcommand{\TextCut}{\model{textcutobject\xspace}}
\newcommand{\SketchGen}{\model{Sketch-Gen\xspace}}
\newcommand{\ImageToLine}{\model{Image-to-Line-Drawings\xspace}}
\newcommand{\AnimeGAN}{\model{AnimeGANv2\xspace}}
\newcommand{\NonLinearBlur}{\model{NonLinear-Blurr-Image\xspace}}
\newcommand{\FotoFilter}{\model{foto\_filter\xspace}}
\newcommand{\KorniaFilter}{\model{kornia-image-filtering\xspace}}
\newcommand{\SketchTo}{\model{sketch2lineart\xspace}}
\newcommand{\DiffFill}{\model{diffusers-image-fill\xspace}}
\newcommand{\ImageToColoringBook}{\model{image2coloringbook\xspace}}

\newcommand{\Gemini}{\model{Gemini}}

\usepackage{helvet}
\newcommand{\model}[1]{{{\small\fontfamily{phv}\selectfont{#1}}\xspace}}
\definecolor{gpt_green}{RGB}{22,163,127} 
\definecolor{gemini_blue}{RGB}{81,134,209} 
\definecolor{sonnet3_brown}{RGB}{204,154,123} 
\definecolor{sonnet35_brown}{RGB}{216, 119, 87} 
\definecolor{qwen_violet}{RGB}{191, 123, 234} 
\definecolor{headerblue}{RGB}{51,122,183}
\definecolor{categorygray}{RGB}{245,245,245}
\definecolor{subjective}{RGB}{255,243,230}
\definecolor{random}{RGB}{230,255,230}
\definecolor{objective}{RGB}{230,240,255}
\definecolor{hard}{RGB}{255,230,230}

\definecolor{pixtral_orange}{RGB}{255, 138, 0}
\definecolor{llama_blue}{RGB}{0, 102, 204}
\definecolor{internvl_blue}{RGB}{0, 153, 255}
\definecolor{llava_red}{RGB}{204, 0, 0}
\definecolor{gemma_blue}{RGB}{0, 128, 255}

\definecolor{lightblue}{rgb}{0.85,0.92,1}

\definecolor{rowLow}{RGB}{255,239,213}      
\definecolor{rowMed}{RGB}{255,255,204}      
\definecolor{rowHigh}{RGB}{210,255,210}     

\newcommand{\geminiflash}{\model{Gemini-\textcolor{gemini_blue}{2.0}-Flash}\xspace}

\newcommand{\geminiflashthinking}{\model{Gemini-\textcolor{gemini_blue}{2.0}-Flash-Thinking}\xspace}
\newcommand{\geminipro}{\model{Gemini-\textcolor{gemini_blue}{2.5}-Pro}\xspace}

\newcommand{\oone}{\model{o\textcolor{gpt_green}{1}}\xspace}

\newcommand{\gpt}{\model{GPT-\textcolor{gpt_green}{4o}}\xspace}
\newcommand{\oonepro}{\model{o1-\textcolor{gpt_green}{Pro}}\xspace}
\newcommand{\gptmini}{\model{GPT-\textcolor{gpt_green}{4o}-mini}\xspace}

\newcommand{\internvlful}{\model{InternVL-\textcolor{internvl_blue}{2.5}-38B}\xspace}
\newcommand{\internvl}{\model{InternVL-\textcolor{internvl_blue}{2.5}}\xspace}

\definecolor{lightcyan}{rgb}{0.88, 1.0, 1.0}
\definecolor{verylightcyan}{rgb}{0.94, 1.0, 1.0}
\definecolor{lightteal}{rgb}{0.6, 0.94, 0.94}

\definecolor{lightpink}{rgb}{1.0, 0.71, 0.76}
\definecolor{verylightpink}{rgb}{1.0, 0.85, 0.9}

\definecolor{lightbrown}{rgb}{0.82, 0.71, 0.55}
\definecolor{verylightbrown}{rgb}{0.9, 0.83, 0.7}
\definecolor{lighttan}{rgb}{0.98, 0.92, 0.84}

\definecolor{lightpeach}{rgb}{1.0, 0.89, 0.77}
\definecolor{lightcoral}{rgb}{0.94, 0.5, 0.5}

\definecolor{lightlavender}{rgb}{0.9, 0.9, 0.98}
\definecolor{lightlilac}{rgb}{0.87, 0.8, 0.9}

\definecolor{lightmint}{rgb}{0.88, 1.0, 0.88}
\definecolor{verylightmint}{rgb}{0.94, 1.0, 0.94}

\newlength{\mpheight}
\definecolor{gpt_green}{RGB}{22,163,127} 
\definecolor{gemini_blue}{RGB}{81,134,209} 

\newcommand{\geminilogo}{{\includegraphics[scale=0.7]{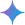}}\xspace}
\newcommand{\gptlogo}{{\includegraphics[scale=0.02]{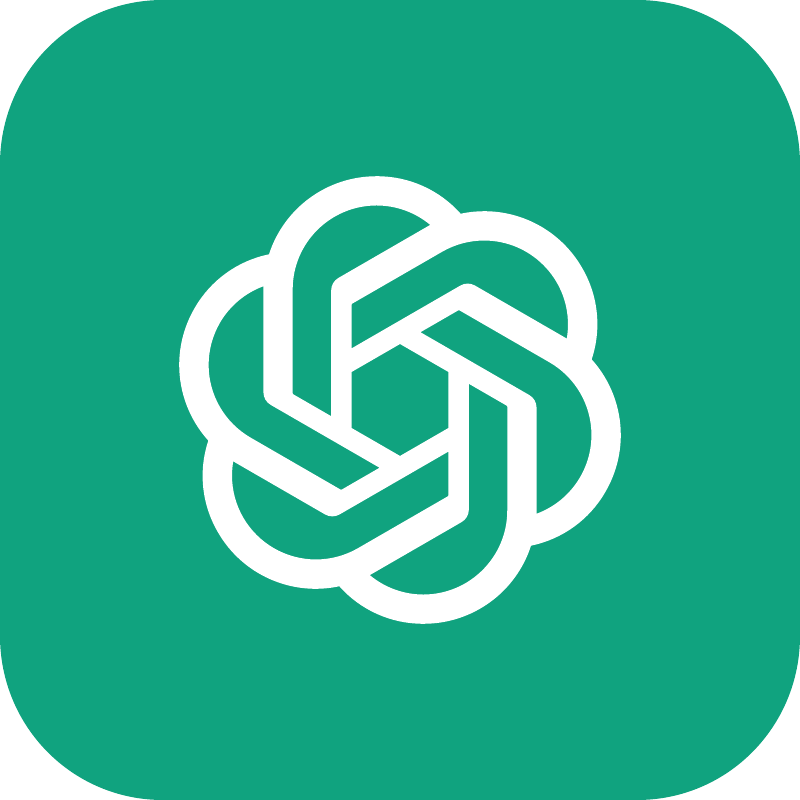}}\xspace}
\newcommand{\openailogo}{{\includegraphics[scale=0.02]{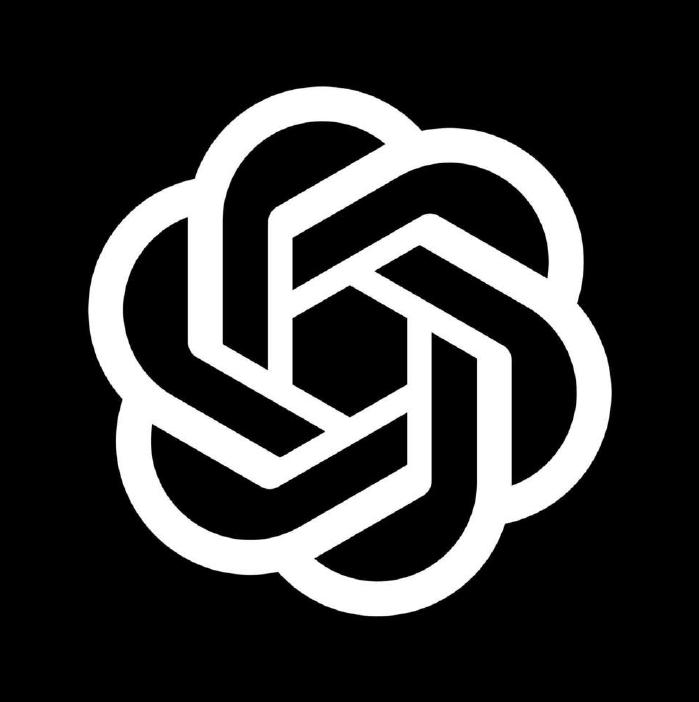}}\xspace}
\newcommand{\seedlogo}{{\includegraphics[scale=0.02]{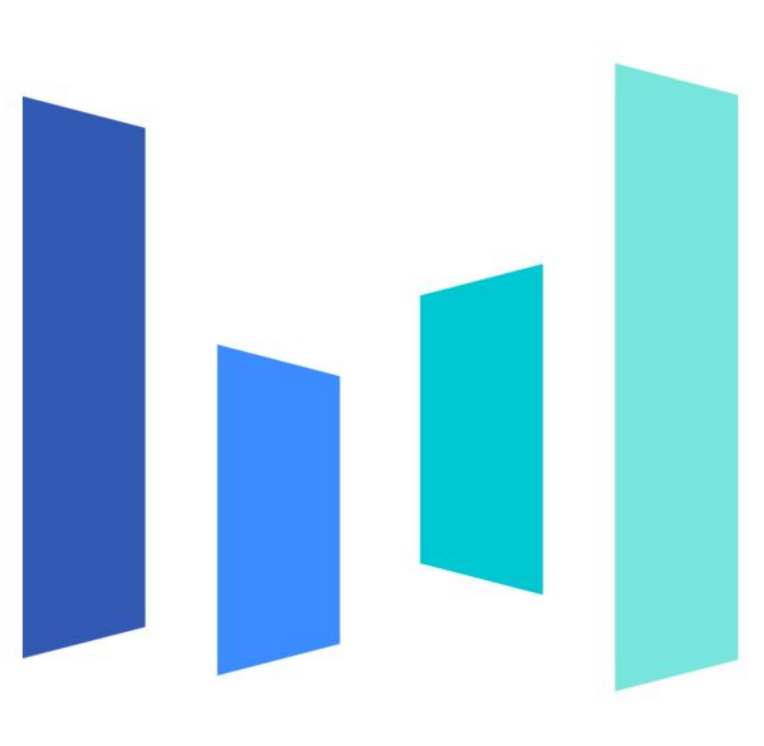}}\xspace}

\newcommand{\huggingface}{\raisebox{-2pt}{\includegraphics[scale=0.011]{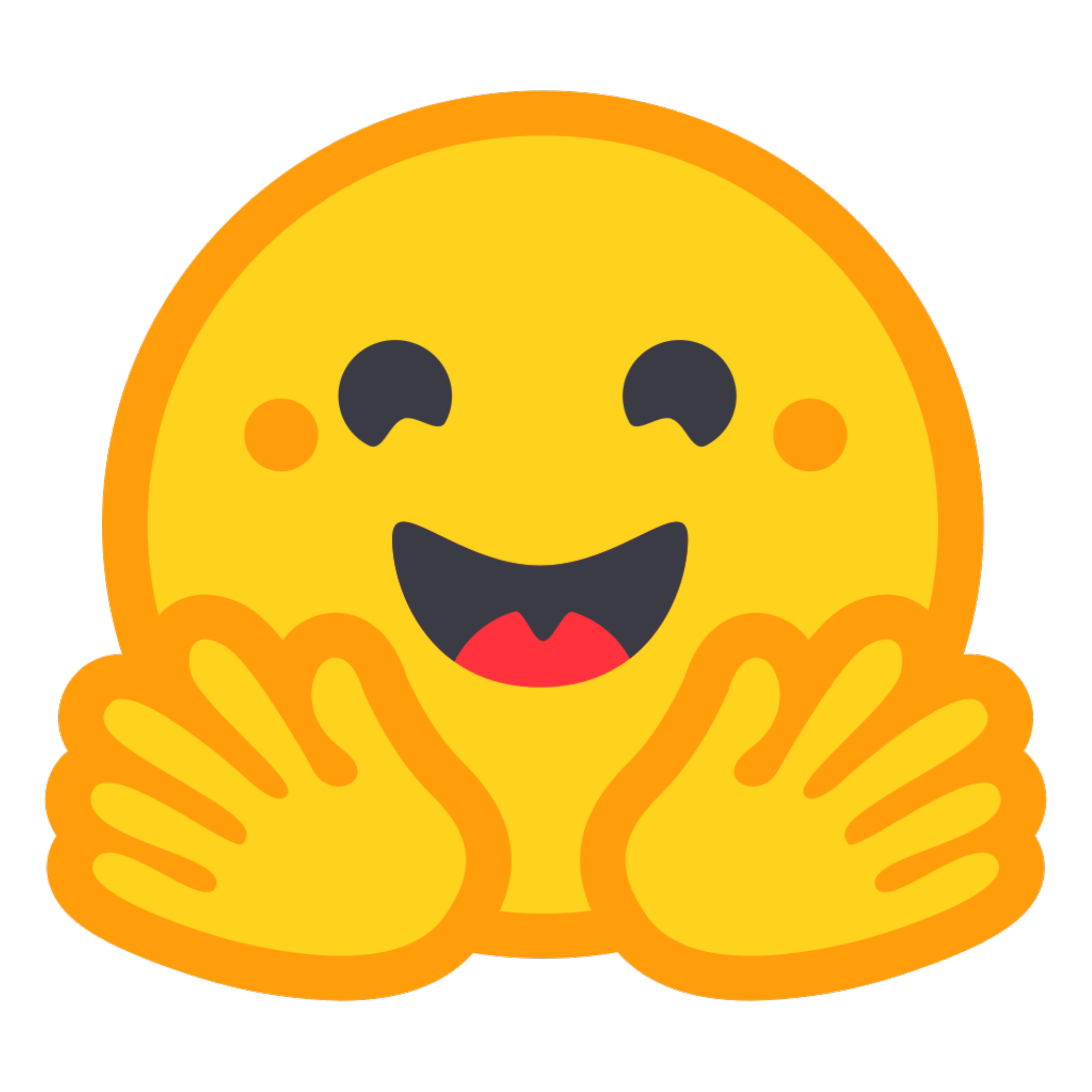}\xspace}}

\newcommand{\papertitle}{
Understanding Generative AI Capabilities \\in Everyday Image Editing Tasks
}

\title{\papertitle}

\author{%
  Mohammad Reza Taesiri$^{\textcolor{ForestGreen}{1}}$\thanks{While MRT is the lead, all first 3 authors made major contributions to code, running experiments, and analyzing results.}\\
  {\small\texttt{mtaesiri@gmail.com}}\\
    \And
  Brandon Collins$^{\textcolor{orange}{2} *}$\\
  {\small\texttt{blc0063@auburn.edu}}\\
  \And
  Logan Bolton$^{\textcolor{orange}{2}*}$\\
  {\small\texttt{logan.bolton@auburn.edu}}\\
  \And
  Viet Dac Lai$^{\textcolor{BrickRed}{3}}$\\
  {\small\texttt{daclai@adobe.com}}\\
 \AND
  Franck Dernoncourt$^{\textcolor{BrickRed}{3}}$\\
  {\small\texttt{dernonco@adobe.com}}\\
  \And
  Trung Bui$^{\textcolor{BrickRed}{3}}$\\
  {\small\texttt{bui@adobe.com}}\\
  \And
  Anh Totti Nguyen$^{\textcolor{orange}{2}}$\\
  {\small\texttt{anh.ng8@gmail.com}}\\
  \AND
    $^{\textcolor{ForestGreen}{1}}${\small \textnormal{University of Alberta}} \quad
    $^{\textcolor{orange}{2}}${\small\textnormal{Auburn University}} \quad
    $^{\textcolor{BrickRed}{3}}${\small\textnormal{Adobe Research}}
}

\begin{document}
\maketitle


\begin{abstract}
Generative AI (GenAI) holds significant promise for automating everyday image editing tasks, especially following the recent release of \gpt{}~\gptlogo on March 25, 2025.
However, what subjects do people most often want edited? What kinds of editing actions do they want to perform (e.g., removing or stylizing the subject)? Do people prefer precise edits with predictable outcomes, or highly creative ones?
By understanding the characteristics of real-world requests and the corresponding edits made by freelance photo-editing wizards, can we draw lessons for improving AI-based editors and determine which types of requests can currently be handled successfully by AI editors?
In this paper, we present a unique study addressing these questions by analyzing \datasetsize{} requests with their associated \totalimagesize{} edits from the recent 12 years on the \href{https://www.reddit.com/r/PhotoshopRequest/}{/r/PhotoshopRequest} Reddit community.
According to human ratings, approximately only 33\% of requests can be fulfilled by the best AI editors (including \gptlogo, \geminilogo, \seedlogo).
Interestingly, AI editors perform worse on low-creativity requests that require precise editing than on more open-ended requests. 
They often struggle to preserve the identity of people and animals, and frequently make non-requested touch-ups.
On the other side of the table, VLM judges (\eg, \oone) perform differently than human judges and may prefer AI edits over human edits.
Code and qualitative examples are available at: \href{https://psrdataset.github.io/}{https://psrdataset.github.io/}.
\end{abstract}

\section{Introduction}
\label{sec:intro}

\begin{figure}[t]
     \centering
        \includegraphics[width=\linewidth]{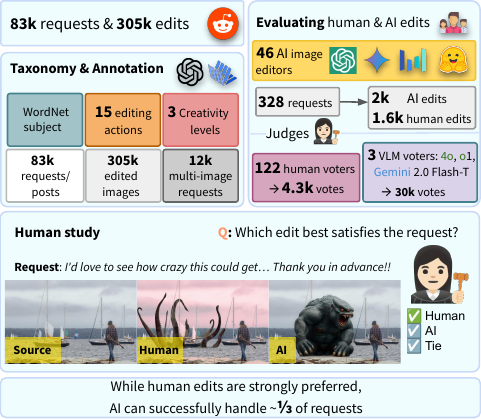}
    \caption{
    We propose PSR, the largest dataset of real-world image-editing requests and human-made edits.
    PSR enables the community (and our work) to identify types of requests that can be automated using existing AIs and those that need improvement.
    PSR is the first dataset to tag all requests with WordNet subjects, real-world editing actions, and creativity levels.
    }
    \label{fig:teaser}
\end{figure}

GenAI for images has gained enormous research interest \cite{huang2024diffusion} and created a 2023 market of \$300M, which is estimated to multiply \cite{marketshare}.
Specifically, text-based image editing is an increasingly high-demand task \cite{shi2024seededit}, especially after the recent \gpt and \geminipro image generators \cite{openai20254o}.
However, four important questions remain open:\\
\textbf{Q1:} What are the \emph{real} everyday image editing requests and needs of users?\\
\textbf{Q2:} According to human judgment, what \% of such requests can be satisfied by existing AIs?
\\
\textbf{Q3:} What are the improvement areas for AI editors compared to human editors?\\
\textbf{Q4:} Are vision language models (VLMs) judging AI-edited images similarly to human judges?

Q1 and Q2 are unanswered partly because many prior datasets (\eg, \cite{brooks2023instructpix2pix,shi2021learning,sheynin2024emu,zhao2025ultraedit}) contain made-up requests written by either human annotators or AIs based on the source image or the (source, target image) pair (see \cref{tab:dataset_comparison}).
Those request distributions may \emph{not} reflect the actual editing needs of users as well as the challenges posed by the real requests, which may have typos, distraction or ambiguity (\eg, ``{\it I'd love to see how crazy this could get... Thank you in advance!!}'' in this \href{https://www.reddit.com/r/PhotoshopRequest/comments/c7f1pa/random_id_love_to_see_how_crazy_this_could_get/}{post}; \cref{fig:teaser}).
On the other hand, some datasets (\eg, \cite{zhang2023magicbrush,brooks2023instructpix2pix,sheynin2024emu,zhao2025ultraedit}) feature images AI-edited and therefore do not represent the real edits by humans.

We aim to answer these four questions by analyzing \totalimagesize{} tuples of (source image, text request, edited image) from the 
\href{https://www.reddit.com/r/PhotoshopRequest/}{/r/PhotoshopRequest} (PSR) Reddit channel, which is the \emph{largest} public online community \cite{feedspotPhotoshopForums,subredditstatsRPhotoshopRequestSubreddit} that shares diverse, everyday image-editing needs with corresponding edits by PSR wizards\footnote{Advanced image editors who are granted to handle paid editing requests in this particular subreddit.
}.
PSR has 1.7M users and receives an average of 141 new requests per day (our statistics for 2025) with a peak as high as 226 per day
\cite{medium_photoshoprequest_2025}.

To answer \textbf{Q1}, \textbf{Q2}, and \textbf{Q3}, our closed-loop study
\textbf{(a)} analyzes 305k tuples, \ie, \datasetsize{} \emph{unique} requests $\times$ 3.67 human-edited images per request;
\textbf{(b)} sends all (request, source image) pairs to image-editing models to collect AI edits;
\textbf{(c)} performs a human study over a set of 328 requests (PSR-328) to collect over 4.5k ratings to compare how 1,644 PSR-wizard edits fare against 2,296 AI edits on the same requests to identify areas where AIs perform well and fall short.
Our work is the first to compare three state-of-the-art (SOTA) image editors: \gpt{}~\gptlogo~\cite{openai2025_4oimage}, \geminiflash~\geminilogo\ \cite{geminiImage}, and \SeedEdit{}~\seedlogo~\cite{shi2024seededit} as well as 46 other AI models on HuggingFace, for a total of \textbf{49} AI editors.
Furthermore, to address \textbf{Q4}, we compare human ratings against those by \textbf{3} SOTA vision-language models (VLMs): \gpt{}, \oone~\openailogo~\cite{openai_o1}, and \geminiflashthinking~\geminilogo~\cite{GoogleDeepMind2024Gemini2}.
Our main findings are:

\begin{compactitem}

    \item  66\% of the time, human judges still \emph{prefer human} edits over AI edits (\cref{sec:human_prefer_human}).    
    
    \item While SOTA VLMs are excellent at regular visual tasks \cite{chen2024iqagpt}, on image-edit judgment, VLMs can be extremely biased, \eg, \oone prefers \gpt edits 85\% of the time, which is in stark contrast to human judgment (\cref{sec:vlms_are_bad_proxy}).

    \item AIs often add extra non-requested changes that improve the aesthetics of the image but also fail to preserve the subject's identity (\cref{sec:ai_improve_aesthetics}).
    
    \item AIs either tie or win 33.35\% of requests, while PSR wizards still outperform AI editors on the remaining 66.65\% (\cref{sec:ai_model_can_handle_one_thrid}).

\end{compactitem}

\begin{table*}[t]
\centering
\caption{PSR is the largest-scale dataset of real-world requests and PSR-wizard edits.
}
\label{tab:dataset_comparison}
\resizebox{\linewidth}{!}{%
\begin{tabular}{rlrrllllrrr}
 \hline
\multirow{2}{*}{Dataset} & & {No. of} & No. of & Source & {Edit} & {Request} & Requests & \multicolumn{3}{c}{Creativity} \\
 & & requests & edits & image & generator & writer & based on & \colorbox{rowLow}{Low} & \colorbox{rowMed}{Med} & \colorbox{rowHigh}{High} \\
\hline
IER~\cite{tan2019expressing} & 2019 & 4k & \textcolor{gray}{4k} & reddit & human \greencheck & human & reddit \greencheck & 78\%  & 15\% &  \textcolor{red}{7\%}  \\
GIER~\cite{shi2020benchmark} & 2020 & 6k & \textcolor{gray}{6k} & reddit & human \greencheck & human & reddit \greencheck & 63\%  & 34\%  & \textcolor{red}{3\%} \\
MA5k-Req~\cite{shi2021learning} & 2021 & 24k & \textcolor{gray}{24k} & real & Ps \cite{fivek} & {Amazon MT} & image pairs & 100\% &  \textcolor{red}{0\%}  & \textcolor{red}{0\%}\\
{InstructPix2Pix}~\cite{brooks2023instructpix2pix} & 2023 & 454k & \textcolor{gray}{454k} &  {SD} \cite{hertz2023prompttoprompt} & {SD} \cite{hertz2023prompttoprompt} & {GPT-3} & image pairs & 12\%  & 44\%  & 42\%  \\
MagicBrush~\cite{zhang2023magicbrush} & 2023 & 10k & \textcolor{gray}{10k} & MS COCO & {DALL-E 2} & {Amazon MT} & source image & 62\% & 25\% & 12\%\\
HIVE~\cite{zhang2024hive} & 2024 & 1.1M & \textcolor{gray}{1.1M} & {SD} \cite{hertz2023prompttoprompt} & {SD} \cite{hertz2023prompttoprompt} & BLIP  & image pairs &  48\% & 30\%  & 22\% \\
EmuEdit \cite{sheynin2024emu} & 2024 & 10M & \textcolor{gray}{10M} & {Emu} \cite{dai2023emu} & {Emu} \cite{dai2023emu} & Llama 2 & source image \& task  & 54\% &  20\% & 26\% \\
AURORA~\cite{krojer2025learning} & 2025 & 280k & \textcolor{gray}{280k} & mixed & mixed & GPT-4o & image pairs & 96\%  &  \textcolor{red}{4\%} & \textcolor{red}{0\%}\\
UltraEdit~\cite{zhao2025ultraedit} & 2025 & 4M & \textcolor{gray}{4M} & MS COCO & SD & GPT-4o, human & source image  & 43\% & 21\%  & 36\%\\
RealEdit~\cite{sushko2025realedit} & 2025 & 57k & 94k & reddit & human \greencheck & human & reddit \greencheck & 58\% & 36\% & \textcolor{red}{6\%} \\
\hline
\textbf{PSR} (ours) & 2025 & \datasetsize{} & \totalimagesize{} & reddit  & human \greencheck & human & reddit \greencheck & 56\%  & 28\%  &  16\% \\
{PSR-328} (ours) & 2025 & 328 & 3.9k & reddit  & human, AIs \greencheck & human & reddit \greencheck & 33\%  & 33\%  &  33\% \\
\end{tabular}%
}
\end{table*}


\section{Related Work}
\label{sec:related_work}

\textbf{Online image editing communities} have been studied to understand user intent, common editing patterns, and challenges in automating image editing~\cite{manuvinakurike2018edit, laput2013pixeltone, shi2020benchmark}.
However, prior image-editing taxonomies \cite{manuvinakurike2018edit} of editing actions often correspond to the low-level functions available in image-editing software (\eg, \action{zoom} or \action{select}) and do \emph{not} correspond to the editing intent of humans who request changes (\eg, human would \emph{not} want to \action{select} an object in an image but rather it is the image editor).
In contrast, we build our taxonomy of actions intended by users who post requests on the PSR channel and therefore our taxonomy is based on Reddit requests instead of based on what functionalities are available in image editing software \cite{manuvinakurike2018edit,brixey2018system}.
Furthermore, existing taxonomies \cite{manuvinakurike2018edit,brixey2018system} contain ambiguous, overlapping labels such as (\action{zoom} vs. \action{crop}), which we resolve in our taxonomy.
Our taxonomy also contains 5 labels that reflect more up-to-date operations: 
\action{super-resolution}, \action{relight}, \action{specialized}, \action{re-color}, and \action{merge}, which do not exist in prior taxonomies \cite{manuvinakurike2018edit,brixey2018system}.


\subsec{Editing datasets}
Synthetic datasets, \eg, InstructPix2Pix~\cite{brooks2023instructpix2pix}, UltraEdit~\cite{zhao2025ultraedit}, and Emu Edit~\cite{sheynin2024emu}, enable a large-scale training-set of AI-generated images but fail to capture the real distribution of both requests and edits made by humans. 
Hybrid datasets like HIVE~\cite{zhang2024hive} (with human feedback) and Seed-Edit-Data~\cite{ge2024seed} (synthetic and human-annotated) balance scale and authenticity. 
Meanwhile, real user-request datasets, such as IER~\cite{tan2019expressing} and GIER~\cite{shi2020benchmark}, offer greater authenticity by collecting genuine editing needs from online communities, but remain constrained in scale and diversity (63--87\% of requests are \colorbox{rowLow}{low}-creativity \cref{tab:dataset_comparison}).
Unlike the above works, our PSR contains \emph{real} image-editing requests by Reddit users and also human-edited images by PSR wizards.

\subsec{The most similar dataset} to PSR is RealEdit~\cite{sushko2025realedit}, which is a \emph{concurrent} work that scrapes 57k requests also from Reddit.
Our dataset is $\sim$3$\times$ larger than RealEdit in the number of (request, source image, edited image) tuples (\cref{tab:dataset_comparison}).
Unlike RealEdit and all previous works, our work is the first to address all four questions Q1--Q4, \ie, we (a) \emph{analyze} everyday image-editing needs submitted by real users; (b) \emph{test} SOTA image editors on these requests; and (c) \emph{compare} how SOTA VLMs judge image edits differently than human judges.

We label each request with subject labels that correspond to WordNet categories (while RealEdit uses 14 manually-defined subjects) and three creativity levels (which do not exist in RealEdit).
PSR features 15 detailed editing actions compared to 6 labels of RealEdit \cite{sushko2025realedit}.

\subsec{Automated evaluation} in image editing adopts both
automated metrics and human evaluation.
Automated metrics provide an objective and scalable way to assess edits by measuring fidelity~\cite{korhonen2012peak,wang2004image}, realism~\cite{radford2021learning}, alignment with given instructions~\cite{zhang2018unreasonable, li2022blip, kim2024augmentation}, or multiple facets of an edit~\cite{ma2024i2ebench}.
Similarly, we use the LAION Aesthetics score to measure the fidelity of images.
Furthermore, to the best of our knowledge, our work is the first work to test SOTA VLMs in rating edited images given a (text request, source image) pair and compare their results with human ratings.

\subsec{Human ratings} While automatic metrics are convenient, human evaluation remains the most reliable method in capturing subjective qualities such as realism, coherence, and user satisfaction.
Public benchmarks facilitate structured human assessments, often using rating scales \cite{kawar2023imagic,basu2023editval,wang2023imagen} or pairwise comparisons ~\cite{jiang2025genai} to evaluate editing performance. 
Similar to GenAI Arena \cite{jiang2025genai}, we also present Win, Lose, and Tie options to human raters.
GenAI Arena compares models based on their images generated from scratch while our work tests image editing.

Prior human studies used a smaller set of images---EditVal \cite{basu2023editval}, TedBench \cite{kawar2023imagic}, and EditBench \cite{wang2023imagen} contain 92, 100, and 240 source images.
In contrast, PSR-\benchmarksize{} test set features 328 real-world requests with controlled creativity levels, uniform subject labels and editing actions.

\begin{figure*}[t]
    \centering
    \includegraphics[width=1\linewidth,clip]{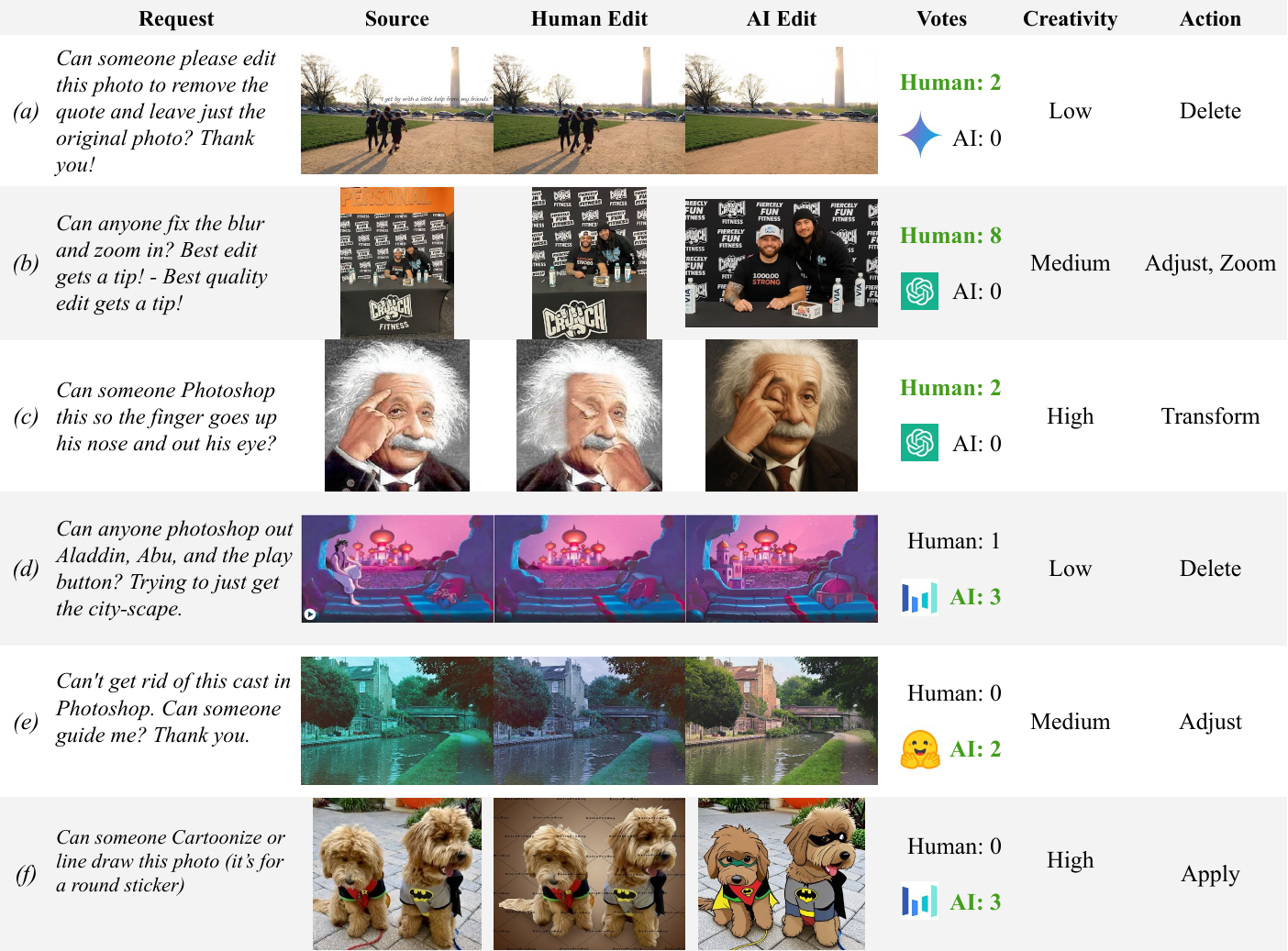}
    \caption{Example cases from PSR dataset where \hl{PSR} wizard edits were preferred by human raters over the AI edits (a-c) and samples where AI edits was preferred (d-f).
    \textbf{(a)}: The human edit completes the request, but the AI \geminilogo edit removes the people, which was not requested.
    \textbf{(b)}: The human edit completes the request, but the AI edit generates a similar image with people resembling those in the source image, although with different identities.
    \textbf{(c)}: The human edit completes the request, but the AI edit does not because the finger does not go through the nose and out the eye.
    \textbf{(d)}: Both edits make the requested removals, but the AI edit makes the image sharper and adds a house in the background (which was not requested).
    \textbf{(e)}: The human edit reduces the color cast but leaves behind a bluish tint and muted tones, while the AI edit successfully restores realistic colors and contrast.
    \textbf{(f)}: The human edit removes the background and applies a soft photo filter that lacks stylization, while the AI edit transforms the dogs into bold, clean cartoons.
    More results are available at:    \href{https://huggingface.co/spaces/PSRDataset/PSR-Battle-Results}{https://huggingface.co/spaces/PSRDataset/PSR-Battle-Results}
    }
    \label{fig:ai_wins_and_failures}
\end{figure*}

\section{PSR Dataset Construction}
\label{sec:data_creation_section}

We source our data from  \href{https://www.reddit.com/r/PhotoshopRequest/}{/r/PhotoshopRequest} (2013–2025) via two approaches: (1) PushShift~\cite{baumgartner2020pushshift} provide   historical data (2013 up to Nov 2022) (2) a custom crawler written for Reddit to download recent data (Oct 2024–Feb 2025).

\subsection{Taxonomy}
\label{sec:data_taxonomoy}

We present a taxonomy of requests where we label each request with the following labels: subject, editing action verb, and creativity level.

The subject identifies the element being modified, the action verb specifies the edit, and the creativity level distinguishes routine tasks from open-ended ones.
This taxonomy enables precise analysis of automated image editing and shows that even routine edits, like object \action{removal}, vary significantly by subject—person, animal, or object.
While low-creativity tasks often suit standard automation tools, high-creativity tasks (\cref{fig:ai_wins_and_failures}) demand models with greater flexibility.

\subsec{Subject}
The subject of an image editing request is the specific element being modified---\eg, in a request to remove a person, the subject is the person. Subjects may include objects, persons, or the entire image. To systematically classify subjects from user instructions, we leverage WordNet’s taxonomy~\cite{fellbaum2010wordnet}. We first extract subjects from raw instructions, then match each to the nearest synset (semantic category) in WordNet’s structured lexical database, providing standardized classification and reduced ambiguity.

\begin{table*}[t]
    \centering
    \renewcommand{\arraystretch}{1.1} 
    \definecolor{lightgray}{gray}{0.97} 
    \rowcolors{2}{white}{lightgray} 
\caption{List of image editing action verbs from our taxonomy with sample user requests. 
See~\cref{sec:supp-prompting_details_tax} for the VLM prompts that are used to generate these action labels.
}
    \resizebox{2\columnwidth}{!}{
    \begin{tabular}{m{2.6cm} m{8cm} m{2.6cm} m{8cm}} 
        \toprule
        \textbf{Editing action} & \textbf{Description and Sample Request} & \textbf{Editing action} & \textbf{Description and Sample Request} \\
        \midrule
        \action{add} & Insert new elements, objects, text, or effects. \textit{e.g. ``Add a copyright watermark to the bottom right.''} &
        \action{adjust}\hfill (\cref{fig:ai_wins_and_failures}b) & Modify properties like tones, contrast, and saturation. \textit{e.g. ``Increase saturation a bit on the elephants.''} \\
        \action{apply} \hfill (\cref{fig:ai_wins_and_failures}f) & Add filters, styles, or effects. \textit{e.g. ``Apply a vintage film effect.''} &
        \action{clone} & Duplicate elements inside the image. \textit{e.g. ``Use cloning tool to blend grass over dirt patches.''} \\
        \action{crop} & Trim edges for a smaller image. \textit{e.g. ``Crop to square format for social media.''} &
        \action{delete}\hfill (\cref{fig:ai_wins_and_failures}a) & Remove elements, objects, or imperfections. \textit{e.g. ``Remove the jacket hanging from the girl's side.''} \\
        \action{replace} & Substitute objects or text. \textit{e.g. ``Please change the pamphlet into a dictionary.''} &
        \action{transform} & Flip, scale, rotate, or skew elements. \textit{e.g. ``Fix the perspective of the building.''} \\
        \action{move} & Reposition elements while keeping the rest unchanged. \textit{e.g. ``Shift the logo 20 pixels up.''} &
        \action{merge} & Combine elements or effects. \textit{e.g. ``Create a panorama from these shots.''} \\
        \action{super-resolution} & Increase resolution for clearer details. \textit{e.g. ``Can someone upscale this image to 4K resolution?''} &
        \action{re-color} & Change the color of an element, object, or text. \textit{e.g. ``Can someone change the dog's fur to black?''} \\
        \action{relight} & Adjust lighting conditions. \textit{e.g. ``Can someone make lighting better / remove shadows?''} &
        \action{zoom}\hfill (\cref{fig:ai_wins_and_failures}b) & Change scale to focus or zoom out. \textit{e.g. ``Zoom in on the man.''} \\
        \action{specialized} & Advanced or composite editing tasks. \textit{e.g. ``Can someone vectorize this logo without background?''} & & \\
        \bottomrule
    \end{tabular}%
    } 
    \label{tab:image_editing_action_verbs}
\end{table*}

\subsec{Editing Action}
Users may describe their requests vaguely, (\eg, ``\textit{Make this look better}'') instead of the more technically precise phrasing such as ``\textit{Improve the lighting to make the subject stand out}''. 
Because users may not know what editing actions are needed to achieve the result or that they want an surprise, out-of-the-box result.
To properly categorize user intent, we develop a diverse list of \numberOfAction{} action verbs that cover various editing actions (\cref{tab:image_editing_action_verbs}).

Prior editing action taxonomies~\cite{manuvinakurike2018edit, tan2019expressing} are tied to low-level tools in popular image editing software, failing to capture high-level user intent and omitting actions like \action{super-resolution}. This limitation motivated us to develop a new taxonomy with comprehensive coverage of modern editing techniques.

To develop our taxonomy, we feed a random subset of 5,000 edit requests into \gptmini~\gptlogo and prompt it to summarize common editing actions.
Additionally, we consult image editing experts in the field to refine our list of actions to accurately reflect image-synthesis tasks in computer vision (\eg, \action{super-resolution}).
\cref{tab:image_editing_action_verbs} presents the final list of action verbs.

\subsec{Creativity levels} 
We categorize requests into three creativity levels:

\begin{compactenum}
    \item \textbf{\colorbox{rowLow}{Low}-creativity} requests such as \emph{``remove a person''} or \emph{``erase an object''} expect a predictable outcome (\cref{fig:ai_wins_and_failures}d).
    
    \item \textbf{\colorbox{rowMed}{Medium}-creativity} requests such as \emph{``change the background''} or \emph{``adjust lighting to look cinematic''} allow some room for interpretation and variability in results (\cref{fig:ai_wins_and_failures}e).
    
    \item \textbf{\colorbox{rowHigh}{High}-creativity} requests, \eg, \textit{``make this image look magical''} or \textit{``transform this into a fantasy scene''}, require creative ideas and can yield widely different outcomes depending on the editor imagination (\cref{fig:ai_wins_and_failures}f).
\end{compactenum}

\begin{figure*}[ht]
    \centering
    \subfloat[Editing action popularity per year]{
        \includegraphics[width=0.8\columnwidth]{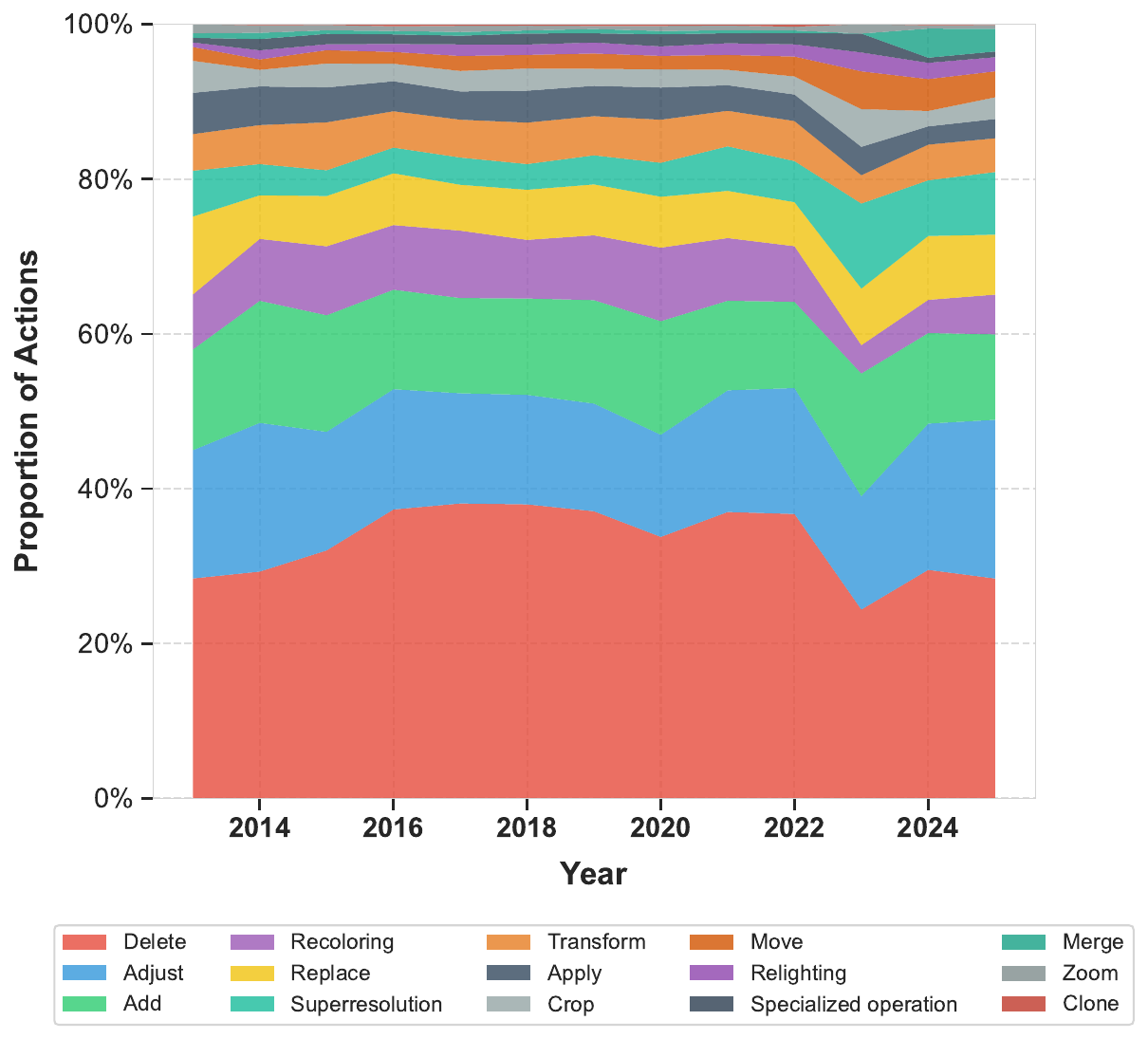}
        \label{fig:action_verbs_over_the_years}
    }
    \hfill
    \subfloat[Most common actions \& subjects]{
        \includegraphics[width=0.8\columnwidth]{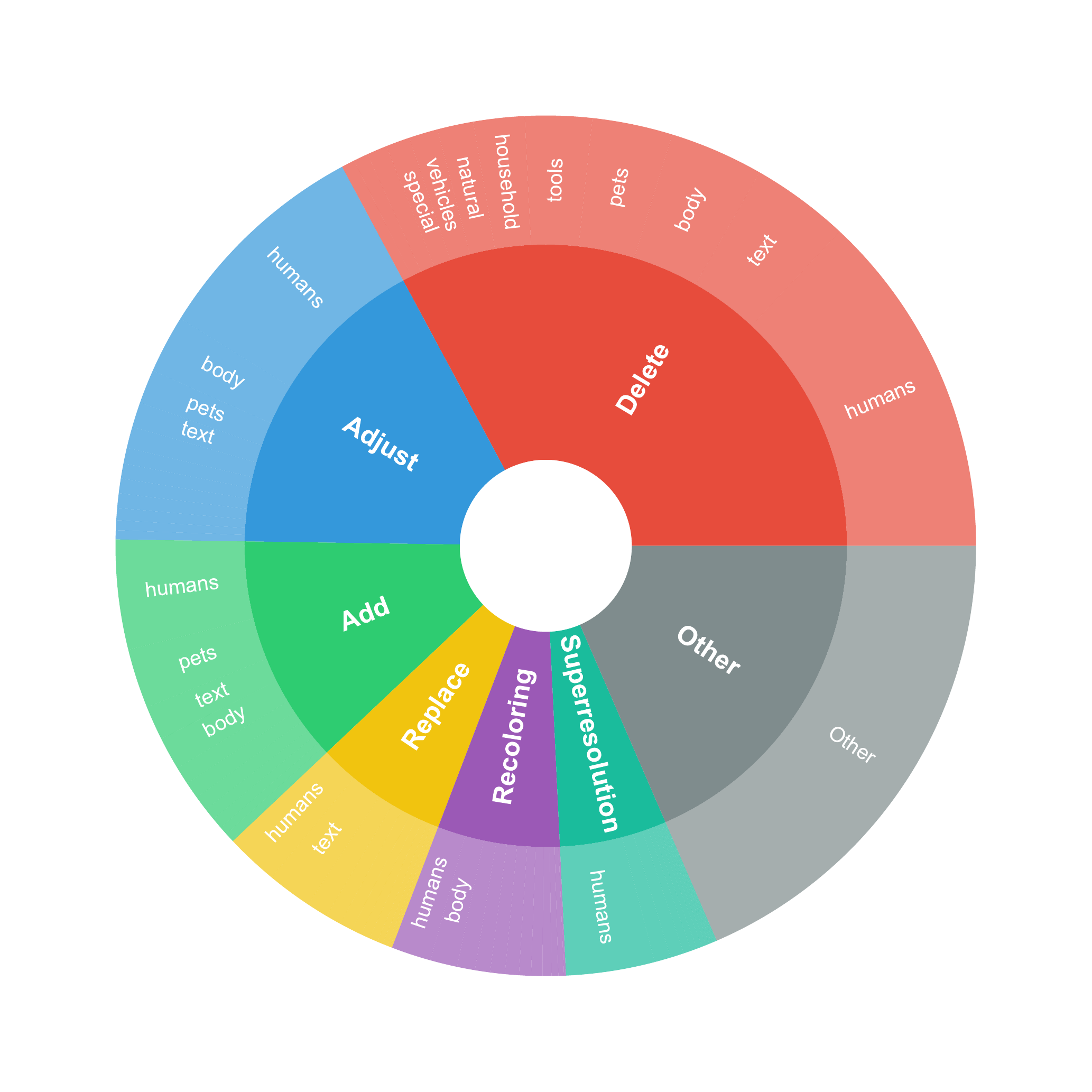}
        \label{fig:action_subject}
    }
    \\
    \subfloat[Action distribution per creativity level]{
        \includegraphics[width=0.8\columnwidth]{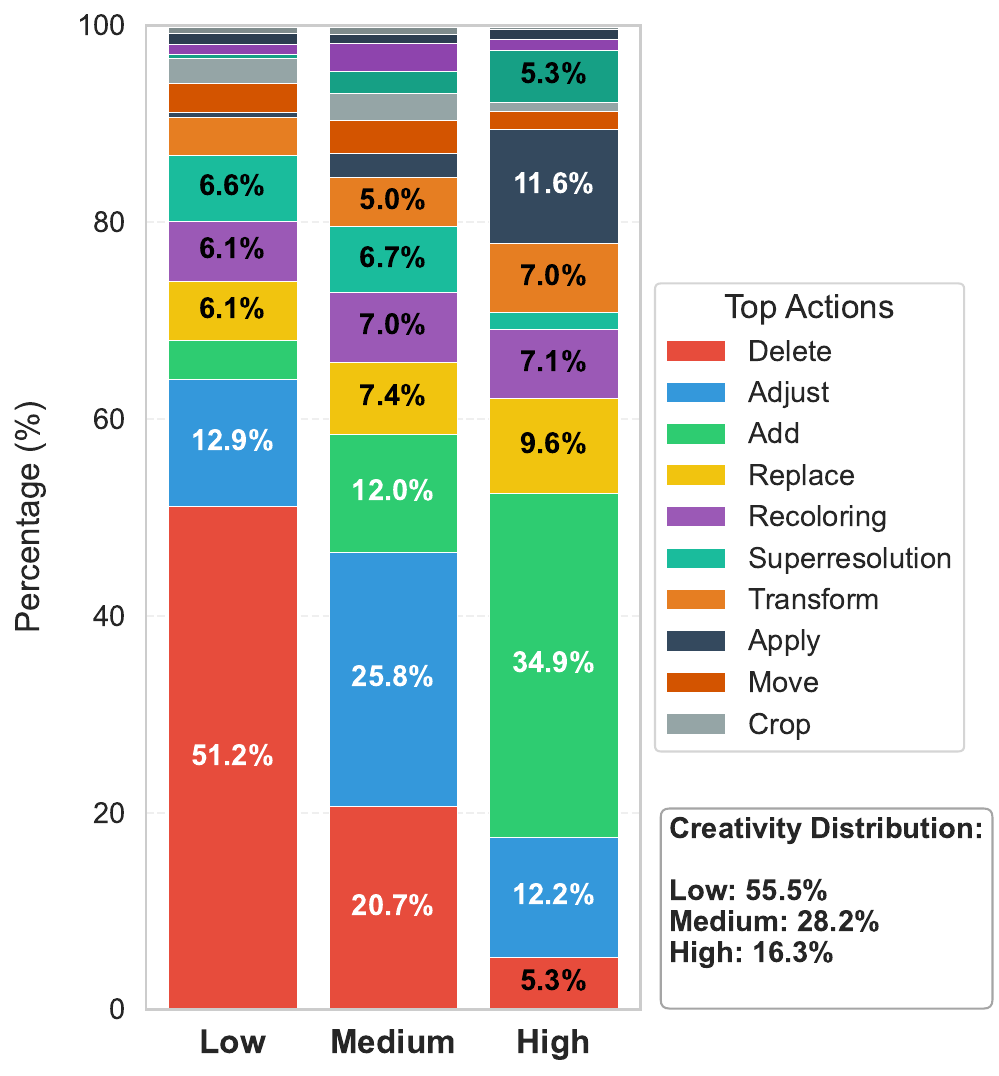}
        \label{fig:creativity_levels}
    }
    \hfill
    \subfloat[Creativity level per year]{
        \includegraphics[width=0.8\columnwidth]{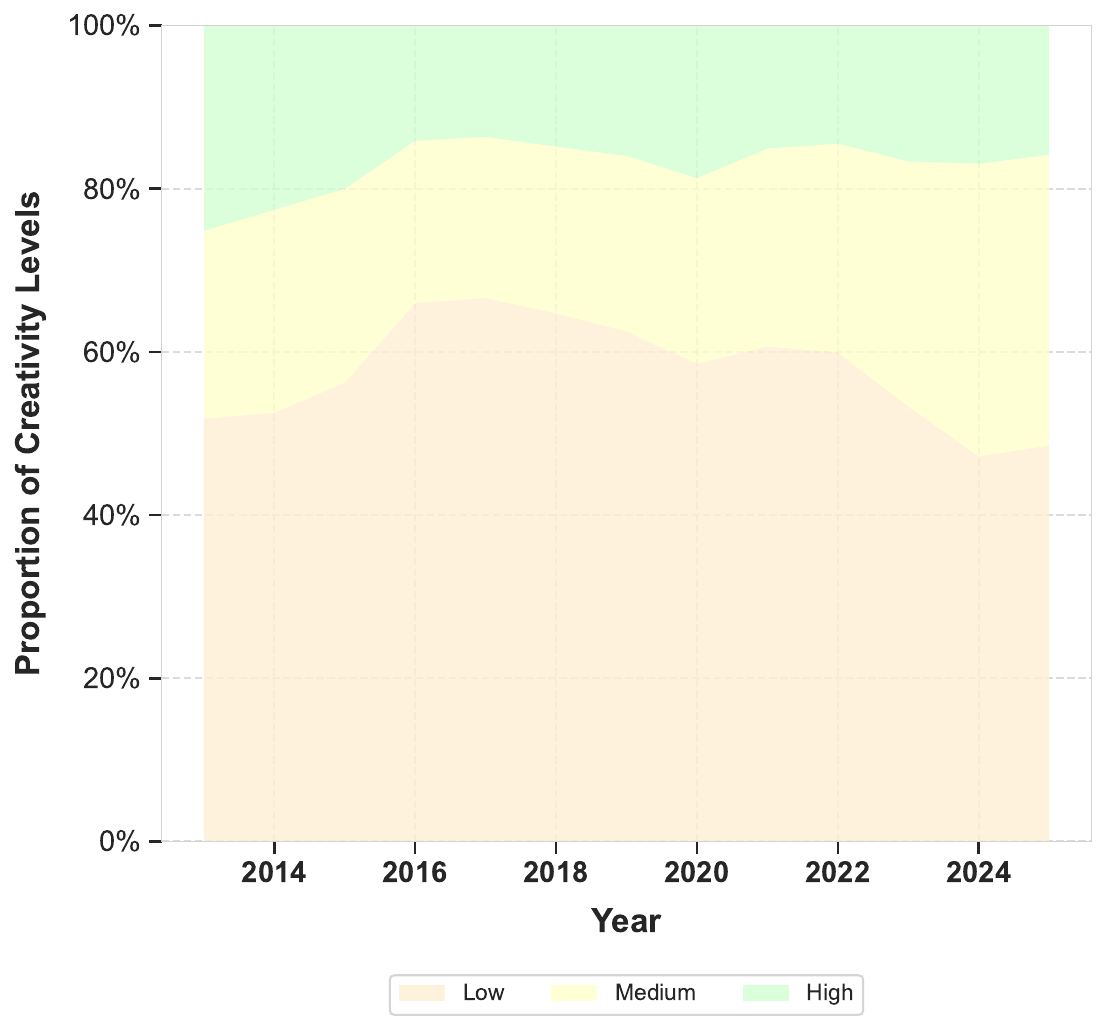}
        \label{fig:creativity_levels_trend_year}
    }
    \caption{
        Over 12 years of Reddit data, \action{\footnotesize delete}, \action{\footnotesize adjust}, and \action{\footnotesize add} are the top-3 most wanted actions (a).
        Specifically, humans, body parts, text, and pets are the most frequent WordNet subjects for such common actions (b).
        While \action{\footnotesize delete} and \action{\footnotesize adjust} are the top-2 most common actions in the \colorbox{rowLow}{low}- and \colorbox{rowMed}{medium}-creativity requests, \action{\footnotesize add} takes up the largest share (34.9\%) in \colorbox{rowHigh}{high}-creativity, \eg, inserting some ``interesting'' background or objects into the scene (c).
        Most requests (55.5\%) require straightforward edits with \colorbox{rowLow}{low} creativity (d).
    }
    \label{fig:appendix_12_years_data}
\end{figure*}

Our creativity classification differentiates between requests for precise technical edits vs. imaginative, open-ended transformations and enables breakdown analysis across requests, actions, and creativity.
For example, the labels in PSR show that \action{delete} is indeed the most common action in  \colorbox{rowLow}{low}-creativity requests (\cref{fig:creativity_levels}).
That is, users often specify an specific element in the image to be deleted.
In contrast, adding a new elements into the image has much larger space for creativity: what object, how, and where to add?
As the result, we find \action{add} to be the most common action in \colorbox{rowHigh}{high}-creativity requests (\cref{fig:creativity_levels}).


\subsection{Dataset Annotation Process}
\label{sec:dataset_annotation_process}

We use \gptmini~\cite{openai2025gpt4omini} and \internvlful~\cite{chen2024expanding} to annotate our dataset. \gptmini generates taxonomy labels, while \internvl handles image captioning and keyword extraction. We prompt \internvl to summarize each image into 36 JSON keys (see \cref{fig:supp-json-schema}), capturing attributes such as image type (e.g., photo or digital art), location, weather, presence of people, and object lists.
To filter out posts unrelated to image editing (e.g., requests for image authentication), we prompt \gptmini to output a binary flag, \emph{image editing relevance}, indicating whether the post involves image editing. Prompting details, including model version and temperature, are in \Cref{sec:supp-prompting_details}.

\subsec{Extracting Subject and Action Verbs} 
We use zero-shot prompting (see \cref{fig:supp-action_verbs_prompt}) to extract actions from the request (\cref{tab:image_editing_action_verbs}).  
\gptmini is provided with a list of valid actions, their descriptions, the input image, and the user-provided request. 
We ask the model to first examine the image and then rewrite the instruction in clear and simplified language to eliminate ambiguity, and finally identify the subjects of the edits along with the corresponding editing actions.

\subsec{Mapping Subjects to WordNet} 
We map the extracted subjects from the previous stage to WordNet’s synsets. Once the subjects are identified, we provide \gptmini with the image, instruction, and subject, instructing it to select the closest WordNet synset based on the given context. Since the generated synset may not always be valid, we perform a search within the WordNet lexical database using NLTK to find the closest matching synset and assign a final synset to the subject.

We use \oonepro~\cite{openai2025o1pro} to summarize WordNet subjects into higher-level semantic categories and organize them into 5 main categories and 12 subcategories (\cref{tab:category_hierarchy}). 
Since the extracted WordNet synsets from the previous step vary in granularity, a reasoning model with long-context capabilities, such as \oonepro, effectively groups these synsets into structured categories. This approach results in a more coherent and meaningful subject categories.

\subsec{Assigning Creativity Levels}
We use few-shot prompting (\cref{fig:supp-creativity_prompt}) to assign creativity levels. \gptmini receives the original image and request, along with examples annotated by creativity level, to classify the input accordingly.

\subsection{PSR dataset statistics}
\label{sec:dataset_statistics}

PSR consists of \datasetAllPostsFinaldataset{} requests and \datasetAllEditsFinaldataset{} edited images, categorized by creativity levels as \colorbox{rowLow}{55.5\%} low, \colorbox{rowMed}{28.2\%} medium, and \colorbox{rowHigh}{16.3\%} high (\cref{fig:creativity_levels_trend_year}). 
It includes 49,134 unique subjects, with 53.5\% of subjects are under the \emph{People and Related} category (\cref{fig:action_subject}).  
The primary action requested by users is \action{delete} (32.9\%), typically involving removing individuals or visual clutter to enhance aesthetics or professionalism.
And these action distribution remain similar over the years (\cref{fig:action_verbs_over_the_years}).
Detailed dataset statistics are provided in~\Cref{sec:supp-dataset_stats}.



\section{Experiment Setup}
\label{sec:exp_setup}

We compare AI edits and human edits to understand the preference of human raters and assess how VLMs can match human raters.

\subsec{AI editors}
We process each request using three generalist SOTA image editing tools: \SeedEdit{}~\seedlogo~\cite{shi2024seededit}, \geminiflash~\geminilogo\ (Image Generation Experimental)~\cite{geminiImage}, and \gpt{}~\gptlogo~\cite{openai2025_4oimage}.\footnote{Due to access constraints and model safeguards, some edits from ~\gptlogo, \seedlogo, and \geminilogo are not available.}
For each request, we generate two images: one using the original instruction (OI) and one using a simplified instruction (SI) generated by \gptmini{}. As user-written instructions often include unnecessary details (\eg, \emph{``Thank you in advance!!''} in \cref{fig:teaser}), \gptmini{} refines them to focus on the core editing task.

Additionally, for each request, we generate three AI-based image edits using \numberOfModels{} off-the-shelf image editing models hosted on HuggingFace Spaces~\huggingface~\cite{huggingface_spaces} (see~\cref{table:supp-edits-per-model} for a list of models).

\subsec{PSR-328 dataset}
Via stratified sampling, we select a random PSR subset that has almost the same amount of images in three levels of creativity groups (114 low, 101 medium, and 113 high).
On average, each request is edited by an average of \textbf{five} Reddit human editors, resulting in a total of 1,644 human edits.
We also generate 7 different AI edits per request. 
In total, our human study has 10,405 unique 4-tuples (source image, request, AI edit, human edit) for evaluation---substantially larger than prior studies~\cite{basu2023editval, wang2023imagen, jiang2025genai}, while keeping the human annotation effort manageable.

\subsec{Human study} 
We conduct a comparative study to assess whether AI-generated or Reddit user edits better satisfy original requests. To ensure unbiased evaluation, randomly paired AI and human edits are shown to human raters, who vote (Win, Lose or Tie) on which best fulfills the request (see \Cref{sec:supp-human-study} for the user interface).

\subsec{Automated metrics and VLM judges} 
We use two automated evaluation methods to complement our human study: LAION \emph{Aesthetic Score} \cite{hentschel2022clip, wang2023exploring, wu2024q} and \emph{VLM-as-a-Judge} \cite{lee2024prometheus, chen2024iqagpt, xiong2024llava, chen2024mllm}.
The Aesthetic Score quantifies visual appeal using a classifier trained on large-scale human preference data while VLM-as-a-Judge uses VLMs to evaluate images by articulating visual qualities and explaining judgments. 
Both metrics act as proxies for human judgment, enabling scalable evaluation of AI- and human-edited images.
We use LAION's Aesthetic Score Predictor~\cite{schuhmann2023improved} for calculating aesthetic metrics, and three VLMs (\gpt~\gptlogo~\cite{openai2024gpt4o}, \oone~\openailogo~\cite{openai_o1}, and \geminiflashthinking~\geminilogo~\cite{GoogleDeepMind2024Gemini2}) as as judges for the image edit rating task.

\begin{table*}[t]
\centering
\caption{Human and VLM evaluation results for image editing preferences across 114 \colorbox{rowLow}{low}-, 101 \colorbox{rowMed}{medium}-, and 113 \colorbox{rowHigh}{high}-creativity requests. 
Human raters often prefer $\mathcal{H}$uman edits over all AI edits, while VLMs show a strong preference for edits made by \SeedEdit{}~\seedlogo{} and \gpt~\gptlogo. \# denotes the number of votes (\ie, pair-wise comparisons).
Bold numbers are the \textbf{highest} \% within each VLM-judge group.
\\ 
AI: the win rate of AI-generated edits (\%).\\ 
$\mathcal{H}$: the win rate of human, PSR-wizard edits (\%). \\
 }
\label{tab:main_results}
\setlength{\tabcolsep}{6pt}
\renewcommand{\arraystretch}{1.2} 
\resizebox{\linewidth}{!}{%
    \begin{tabular}{l rrrr : rrrr : rrrr : rrrr}
                & \multicolumn{4}{c:}{a. Human raters}              %
                & \multicolumn{4}{:c:}{b. \gptlogo{} \gpt judge} %
                & \multicolumn{4}{:c:}{c. \openailogo{} \oone judge} %
                & \multicolumn{4}{:c:}{d. \geminilogo{} \geminiflashthinking{}} \\[2pt]
    \cmidrule(lr){2-5} \cmidrule(lr){6-9} \cmidrule(lr){10-13} \cmidrule(lr){14-17}
    \textbf{Group}
                & \# 
                & $\mathcal{H}$uman 
                & AI 
                & Tie                           
                & \# & $\mathcal{H}$ & AI & Tie                          
                & \# & $\mathcal{H}$ & AI & Tie                           
                & \# & $\mathcal{H}$ & AI & Tie \\ 
\hline

All data     & \cellcolor{verylightpink}4,359 & \cellcolor{verylightpink}\textbf{66.0} & \cellcolor{verylightpink}25.8 &\cellcolor{verylightpink}8.2 & \cellcolor{lightgray}10,313 & \cellcolor{lightgray}42.1 & \cellcolor{lightgray}\textbf{ 52.4 }& \cellcolor{lightgray}5.5 & \cellcolor{lightgray}10,352 & \cellcolor{lightgray}\textbf{51.6} & \cellcolor{lightgray}47.5 & \cellcolor{lightgray}0.9 & \cellcolor{lightgray}10,354 & \cellcolor{lightgray}41.9 & \cellcolor{lightgray}\textbf{ 52.5} & \cellcolor{lightgray}5.5 \\
\hline

\seedlogo~\SeedEdit   & \cellcolor{lightblue}886 & \cellcolor{lightblue}\textbf{53.6} &\cellcolor{lightblue} 37.8 & \cellcolor{lightblue}8.6 & \cellcolor{lightlilac}1,735 & \cellcolor{lightlilac}31.9 & \cellcolor{lightlilac}\textbf{60.6} & \cellcolor{lightlilac}7.5 & \cellcolor{lightlilac}1,744 & \cellcolor{lightlilac}42.7 & \cellcolor{lightlilac}\textbf{56.6} &\cellcolor{lightlilac} 0.7 & \cellcolor{lightlilac}1,744 & \cellcolor{lightlilac}36.1 & \cellcolor{lightlilac}\textbf{55.0} & \cellcolor{lightlilac}8.9 \\
\gptlogo~\gpt             & \cellcolor{lightblue}1,014 & \cellcolor{lightblue}\textbf{61.5} & \cellcolor{lightblue}32.8 & \cellcolor{lightblue}5.6 & \cellcolor{lightlilac}2,609 & \cellcolor{lightlilac}11.0 & \cellcolor{lightlilac}\textbf{85.3} & \cellcolor{lightlilac}3.7 & \cellcolor{lightlilac}2,618 & \cellcolor{lightlilac}15.9 & \cellcolor{lightlilac}\textbf{83.9} & \cellcolor{lightlilac}0.3 & \cellcolor{lightlilac}2,623 & \cellcolor{lightlilac}17.3 &\cellcolor{lightlilac} \textbf{81.2} &\cellcolor{lightlilac} 1.4 \\
\geminilogo~\Gemini                   & \cellcolor{lightblue}681 & \cellcolor{lightblue}\textbf{70.0} & \cellcolor{lightblue}20.6 & \cellcolor{lightblue}9.4 & \cellcolor{lightlilac} 2,518 & \cellcolor{lightlilac} \textbf{ 51.5} & \cellcolor{lightlilac} 42.6 &  \cellcolor{lightlilac} 5.9 & \cellcolor{lightlilac} 2,524 & \cellcolor{lightlilac} \textbf{ 63.9 }& \cellcolor{lightlilac} 35.1 & \cellcolor{lightlilac} 1.0 & \cellcolor{lightlilac} 2,524 &\cellcolor{lightlilac}  46.8 & \cellcolor{lightlilac} \textbf{ 46.9} &  \cellcolor{lightlilac} 6.3 \\
\huggingface{}HF           & \cellcolor{lightblue}1,778 & \cellcolor{lightblue}\textbf{73.2} &\cellcolor{lightblue} 17.8 & \cellcolor{lightblue}9.1 &  \cellcolor{lightlilac} 3,451 & \cellcolor{lightlilac} \textbf{63.9} & \cellcolor{lightlilac} 30.5 &\cellcolor{lightlilac}  5.5 & \cellcolor{lightlilac} 3,466 & \cellcolor{lightlilac} \textbf{74.0 }&\cellcolor{lightlilac}  24.6 & \cellcolor{lightlilac} 1.4 & \cellcolor{lightlilac} 3,463 & \cellcolor{lightlilac} \textbf{60.1} & \cellcolor{lightlilac} 33.6 & \cellcolor{lightlilac} 6.3 \\
\midrule
\multicolumn{17}{l}{%
  \qquad\qquad\qquad\qquad\textbf{Creativity}\quad\qquad\qquad\qquad\qquad\qquad
  \sq{rowLow}\,Low\qquad\qquad\qquad\qquad\quad\quad
  \sq{rowMed}\,Medium\qquad\qquad\qquad\qquad\quad
  \sq{rowHigh}\,High
}\\
\midrule
\rowcolor{rowLow}Low & 1,485 & \textbf{70.1} & 21.1 & 8.8 & 3,468 & 45.2 & \textbf{48.2} & 6.5 & 3,474 & \textbf{56.9} & 41.9 & 1.2 & 3,486 & 42.4 &\textbf{ 44.6} & 13.1 \\

\rowcolor{rowMed}Medium & 1,282 & \textbf{67.6} & 25.3 & 7.2 & 3,207 & 39.6 & \textbf{56.6} & 3.7 & 3,222 & \textbf{50.9} & 48.7 & 0.4 & 3,218 & 39.3 & \textbf{57.9} & 2.8 \\
\rowcolor{rowHigh}High & 1,587 & \textbf{60.9} & 30.7 & 8.4 & 3,632 & 41.3 & \textbf{52.6} & 6.1 & 3,650 & 47.0 &\textbf{ 52.0} & 1.1 & 3,644 & 43.8 & \textbf{55.5} & 0.7 \\
\midrule
\cellcolor{rowLow} \seedlogo~\SeedEdit   & 335 & \textbf{63.3 }& 29.3 & 7.5 & 660 & 40.5 & \textbf{51.4} & 8.2 & 665 & \textbf{53.5} & 45.4 & 1.1 & 666 & 39.9 & 40.7 & 19.4 \\
\cellcolor{rowLow} \gptlogo~\gpt  & 352 & \textbf{71.6} & 22.2 & 6.2 & 869 & 13.7 & \textbf{81.8 }& 4.5 & 870 & 21.4 & \textbf{78.6} & 0.0 & 874 & 24.5 & \textbf{71.7} & 3.8 \\
\cellcolor{rowLow} \geminilogo~\Gemini  & 213 & \textbf{70.9} & 19.7 & 9.4 & 783 & \textbf{55.8} & 36.7 & 7.5 & 779 & \textbf{73.3} & 25.0 & 1.7 & 782 & \textbf{46.2} & 37.9 & 16.0 \\
\cellcolor{rowLow} \huggingface{}HF  & 585 & \textbf{72.8} & 16.2 & 10.9 & 1,156 & \textbf{64.5} & 29.1 & 6.4 & 1,160 & \textbf{74.6} & 23.6 & 1.8 & 1,164 & 54.6 & 30.9 & 14.4 \\
\midrule
\cellcolor{rowMed} \seedlogo~\SeedEdit   & 159 & \textbf{54.1} & 39.0 & 6.9 & 349 & 25.2 & \textbf{70.5} & 4.3 & 350 & 46.3 &\textbf{ 53.4} & 0.3 & 350 & 31.1 & \textbf{62.9} & 6.0 \\
\cellcolor{rowMed} \gptlogo~\gpt  & 359 & \textbf{61.8} & 34.5 & 3.6 & 938 & 9.5 & \textbf{88.0 }& 2.6 & 942 & 15.2 & \textbf{84.5 }& 0.3 & 942 & 13.2 & \textbf{86.4} & 0.4 \\
\cellcolor{rowMed} \geminilogo~\Gemini  & 211 & \textbf{73.9} & 17.5 & 8.5 & 869 & \textbf{51.7 }& 44.3 & 4.0 & 876 & \textbf{60.3 }& 39.4 & 0.3 & 875 & 47.5 & \textbf{49.3} & 3.2 \\
\cellcolor{rowMed} \huggingface{}HF  & 553 & \textbf{72.7} & 18.3 & 9.0 & 1,051 & \textbf{61.4} & 34.3 & 4.4 & 1,054 & \textbf{76.7 }& 22.8 & 0.6 & 1,051 & \textbf{58.6} & 37.8 & 3.6 \\
\midrule
\cellcolor{rowHigh} \seedlogo~\SeedEdit   & 392 & \textbf{45.2} & 44.6 & 10.2 & 724 & 27.5 & \textbf{64.1} & 8.4 & 727 & 31.1 & \textbf{68.2} & 0.7 & 726 & 34.8 & 6\textbf{4.5 }& 0.7 \\
\cellcolor{rowHigh} \gptlogo~\gpt  & 303 &\textbf{ 49.5} & 43.2 & 7.3 & 802 & 9.9 & \textbf{85.9} & 4.2 & 806 & 10.7 & \textbf{88.8} & 0.5 & 807 & 14.4 & \textbf{85.5 }& 0.1 \\
\cellcolor{rowHigh} \geminilogo~\Gemini  & 256 & \textbf{66.0} & 23.8 & 10.2 & 865 & \textbf{47.4 }& 46.2 & 6.4 & 868 & \textbf{59.0} & 39.9 & 1.2 & 866 & 46.4 & \textbf{52.8} & 0.8 \\
\cellcolor{rowHigh} \huggingface{}HF  & 636 & \textbf{73.9} & 18.9 & 7.2 & 1,241 &\textbf{ 65.4} & 28.8 & 5.7 & 1,249 & \textbf{71.3 }& 27.1 & 1.6 & 1,245 & \textbf{66.3} & 32.6 & 1.0 \\

    \bottomrule
    \end{tabular}
}
\end{table*}


\section{Experimental Results}
\label{sec:results}

\subsection{Human edits are strongly preferred over AI edits by human raters}
\label{sec:human_prefer_human}

We collect 4,359 votes from 122 users and \cref{tab:main_results} shows the results. 
On average, human edits are more preferred (66.0\% of the time), while 25.8\% of the time AI edits win, and they tie in the remaining 8.2\%
(\cref{tab:main_results}; \colorbox{verylightpink}{25.8\%}).
This result is important given the global interest in the image generation capabilities of \geminiflash{} and \gpt \cite{geminiImage}.

\subsec{AI-win-rates by models}
Human edits are more preferred compared to every single AI editor (\cref{tab:main_results}; Human win rates $\geq$ \colorbox{lightblue}{53\%}).
\seedlogo has the highest overall win rate (37.8\%) over human editors, followed by \gptlogo (32.8\%), and then \geminilogo (20.6\%).
HuggingFace models \huggingface{} perform the worst (\cref{tab:main_results}; 17.8\%).

\subsec{AI-win rates by creativity levels}
Human edits consistently are more preferred across all three levels. 
However, there is a clear trend that AI edits are more preferred as the requests require higher creativity.
That is, the gap human-vs-AI becomes smaller as the tasks are more open-ended: 70.1\% vs.\ 21.1\% for \colorbox{rowLow}{low} creativity, (67.6\% vs.\ 25.3\%) for \colorbox{rowMed}{medium} creativity, and (60.9\% vs.\ 30.7\%) for \colorbox{rowHigh}{high} creativity.


\subsec{Performance by editing actions}
On average over all 49 models (\cref{tab:verb-win-rates}), AIs win the most on the following editing requests: \action{merge} (30.9\%), \action{apply} (30.6\%), and \action{add} (30.6\%) and the least on \action{zoom} (10.7\%), \action{crop} (15.5\%), and \action{move} (20.2\%).
For a detailed breakdown, see \Cref{sec:supp-win_rate_breakdown}.

\subsec{Qualitative analysis} 
We also conduct a qualitative analysis (\Cref{sec:supp-why_ai_wins_or_lose}) to identify patterns in cases where AI edits succeed or fail. 
From analyzing 206 cases where AI edits \textbf{\textcolor{ForestGreen}{win}} votes, we find that 72\% of the time, human edits poorly follow the instructions and AIs more closely follow the requests.
In contrast, after analyzing 400 cases where AI \textbf{\textcolor{BrickRed}{loses}}, we find that 43\% of the time, AIs misinterpret the request.
In the remaining loss cases, AIs introduce unintended changes, artifacts or facial distortions.
A key issue is their failure to preserve identity (\cref{fig:aesthetic_scores_box_plot}a--b).

\begin{figure*}[h]
  \centering
  \includegraphics[width=\linewidth]{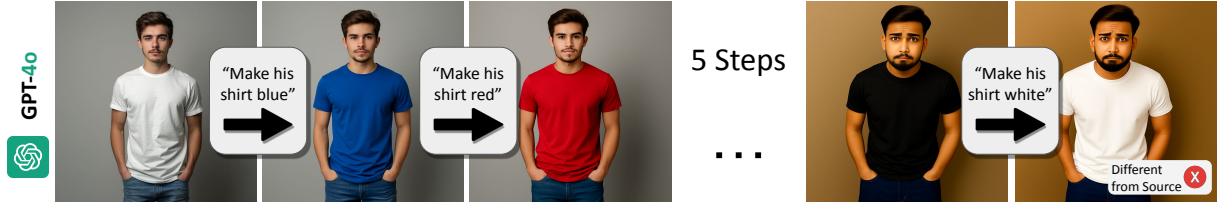}
\caption{
AIs can significantly alter both a person’s identity and the overall image quality through iterative image-editing requests.
\gpt{} was repeatedly instructed to modify the shirt color, with each step's output serving as the input for the next iteration.
Over iterations, facial identity, body shape, and the background shift away from the original person (more details in~\Cref{sec:appendix-face_diff}).
}
\label{fig:gpt_facial_fails}
\end{figure*}

To quantify the issue of failing to preserve identity, we perform a controlled experiment where we repeatedly ask AI editors to only change the shirt color of a person of varying gender and ages.
However, both \gptlogo and \geminilogo models tend to change the facial identity and even body shape over a sequence of multi-turn requests (\cref{fig:gpt_facial_fails} and \Cref{sec:appendix-face_diff}).

\subsection{In rating edited images, VLMs are a poor proxy for human raters}
\label{sec:vlms_are_bad_proxy}

Evaluating edited images is naturally a multimodal task that is challenging because it requires understanding the (text request, source image) pair and analyzing how the changes in the edited image satisfies the request.
Assessing how SOTA VLMs perform at edit rating is important for (1) benchmarking and advancing future VLMs; and (2) automating the rating efforts currently performed by humans.

\subsec{Experiment}
We collect over 10k ratings from each of three separate VLMs that serve as judges between human edits and AI edits.
We replicate the same setup as in~\cref{sec:human_prefer_human}. Specifically, given a textual request, a source image, and two edited images, we ask each VLM to judge which edit better satisfies the request. In this setup, the images are labeled as Source, Edit A, and Edit B. Human-generated and AI-generated edits are randomly assigned to labels A and B. The VLM then evaluates the edits and delivers a verdict of either \emph{Edit A is better}, \emph{Edit B is better}, or \emph{Tie}, indicating that both edits equally satisfy the request. The system message and prompting method are detailed in~\cref{fig:supp-judge_prompt} and~\cref{fig:supp-judge_prompt_python_code}.


\subsec{Result} 
While humans strongly prefer human-created edits (\cref{sec:human_prefer_human}), the trend is mixed for all VLMs.
On average, over all human \& AI edits, the three VLM judges (\gpt, \oone, and \geminiflash{}) choose AI edits at a \textbf{near-random chance}, \ie, 52.4\%, 51.6\% and 52.5\% of the time (\cref{tab:main_results}).
Interestingly, all three VLM judges prefer edits by \gpt at an extremely high rate ($\geq$ 81\%).
Additionally, Cohen's $\kappa$ scores between humans and every AI editor (\Cref{tab:new_kappa})  further confirm weak agreement between human and VLM ratings.
Among the three VLMs, \oone{} is the most consistent with humans but still scores a low $\kappa = 0.22$.

When examining VLM ratings by individual \colorbox{lightlilac}{model groups}, we observe a clear preference among VLMs for edits produced by \SeedEdit{} and \gpt{}.
Notably, \oone{} selects \gpt{} edits 83.9\% of the time. 
In contrast, when comparing human edits vs. those by \geminiflash{} or Hugging Face models, VLMs generally favor human edits. 
Still, agreement with human judgments is low, with Cohen’s $\kappa$ ranging from $0.14$ to $0.25$—nearing random in some cases (\Cref{tab:new_kappa_model_group}).

\subsec{Qualitative analysis}
Analyzing the textual chain-of-thought responses of VLM judges, we find that they are often \emph{blind} to critical image details \cite{Rahmanzadehgervi_2024_ACCV} and miss differences between image pairs (\Cref{sec:supp-vlm-as-judge_blind}). 
VLMs also overlook key aspects, such as changes in characters' identities, or hallucinate nonexistent elements (\Cref{sec:supp-judge-fails}). 
These issues underscore the ongoing challenges of using VLMs for judging image edits.

\begin{figure}[ht]
    \centering
            \centering
            \includegraphics[width=\columnwidth]{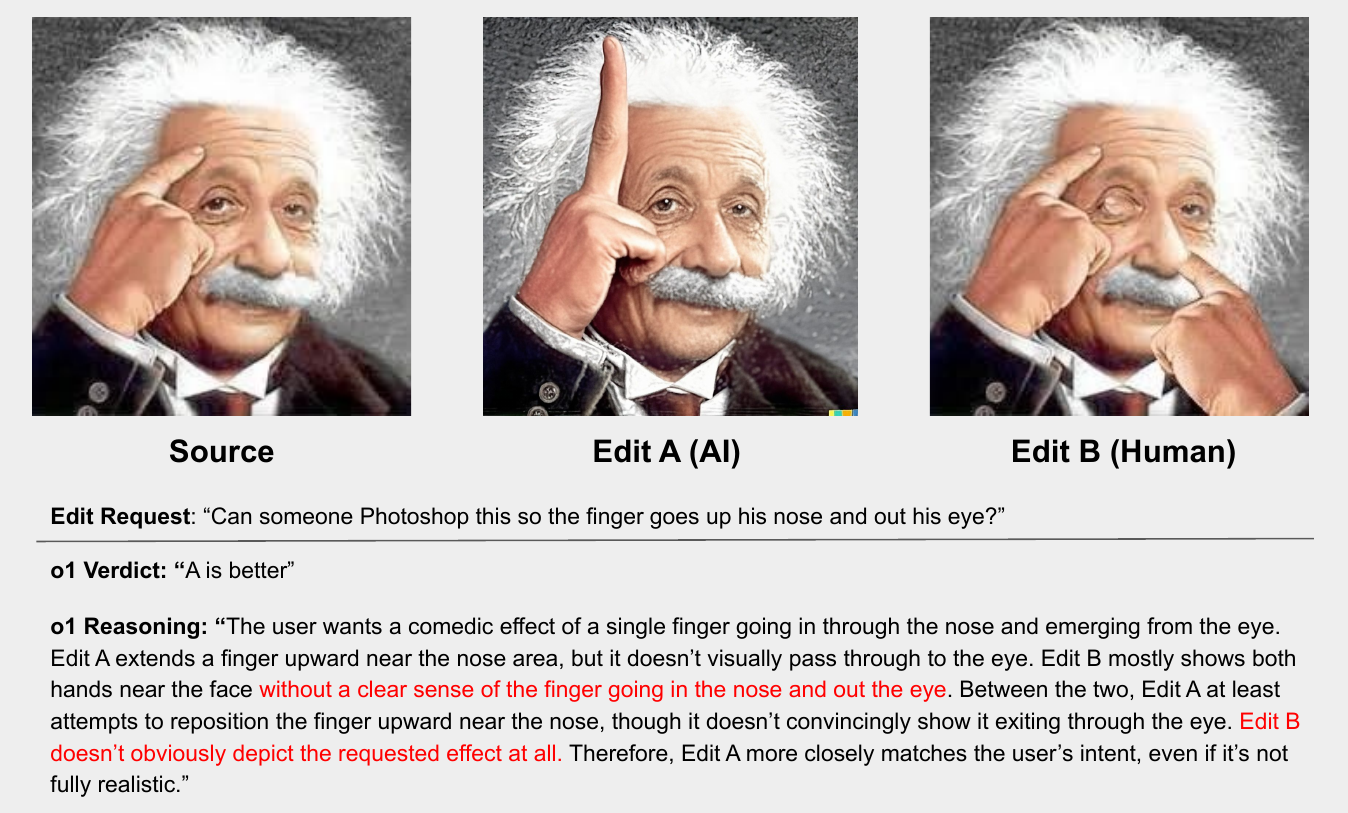}
\caption{\oone{} judge occasionally fails to notice details in edited images, here, overlooking the position of the hand and the configuration of the fingers.}
    \label{fig:o1-hallucination}
\end{figure}

\subsection{AI editors often improve aesthetics even when not requested}
\label{sec:ai_improve_aesthetics}

We find a strong pattern among all AI editors is their tendency to enhance image aesthetics, even without explicit instruction. 

\begin{figure*}[ht]
  \centering
  \begin{minipage}[t]{0.495\textwidth}
    \centering
    \begin{subfigure}[t]{0.95\textwidth}
      \includegraphics[width=\linewidth]{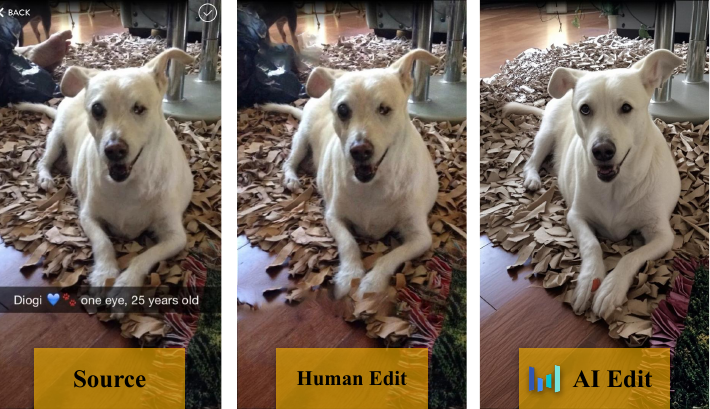}
      \caption{Request: \emph{Could someone please remove the text and the foot in the background?}
      }   
      \label{fig:aesthetic_a}
    \end{subfigure}\vspace{0.5em}

  \end{minipage}\hfill
  \begin{minipage}[t]{0.495\textwidth}
    \begin{subfigure}[t]{0.95\textwidth}
      \includegraphics[width=\linewidth]{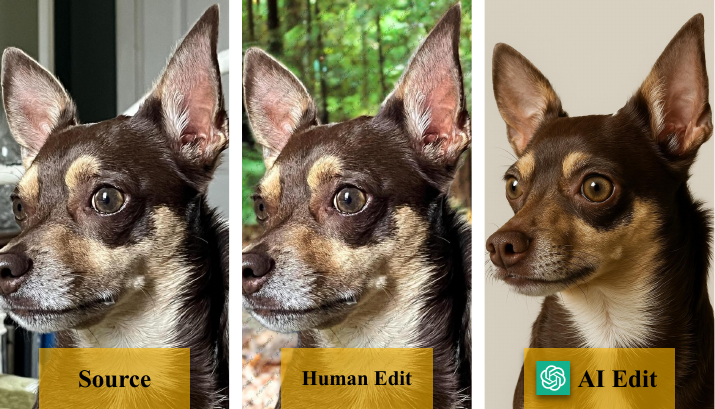}
      \caption{Request: \emph{Could anybody make the background simple and neutral?}}   
      \label{fig:aesthetic_b}
    \end{subfigure}
    
  \end{minipage}

  \caption{%
    AI models tend to increase overall image aesthetics. 
    While both requests (a) and (b) ask for changes in the background, AI editors tend to enhance the subject's facial features.
    SeedEdit \seedlogo{} modifies the dog's eye---despite it not being requested---whereas the human edit keeps the eye intact (a).
    \gpt~\gptlogo noticeably changes the dog's ears and the fur (b).
  }
  \label{fig:aesthetic_scores_box_plot}
\end{figure*}

\subsec{Qualitative analysis}
For example, they often touch up human faces, making skin appear smoother and more polished (data not shown due to concerns of revealing identity of real people).
Similarly, AI models enhance pets’ facial features.
For example, when instructed simply to remove the text and foot of a dog (\cref{fig:aesthetic_a}) or image background (\cref{fig:aesthetic_b}), \SeedEdit{} and \gpt{} perform the requested change \emph{and also} improve the dog's overall facial aesthetics (\cref{fig:aesthetic_b}) and even restore damaged eye (\cref{fig:aesthetic_a}), which is not requested.

\begin{figure*}[ht]
  \centering
  \begin{minipage}[t]{0.34\textwidth}
    \centering
    \begin{subfigure}[t]{\textwidth}
      \includegraphics[width=\linewidth]{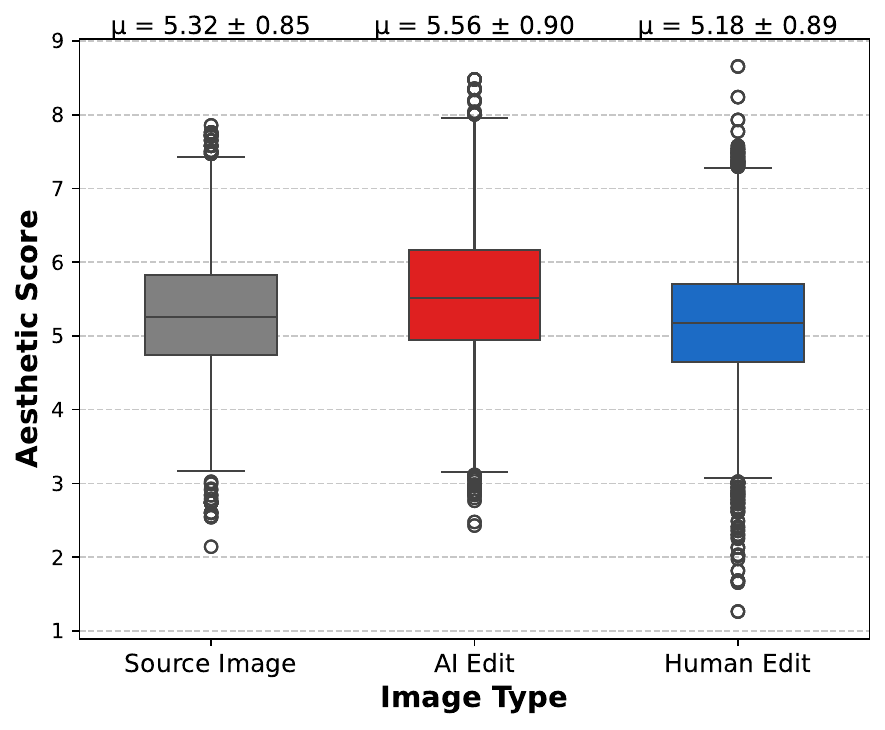}
      \caption{Aesthetic scores of source, AI edits, and human edits.}   
      \label{fig:aesthetic_c}
    \end{subfigure}\vspace{0.5em}
    
  \end{minipage}\hfill
  \begin{minipage}[t]{0.635\textwidth}
    \centering
    \begin{subfigure}[t]{\textwidth}
            \centering
            \includegraphics[width=1.0\linewidth]{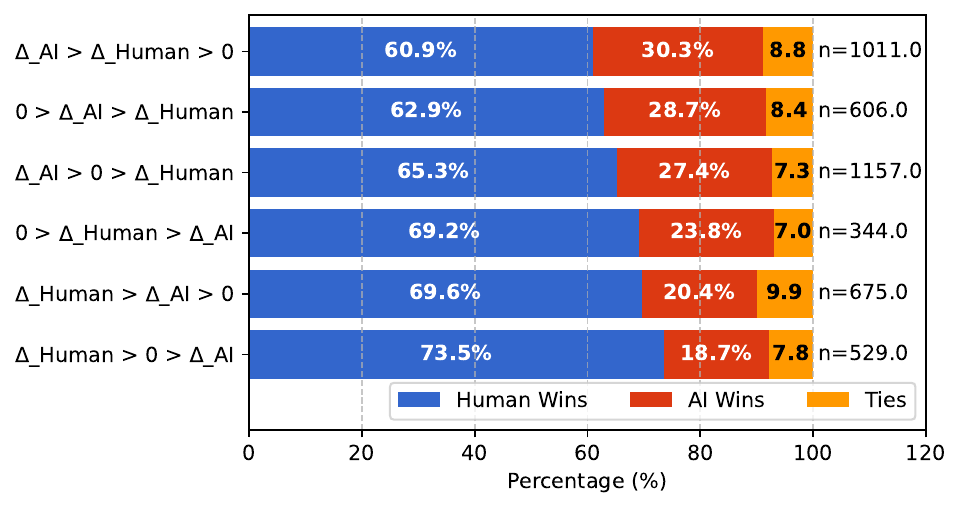}
        
        \label{fig:supp-aesthetic_category_win_rates}
        
      
      \caption{Win rate (\%) categorized by increase ($\Delta$) in aesthetics scores between the (human edit, source image) and (AI edit, source image). }
      \label{fig:win_rate_aesthetics}
    \end{subfigure}
    
  \end{minipage}
    \caption{
    AI edits have higher LAION aesthetics scores than the source image and human edits (a).
    AI edits are \colorbox{red!90!black}{\textcolor{white}{more likely to win}} when they contain a greater increase ($\Delta$) in aesthetics score (b). 
    }
  \label{fig:aesthetic_scores}
\end{figure*}

\subsec{Quantitative analysis}
We compute LAION aesthetic scores for all human and AI edits in PSR-328. 
On average, AI-edited images have higher aesthetics scores ($\mu = 5.56$) compared to human-edited images ($\mu = 5.18$) and even source images ($\mu = 5.32$; \cref{fig:aesthetic_c}).
The scores confirm that AI editors tend to increase image aesthetics.


AI-generated edits typically have higher aesthetic scores than human edits, regardless of the rating outcome (AI wins, Human wins or Tie; \cref{fig:aesthetics_by_rating}).
Categorizing the AI win rates by the increase ($\Delta$) in aesthetics scores between the source and edit images, we find a strong correlation between the aesthetics score gain and the increase in AI win rates---as the $\Delta$ in aesthetics of AI edits increase and surpass that of humans ($\Delta_\text{Human}$), AI edits tend to win more
(\cref{fig:win_rate_aesthetics}; 18.7\% $\to$ 30.3\%).


\subsection{AI editors can satisfactorily handle $\frac{1}{3}$ of all PSR requests}
\label{sec:ai_model_can_handle_one_thrid}


Given the increase in popularity of SOTA AI editors \cite{openai2024gpt4o,geminiImage}, it is important to estimate how much of the real-world requests (here, in PSR) can be satisfactorily handled by existing AI editors.
The answer might inform the area for improvement for future AIs.

\subsec{Experiment} Given the requests, AI edits and their ratings provided by humans, we first compute the win rates for humans and AI editors specifically for each edition action.
For each request, if it is rated ``Tie'' or ``AI wins'' by human raters, we consider it to be \textbf{satisfactorily handled} by AI editors (\cref{tab:verb-win-rates}; AI win+Tie).

\subsec{Results by editing actions}
Given this definition, the top-5 editing actions (see \cref{tab:verb-win-rates}) that AI editors handle the most satisfactorily are \action{clone} (48.0\%), \action{merge} (39.3\%), \action{apply} (39.3), \action{add} (38.0\%), and \action{delete} (34.9\%).
This is consistent with the fact that \action{apply}, \action{add}, and \action{delete} are the most common editing actions in large-scale training sets in the literature \cite{dai2023emu,brooks2023instructpix2pix} since the training-set examples can be synthesized using image inpainters or style transfer models. 
Coincidentally, \action{add} and \action{delete} are among the top-5 most popular requests in PSR (\cref{fig:appendix_12_years_data}).

To estimate the \% of requests that can be handled satisfactorily by AIs over all editing actions, we
multiply the combined AI win+Tie rates by the proportion that each action group contributes to the overall dataset (\cref{fig:appendix_12_years_data}):
$\sum_{v=1}^{v} D_v \times AI_v = 33.35\% $ 
where $D_v$ represents the proportion of dataset requests associated with action verb $v$, and $AI_v$ denotes the combined percentage of AI wins and ties for edits with verb $v$.
That is, \textbf{33.35\% of all image-editing requests can be handled by existing AI editors}.


\section{Discussion, Limitations, and Conclusion}
\label{sec:discussion}

\subsec{Limitations}
Our dataset annotation and taxonomy rely on multiple large language models to perform labeling, which may introduce biases or inaccuracies. 
Additionally, some AI image editors were unavailable publicly (\eg, Emu-Edit~\cite{sheynin2024emu}, OmniEdit~\cite{zhao2024ultraedit}, etc.) during evaluation and are thus excluded from our human study.

In this study, we compare generative AI and human edits to understand the gap between current AI capabilities and  user needs. AI tools excel at tasks like object removal and outpainting, effectively extending images and filling in missing details. However, in real-world use, current models can adequately handle only about one-third of user requests. Their main limitations are unintended modifications outside the target region and inadvertent changes to essential features, such as the subject’s identity.

\subsection*{Acknowledgment}
We thank Peng Wang from the SeedEdit team that has kindly run their models on our PSR-328 to enable this study.
We thank Hung H. Nguyen, Pooyan Rahmanzadehgervi, Tin Nguyen, and Giang Nguyen at Auburn University for feedback and discussions of the earlier results.
We thank Carlene Gonzalez and Natalie Au Yeung from Adobe for their insightful feedback on the action verb category.
AN was supported by the NSF Grant No. 2145767, and donations from NaphCare Foundation \& Adobe Research.

\clearpage
\appendix


\newcommand{\beginsupplementary}{%
    \setcounter{table}{0}
    \renewcommand{\thetable}{A\arabic{table}}%
    
    \setcounter{figure}{0}
    \renewcommand{\thefigure}{A\arabic{figure}}%
    
    \setcounter{section}{0}
    \renewcommand{\thesection}{A\arabic{section}}
    \renewcommand{\thesubsection}{\thesection.\arabic{subsection}}
}

\beginsupplementary%
\appendix

\newcommand{\toptitlebar}{
    \hrule height 4pt
    \vskip 0.25in
    \vskip -\parskip%
}
\newcommand{\bottomtitlebar}{
    \vskip 0.29in
    \vskip -\parskip%
    \hrule height 1pt
    \vskip 0.09in%
}

\newcommand{\suptitle}{Appendix for:\\\papertitle}

\newcommand{\maketitlesupp}{
    \newpage
    \onecolumn
        \null
        \vskip .375in
        \begin{center}
            \toptitlebar
            {\Large \bf \suptitle\par}
            \bottomtitlebar
            \vspace*{24pt}
            {
                \large
                \lineskip=.5em
                \par
            }
            \vskip .5em
            \vspace*{12pt}
        \end{center}
}

\maketitlesupp%

\crefformat{section}{Appendix~A#1}
\Crefformat{section}{Appendix~A#1}

\section{Dataset Statistics}
\label{sec:supp-dataset_stats}

We collect \datasetAllPosts{} posts and \datasetAllEdits{} edited images from Reddit, with \datasetAllPushShiftsPosts{} posts sourced from PushShift between 2013 and 2022 and \datasetAllRecentPosts{} gathered between October 2024 and early 2025. After processing, the final dataset includes \datasetAllPostsFinaldataset{} posts and \datasetAllEditsFinaldataset{} edited images, with \datasetAllSingle{} single-image and \datasetAllMulti{} multiple-image requests. \cref{tab:dataset_statistics} shows the breakdown of the dataset composition, and \Cref{fig:appendix_12_years_data} illustrates key statistics and trends within the data.

\begin{figure}[ht]
    \centering
    \includegraphics[width=0.5\linewidth]{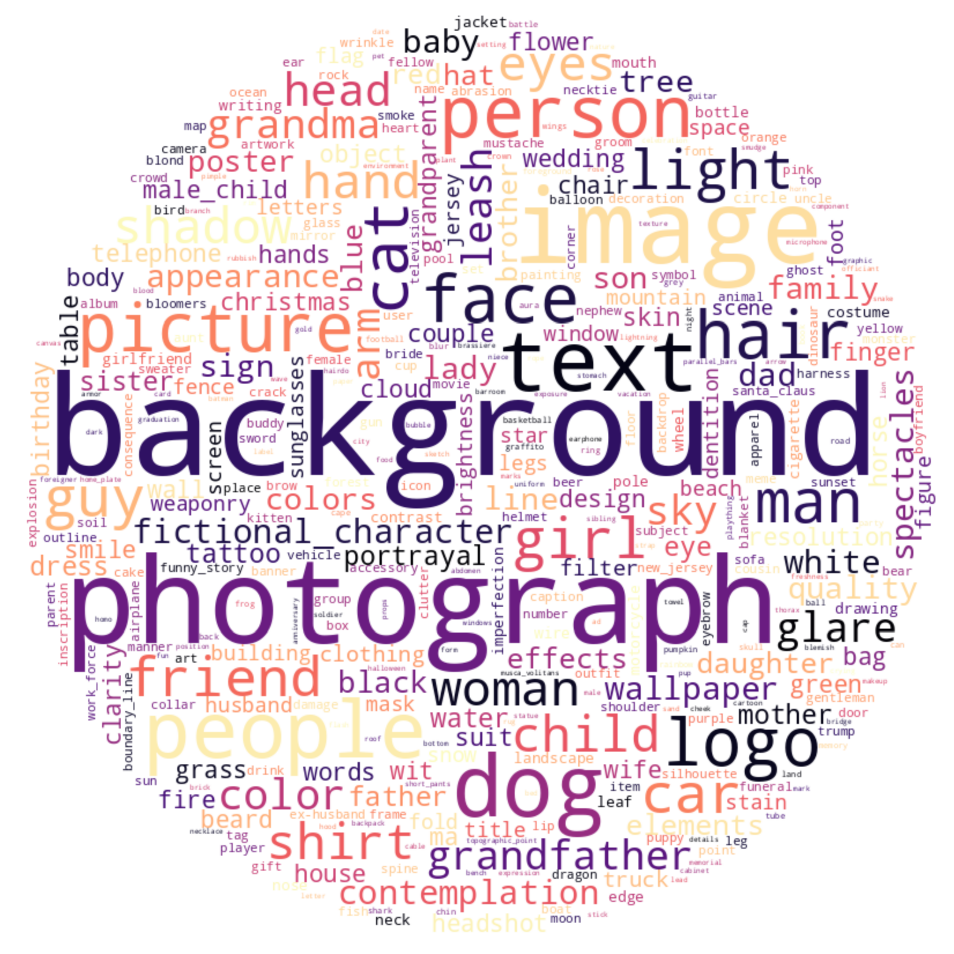}
    \caption{Word cloud visualization of image editing request subjects, mapped to WordNet synsets.}
    \label{fig:supp-SubjectWordCloud}
\end{figure}

\begin{figure}[htbp]
    \centering
    \includegraphics[width=\textwidth]{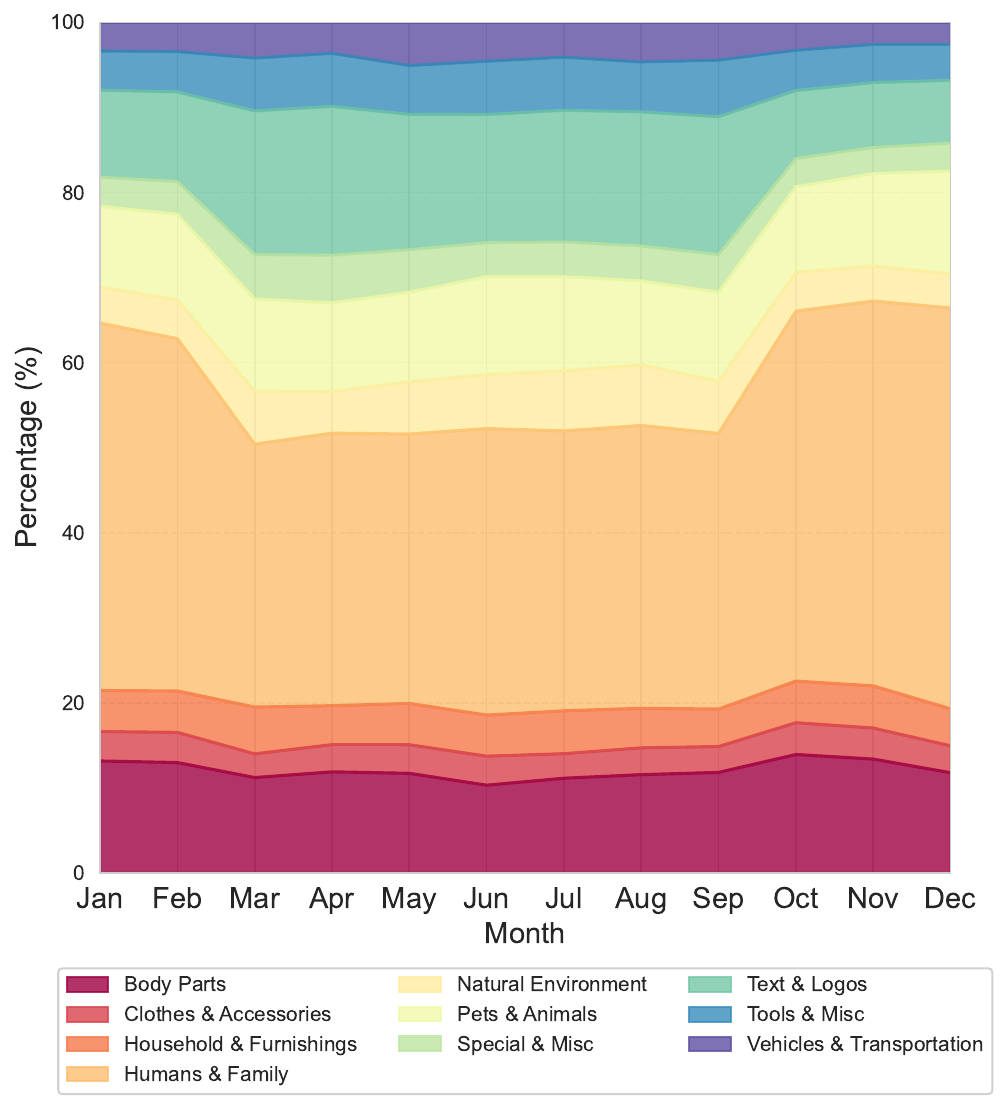}
    \caption{Monthly distribution of subject categories in image editing requests, highlighting trends and variations over time}
    \label{fig:supp-subject-sub-category-monthly}
\end{figure}

Our dataset contains a diverse set of subjects, with 49,134 unique subjects across all requests.
The \emph{People and Related} category is the most common, accounting for 53.5\% of the requests (\cref{tab:category_hierarchy}).
Subject trends exhibit seasonal variations (\cref{fig:supp-subject-sub-category-monthly}), with family-related requests peaking around the holiday season and New Year.  Additionally, \cref{fig:supp-SubjectWordCloud} illustrates the most common subjects.

\begin{figure}[ht]
    \centering
    \includegraphics[width=\linewidth]{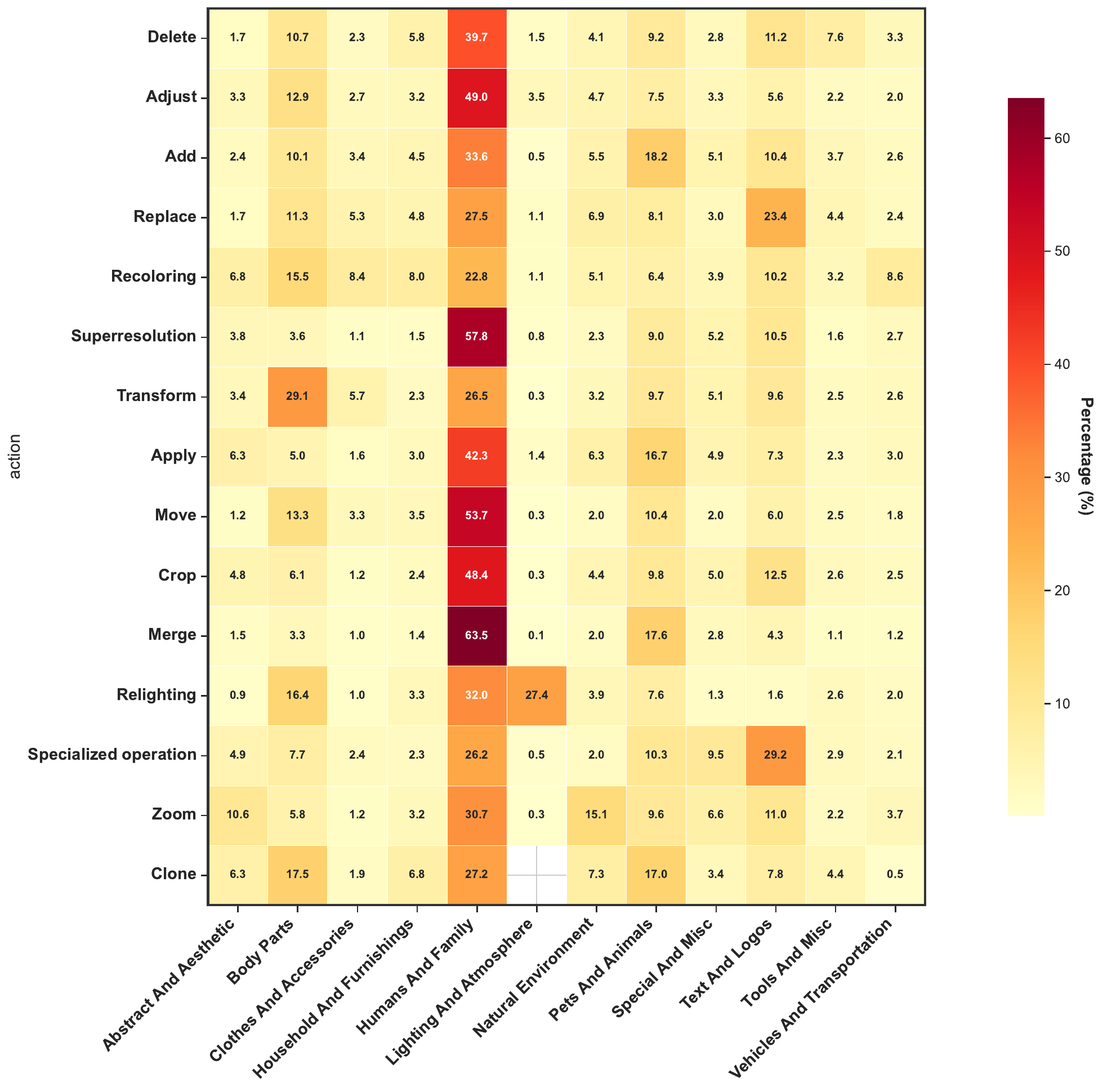}
    \caption{Distribution of Actions Across Different Subject Subcategories}
    \label{fig:supp-Action-Subcategory-Distribution}
\end{figure}

\begin{figure}[htbp]
    \centering
    \includegraphics[width=\textwidth]{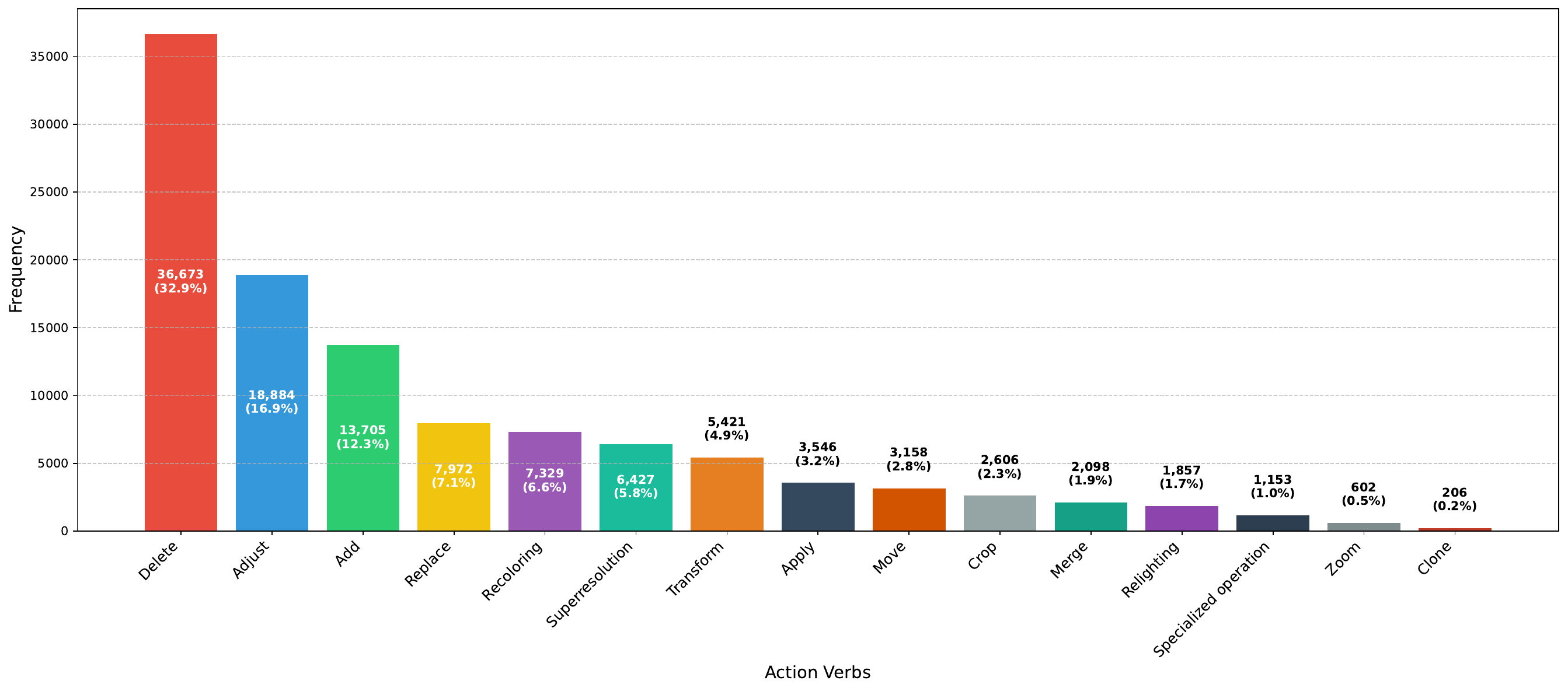}
    \caption{Distribution of different action verbs in our dataset}
    \label{fig:supp-action_verb_dist}
\end{figure}

The most common action is \action{delete}, accounting for 32.9\% of all edits (\cref{fig:supp-action_verb_dist}), a trend that has remained dominant over the years (\cref{fig:action_verbs_over_the_years}). These involve removing \emph{people}, such as photobombers, for aesthetic or personal reasons. Other deletion requests involve removing \emph{objects}, like poles, bags, or signs, in order to reduce visual clutter as well as eliminating \emph{facial imperfections}, such as acne or wrinkles for social media or professional use. The most frequent types of actions applied to subjects are illustrated in \cref{fig:action_subject} (See \cref{fig:supp-Action-Subcategory-Distribution} for more details).

\cref{fig:creativity_levels} shows that 55.5\% of user requests fall into the low-creativity category—indicating that most modifications allow little creative input. In this group, the \action{delete} action predominates, accounting for 51.2\% of requests. Conversely, high-creativity requests, which involve more complex transformations, are mainly associated with the \action{add} action (35.9\%). Meanwhile, the medium-creativity category displays a more balanced distribution, with \action{add} and \action{delete} actions representing 25.8\% and 20.7\% of requests, respectively. This distribution suggests that users generally prioritize simple modifications over intricate creative edits.

\begin{table}[htb]
  \centering
  \begin{minipage}[t]{0.47\linewidth}
    \centering
    \caption{Distribution of user requests and edited images across different data sources in our PSR dataset. }
    \label{tab:dataset_statistics}
    \resizebox{\linewidth}{!}{%
      \begin{tabular}{lrr}
        \toprule
        Category & Count & Coverage (\%) \\
        \midrule
        \multicolumn{3}{l}{\textbf{Request Statistics}} \\
        \midrule
        Total Requests           & \datasetAllPostsFinaldataset{} & 100.0 \\
        \quad Single-image       & \datasetAllSingle{}            & 84.4  \\
        \quad Multiple-image     & \datasetAllMulti{}             & 15.6  \\
        Historical (Pushshift)   & \datasetAllPushShiftsPosts{}   & 70.6  \\
        Recent Data (2024--2025) & \datasetAllRecentPosts{}       & 29.4  \\
        \midrule
        \multicolumn{3}{l}{\textbf{Edit Statistics}} \\
        \midrule
        Total Edits              & \datasetAllEditsFinaldataset{} & 100.0 \\
        Historical (Pushshift)   & \datasetAllPushShiftsEdits     & 29.6  \\
        Recent Data (2024--2025) & \datasetAllRecentEdits         & 70.4  \\
        \bottomrule
      \end{tabular}%
    }
  \end{minipage}
  \hfill
  \begin{minipage}[t]{0.47\linewidth}
    \centering
    \caption{Distribution of main and subcategories of subjects in image-editing requests. }
    \label{tab:category_hierarchy}
    \resizebox{\linewidth}{!}{%
      \begin{tabular}{lrr}
        \toprule
        Category & Count & (\%) \\
        \midrule
        \textbf{People \& Related}           & \textbf{44,416} & \textbf{53.5} \\
        \hspace{1em}Humans \& Family         & 31,914          & 38.5 \\
        \hspace{1em}Body Parts               & 9,840           & 11.9 \\
        \hspace{1em}Clothes \& Accessories   & 2,662           & 3.2 \\
        \addlinespace
        \textbf{Text Branding \& Abstract}   & \textbf{14,114} & \textbf{17.0} \\
        \hspace{1em}Text \& Logos            & 8,759           & 10.6 \\
        \hspace{1em}Special \& Misc          & 2,979           & 3.6 \\
        \hspace{1em}Abstract \& Aesthetic    & 2,376           & 2.9 \\
        \addlinespace
        \textbf{Inanimate Objects}           & \textbf{10,525} & \textbf{12.7} \\
        \hspace{1em}Tools \& Misc            & 4,021           & 4.8 \\
        \hspace{1em}Household \& Furnishings & 3,776           & 4.6 \\
        \hspace{1em}Vehicles \& Transportation & 2,728         & 3.3 \\
        \addlinespace
        \textbf{Animals}                     & \textbf{8,411}  & \textbf{10.1} \\
        \hspace{1em}Pets \& Animals          & 8,411           & 10.1 \\
        \addlinespace
        \textbf{Environment \& Background}   & \textbf{5,510}  & \textbf{6.6} \\
        \hspace{1em}Natural Environment      & 3,904           & 4.7 \\
        \hspace{1em}Lighting \& Atmosphere   & 1,606           & 1.9 \\
        \addlinespace
        \midrule
        \textbf{Total}                       & \textbf{82,976} & \textbf{100.0} \\
        \bottomrule
      \end{tabular}%
    }
  \end{minipage}
\end{table}

\clearpage
\section{Model Inference and Prompting Details}
\label{sec:supp-prompting_details}

In this section, we provide details about the prompts used, the model versions, and the temperature.

\begin{table}[h]
    \centering
    \caption{Model Configuration Details}
    \begin{tabular}{l l}
        \toprule
        \textbf{Model} & \textbf{Details} \\
        \midrule
        \internvlful & Version: \textit{OpenGVLab/InternVL2\_5-38B-AWQ}\\
                         & Temperature: 0.7 \\
                         & Hosted internally using \href{https://github.com/InternLM/lmdeploy}{lmdeploy} \\
        \midrule
        \gptmini & Version: \textit{gpt-4o-mini-2024-07-18-global-batch} \\
                    & Batch API \\
                    & Temperature: 0 (default) \\
                    & Using the API from Azure OpenAI Service \\
        \midrule
        \oone & Version: \textit{o1-2024-12-17} \\
           & Temperature: N/A \\
          & Reasoning Effort: High \\
           & Using the API from Azure OpenAI Service\\
        \midrule
        \geminiflashthinking & Version: \textit{gemini-2.0-flash-thinking-exp-01-21} \\
                                  & Temperature: 0.7 (default) \\
                                  & Using the API from Google AI Studio \\
        \bottomrule
    \end{tabular}
    \label{tab:supp-model_details}
\end{table}

\clearpage
\section{Prompting Details for Taxonomy Construction}
\label{sec:supp-prompting_details_tax}

\begin{figure}[htbp]
  \centering
\begin{tcblisting}{
  title=JSON schema for summarizing an image to JSON (Image-to-JSON),
  colback=black!5!white,
  colframe=black,
  fonttitle=\bfseries\color{white},
  coltitle=black,
  listing only,
  listing options={
    language=json,
    basicstyle=\ttfamily\footnotesize,
    keywordstyle=\color{orange},
    stringstyle=\color{blue},
    identifierstyle=\color{black},
    showstringspaces=false,
    tabsize=2
  }
}
{
  "description": "Brief description of the main content",
  "image_type": "photograph/digital-art/illustration/screenshot/meme",
  "setting": "indoor/outdoor/digital/mixed",
  "location": "beach/office/park/etc",
  "time_of_day": "day/night/unknown",
  "weather": "sunny/cloudy/rainy/not-applicable",
  "has_people": false,
  "people_count": 0,
  "has_adults": false,
  "has_children": false,
  "has_elderly": false,
  "has_groups": false,
  "has_animals": false,
  "has_dogs": false,
  "has_cats": false,
  "has_birds": false,
  "has_wildlife": false,
  "other_animals": [],
  "foreground_objects": [],
  "background_objects": [],
  "prominent_objects": [],
  "dominant_colors": [],
  "lighting": "bright/dim/dark/natural/artificial",
  "has_text": false,
  "text_content": "",
  "text_language": "",
  "mood": [],
  "atmosphere": "",
  "is_nsfw": false,
  "is_violent": false,
  "has_gore": false,
  "has_nudity": false,
  "is_sensitive": false,
  "image_quality": "high/medium/low",
  "orientation": "landscape/portrait/square",
  "tags": [],
  "ai_confidence": "high/medium/low"
}
\end{tcblisting}
  \caption{JSON schema for image metadata classification used with \internvlful}
  \label{fig:supp-json-schema}
\end{figure}

\begin{figure}[htbp]
  \centering
\begin{tcblisting}{
  title    = System message for extracting the request’s metadata,
  colback  = black!5!white,
  colframe = black,
  fonttitle=\bfseries\color{white},
  coltitle = black,
  listing only,
  listing options={
    language     = python,
    basicstyle   = \ttfamily\scriptsize,   % ← smaller font here
    keywordstyle = \color{orange},
    stringstyle  = \color{blue},
    identifierstyle=\color{black},
    showstringspaces=false,
    tabsize=2,
    breaklines=true
  }
}
def create_system_message() -> str:
    return """You are an AI system that analyzes image editing requests. Given a textual instruction and an image, evaluate the clarity, complexity, and appropriateness of the editing request. Assess the instruction's ambiguity (1-5 scale, where 1 is crystal clear and 5 is completely vague), complexity level (1-5 scale, where 1 is basic editing and 5 is expert-level), and check for any inappropriate or NSFW content. Verify if the image is valid and usable, and determine if the request is actually related to image editing. Provide your analysis in the following JSON format, including specific reasoning for each field:
{
  "original_instruction":"Preserved exactly as given to maintain reference point - no modifications or interpretations",
  "rewritten_instruction": {
    "text": "Clear, structured version of the original instruction",
    "reasoning": "Clarified version that removes ambiguity, fills in implied steps, and provides specific direction. Should be actionable without additional context"
  },
  "missing_details": {
    "items": ["List specific information that would be needed to complete the task but wasn't provided in the original instruction"],
    "reasoning": "Identifies gaps that would need to be filled to successfully complete the task"
  },
  "external_references": {
    "value": "True: References external links or comments | False: Self-contained instruction",
    "reasoning": "Identifies if critical information is located outside the main instruction"
  },
  "nsfw_analysis": {
    "value": "True: Contains adult/mature themes | False: Safe for general audience",
    "reasoning": "Evaluates if content contains mature themes, nudity, or adult subject matter"
  },
  "inappropriate_content": {
    "value": "True: Contains harmful/offensive/inappropriate content | False: Appropriate content",
    "reasoning": "Identifies presence of: 1) Harmful content (violence, hate speech, extreme gore) 2) Offensive content (discriminatory themes, extreme political content, severe profanity) 3) Inappropriate but non-harmful content (crude humor, mild toilet humor, silly/whimsical inappropriate gestures, playful trolling). Note: Mild humorous or whimsical content that might be considered 'silly inappropriate' (like tongue-in-cheek jokes, mild pranks, or playful memes) should be marked False unless they cross into actually offensive territory. Consider context and intent - distinguish between harmful inappropriate vs harmless fun"
  },
  "image_editing_relevance": {
    "value": "True: Related to image manipulation | False: Unrelated to image editing",
    "reasoning": "Confirms if the instruction pertains to image editing rather than other topics"
  },
  "image_validity": {
    "value": "True: Image is usable | False: Image is blank/corrupted/missing",
    "reasoning": "Verifies if provided image is suitable for editing"
  },
}
"""
\end{tcblisting}
  \caption{Prompt for extracting basic information from the request}
  \label{fig:supp-basic_info}
\end{figure}

\begin{figure}[htbp]
  \centering
\begin{tcblisting}{
  title=System message for extracting action verbs from the request,
  colback=black!5!white,
  colframe=black,
  fonttitle=\bfseries\color{white},
  coltitle=black,
  listing only,
  listing options={
    language=python,
    basicstyle=\ttfamily\small,
    keywordstyle=\color{orange},
    stringstyle=\color{blue},
    identifierstyle=\color{black},
    showstringspaces=false,
    tabsize=2,
    breaklines=true 
  }
}
def create_system_message(categories_desc) -> str:
    return f"""Analyze the following image editing instruction and identify which of these specific actions it contains.

Available categories with examples:
{categories_desc}

ANALYSIS GUIDELINES:
    1. Evaluate both the original and clarified requests
    2. Only include actions that are:
        - Explicitly stated OR
        - Logically necessary to achieve the described result
    3. Consider the final image's appearance to identify implicit actions
    4. Exclude actions that are:
        - Only potentially useful but not required
        - Vaguely related but not essential
        - Could be used as alternatives

Focus only on actual image manipulation actions that match our predefined categories.
Return your response as a JSON object with an 'actions' array containing only valid categories from the list provided.
"""
\end{tcblisting}
  \caption{Prompt for extracting action verbs from an editing request}
  \label{fig:supp-action_verbs_prompt}
\end{figure}

\begin{figure}[htbp]
  \centering
\begin{tcblisting}{
  title=System message for assigning a WordNet synset to the subject of an image editing request ,
  colback=black!5!white,
  colframe=black,
  fonttitle=\bfseries\color{white},
  coltitle=black,
  listing only,
  listing options={
    language=python,
    basicstyle=\ttfamily\small,
    keywordstyle=\color{orange},
    stringstyle=\color{blue},
    identifierstyle=\color{black},
    showstringspaces=false,
    tabsize=2,
    breaklines=true 
  }
}
def create_system_message() -> str:
    return """You are an AI system designed to analyze image editing requests and map the **subject** of the edit to its corresponding WordNet synset. The **action verb is not mapped**-only the subject needs to be processed.

### **Processing Pipeline:**
1. **Candidate Keyword Selection** - Extract the most relevant subject from the instruction and image. Generate a list of candidate keywords that are **highly likely to already exist as a WordNet synset**. Prioritize concrete nouns and commonly recognized entities.
2. **WordNet Synset Matching** - Use the refined candidate keywords to query WordNet, -confirming and selecting the best-fitting synset(s) for the subject.

### **Inputs Provided:**
- A textual instruction describing the image edit.
- An image associated with the request.
- The previously extracted **subject** (not the action verb).

### **Expected Output:**
Generate a structured list of **candidate search keywords** that:
- Are **highly probable** to already exist as a WordNet synset.
- Accurately represent the subject in the context of the image editing request.
- Can be directly used for WordNet lookup to retrieve the most relevant synset.

### **Important Constraints:**
- **Do not process or map the action verb**-focus solely on the subject.
- Ensure the **keywords are already strong candidates for WordNet synsets** before attempting lookup.
- Prefer concrete, commonly used nouns that align well with WordNe's structure.
"""
\end{tcblisting}
  \caption{Prompt for assigning WordNet synsets to subjects of editing requests}
  \label{fig:supp-wordnet_prompt}
\end{figure}

\clearpage

\begin{figure}[H]        
  \centering
  \begin{tcblisting}{
    title     = Full Details for Subcategories (Part 1),
    colback   = black!5!white,
    colframe  = black,
    fonttitle = \bfseries\color{white},
    listing only,
    enhanced,
    breakable,           % <-- lets the box itself split if it needs to
    listing options = {
      language      = python,
      basicstyle    = \ttfamily\fontsize{7pt}{7pt}\selectfont,
      keywordstyle  = \color{orange},
      stringstyle   = \color{blue},
      showstringspaces = false,
      tabsize       = 2,
      breaklines    = true
    }
  }
categories_desc = {
    "categories": [
        {
            "category": "Recoloring",
            "definition": "Change the color of an element, object, or text inside the image, but not the whole image",
            "samples": [
                "Can anyone change the dog's fur to black?",
                "Could somebody change the turquoise on the vanity and mirror to white?",
                "Can someone show me how this truck looks in 3 different colors?",
                "Can someone colorize and touch up my grandma?",
                "Can someone please make this grayscale with only the house blue?",
            ],
        },
        {
            "category": "Relighting",
            "definition": "Improve or change the lighting conditions of the scene such as the temperature, color, direction or position of the light source",
            "samples": [
                "pls get rid of the green light or change it to another colour",
                "Can someone relight this photo, removing all harsh shadows",
                "Can someone make lighting better / remove shadows?",
            ],
        },
        {
            "category": "Superresolution",
            "definition": "Modify image so that that the image has a higher resolution and showing clearer, fine details",
            "samples": [
                "How can I increase the pixel count on this picture?",
                "Would a kind soul be able to clean this up with a higher resolution?",
                "Can someone upscale this image to 4K resolution?",
            ],
        },
        {
            "category": "Adjust",
            "definition": "Enhance or correct an entire image's overall apperance by modifying its common properties",
            "samples": [
                "Increase saturation a bit on the elephants",
                "Brighten the shadows by 40% in the portrait",
                "Can someone adjust the lighting/contrast on this?",
            ],
        },
  \end{tcblisting}
  \caption{Full descriptions for each type of action verb in the data set (Part 1/3).}
  \label{fig:verb-description}
\end{figure}

\begin{figure}[H]\ContinuedFloat
  \centering
  \begin{tcblisting}{
    title     = Full Details for Subcategories (Part 2),
    colback   = black!5!white,
    colframe  = black,
    fonttitle = \bfseries\color{white},
    listing only,
    enhanced,
    breakable,
    listing options = {
      language      = python,
      basicstyle    = \ttfamily\fontsize{7pt}{7pt}\selectfont,
      keywordstyle  = \color{orange},
      stringstyle   = \color{blue},
      showstringspaces = false,
      tabsize       = 2,
      breaklines    = true
    }
  }
        {
            "category": "Delete",
            "definition": "Remove unwanted elements, text, objects, people, or imperfections from the image",
            "samples": [
                "Remove the jacket hanging from the girl's side",
                "Delete the distracting signpost in the background",
                "Please remove the 3rd girl from the left in light blue!",
            ],
        },
        {
            "category": "Crop",
            "definition": "Trim the edges of an image to make a smaller image to meet specific size requirements",
            "samples": [
                "Crop the photo to eliminate the space to the left and right",
                "Crop to square format for social media",
            ],
        },
        {
            "category": "Add",
            "definition": "Insert new elements, objects, text, or effects that weren't in the original image",
            "samples": [
                "Insert a ball hitting the tennis racket",
                "Add a copyright watermark to the bottom right",
                "Can someone put a believable tattoo on my daughter?",
            ],
        },
        {
          "category": "Replace",
          "definition": "Substitute objects or text in the image with something else while keeping the rest of the image unchanged",
          "samples": [
              "Please change the pamphlet she is holding into a dictionary",
              "I hate the backround. I would like a neat, white background",
              "Can someone replace the ball with a planet?",
          ],
      },
      {
          "category": "Apply",
          "definition": "Add filters, styles, or effects that modify the overall appearance of the image",
          "samples": [
              "Add a Gaussian blur to the background",
              "Apply a vintage film effect",
              "Is someone able to help me turn this into a cartoon",
          ],
      },
  \end{tcblisting}
  \caption{Full descriptions for each type of action verb in the data set (Part 2/3).}
  \label{fig:verb-description-b}
\end{figure}

\begin{figure}[H]\ContinuedFloat
  \centering
  \begin{tcblisting}{
    title     = Full Details for Subcategories (Part 3),
    colback   = black!5!white,
    colframe  = black,
    fonttitle = \bfseries\color{white},
    listing only,
    enhanced,
    breakable,
    listing options = {
      language      = python,
      basicstyle    = \ttfamily\fontsize{7pt}{7pt}\selectfont,
      keywordstyle  = \color{orange},
      stringstyle   = \color{blue},
      showstringspaces = false,
      tabsize       = 2,
      breaklines    = true
    }
  }
      {
          "category": "Zoom",
          "definition": "Adjust the image scale to zoom in on a specific area or add new content to mimic a zoom-out action",
          "samples": [
              "Zoom in on the man",
              "Zoom out 50% to show more context",
              "I'm happy to tip if someone is able to zoom this out",
          ],
      },
      {
          "category": "Transform",
          "definition": "Change the geometric properties (flip, scale, rotate, skew, perspective, distort, warp) of the image or objects",
          "samples": [
              "Flip the photo horizontally",
              "Fix the perspective of the building",
              "Please rotate the box in my hand",
          ],
      },
      {
          "category": "Move",
          "definition": "Change the position of existing elements within the image while keeping the rest of the image unchanged",
          "samples": [
              "Move the white framed picture to the blue wall",
              "Shift the logo 20 pixels up",
              "Will someone please edit my friend closer to me",
          ],
      },
      {
          "category": "Clone",
          "definition": "Make more copies of some existing elements inside the image",
          "samples": [
              "Can someone clone my cat",
              "Use a cloning tool to blend grass to cover any patches of dirt",
              "Can someone multiply me and make it look like my arms are interlocked?",
          ],
      },
      {
          "category": "Merge",
          "definition": "Combine multiple elements or effects from multiple images into a cohesive final image",
          "samples": [
              "Can someone combine these 2 photos?",
              "Please combine so I'm kissing this moose!",
              "Create a panorama from these shots",
          ],
      },
      {
          "category": "Specialized operation",
          "definition": "Specialized or composite editing operations that don't fit into standard categories",
          "samples": [
              "Can someone vectorize this logo for me without background?",
              "Convert to JPEG format",
              "Can someone make a collage of 12 photos",
          ],
      },
  ]
}
\end{tcblisting}
  \caption{Full descriptions for each type of action verb in the data set (Part 3/3).}
  \label{fig:verb-description-c}
\end{figure}

\clearpage

\begin{figure}[htbp]
  \centering
\begin{tcblisting}{
  title=System message for creativity level assignment,
  colback=black!5!white,
  colframe=black,
  fonttitle=\bfseries\color{white},
  coltitle=black,
  listing only,
  listing options={
    language=python,
    basicstyle=\ttfamily\small,
    keywordstyle=\color{orange},
    stringstyle=\color{blue},
    identifierstyle=\color{black},
    showstringspaces=false,
    tabsize=1,
    breaklines=true 
  }
}
def create_system_message() -> str:
    return """As an AI assistant, your task is to assign a creativity score to edited images based on the diversity of acceptable final versions for a given request. The creativity is measured by considering how many different ways the image could be edited to fulfill the request.

- **Low Creativity:** The request leads to similar edited images with limited variations.
- **Medium Creativity:** The request allows for some variation but not an extensive range.
- **High Creativity:** The request can be fulfilled in many different ways leading to very different images.

When evaluating, think about the range of possible acceptable edited images for the request.

**Examples:**

1. **Request:** "Remove the red-eye effect from this photo."
   - The edits will be similar, focusing on correcting the eyes.
   - **Creativity Score:** Low.

2. **Request:** "Transform this portrait into a work of abstract art."
   - There are countless ways to interpret and edit the image.
   - **Creativity Score:** High.

3. **Request:** "Adjust the brightness and contrast to enhance the image."
   - There are some variations in how this can be done.
   - **Creativity Score:** Medium.

4. **Request:** "Crop the image to focus on the main subject."
   - Limited variations in the final image.
   - **Creativity Score:** Low.

5. **Request:** "Add a dramatic sky to this landscape photo."
   - Several ways to interpret a 'dramatic sky.'
   - **Creativity Score:** Medium.

6. **Request:** "Reimagine this landscape in a fantasy setting."
   - Numerous possibilities for how the image could be edited.
   - **Creativity Score:** High.

Provide the creativity score (**Low**, **Medium**, or **High**) along with a brief explanation for your assessment.
"""
\end{tcblisting}
  \caption{Prompt for assigning a creativity level to requests}
  \label{fig:supp-creativity_prompt}
\end{figure}

\begin{figure}[htbp]
  \centering
\begin{tcblisting}{
  title=System message for comparing different edited images using VLMs,
  colback=black!5!white,
  colframe=black,
  fonttitle=\bfseries\color{white},
  coltitle=black,
  listing only,
  listing options={
    language=python,
    basicstyle=\ttfamily\small,
    keywordstyle=\color{orange},
    stringstyle=\color{blue},
    identifierstyle=\color{black},
    showstringspaces=false,
    tabsize=1,
    breaklines=true 
  }
}
def create_system_message() -> str:
    return """
You are an image editing evaluation assistant that helps users determine which edited version better fulfills their request. When presented with an original image, an editing request, and two edited versions (A and B), carefully analyze how each edit implements the requested changes. First, examine the specific editing request and how it relates to the original image. Then analyze each edited version, noting strengths, weaknesses, and how closely they match the user's intent. Provide clear reasoning that considers technical quality, aesthetic appeal, and faithfulness to the request. Finally, deliver your verdict in one of three ways: "Image A is better" if the first edit is superior, "Image B is better" if the second edit is superior, or "Tie, both edits are equally good" if they are comparable in quality and adherence to the request.
"""
\end{tcblisting}
  \caption{Prompt for judging different edited images using VLMs.}
  \label{fig:supp-judge_prompt}
\end{figure}

\begin{figure}[htbp]
  \centering
\begin{tcblisting}{
  title=Sample Python code for sending multiple images to be evaluated by a judge,
  colback=black!5!white,
  colframe=black,
  fonttitle=\bfseries\color{white},
  coltitle=black,
  listing only,
  listing options={
    language=python,
    basicstyle=\ttfamily\small,
    keywordstyle=\color{orange},
    stringstyle=\color{blue},
    identifierstyle=\color{black},
    showstringspaces=false,
    tabsize=1,
    breaklines=true 
  }
}
        response = client.chat.completions.create(
            model="MODEL_NAME",
            messages=[
                {"role": "system", "content": system_message},
                {
                    "role": "user",
                    "content": [
                        {
                            "type": "text",
                            "text": f"Analyze the following image editing request and compare the edits:\n\nUser instruction: {original_instruction}",
                        }
                    ],
                },
                {
                    "role": "user",
                    "content": [
                        {"type": "text", "text": "Source Image"},
                        {
                            "type": "image_url",
                            "image_url": {"url": f"{source_image_base64}"},
                        },
                    ],
                },
                {
                    "role": "user",
                    "content": [
                        {"type": "text", "text": "Edit A"},
                        {
                            "type": "image_url",
                            "image_url": {"url": f"{edit1_image_base64}"},
                        },
                    ],
                },
                {
                    "role": "user",
                    "content": [
                        {"type": "text", "text": "Edit B"},
                        {
                            "type": "image_url",
                            "image_url": {"url": f"{edit2_image_base64}"},
                        },
                    ],
                },
            ],
            tools=[tool],
            tool_choice="auto",
            max_completion_tokens=8192,
        )
\end{tcblisting}
\caption{Example code demonstrating how to submit multiple images in one message to a vision-language model (VLM) for judgment in the \emph{"VLMs as a Judge} experiment.}
  \label{fig:supp-judge_prompt_python_code}
\end{figure}


\clearpage
\section{Image Generation Details}
\label{sec:supp-generation_details}

In this section, we provide details about how we collected the images for all the models.


\begin{table}[h]
    \centering
    \caption{Image Generation Details}
    \resizebox{\textwidth}{!}{%
        \begin{tabular}{l p{0.8\textwidth}}
            \toprule
            \textbf{Model} & \textbf{Details} \\
            \midrule
            \gpt & We used ChatGPT's web interface to generate images. \\
            \midrule
            \geminiflash{} & We used the official API to generate images. \\
            \midrule
            \SeedEdit{} & We sent images and prompts to SeedEdit's authors to have images generated locally. \\
            \midrule
            Hugging Face ( \huggingface) & We accessed models through Hugging Face Spaces and inserted the image and prompt (if applicable). Masks and bounding boxes were added manually if needed. See~\Cref{sec:supp-full_list_of_models} for full list of the models. \\
            \bottomrule
        \end{tabular}
    }
    \label{tab:supp-model_details_image_generation}
\end{table}

\clearpage
\section{Human Study}
\label{sec:supp-human-study}

In this section, we provide details about the human study. A total of 122 different people participated in our study, from North America, representing two universities and one institution. One-third of the participants are professional image editors and are familiar with image editing techniques.

\begin{figure}[ht]
    \centering
    \includegraphics[width=0.8\linewidth]{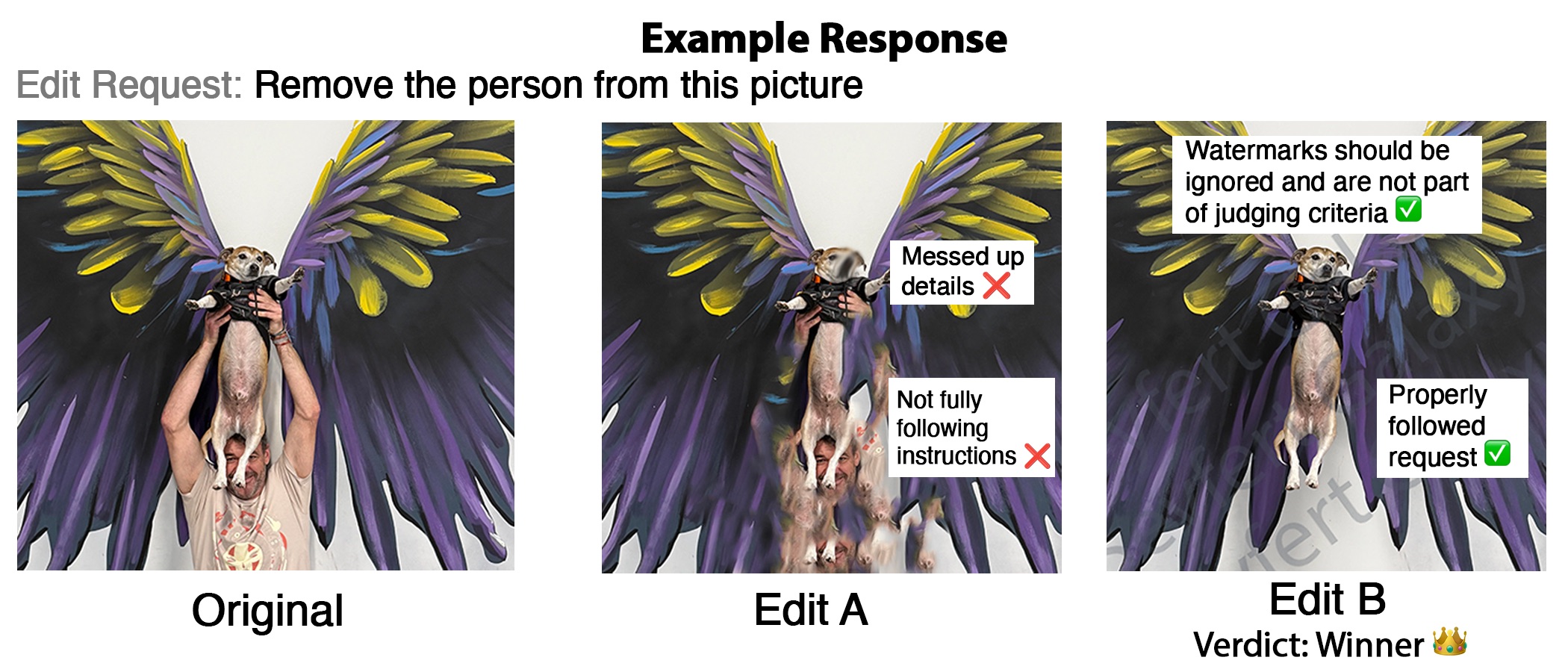}
    \caption{The introduction screen for the human study guides users on how to rate images based on the user’s request.}
    \label{fig:supp-human-study-intro-screen}
\end{figure}

\vspace{40pt}

\begin{figure}[ht]
    \centering
    \includegraphics[width=0.8\linewidth]{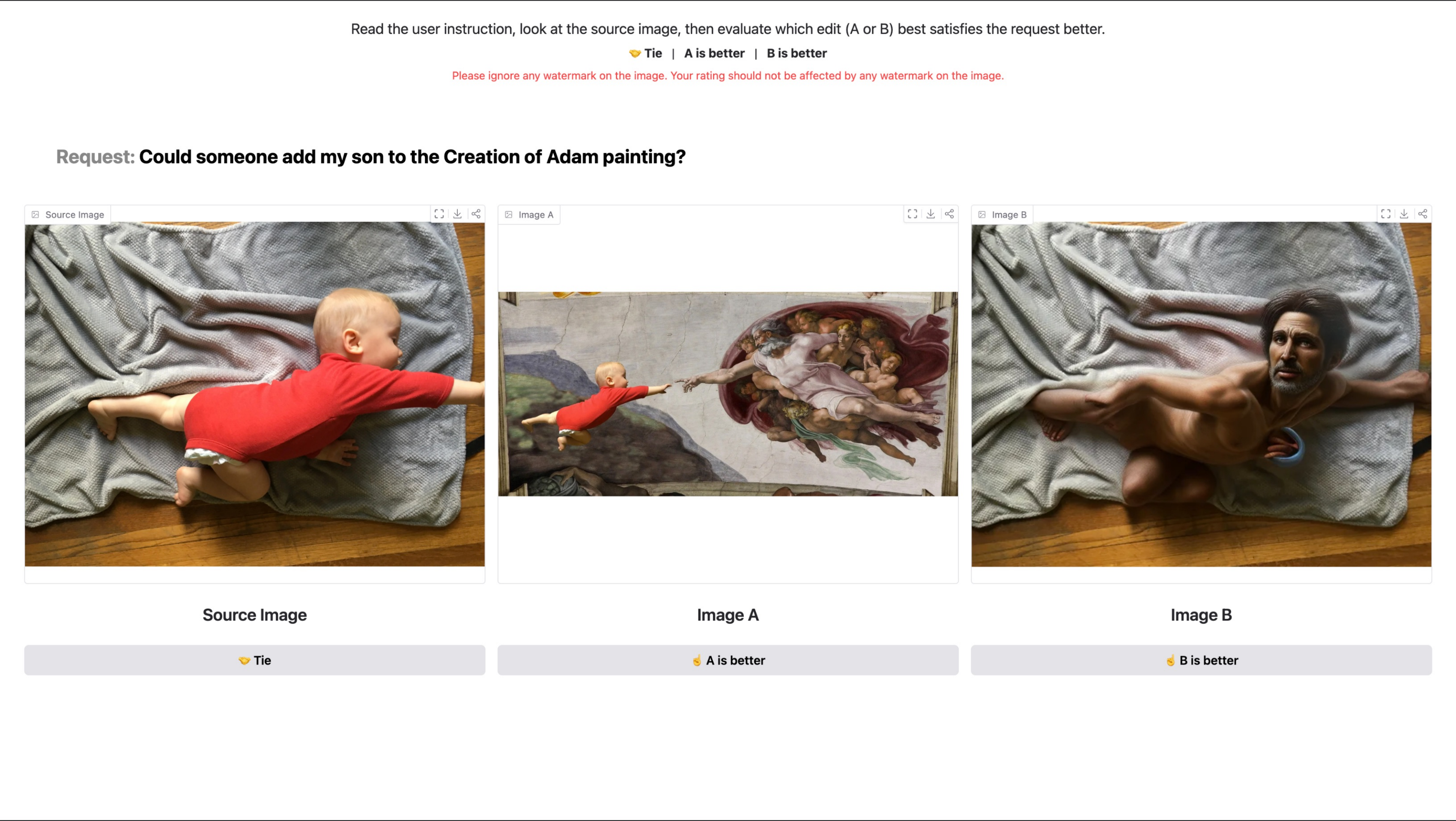}
    \caption{The user interface for the human study displays the original image, the user-provided request, and two edits. Users must decide which edit best satisfies the user’s request.}
    \label{fig:supp-human-study-sample-screen}
\end{figure}

Before starting the survey, users are shown an example of how to judge quality through \cref{fig:supp-human-study-intro-screen}. Following the survey introduction, the user is shown the original image, the edit request, and the two edited images (\cref{fig:supp-human-study-sample-screen}). The user then chooses if image A or image B is better, or if they are tied in quality.

\clearpage
\subsection{Details About AI-based Image Editing Tools}
\label{sec:supp-full_list_of_models}

\begin{table}[ht]
\centering
\caption{Number of edits for unique posts generated via different AI tools.}
\label{table:supp-edits-per-model}
\resizebox{0.9\textwidth}{!}{
    \begin{tabular}{l r l r}
    \toprule
    Model & Count & Model & Count \\
    \midrule
    \href{https://openai.com/index/introducing-4o-image-generation/}{\gpt{}} & 328 & \href{https://huggingface.co/spaces/fffiloni/diffusers-image-outpaint}{\model{fffiloni/diffusers-image-outpaint}} & 5 \\
    \href{https://ai.google.dev/gemini-api/docs/image-generation}{\geminiflash{}} & 327 & \href{https://huggingface.co/spaces/ameerazam08/Diffusion-Eraser}{\model{ameerazam08/Diffusion-Eraser}} & 4 \\
    \href{https://team.doubao.com/en/special/seededit}{\SeedEdit{}} & 274 & \href{https://huggingface.co/spaces/black-forest-labs/FLUX.1-Fill-dev}{\FluxFillDev{}} & 3 \\
    \href{https://huggingface.co/spaces/multimodalart/cosxl}{\CosXL} & 162 & \href{https://huggingface.co/spaces/SkalskiP/FLUX.1-inpaint-dev}{\FluxInpaintDev{}} & 3 \\
    \href{https://huggingface.co/spaces/timbrooks/instruct-pix2pix}{\InstructPix{}} & 153 & \href{https://huggingface.co/spaces/OzzyGT/diffusers-image-fill}{\DiffFill{}} & 3 \\
    \href{https://huggingface.co/spaces/AI4Editing/MagicQuill}{\Magic} & 119 & \href{https://huggingface.co/spaces/1aurent/ic_light}{\ICLight{}} & 3 \\
    \href{https://huggingface.co/spaces/editing-images/ledits}{\Ledits} & 81 & \href{https://huggingface.co/spaces/mukaist/textcutobject}{\TextCut{}} & 3 \\
    \href{https://huggingface.co/spaces/ameerazam08/FLUX.1-dev-Inpainting-Model-Beta-GPU}{\FluxInpaint} & 52 & \href{https://huggingface.co/spaces/finegrain/finegrain-object-cutter}{\FineCutter{}} & 3 \\
    \href{https://huggingface.co/spaces/aryadytm/remove-photo-object}{\RemoveObject{}} & 48 & \href{https://huggingface.co/spaces/KBlueLeaf/Sketch-Gen}{\SketchGen{}} & 2 \\
    \href{https://huggingface.co/spaces/briaai/BRIA-Eraser-API}{\BriaEraser{}} & 47 & \href{https://huggingface.co/spaces/akhaliq/AnimeGANv2}{\AnimeGAN{}} & 2 \\
    \href{https://huggingface.co/spaces/finegrain/finegrain-object-eraser}{\FineEraser{}} & 45 & \href{https://huggingface.co/spaces/awacke1/Image-to-Line-Drawings}{\ImageToLine{}} & 2 \\
    \href{https://huggingface.co/spaces/modelscope/ReplaceAnything}{\ReplaceAny{}} & 36 & \href{https://huggingface.co/spaces/kornia/kornia-image-filtering}{\KorniaFilter{}} & 1 \\
    \href{https://huggingface.co/spaces/diffusers/stable-diffusion-xl-inpainting}{\StableDiff{}} & 29 & \href{https://huggingface.co/spaces/islamuddin/NonLinear-Blurr-Image}{\NonLinearBlur{}} & 1 \\
    \href{https://huggingface.co/spaces/InstantX/flux-IP-adapter}{\FluxIP{}} & 27 & \href{https://huggingface.co/spaces/openfree/ColorRevive}{\model{openfree/ColorRevive}} & 1 \\
    \href{https://huggingface.co/spaces/sczhou/CodeFormer}{\CodeFormer{}} & 22 & \href{https://huggingface.co/spaces/fffiloni/InstantIR}{\model{fffiloni/InstantIR}} & 1 \\
    \href{https://huggingface.co/spaces/modelscope/old_photo_restoration}{\PhotoRestore{}} & 19 & \href{https://huggingface.co/spaces/briaai/BRIA-Generative-Fill-API}{\BRIAFillAPI{}} & 1 \\
    \href{https://huggingface.co/spaces/multimodalart/flux-fill-outpaint}{\FluxFillOut{}} & 16 & \href{https://huggingface.co/spaces/ShawnLJW/image2coloringbook}{\ImageToColoringBook{}} & 1 \\
    \href{https://huggingface.co/spaces/editing-images/leditsplusplus}{\LeditsPlus{}} & 15 & \href{https://huggingface.co/spaces/yizhangliu/ImgCleaner}{\model{yizhangliu/ImgCleaner}} & 1 \\
    \href{https://huggingface.co/spaces/marcosv/InstructIR}{\model{marcosv/InstructIR}} & 10 & \href{https://huggingface.co/spaces/fgpzen/remove-photo-object}{\RemoveObject{}} & 1 \\
    \href{https://huggingface.co/spaces/fffiloni/text-guided-image-colorization}{\Colorize{}} & 10 & \href{https://huggingface.co/spaces/briaai/BRIA-2.3-Inpainting}{\BRIAInpainting{}} & 1 \\
    \href{https://huggingface.co/spaces/PeepDaSlan9/B2BMGMT_Sharpening}{\BSharp{}} & 10 & \href{https://huggingface.co/spaces/not-lain/background-removal}{\model{not-lain/background-removal}} & 1 \\
    \href{https://huggingface.co/spaces/briaai/BRIA-2.2-ControlNet-Recoloring}{\BriaRecolor{}} & 9 & \href{https://huggingface.co/spaces/schirrmacher/ormbg}{\model{schirrmacher/ormbg}} & 1 \\
    \href{https://huggingface.co/spaces/drmurataltun/foto_filter}{\FotoFilter{}} & 7 & \href{https://huggingface.co/spaces/tori29umai/sketch2lineart}{\SketchTo{}} & 1 \\
    \href{https://huggingface.co/spaces/turboedit/turbo_edit}{\TurboEdit{}} & 6 \\
    \bottomrule
    \end{tabular}
}
\end{table}

\begin{table}[htp]
\centering
\caption{Model Performance Comparison (Win Rate as Judged by Humans \%)}
\label{tab:model-winrates}
\resizebox{\textwidth}{!}{
\begin{tabular}{lcccc}
\hline
\toprule\addlinespace
Model & Total Matches & Human Edit Win Rate & AI Edit Win Rate & Tie Rate \\
\midrule
\RemoveObject & 3 & 0.0 & 100.0 & 0.0 \\
schirrmacher/ormbg & 2 & 0.0 & 50.0 & 50.0 \\
\FineCutter & 1 & 0.0 & 100.0 & 0.0 \\
\AnimeGAN & 6 & 16.7 & 83.3 & 0.0 \\
\ImageToColoringBook & 5 & 20.0 & 20.0 & 60.0 \\
\model{ameerazam08/Diffusion-Eraser} & 10 & 40.0 & 30.0 & 30.0 \\
\BriaEraser & 65 & 43.1 & 35.4 & 21.5 \\
\model{openfree/ColorRevive} & 2 & 50.0 & 0.0 & 50.0 \\
\KorniaFilter & 2 & 50.0 & 50.0 & 0.0 \\
\TextCut & 4 & 50.0 & 25.0 & 25.0 \\
\ICLight & 4 & 50.0 & 50.0 & 0.0 \\
\model{fffiloni/diffusers-image-outpaint} & 8 & 50.0 & 25.0 & 25.0 \\
\SeedEdit(Simplified Instruction) & 443 & 53.5 & 39.5 & 7.0 \\
\SeedEdit(Original Instruction) & 442 & 53.6 & 36.2 & 10.2 \\
\RemoveObject & 60 & 56.7 & 21.7 & 21.7 \\
\gpt(Original Instruction) & 512 & 61.1 & 33.6 & 5.3 \\
\gpt(Simplified Instruction) & 502 & 62.0 & 32.1 & 6.0 \\
\CodeFormer & 46 & 63.0 & 26.1 & 10.9 \\
\FluxInpaint & 101 & 63.4 & 25.7 & 10.9 \\
\ReplaceAny & 67 & 65.7 & 20.9 & 13.4 \\
\TurboEdit & 9 & 66.7 & 33.3 & 0.0 \\
\BRIAInpainting & 3 & 66.7 & 33.3 & 0.0 \\
\FluxFillDev & 6 & 66.7 & 16.7 & 16.7 \\
\FineEraser & 63 & 66.7 & 20.6 & 12.7 \\
\SketchGen & 3 & 66.7 & 33.3 & 0.0 \\
\geminiflash(Simplified Instruction) & 341 & 67.2 & 21.1 & 11.7 \\
\PhotoRestore & 53 & 67.9 & 28.3 & 3.8 \\
\Magic & 215 & 68.4 & 21.9 & 9.8 \\
\BriaRecolor & 13 & 69.2 & 15.4 & 15.4 \\
\DiffFast & 17 & 70.6 & 23.5 & 5.9 \\
\ImageToLine & 7 & 71.4 & 14.3 & 14.3 \\
\FluxIP & 43 & 72.1 & 20.9 & 7.0 \\
\model{marcosv/InstructIR} & 22 & 72.7 & 9.1 & 18.2 \\
\geminiflash(Original Instruction) & 340 & 72.9 & 20.0 & 7.1 \\
\BSharp & 12 & 75.0 & 0.0 & 25.0 \\
\FluxFillOut & 21 & 76.2 & 19.0 & 4.8 \\
\CosXL & 334 & 78.7 & 14.7 & 6.6 \\
\model{yizhangliu/ImgCleaner} & 5 & 80.0 & 20.0 & 0.0 \\
\FotoFilter & 11 & 81.8 & 18.2 & 0.0 \\
\Colorize & 24 & 83.3 & 16.7 & 0.0 \\
\InstructPix & 281 & 84.0 & 9.6 & 6.4 \\
\Ledits & 156 & 84.6 & 10.3 & 5.1 \\
\LeditsPlus & 25 & 88.0 & 8.0 & 4.0 \\
\StableDiff & 51 & 90.2 & 7.8 & 2.0 \\
\FluxInpaintDev & 13 & 92.3 & 0.0 & 7.7 \\
\model{fffiloni/InstantIR} & 1 & 100.0 & 0.0 & 0.0 \\
\model{not-lain/background-removal} & 1 & 100.0 & 0.0 & 0.0 \\
\BRIAFillAPI & 1 & 100.0 & 0.0 & 0.0 \\
\SketchTo & 1 & 100.0 & 0.0 & 0.0 \\
\NonLinearBlur & 2 & 100.0 & 0.0 & 0.0 \\
\hline
\end{tabular}
}
\end{table}

\clearpage
\subsection{Breakdown of Model Performance by Different Categories}
\label{sec:supp-win_rate_breakdown}

\begin{table}[htp]
\centering
\caption{Win rate breakdown by different models with at least 30 matchups. SeedEdit leads using Simplified Instructions (SI) with a +3.3\% absolute improvement over the Original Instructions (OI).}
\resizebox{0.7\columnwidth}{!}{%
    \begin{tabular}{lrrrr}
        \toprule
        \textbf{Model} & \textbf{AI} & \textbf{Human} & \textbf{Tie} & \textbf{Count} \\
        \midrule
    \href{https://team.doubao.com/en/special/seededit}{\SeedEdit{} (SI)} \seedlogo & 39.5 & 53.5 & 7.0 & 443 \\
    \href{https://team.doubao.com/en/special/seededit}{\SeedEdit{} (OI)} \seedlogo & 36.2 & 53.6 & 10.2 & 442 \\
    \href{https://huggingface.co/spaces/briaai/BRIA-Eraser-API}{\BriaEraser{}} & 35.4 & 43.1 & 21.5 & 65 \\
    \gpt (OI) & 33.6 & 61.1 & 5.3 & 512 \\
    \gpt (SI) & 32.1 & 62.0 & 6.0 & 502 \\
    \href{https://huggingface.co/spaces/modelscope/old_photo_restoration}{Old Photo Restoration} & 28.3 & 67.9 & 3.8 & 53 \\
    \href{https://huggingface.co/spaces/SkalskiP/FLUX.1-inpaint}{FLUX.1-Inpainting} & 25.7 & 63.4 & 10.9 & 101 \\
    \href{https://huggingface.co/spaces/AI4Editing/MagicQuill}{MagicQuill} & 21.9 & 68.4 & 9.8 & 215 \\
    \href{https://developers.googleblog.com/en/experiment-with-gemini-20-flash-native-image-generation/}{Gemini 2.0 Flash (OI)} \geminilogo & 21.1 & 67.2 & 11.7 & 341 \\
    \href{https://huggingface.co/spaces/aryadytm/remove-photo-object}{Remove-Photo-Object} & 21.7 & 56.7 & 21.7 & 60 \\
    \href{https://developers.googleblog.com/en/experiment-with-gemini-20-flash-native-image-generation/}{Gemini 2.0 Flash (SI)} \geminilogo & 20.0 & 72.9 & 7.1 & 340 \\
    \href{https://huggingface.co/spaces/modelscope/ReplaceAnything}{ReplaceAnything} & 20.9 & 65.7 & 13.4 & 67 \\
    \href{https://huggingface.co/spaces/finegrain/finegrain-object-eraser}{Finegrain-Object-Eraser} & 20.6 & 66.7 & 12.7 & 63 \\
    \href{https://huggingface.co/spaces/multimodalart/cosxl}{\CosXL} & 14.7 & 78.7 & 6.6 & 334 \\
    \href{https://huggingface.co/spaces/editing-images/ledits}{\Ledits} & 10.3 & 84.6 & 5.1 & 156 \\
    \href{https://huggingface.co/spaces/timbrooks/instruct-pix2pix}{\InstructPix{}} & 9.6 & 84.0 & 6.4 & 281 \\
    \href{https://huggingface.co/spaces/diffusers/stable-diffusion-xl-inpainting}{Stable-Diffusion-XL-Inpainting} & 7.8 & 90.2 & 2.0 & 51 \\
        \bottomrule
    \end{tabular}
}
\label{tab:model-performance}
\end{table}

\begin{table}[htp]
\centering
\caption{\% of votes for human edits and AI edits and for the ``Tie'' options, categorized by editing actions.
``AI Win+Tie'' represents the sum of the AI and Tie columns, indicating the percentage of \% requests that can already be handled by the 49 AI models.}
\label{tab:verb-win-rates}
\begin{tabular}{lrrrrr}
\toprule
 Action & Human Wins & AI Wins & Tie & AI Win+Tie & no. of votes \\
\midrule
\textbf{{\action{Adjust}}} & 67.7 & 25.1 & 7.2 & 32.3 & 1293 \\
\textbf{{\action{Delete}}} & 65.1 & 26.3 & 8.6 & 34.9 & 1123 \\
\textbf{{\action{Add}}} & 62.0 & 30.6 & 7.4 & 38.0 & 1114 \\
\textbf{{\action{Recoloring}}} & 68.4 & 25.2 & 6.3 & 31.6 & 599 \\
\textbf{{\action{Replace}}} & 71.2 & 22.3 & 6.6 & 28.8 & 548 \\
\textbf{{\action{Apply}}} & 60.4 & 30.6 & 9.0 & 39.6 & 399 \\
\textbf{{\action{Transform}}} & 70.1 & 20.8 & 9.0 & 29.9 & 355 \\
\textbf{{\action{Superresolution}}} & 70.2 & 23.6 & 6.2 & 29.8 & 352 \\
\textbf{{\action{Merge}}} & 60.7 & 30.9 & 8.4 & 39.3 & 191 \\
\textbf{{\action{Relighting}}} & 66.9 & 24.2 & 9.0 & 33.1 & 178 \\
\textbf{{\action{Move}}} & 68.1 & 20.2 & 11.7 & 31.9 & 163 \\
\textbf{{\action{Crop}}} & 81.4 & 15.5 & 3.1 & 18.6 & 129 \\
\textbf{{\action{Specialized operation}}} & 62.9 & 26.7 & 10.3 & 37.1 & 116 \\
\textbf{{\action{Zoom}}} & 85.3 & 10.7 & 4.0 & 14.7 & 75 \\
\textbf{{\action{Clone}}} & 52.0 & 28.0 & 20.0 & 48.0 & 25 \\
\bottomrule
\end{tabular}
\end{table}


\begin{table}[ht]
\centering
\caption{\SeedEdit{} \seedlogo{} performance (win rate \%), categorized by different action verb types }
\label{tab:seededit-action-verb}
\begin{tabular}{lrrrrr}
\toprule
 Action & Human Wins & AI Wins & Tie & AI Win+Tie & Count \\
\midrule
\textbf{{\action{Add}}} & 48.0 & 42.0 & 10.0 & 52.0 & 281 \\
\textbf{{\action{Delete}}} & 53.5 & 38.0 & 8.5 & 46.5 & 271 \\
\textbf{{\action{Adjust}}} & 54.3 & 41.1 & 4.6 & 45.7 & 197 \\
\textbf{{\action{Recoloring}}} & 63.1 & 31.1 & 5.8 & 36.9 & 103 \\
\textbf{{\action{Replace}}} & 57.0 & 34.4 & 8.6 & 43.0 & 93 \\
\textbf{{\action{Apply}}} & 41.4 & 54.0 & 4.6 & 58.6 & 87 \\
\textbf{{\action{Transform}}} & 62.7 & 33.9 & 3.4 & 37.3 & 59 \\
\textbf{{\action{Move}}} & 68.6 & 14.3 & 17.1 & 31.4 & 35 \\
\textbf{{\action{Relighting}}} & 64.3 & 28.6 & 7.1 & 35.7 & 28 \\
\textbf{{\action{Merge}}} & 37.0 & 55.6 & 7.4 & 63.0 & 27 \\
\textbf{{\action{Superresolution}}} & 52.2 & 30.4 & 17.4 & 47.8 & 23 \\
\textbf{{\action{Specialized operation}}} & 72.7 & 22.7 & 4.5 & 27.3 & 22 \\
\textbf{{\action{Crop}}} & 88.9 & 5.6 & 5.6 & 11.1 & 18 \\
\textbf{{\action{Zoom}}} & 60.0 & 20.0 & 20.0 & 40.0 & 5 \\
\textbf{{\action{Clone}}} & 0.0 & 100.0 & 0.0 & 100.0 & 1 \\
\bottomrule
\end{tabular}
\end{table}


\begin{table}[ht]
\centering
\caption{\gpt{} \gptlogo performance (win rate \%), categorized by different action verb types }
\label{tab:gpt-action-verb}
\begin{tabular}{lrrrrr}
\toprule
 Action & Human Wins & AI Wins & Tie & AI Win+Tie & Count \\
\midrule
\textbf{{\action{Adjust}}} & 59.8 & 35.4 & 4.8 & 40.2 & 336 \\
\textbf{{\action{Delete}}} & 68.6 & 27.7 & 3.7 & 31.4 & 271 \\
\textbf{{\action{Add}}} & 50.7 & 44.3 & 5.0 & 49.3 & 219 \\
\textbf{{\action{Recoloring}}} & 63.9 & 30.4 & 5.7 & 36.1 & 158 \\
\textbf{{\action{Replace}}} & 61.5 & 32.3 & 6.2 & 38.5 & 130 \\
\textbf{{\action{Superresolution}}} & 70.3 & 27.9 & 1.8 & 29.7 & 111 \\
\textbf{{\action{Transform}}} & 64.9 & 28.7 & 6.4 & 35.1 & 94 \\
\textbf{{\action{Apply}}} & 53.9 & 40.8 & 5.3 & 46.1 & 76 \\
\textbf{{\action{Relighting}}} & 51.0 & 37.3 & 11.8 & 49.0 & 51 \\
\textbf{{\action{Merge}}} & 54.2 & 39.6 & 6.2 & 45.8 & 48 \\
\textbf{{\action{Move}}} & 51.5 & 42.4 & 6.1 & 48.5 & 33 \\
\textbf{{\action{Crop}}} & 58.6 & 37.9 & 3.4 & 41.4 & 29 \\
\textbf{{\action{Specialized operation}}} & 39.3 & 42.9 & 17.9 & 60.7 & 28 \\
\textbf{{\action{Zoom}}} & 92.0 & 8.0 & 0.0 & 8.0 & 25 \\
\textbf{{\action{Clone}}} & 58.3 & 33.3 & 8.3 & 41.7 & 12 \\
\bottomrule
\end{tabular}
\end{table}


\begin{table}[ht]
\centering
\caption{\geminiflash{} performance (win rate \%), categorized by different action verb types }
\label{tab:gemini-action-verb}
\begin{tabular}{lrrrrr}
\toprule
 Action & Human Wins & AI Wins & Tie & AI Win+Tie & Count \\
\midrule
\textbf{{\action{Adjust}}} & 73.6 & 15.9 & 10.6 & 26.4 & 208 \\
\textbf{{\action{Add}}} & 69.0 & 22.8 & 8.2 & 31.0 & 158 \\
\textbf{{\action{Delete}}} & 74.2 & 21.3 & 4.5 & 25.8 & 155 \\
\textbf{{\action{Recoloring}}} & 65.4 & 24.3 & 10.3 & 34.6 & 107 \\
\textbf{{\action{Replace}}} & 67.1 & 23.5 & 9.4 & 32.9 & 85 \\
\textbf{{\action{Transform}}} & 74.6 & 12.7 & 12.7 & 25.4 & 71 \\
\textbf{{\action{Apply}}} & 63.6 & 21.8 & 14.5 & 36.4 & 55 \\
\textbf{{\action{Superresolution}}} & 71.2 & 21.2 & 7.7 & 28.8 & 52 \\
\textbf{{\action{Merge}}} & 71.4 & 25.7 & 2.9 & 28.6 & 35 \\
\textbf{{\action{Relighting}}} & 80.0 & 11.4 & 8.6 & 20.0 & 35 \\
\textbf{{\action{Move}}} & 57.6 & 24.2 & 18.2 & 42.4 & 33 \\
\textbf{{\action{Crop}}} & 75.0 & 16.7 & 8.3 & 25.0 & 24 \\
\textbf{{\action{Specialized operation}}} & 66.7 & 23.8 & 9.5 & 33.3 & 21 \\
\textbf{{\action{Zoom}}} & 83.3 & 11.1 & 5.6 & 16.7 & 18 \\
\textbf{{\action{Clone}}} & 71.4 & 14.3 & 14.3 & 28.6 & 7 \\
\bottomrule
\end{tabular}
\end{table}


\begin{table}[ht]
\centering
\caption{HuggingFace performance (win rate \%), categorized by different action verb types }
\label{tab:huggingface-action-verb}
\begin{tabular}{lrrrrr}
\toprule
 Action & Human Wins & AI Wins & Tie & AI Win+Tie & Count \\
\midrule
\textbf{{\action{Adjust}}} & 75.2 & 16.5 & 8.3 & 24.8 & 552 \\
\textbf{{\action{Add}}} & 73.7 & 19.7 & 6.6 & 26.3 & 456 \\
\textbf{{\action{Delete}}} & 66.9 & 19.7 & 13.4 & 33.1 & 426 \\
\textbf{{\action{Replace}}} & 83.3 & 11.7 & 5.0 & 16.7 & 240 \\
\textbf{{\action{Recoloring}}} & 75.3 & 19.5 & 5.2 & 24.7 & 231 \\
\textbf{{\action{Apply}}} & 71.3 & 17.7 & 11.0 & 28.7 & 181 \\
\textbf{{\action{Superresolution}}} & 72.3 & 20.5 & 7.2 & 27.7 & 166 \\
\textbf{{\action{Transform}}} & 74.8 & 13.7 & 11.5 & 25.2 & 131 \\
\textbf{{\action{Merge}}} & 67.9 & 19.8 & 12.3 & 32.1 & 81 \\
\textbf{{\action{Relighting}}} & 73.4 & 18.8 & 7.8 & 26.6 & 64 \\
\textbf{{\action{Move}}} & 82.3 & 9.7 & 8.1 & 17.7 & 62 \\
\textbf{{\action{Crop}}} & 93.1 & 6.9 & 0.0 & 6.9 & 58 \\
\textbf{{\action{Specialized operation}}} & 71.1 & 20.0 & 8.9 & 28.9 & 45 \\
\textbf{{\action{Zoom}}} & 85.2 & 11.1 & 3.7 & 14.8 & 27 \\
\textbf{{\action{Clone}}} & 20.0 & 20.0 & 60.0 & 80.0 & 5 \\
\bottomrule
\end{tabular}
\end{table}

\clearpage


\begin{table}[htbp]
\centering
\caption{Human preference win rate (\%) by ``main category'' (all models). AI+Tie denotes share currently handled by AI.}
\label{tab:main-cat-win-rates}
\begin{tabular}{lrrrrr}
\toprule
 Category & Human Wins & AI Wins & Tie & AI Win+Tie & Count \\
\midrule
\textbf{{\category{People And Related}}} & 68.5 & 24.3 & 7.2 & 31.5 & 1992 \\
\textbf{{\category{Inanimate Objects}}} & 67.0 & 23.1 & 9.8 & 33.0 & 864 \\
\textbf{{\category{Animals}}} & 59.0 & 32.9 & 8.1 & 41.0 & 642 \\
\textbf{{\category{Environment And Background}}} & 63.5 & 28.1 & 8.4 & 36.5 & 452 \\
\textbf{{\category{Text Branding And Abstract}}} & 65.1 & 25.2 & 9.7 & 34.9 & 404 \\
\bottomrule
\end{tabular}
\end{table}

\begin{table}[htbp]
\centering
\caption{\SeedEdit{}  performance (win rate \%), categorized by different subjects}
\label{tab:seededit-main-category}
\begin{tabular}{lrrrrr}
\toprule
 Category & Human Wins & AI Wins & Tie & AI Win+Tie & Count \\
\midrule
\textbf{{\category{People And Related}}} & 54.2 & 37.4 & 8.4 & 45.8 & 321 \\
\textbf{{\category{Inanimate Objects}}} & 61.7 & 30.6 & 7.7 & 38.3 & 196 \\
\textbf{{\category{Animals}}} & 41.2 & 49.1 & 9.7 & 58.8 & 165 \\
\textbf{{\category{Environment And Background}}} & 51.0 & 40.2 & 8.8 & 49.0 & 102 \\
\textbf{{\category{Text Branding And Abstract}}} & 58.4 & 32.7 & 8.9 & 41.6 & 101 \\
\bottomrule
\end{tabular}
\end{table}

\begin{table}[htbp]
\centering
\caption{\geminiflash{}  performance (win rate \%), categorized by different subjects}
\label{tab:gemini-main-category}
\begin{tabular}{lrrrrr}
\toprule
 Category & Human Wins & AI Wins & Tie & AI Win+Tie & Count \\
\midrule
\textbf{{\category{People And Related}}} & 72.9 & 19.2 & 7.9 & 27.1 & 328 \\
\textbf{{\category{Inanimate Objects}}} & 62.7 & 21.2 & 16.1 & 37.3 & 118 \\
\textbf{{\category{Animals}}} & 69.7 & 20.2 & 10.1 & 30.3 & 99 \\
\textbf{{\category{Text Branding And Abstract}}} & 72.1 & 22.1 & 5.9 & 27.9 & 68 \\
\textbf{{\category{Environment And Background}}} & 67.2 & 25.4 & 7.5 & 32.8 & 67 \\
\bottomrule
\end{tabular}
\end{table}

\begin{table}[htbp]
\centering
\caption{\gpt{}  performance (win rate \%), categorized by different subjects}
\label{tab:gpt-main-category}
\begin{tabular}{lrrrrr}
\toprule
 Category & Human Wins & AI Wins & Tie & AI Win+Tie & Count \\
\midrule
\textbf{{\category{People And Related}}} & 63.6 & 31.3 & 5.1 & 36.4 & 514 \\
\textbf{{\category{Inanimate Objects}}} & 61.8 & 31.4 & 6.8 & 38.2 & 191 \\
\textbf{{\category{Animals}}} & 53.1 & 43.4 & 3.4 & 46.9 & 145 \\
\textbf{{\category{Environment And Background}}} & 62.7 & 26.5 & 10.8 & 37.3 & 83 \\
\textbf{{\category{Text Branding And Abstract}}} & 61.7 & 33.3 & 4.9 & 38.3 & 81 \\
\bottomrule
\end{tabular}
\end{table}

\begin{table}[htbp]
\centering
\caption{HuggingFace performance (win rate \%), categorized by different subjects}
\label{tab:huggingface-main-category}
\begin{tabular}{lrrrrr}
\toprule
 Category & Human Wins & AI Wins & Tie & AI Win+Tie & Count \\
\midrule
\textbf{{\category{People And Related}}} & 75.4 & 16.9 & 7.7 & 24.6 & 829 \\
\textbf{{\category{Inanimate Objects}}} & 74.1 & 15.3 & 10.6 & 25.9 & 359 \\
\textbf{{\category{Animals}}} & 70.8 & 20.2 & 9.0 & 29.2 & 233 \\
\textbf{{\category{Environment And Background}}} & 69.0 & 23.5 & 7.5 & 31.0 & 200 \\
\textbf{{\category{Text Branding And Abstract}}} & 68.2 & 17.5 & 14.3 & 31.8 & 154 \\
\bottomrule
\end{tabular}
\end{table}


\clearpage
\subsection{Analysis of Aesthetic Scores}
\label{sec:supp-analysis_of_Aesthetic_Score}

\begin{figure*}[ht]
    \centering
    \includegraphics[width=0.85\linewidth]{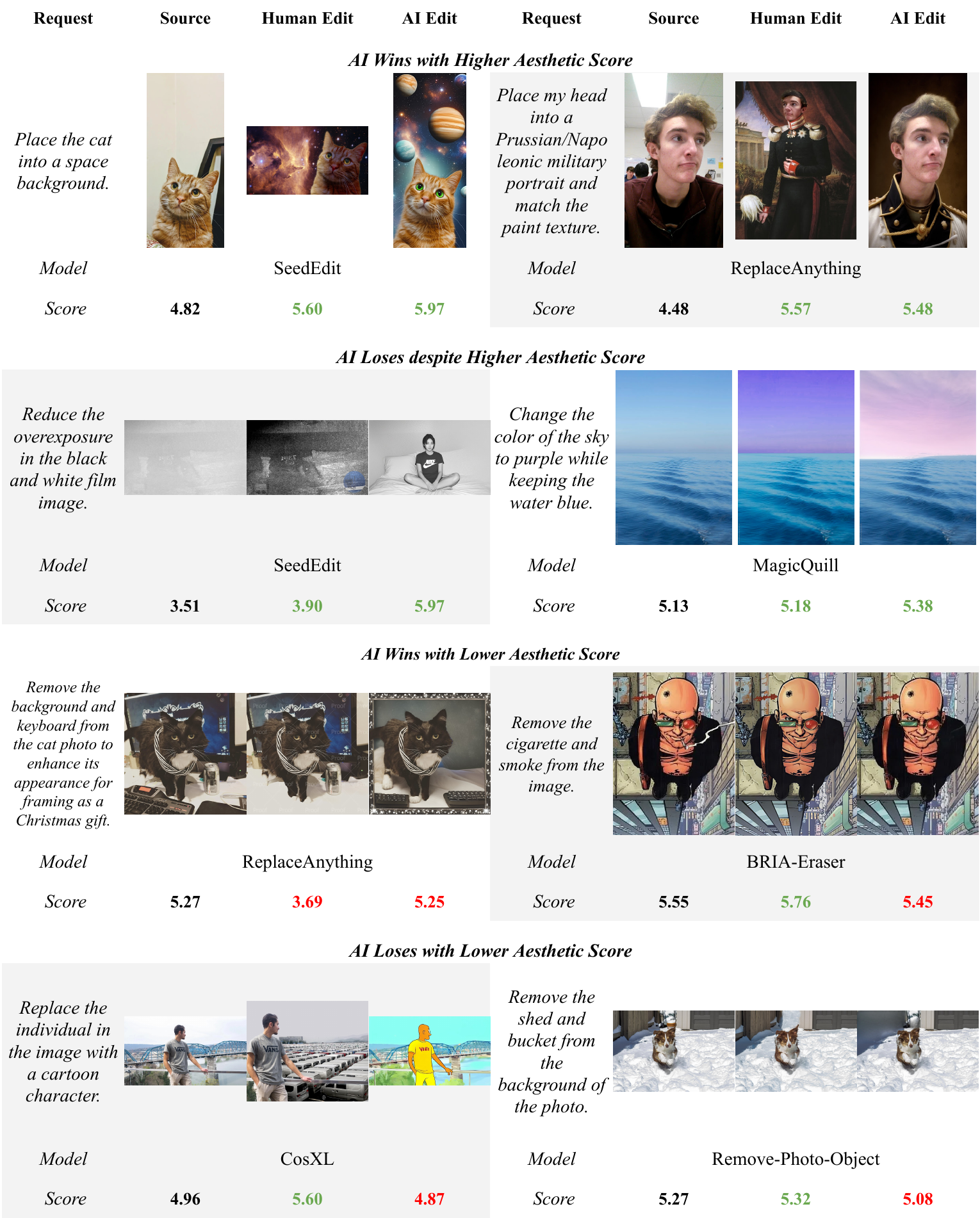}
    \caption{Samples showing that the aesthetic score is not a reliable proxy for human evaluation. The AI edit may win or lose user preference, while the aesthetic score can either increase or decrease relative to the source image.}
    \label{fig:supp-sample_ai_win_lose_different_aes_scores}
\end{figure*}


\begin{figure}[ht]
    \centering
    \includegraphics[width=\linewidth]{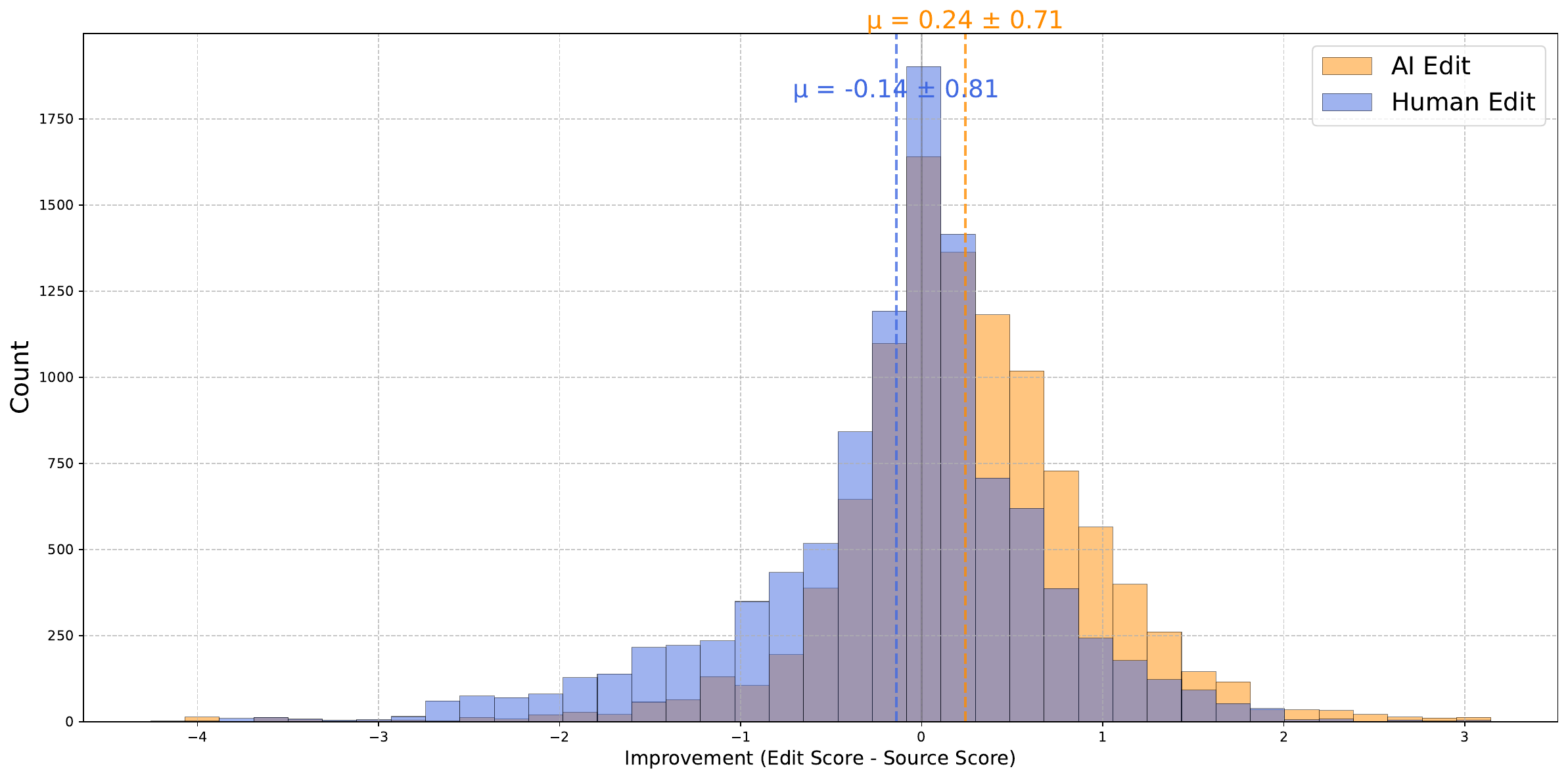}
    \caption{Histogram of Aesthetic Improvement Score (Edited Image Score - Source Image Score)}
    \label{fig:supp-DistributionofAestheticImprovementScore}
\end{figure}

\begin{figure}[ht]
\includegraphics[width=\linewidth]{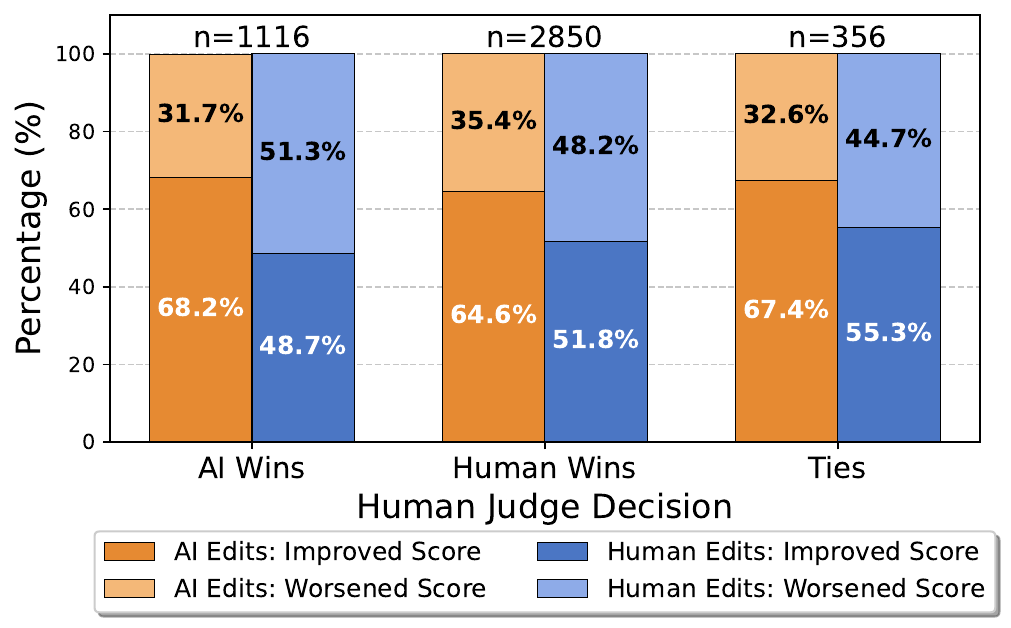}
      \caption{Distribution of aesthetic score changes by different rating outcomes.}   
      \label{fig:aesthetics_by_rating}
\end{figure}

\clearpage
\subsection{Additional Results for VLMs-as-a-Judge}
\label{sec:supp-vlm-as-judge}

\begin{table}[htp]
    \centering
    \caption{Cohen's $\kappa$ for agreement between human judgment and VLMs, showing generally poor agreement between VLMs and human preferences}
    \label{tab:new_kappa}
\begin{tabular}{ll}
\textbf{Judge} & \textbf{$\kappa$} \\
\midrule
\gpt & 0.195\;(\textit{n}=4332) \\
\oone & 0.229\;(\textit{n}=4350) \\
\geminiflash{} & 0.177\;(\textit{n}=4348) \\
\bottomrule
\end{tabular}
\end{table}

\begin{table}[htp]
    \centering
    \caption{Cohen's $\kappa$ for agreement between human judgment and VLMs across different model groups. \oone{} strongly prefers \gpt{} edits, resulting in very poor agreement with humans.}
    \label{tab:new_kappa_model_group}
    \begin{tabular}{l*{3}{rr}}
        & \multicolumn{2}{c}{\gpt} 
        & \multicolumn{2}{c}{\oone} 
        & \multicolumn{2}{c}{\geminiflash{}} \\
        Model & $\kappa$ & $n$ & $\kappa$ & $n$ & $\kappa$ & $n$ \\
        \midrule
        \seedlogo~\SeedEdit          & 0.226 &  882 & 0.287 &  885 & 0.203 &  882 \\
        \gptlogo~\gpt                & 0.047 & 1006 & 0.054 & 1013 & 0.056 & 1014 \\
        \geminilogo~\Gemini          & 0.141 &  676 & 0.250 &  679 & 0.144 &  680 \\
        \huggingface{}HF             & 0.203 & 1768 & 0.195 & 1773 & 0.172 & 1772 \\
        \bottomrule
    \end{tabular}
\end{table}

\clearpage
\subsubsection{Sample Reasoning Outputs from VLM Judges}
In this section, we provide sample reasoning outputs for each VLM across various agreement settings, compared to human evaluations.

\begin{figure}[htp]
    \centering
    \includegraphics[width=1\linewidth, trim={0 50pt 0 50pt}, clip]{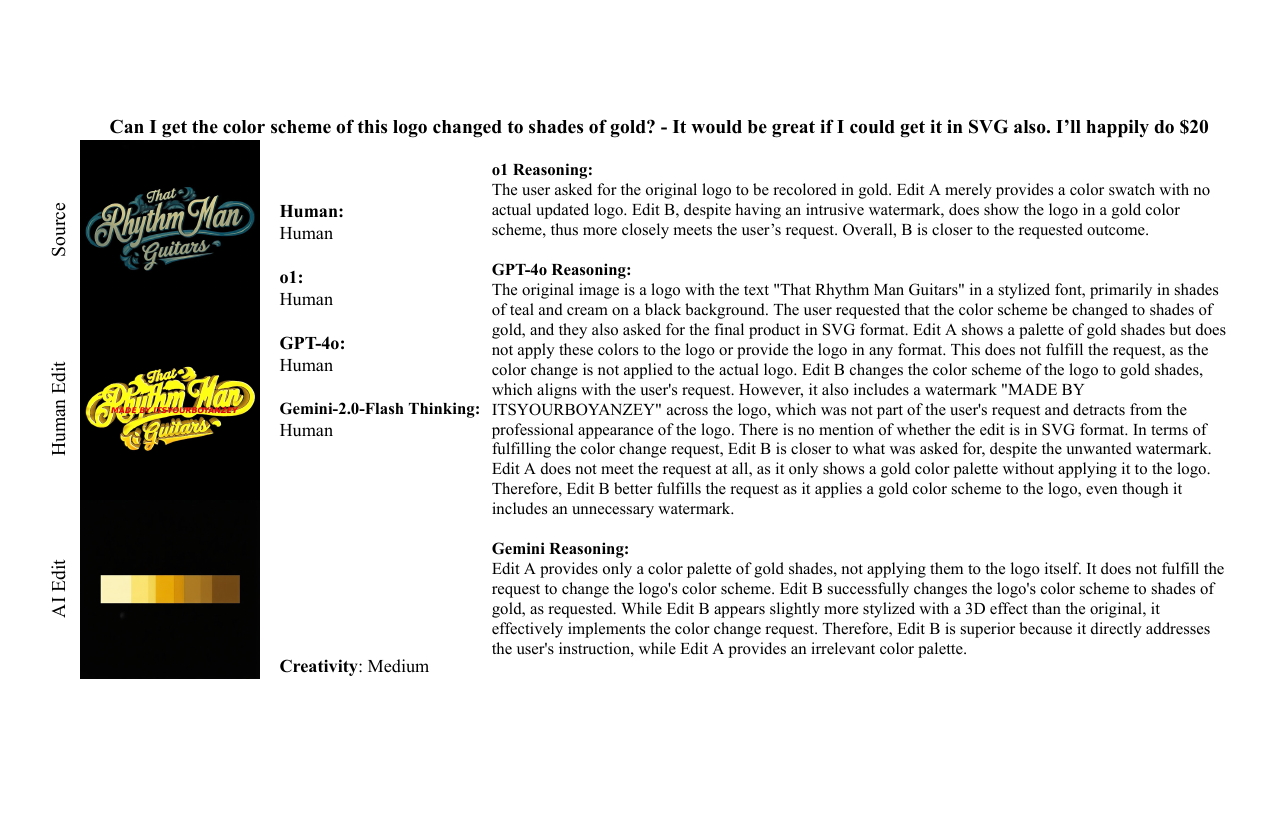}
    \caption{Sample reasoning of VLMs-as-a-judge for the case where all three VLMs agree with the human decision.}
    \label{fig:supp:sample_judge_reasoning_1}
\end{figure}

\begin{figure}[htp]
    \centering
    \includegraphics[width=1\linewidth, trim={0 50pt 0 50pt}, clip]{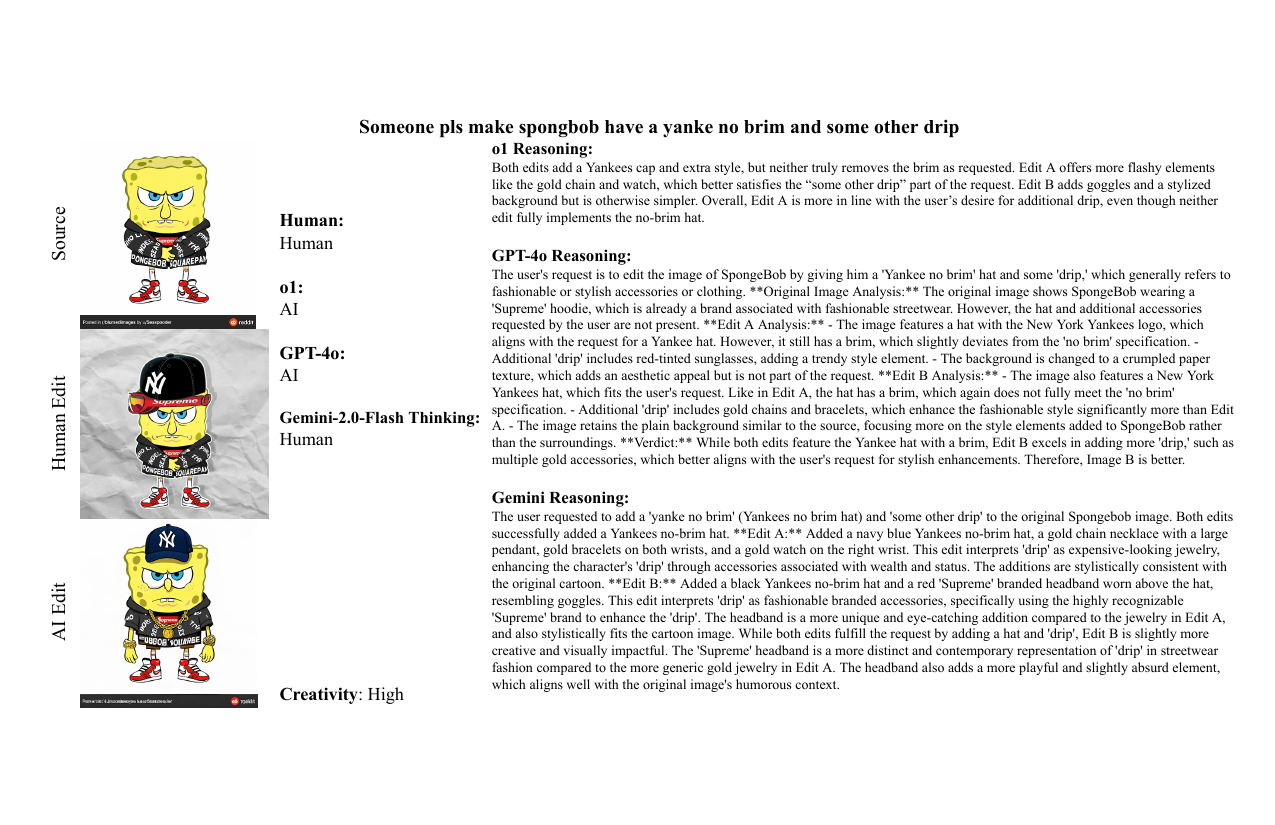}
\caption{Sample reasoning of VLMs-as-a-judge for the case with mixed results, where some of the VLMs agree with the human decision.}
    \label{fig:supp:sample_judge_reasoning_2}
\end{figure}

\begin{figure}[htp]
    \centering
    \includegraphics[width=1\linewidth, trim={0 50pt 0 50pt}, clip]{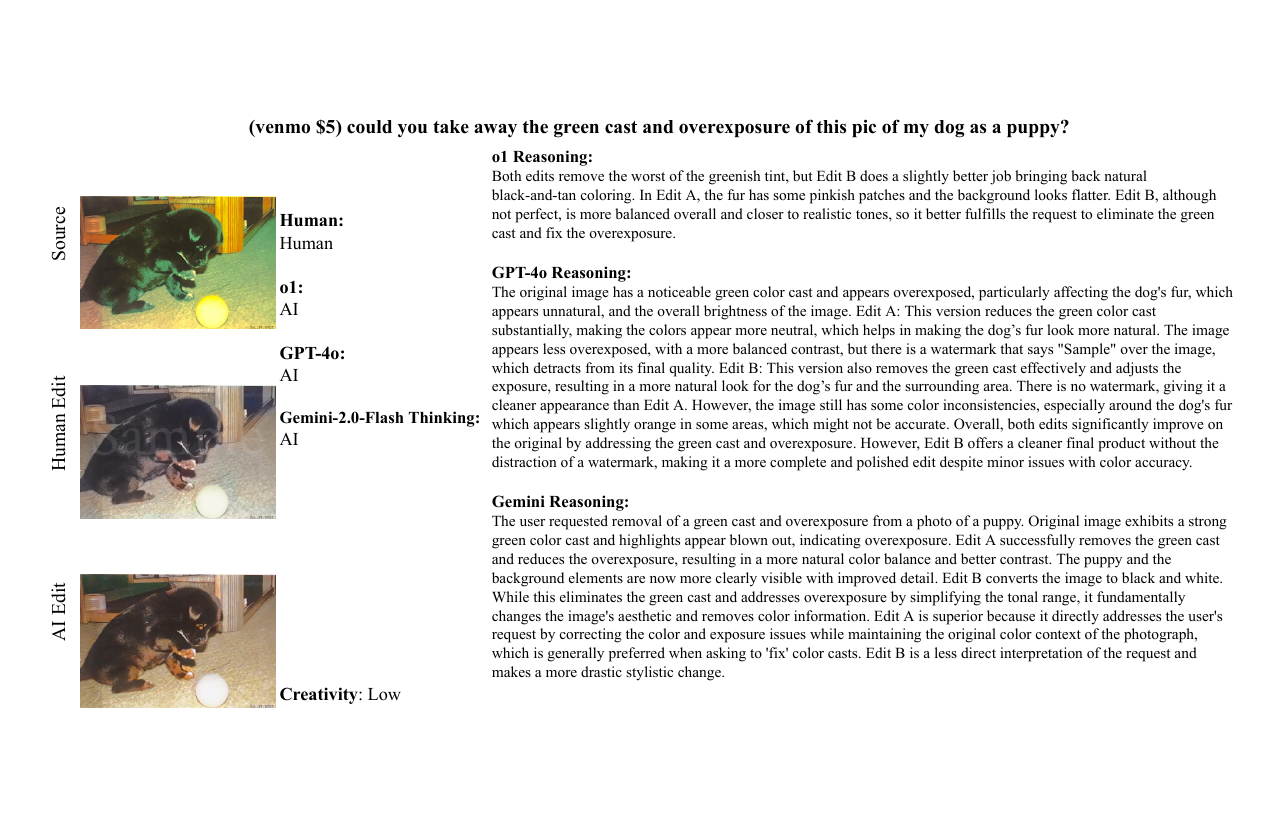}
    \caption{Sample reasoning of VLMs-as-a-judge for the case where all three VLMs disagree with the human decision.}
    \label{fig:supp:sample_judge_reasoning_3}
\end{figure}

\begin{figure}[htp]
    \centering
    \includegraphics[width=1\linewidth, trim={0 50pt 0 50pt}, clip]{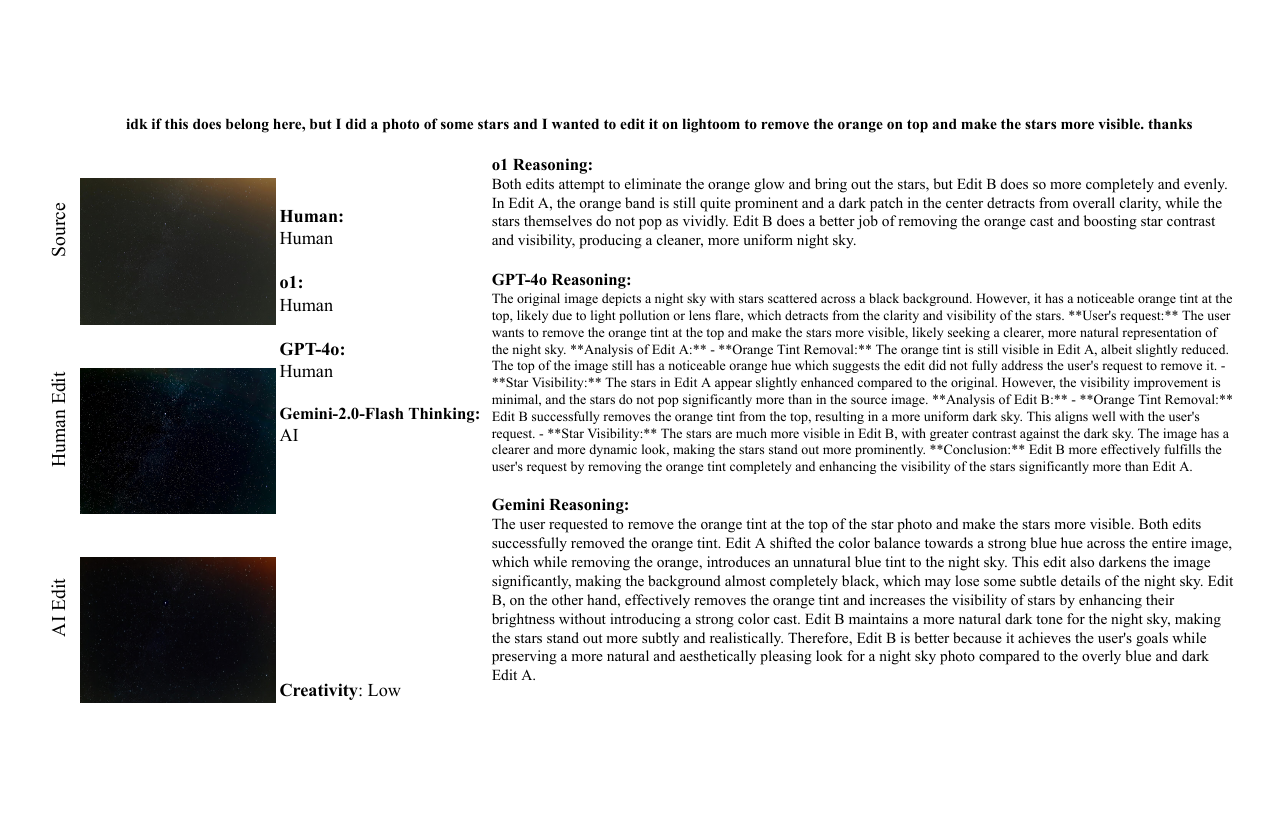}
    \caption{Sample reasoning of VLMs-as-a-judge for the case with mixed results, where some of the VLMs agree with the human decision.}
    \label{fig:supp:sample_judge_reasoning_4}
\end{figure}

\begin{figure}[htp]
    \centering
    \includegraphics[width=1\linewidth, trim={0 50pt 0 50pt}, clip]{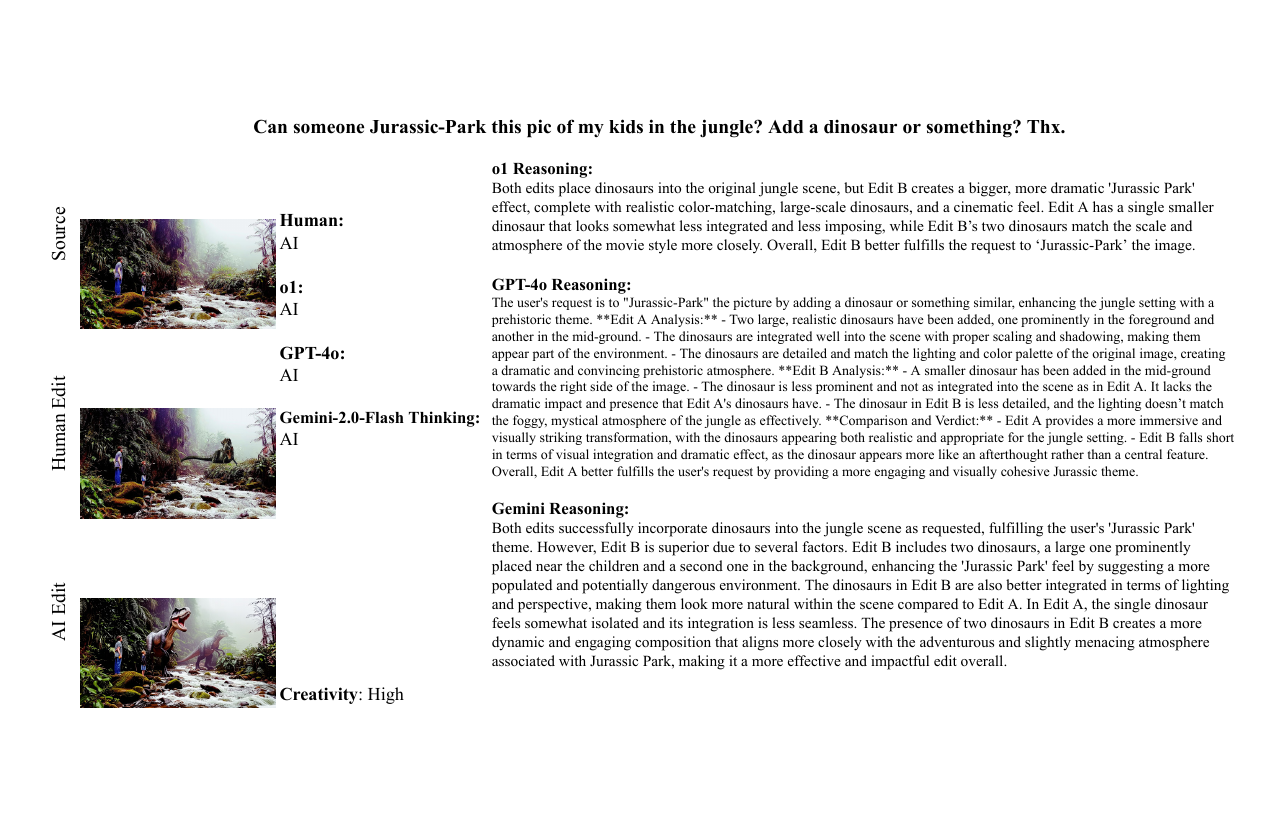}
    \caption{Sample reasoning of VLMs-as-a-judge for the case where all three VLMs agree with the human decision.}
    \label{fig:supp:sample_judge_reasoning_5}
\end{figure}

\begin{figure}[htp]
    \centering
    \includegraphics[width=1\linewidth, trim={0 50pt 0 50pt}, clip]{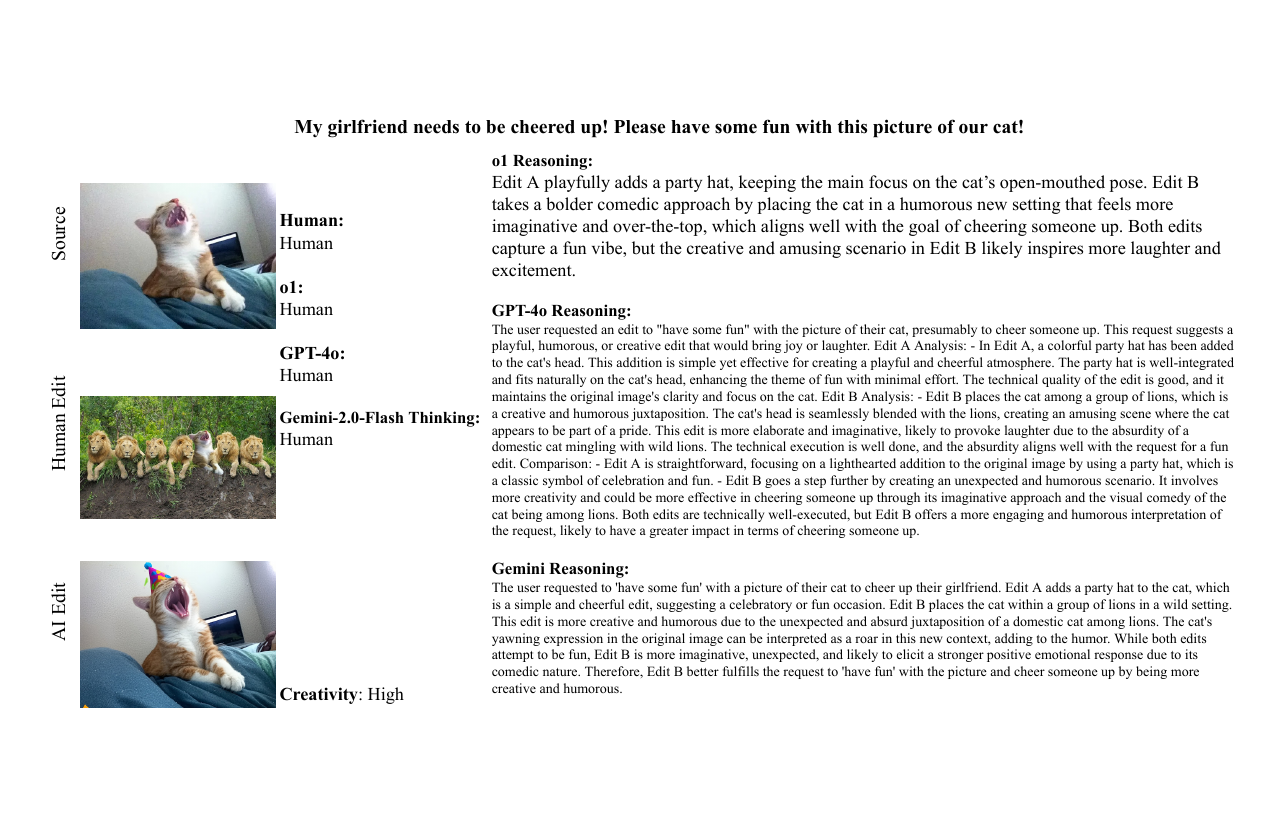}
    \caption{Sample reasoning of VLMs-as-a-judge for the case where all three VLMs agree with the human decision.}
    \label{fig:supp:sample_judge_reasoning_6}
\end{figure}

\begin{figure}[htp]
    \centering
    \includegraphics[width=1\linewidth, trim={0 80pt 0 50pt}, clip]{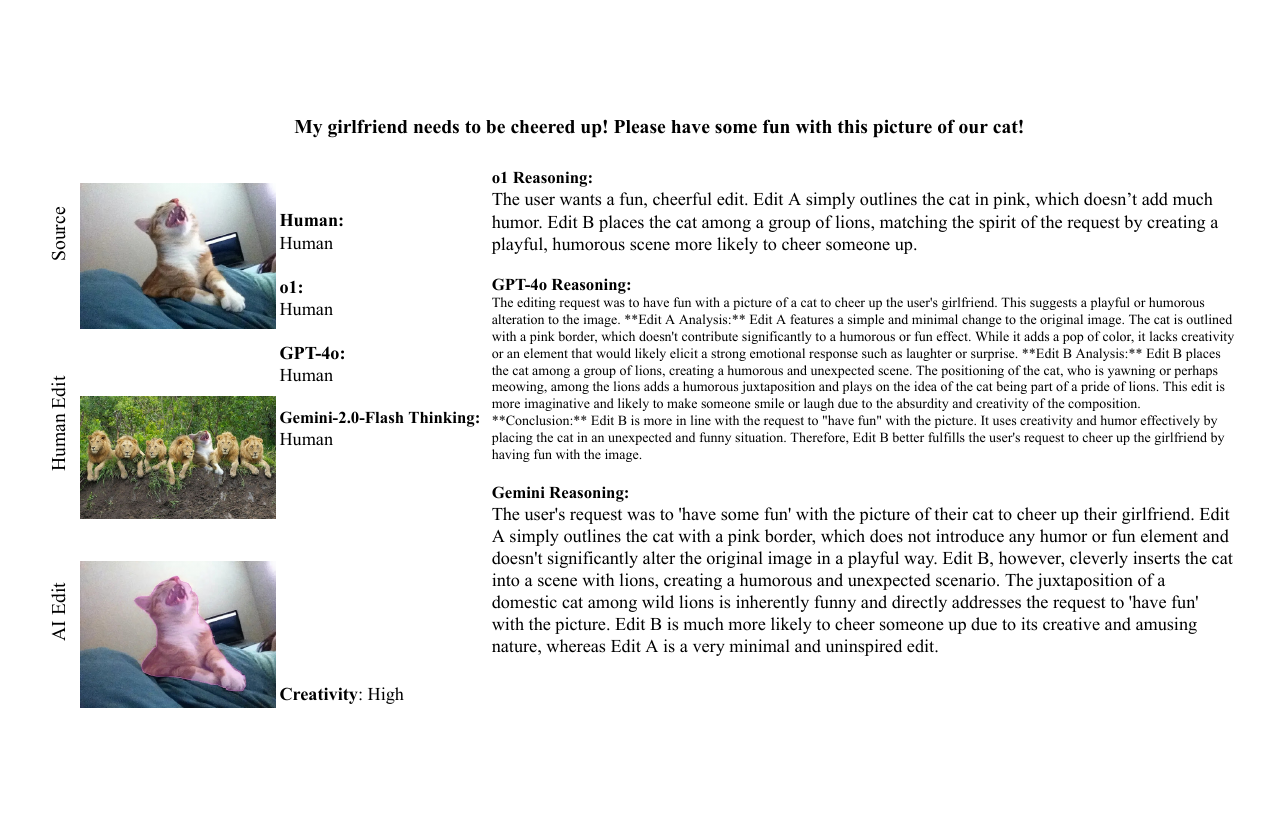}
    \caption{Sample reasoning of VLMs-as-a-judge for the case where all three VLMs agree with the human decision.}
    \label{fig:supp:sample_judge_reasoning_7}
\end{figure}

\begin{figure}[htp]
    \centering
    \includegraphics[width=1\linewidth, trim={0 50pt 0 50pt}, clip]{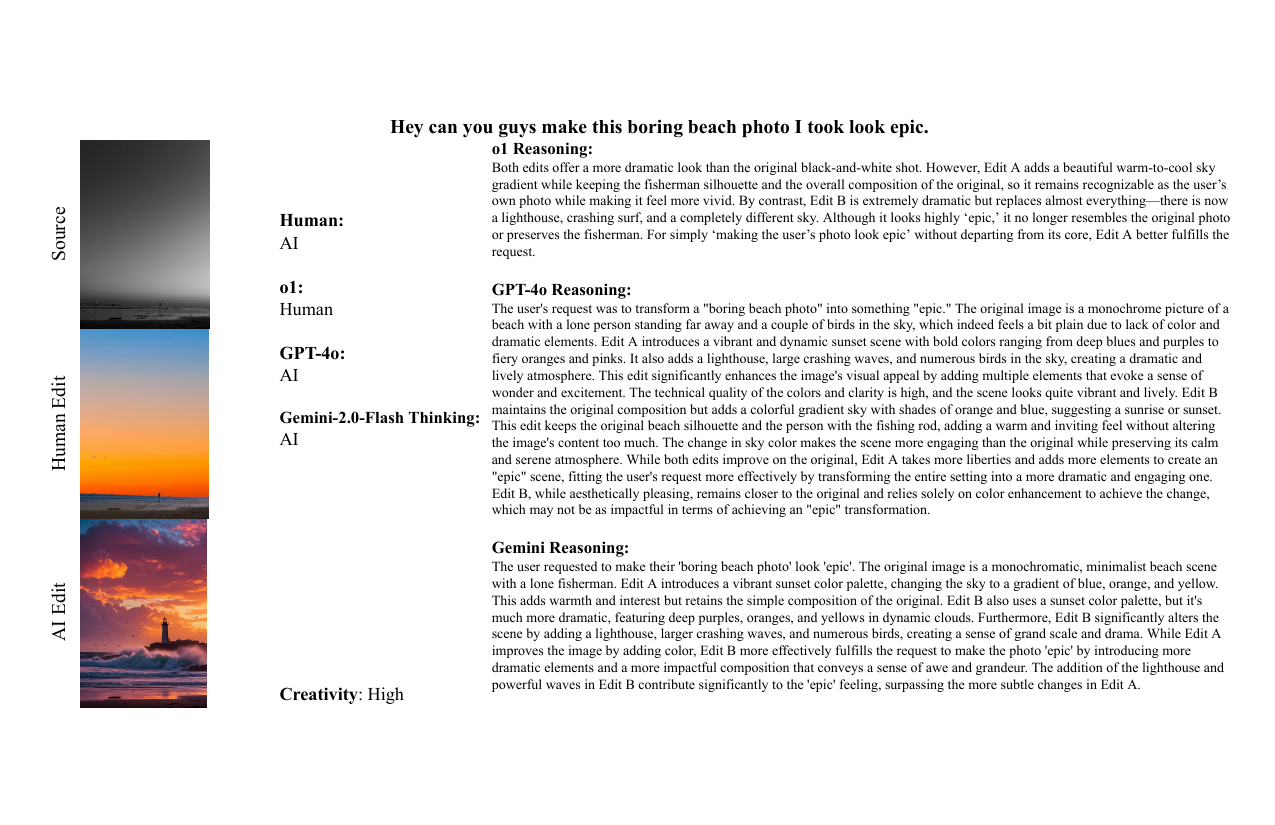}
    \caption{Sample reasoning of VLMs-as-a-judge for the case with mixed results, where some of the VLMs agree with the human decision.}
    \label{fig:supp:sample_judge_reasoning_8}
\end{figure}

\clearpage
\subsubsection{VLM Judges Fail to Detect Changes and Alterations in Subject Identity}
\label{sec:supp-vlm-as-judge_blind}

\begin{figure}[htp]
    \centering
    \includegraphics[width=1\linewidth, trim={0 50pt 0 50pt}, clip]{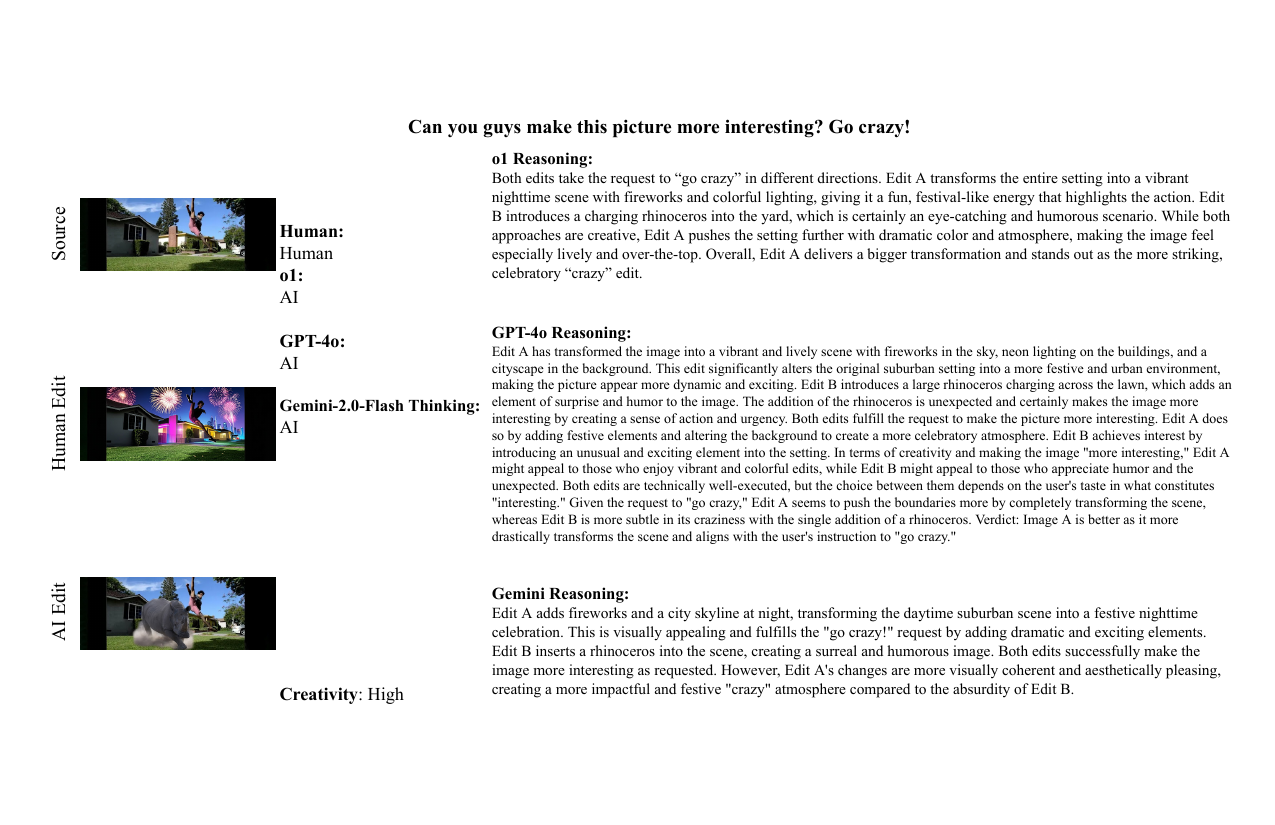}
\caption{Sample reasoning by VLMs-as-a-judge  for cases where all VLMs disagree with human judgments, primarily due to VLMs ignoring subtle differences between images.}
    \label{fig:supp:blind_judge_0}
\end{figure}

\begin{figure}[htp]
    \centering
    \includegraphics[width=1\linewidth, trim={0 50pt 0 50pt}, clip]{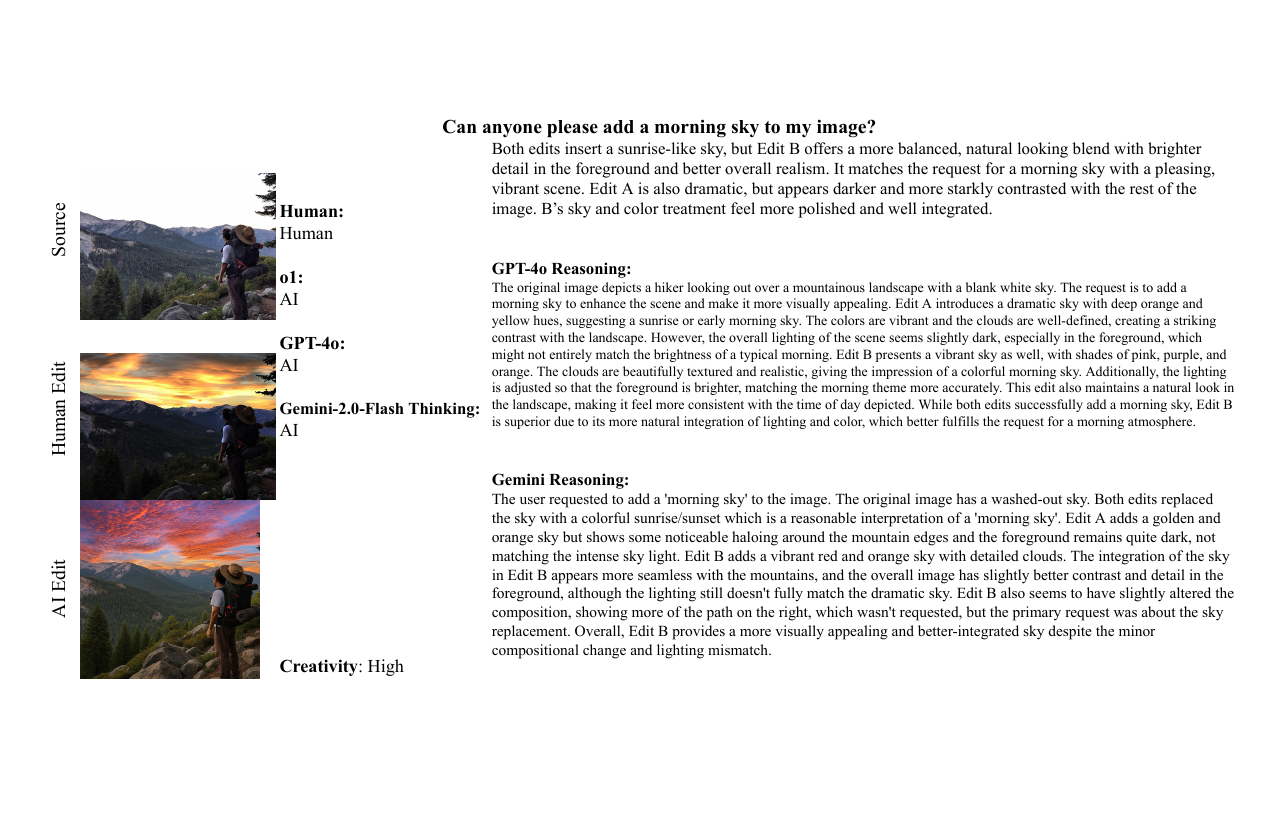}
\caption{Sample reasoning by VLMs-as-a-judge  for cases where all VLMs disagree with human judgments, primarily due to VLMs ignoring subtle differences between images.}
    \label{fig:supp:blind_judge_1}
\end{figure}

\begin{figure}[htp]
    \centering
    \includegraphics[width=1\linewidth, trim={0 50pt 0 50pt}, clip]{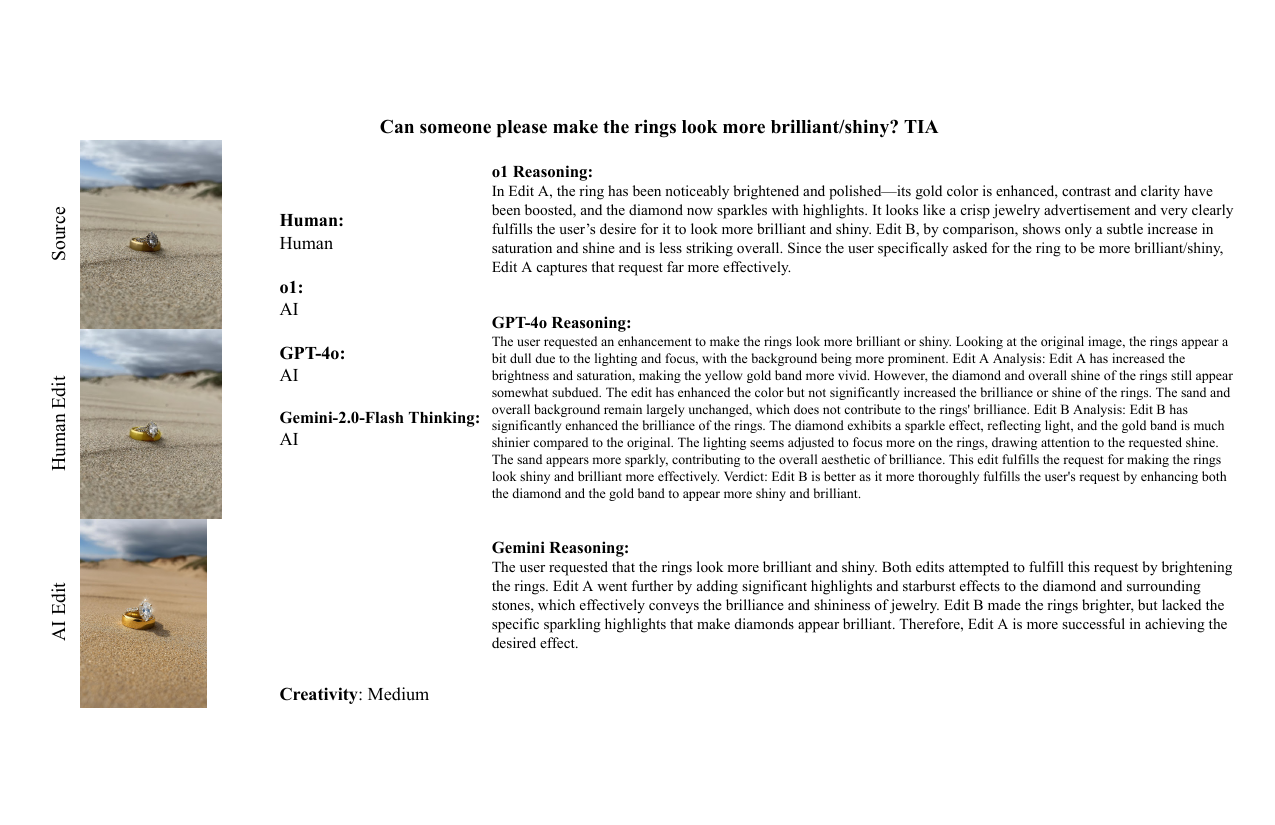}
\caption{Sample reasoning by VLMs-as-a-judge  for cases where all VLMs disagree with human judgments, primarily due to VLMs ignoring subtle differences between images.}
    \label{fig:supp:blind_judge_2}
\end{figure}

\begin{figure}[htp]
    \centering
    \includegraphics[width=1\linewidth, trim={0 50pt 0 50pt}, clip]{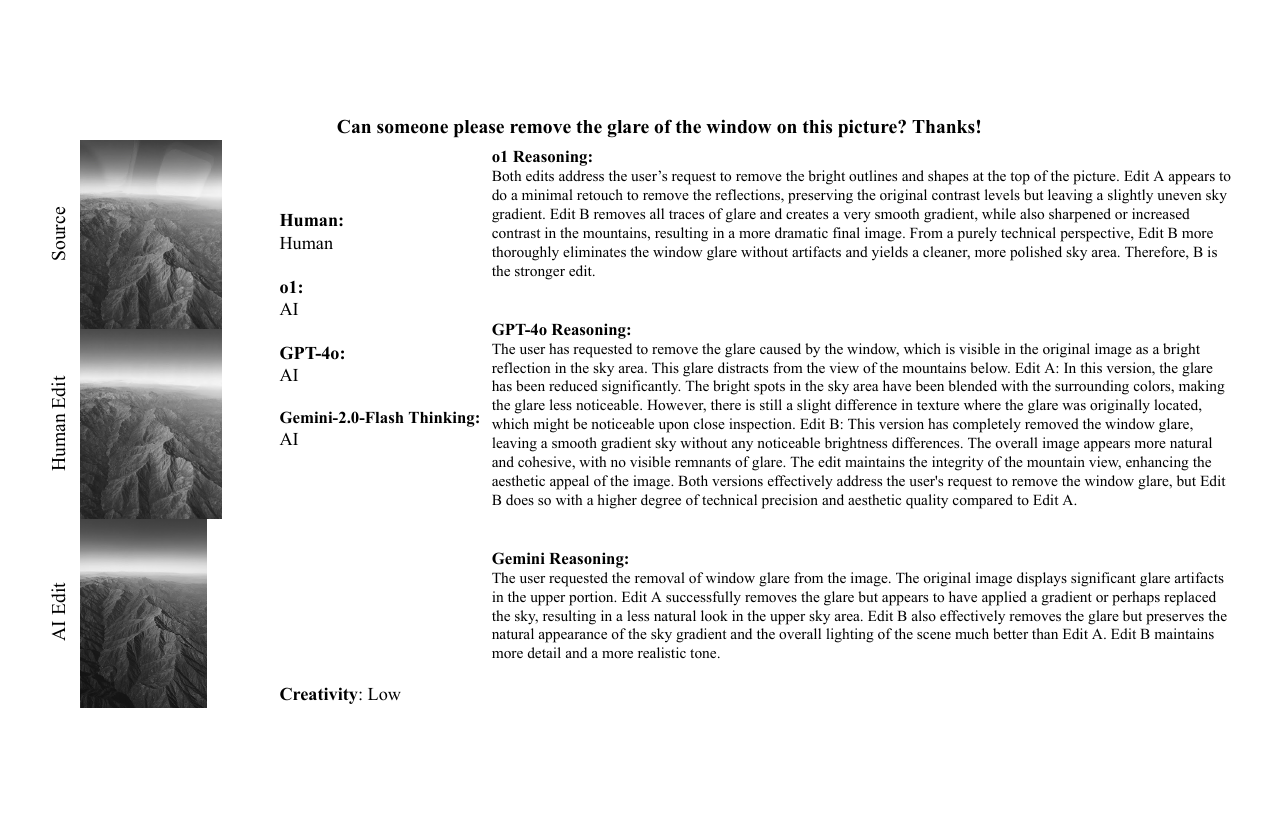}
\caption{Sample reasoning by VLMs-as-a-judge  for cases where all VLMs disagree with human judgments, primarily due to VLMs ignoring subtle differences between images.}
    \label{fig:supp:blind_judge_3}
\end{figure}

\clearpage
\subsubsection{Hallucinations by VLMs Acting as Judges}
\label{sec:supp-judge-fails}

\begin{figure}[htp]
    \centering
    \begin{tcolorbox}[
        colback=gray!15,
        colframe=gray!50!black,
        title=o1 VLM-as-a-Judge Hallucinations,
        width=0.7\columnwidth,
        arc=2mm,
        boxrule=1pt
    ]
        \centering
        \begin{minipage}{\columnwidth}
            \centering
            \includegraphics[width=\columnwidth]{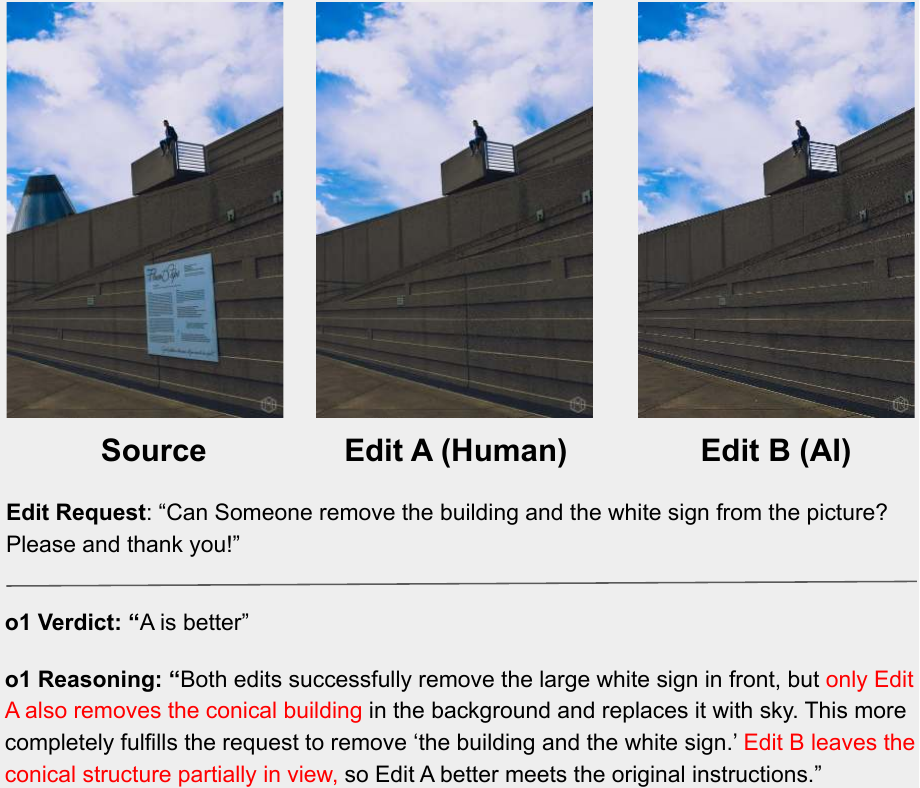}
        \end{minipage}
    \end{tcolorbox}
\caption{o1 occasionally fails to notice image details when judging different edits. In this example, it thinks only one of the images removed the conical structure, while in reality, both edits removed the building.}
    \label{fig:o1-Einstein}
\end{figure}

\clearpage
\subsection{Qualitative Analysis of AI Wins and Losses to Identify Patterns}
\label{sec:supp-why_ai_wins_or_lose}

\textbf{When AI Outperforms Human Preference (206 top-rated edits)}
AI wins primarily due to more accurately reflecting user requests:
\begin{itemize}
    \item Overall, 72\% of winning AI edits align closely with user instructions.
    \item Breakdown by model:
    \begin{itemize}
        \item \gpt: 73\% of 62 images
        \item \geminiflash{}: 72\% of 32 images
        \item \SeedEdit{}: 74\% of 76 images
        \item Hugging Face models (\huggingface): 69\% of 36 images
    \end{itemize}
\end{itemize}

\textbf{Common Patterns of AI Failures (400 sampled losses)}
Primary reasons for AI losing include:
\begin{itemize}
    \item \textbf{Misunderstanding user prompts (43\% overall)}:
    \begin{itemize}
        \item \gpt: 16\%
        \item \geminiflash{}: 58\%
        \item \SeedEdit{}: 46\%
        \item Hugging Face models (\huggingface): 50\%
    \end{itemize}
    
    \item \textbf{Unwanted artifacts or distortions}:
    \begin{itemize}
        \item Facial identity distortions were prominent in GPT-4o (78\%) compared to others:
        \begin{itemize}
            \item \geminiflash{}: 14\%
            \item \SeedEdit{}: 23\%
            \item Hugging Face models (\huggingface): 12\%
        \end{itemize}
        \item Other unrequested changes were found in 14\% to 23\% of edits across models.
    \end{itemize}
\end{itemize}

\textbf{Overall Insight}
AI models excel when accurately interpreting user requests, yet significant challenges remain in understanding instructions and minimizing unintended modifications. The severity of these issues varies significantly across different AI models.

\clearpage
\subsection{AI Edit Observations}
\label{sec:ai-extra-edits}

We qualitatively observe that many of the models - in particular text-image models like \SeedEdit{} or \CosXL{} - frequently add irrelevant changes to the original image. We manually inspect the top 50 highest win rate AI edited images and find that 22\% of them contain edits that are irrelevant to the original edit request.

\begin{figure*}[htbp]
  \centering
  \begin{minipage}[t]{0.7\textwidth}
      \begin{minipage}[t]{0.27\textwidth}
        \centering
        \textbf{Original}\\[2mm]
        \includegraphics[width=\textwidth]{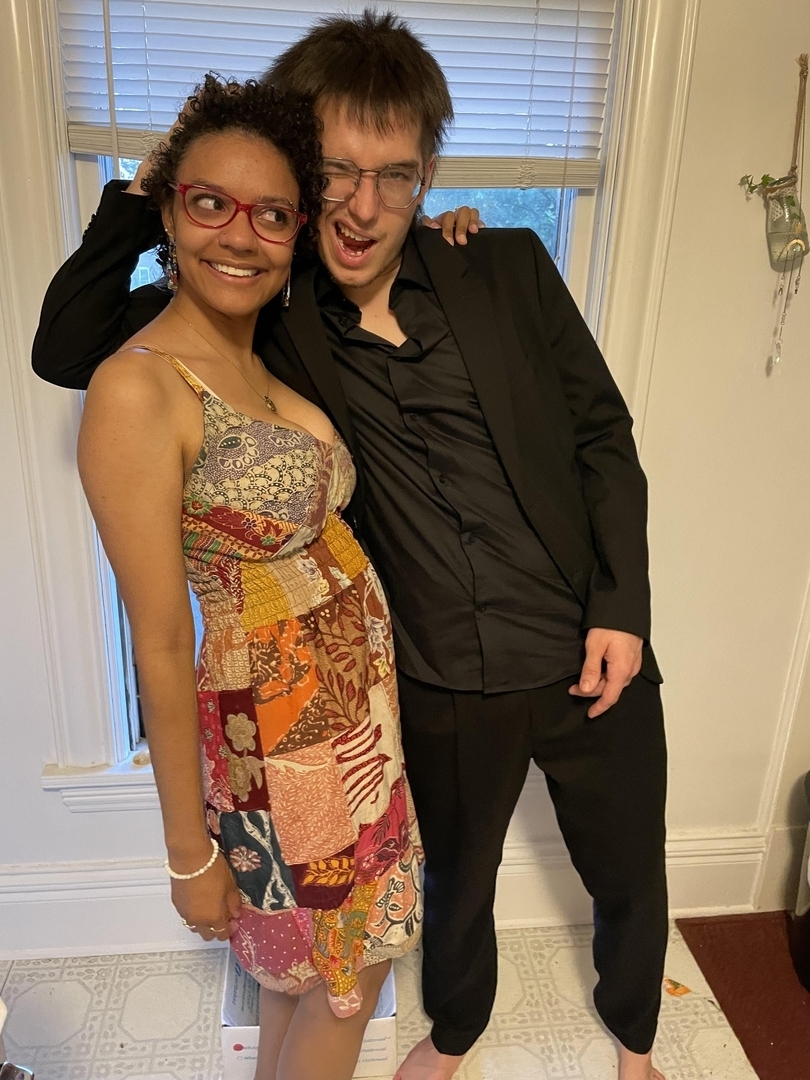}
      \end{minipage}\hfill
      \begin{minipage}[t]{0.27\textwidth}
        \centering
        \textbf{Human Edit}\\[2mm]
        \includegraphics[width=\textwidth]{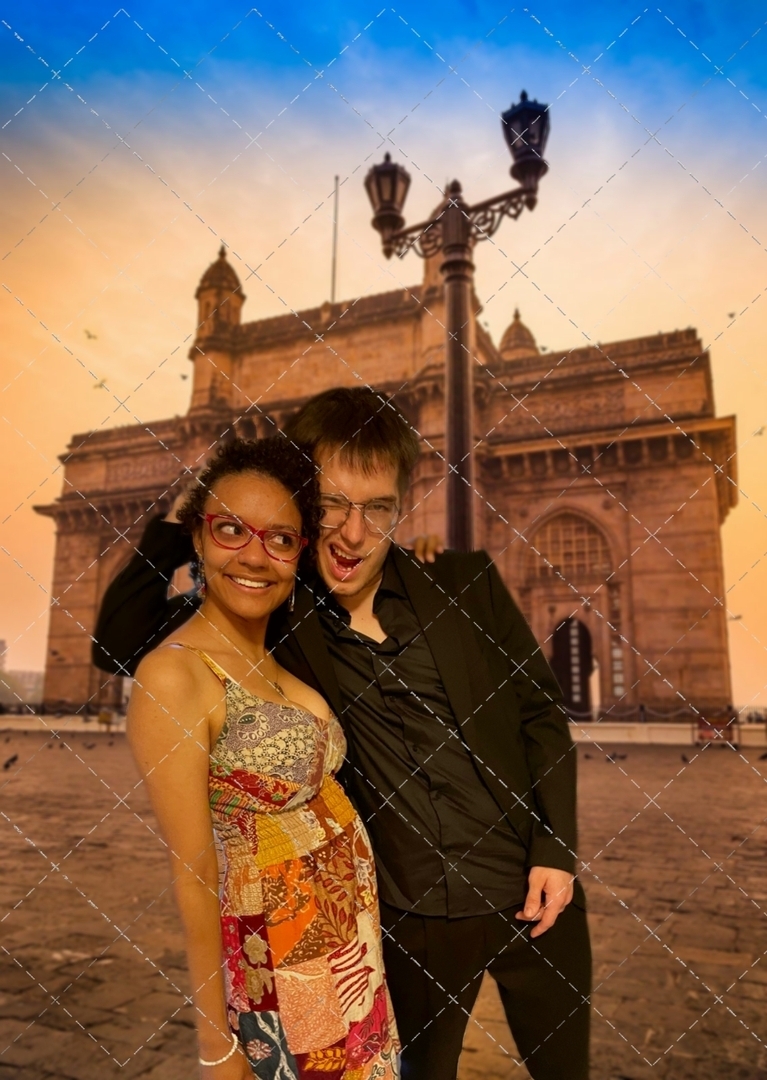}
      \end{minipage}\hfill
      \begin{minipage}[t]{0.27\textwidth}
        \centering
        \textbf{\CosXL}\\[2mm]
        \includegraphics[width=\textwidth]{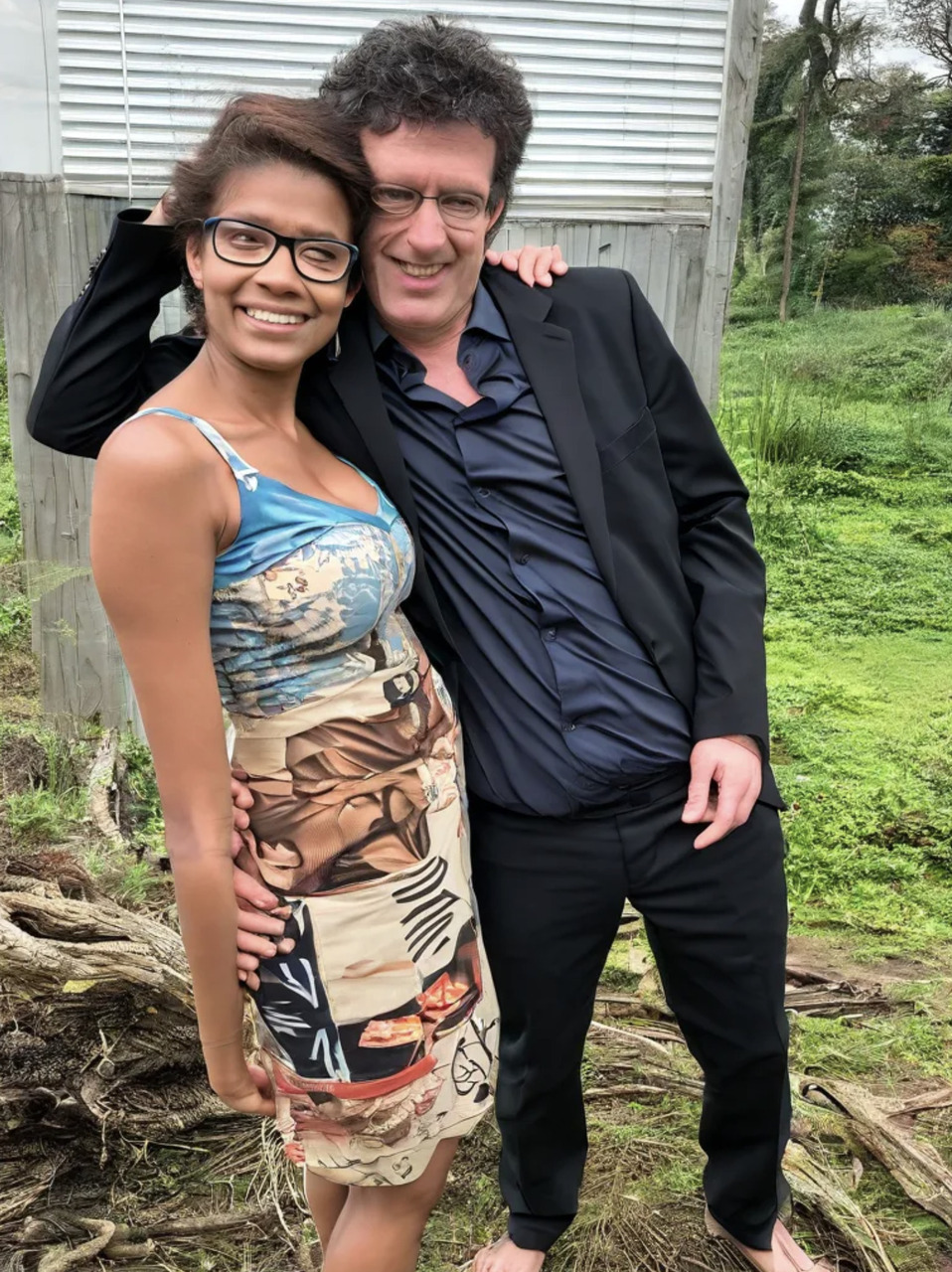}
      \end{minipage}\hfill
  \end{minipage}
  \caption{This edit adds unnecessary changes to the peoples' face, hands and clothes despite only needing to edit the background. \textit{User request: Place the two individuals in a different background landscape.}}
\end{figure*}

\begin{figure*}[htbp]
  \centering
  \begin{minipage}[t]{0.9\textwidth}
      \begin{minipage}[t]{0.32\textwidth}
        \centering
        \textbf{Original}\\[2mm]
        \includegraphics[width=\textwidth]{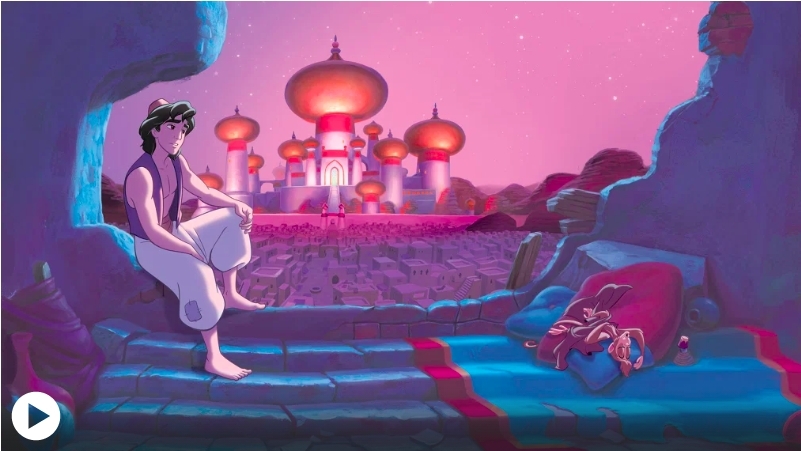}
      \end{minipage}\hfill
      \begin{minipage}[t]{0.32\textwidth}
        \centering
        \textbf{Human Edit}\\[2mm]
        \includegraphics[width=\textwidth]{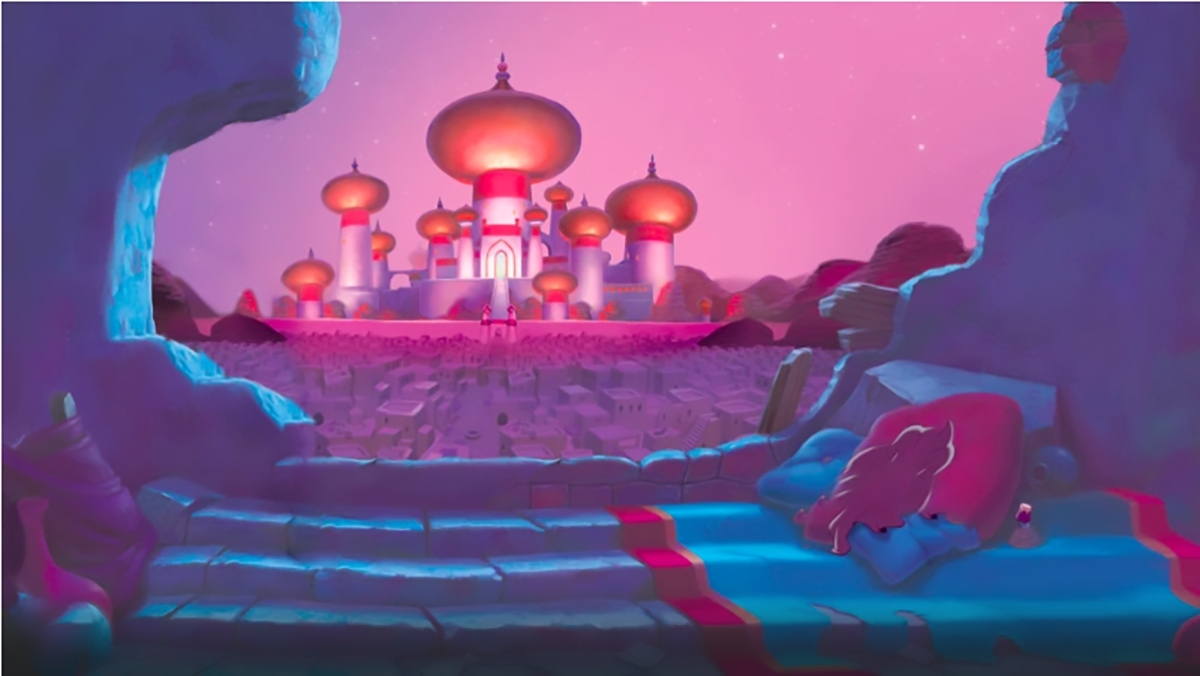}
      \end{minipage}\hfill
      \begin{minipage}[t]{0.32\textwidth}
        \centering
        \textbf{\SeedEdit}\\[2mm]
        \includegraphics[width=\textwidth]{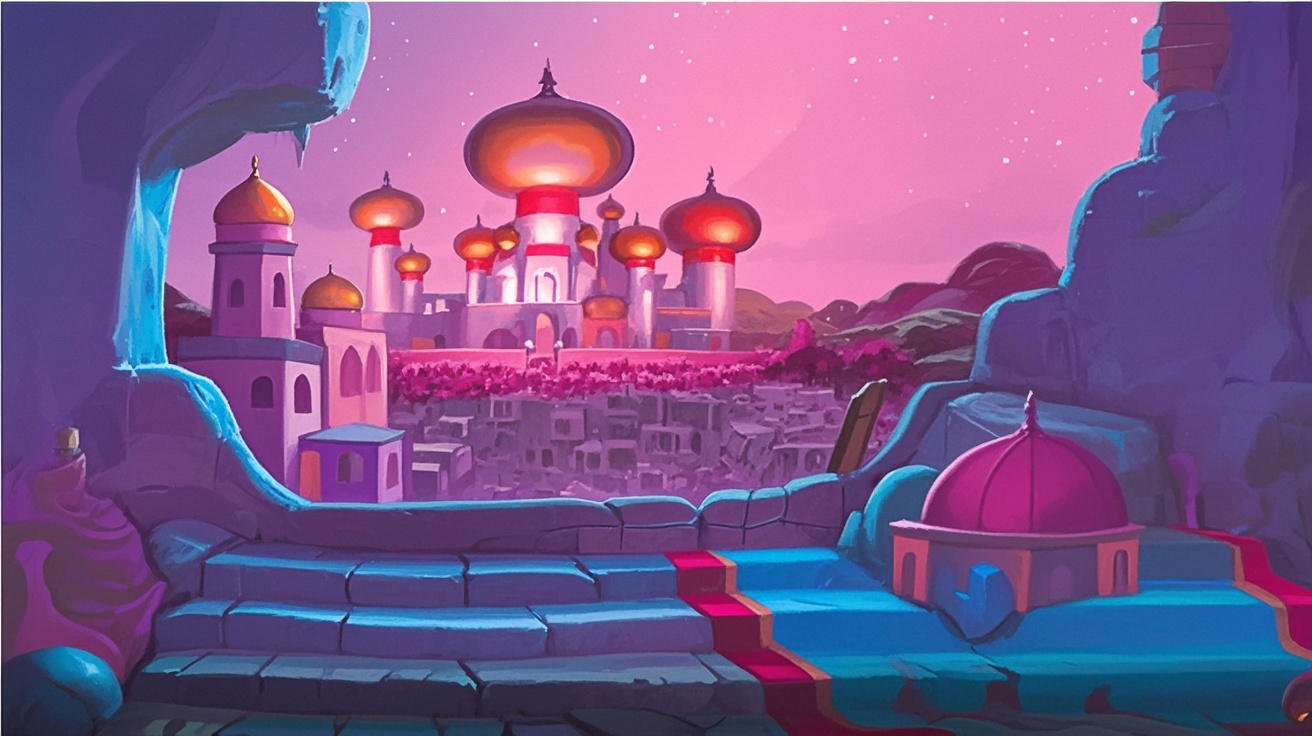}
      \end{minipage}\hfill
  \end{minipage}
  \caption{The request only asks for portions of the image to be removed, but the model adds additional buildings to the city background. \textit{User request: Remove Aladdin, Abu, and the play button to focus on the city-scape.}}
\end{figure*}

\begin{figure*}[htbp]
  \centering
  \begin{minipage}[t]{0.7\textwidth}
      \begin{minipage}[t]{0.27\textwidth}
        \centering
        \textbf{Original}\\[2mm]
        \includegraphics[width=\textwidth]{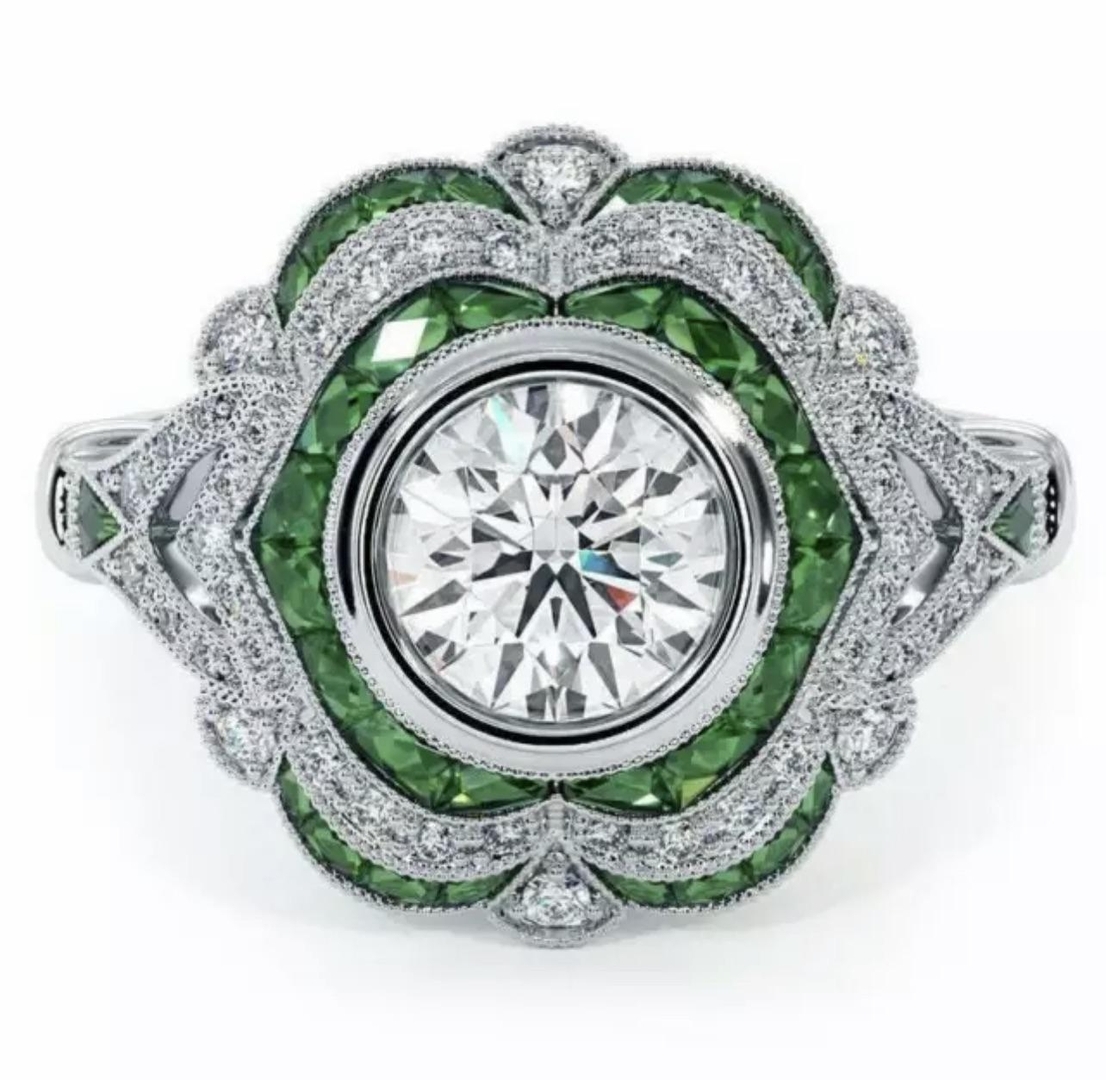}
      \end{minipage}\hfill
      \begin{minipage}[t]{0.27\textwidth}
        \centering
        \textbf{Human Edit}\\[2mm]
        \includegraphics[width=\textwidth]{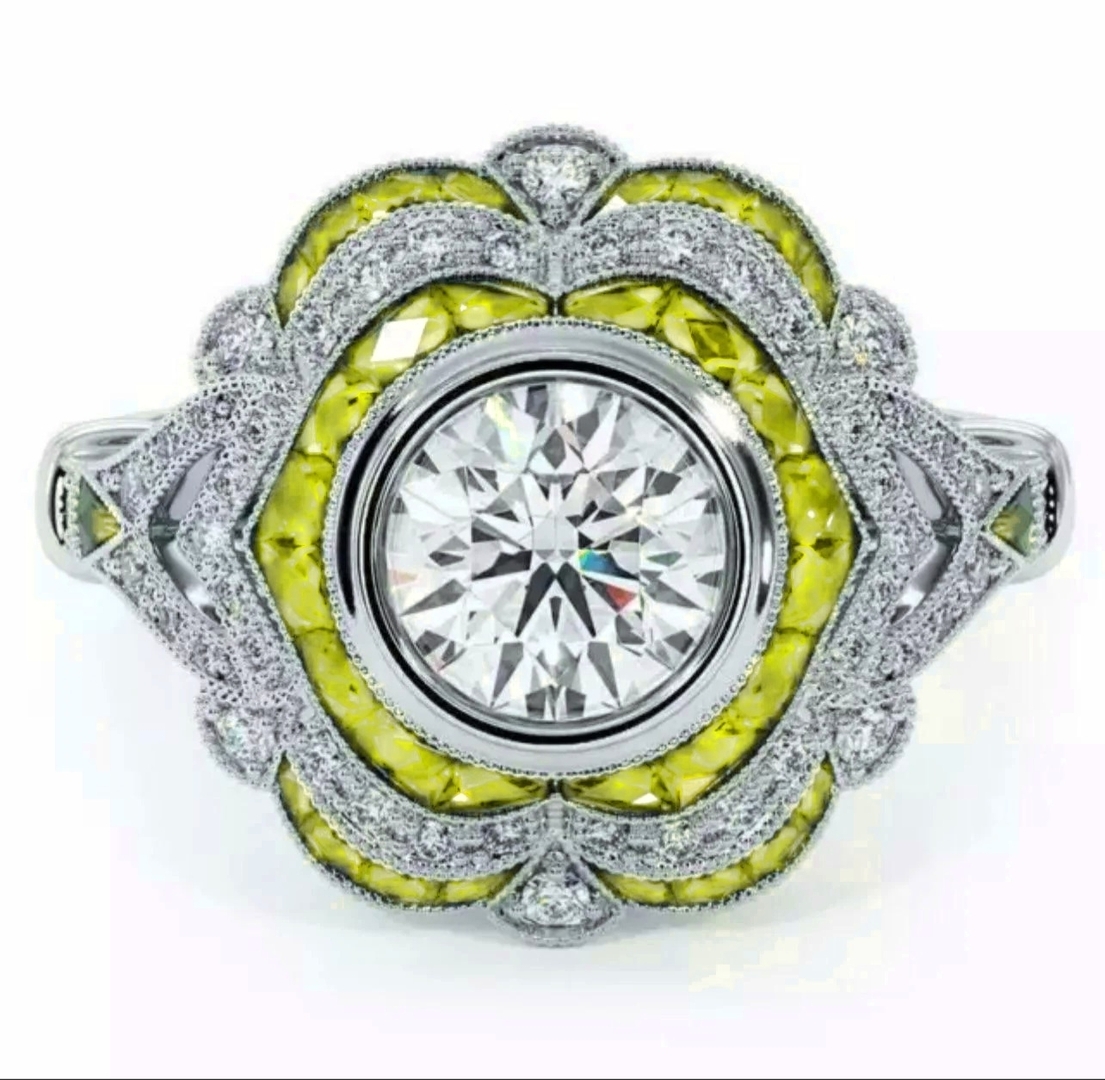}
      \end{minipage}\hfill
      \begin{minipage}[t]{0.27\textwidth}
        \centering
        \textbf{\Magic}\\[2mm]
        \includegraphics[width=\textwidth]{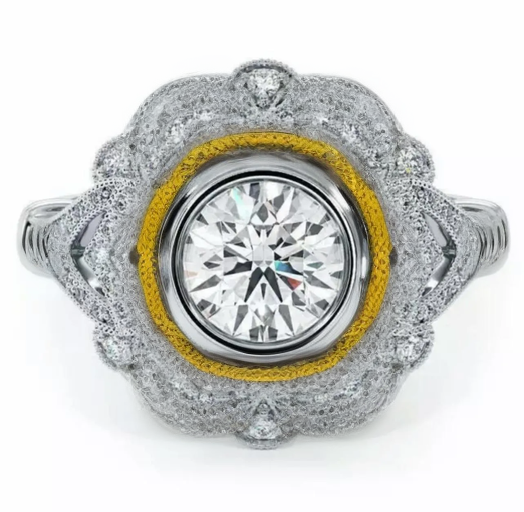}
      \end{minipage}\hfill
  \end{minipage}
  \caption{The model changes the structure of the ring even though the request says to only change the color. \textit{User request: Change the color of the green stones to pale yellow.}}
\end{figure*}







\clearpage
\section{Additional Samples}
\label{sec:supp-additional_samples}

\begin{figure*}[htbp]
  \centering
  \begin{minipage}[t]{0.18\textwidth}
    \centering
    \textbf{Original}\\[2mm]
    \includegraphics[width=\textwidth]{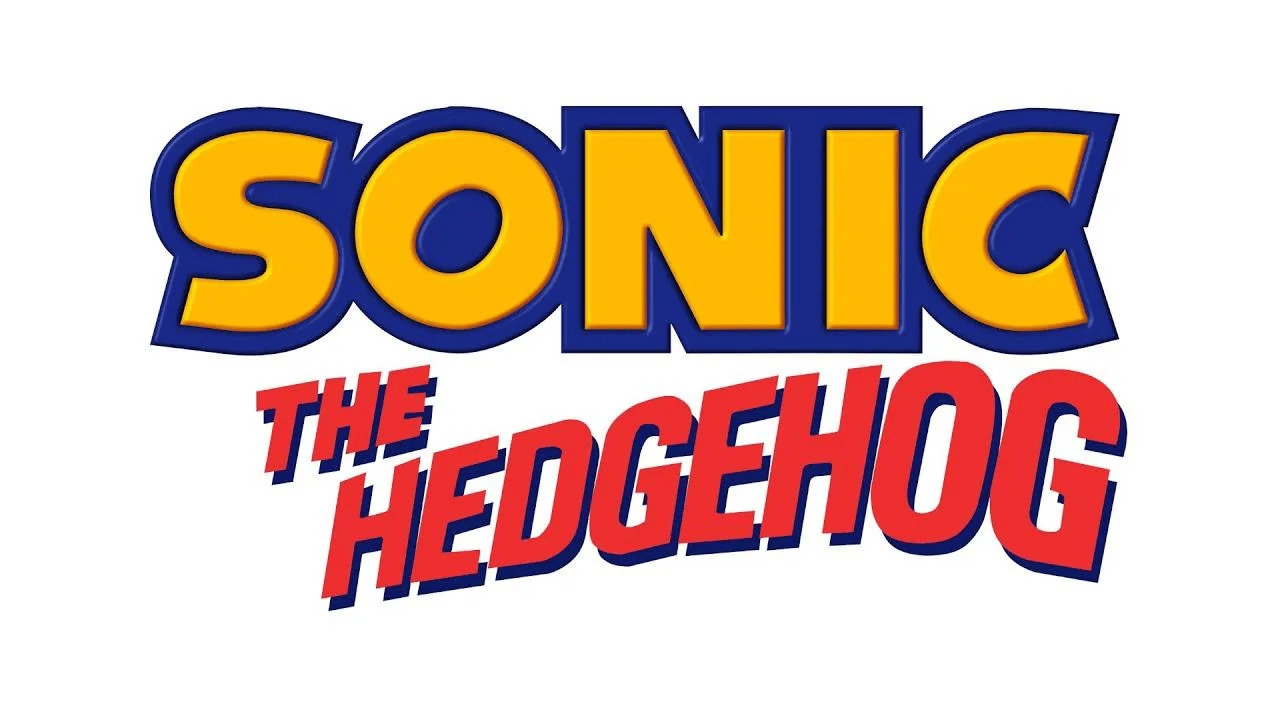}
  \end{minipage}\hfill
  \begin{minipage}[t]{0.22\textwidth}
    \centering
    \textbf{Human Edit}\\[2mm]
    \includegraphics[width=\textwidth]{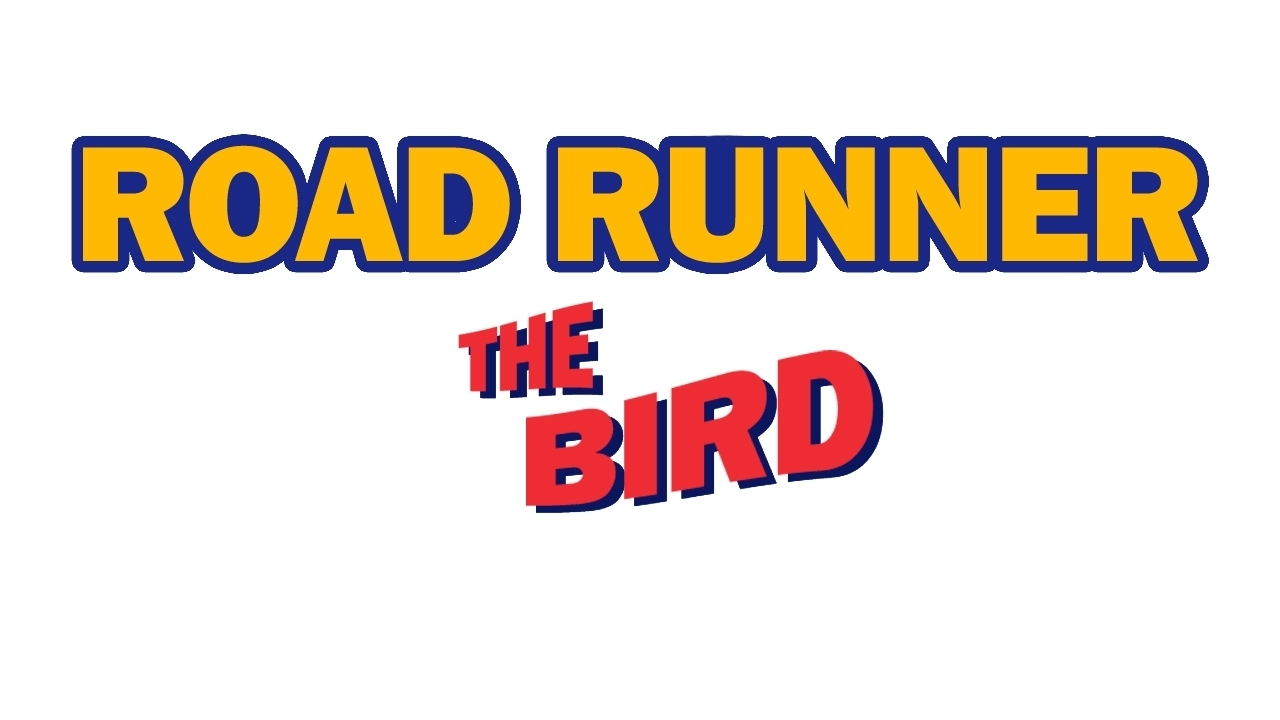}
  \end{minipage}\hfill
  \begin{minipage}[t]{0.18\textwidth}
    \centering
    \model{MagicQuill}\\[2mm]
    \includegraphics[width=\textwidth]{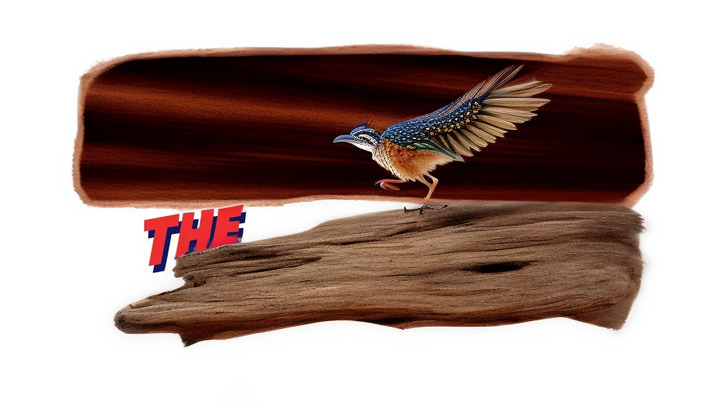}
  \end{minipage}\hfill
  \begin{minipage}[t]{0.15\textwidth}
    \centering
    \model{\Ledits}\\[2mm]
    \includegraphics[width=\textwidth]{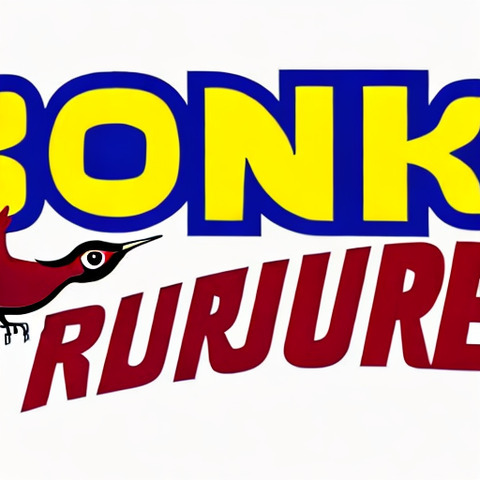}
  \end{minipage}\hfill
  \begin{minipage}[t]{0.23\textwidth}
    \centering
    \model{\geminiflash{}}\\[2mm]
    \includegraphics[width=\textwidth]{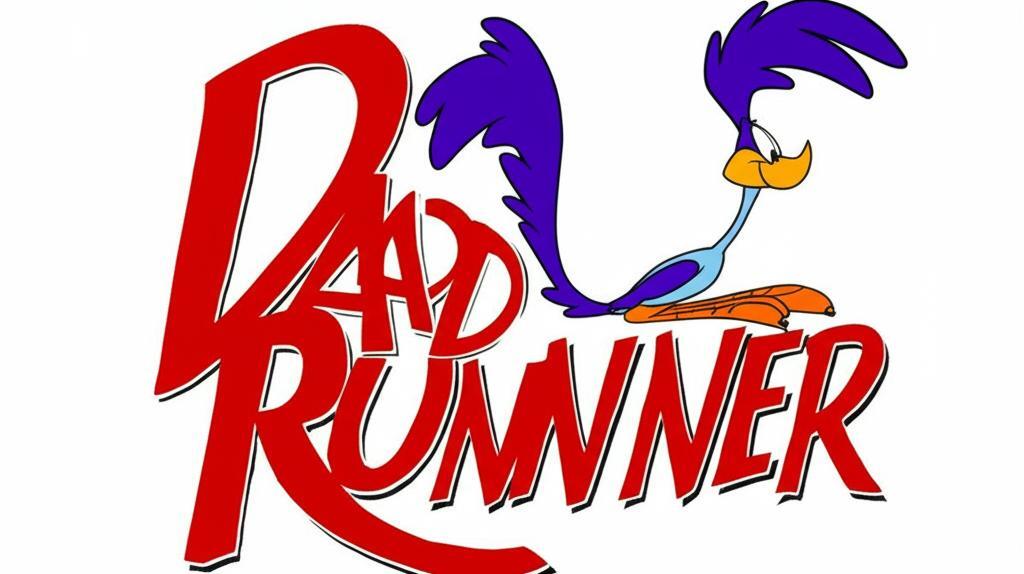}
  \end{minipage}\hfill

  \caption{Models fail at edits requiring text replacement. User request: \textit{Replace `SONIC' with `ROAD RUNNER' and `HEDGEHOG' with `BIRD'.}}
\end{figure*}



\begin{figure*}[htbp]
  \centering
  \begin{minipage}[t]{0.17\textwidth}
    \centering
    \textbf{Original}\\[2mm]
    \includegraphics[width=\textwidth]{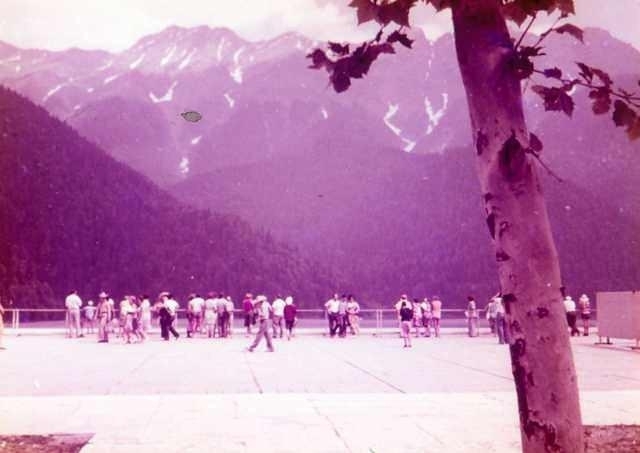}
  \end{minipage}\hfill
  \begin{minipage}[t]{0.17\textwidth}
    \centering
    \textbf{Human Edit}\\[2mm]
    \includegraphics[width=\textwidth]{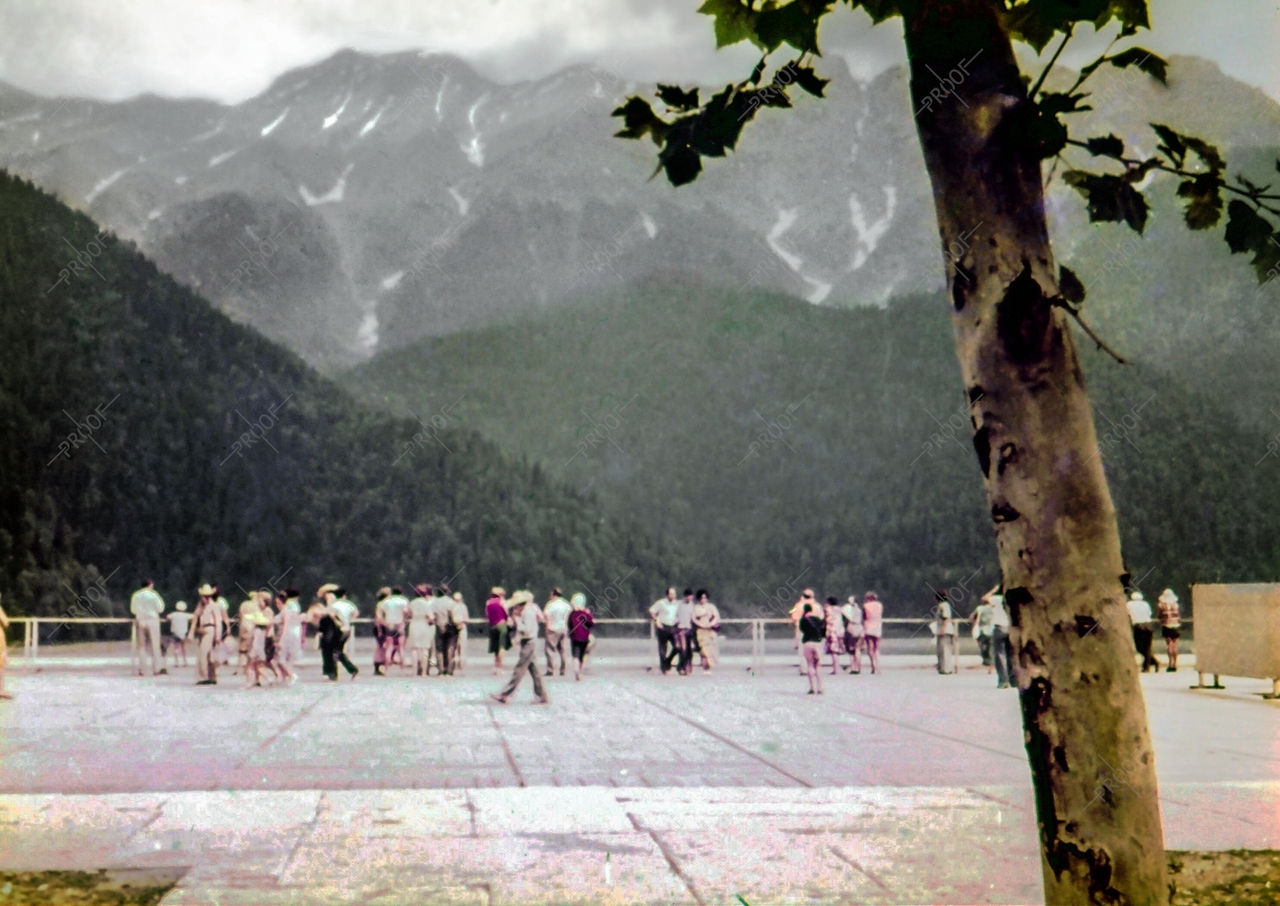}
  \end{minipage}\hfill
  \begin{minipage}[t]{0.20\textwidth}
    \centering
    {\tiny\scalebox{0.85}{\mbox{\model{Old Photo Restoration}}}}\\[1mm]
    \includegraphics[width=\textwidth]{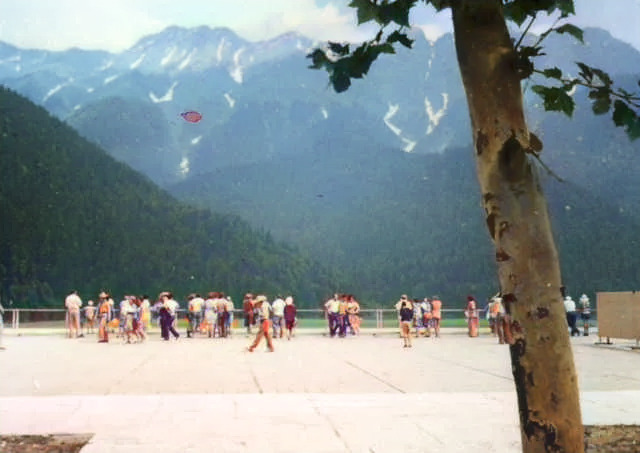}
  \end{minipage}\hfill
  \begin{minipage}[t]{0.14\textwidth}
    \centering
    {\tiny\scalebox{0.5}{\mbox{\model{\BriaRecolor}}}}\\[1mm]
    \includegraphics[width=\textwidth]{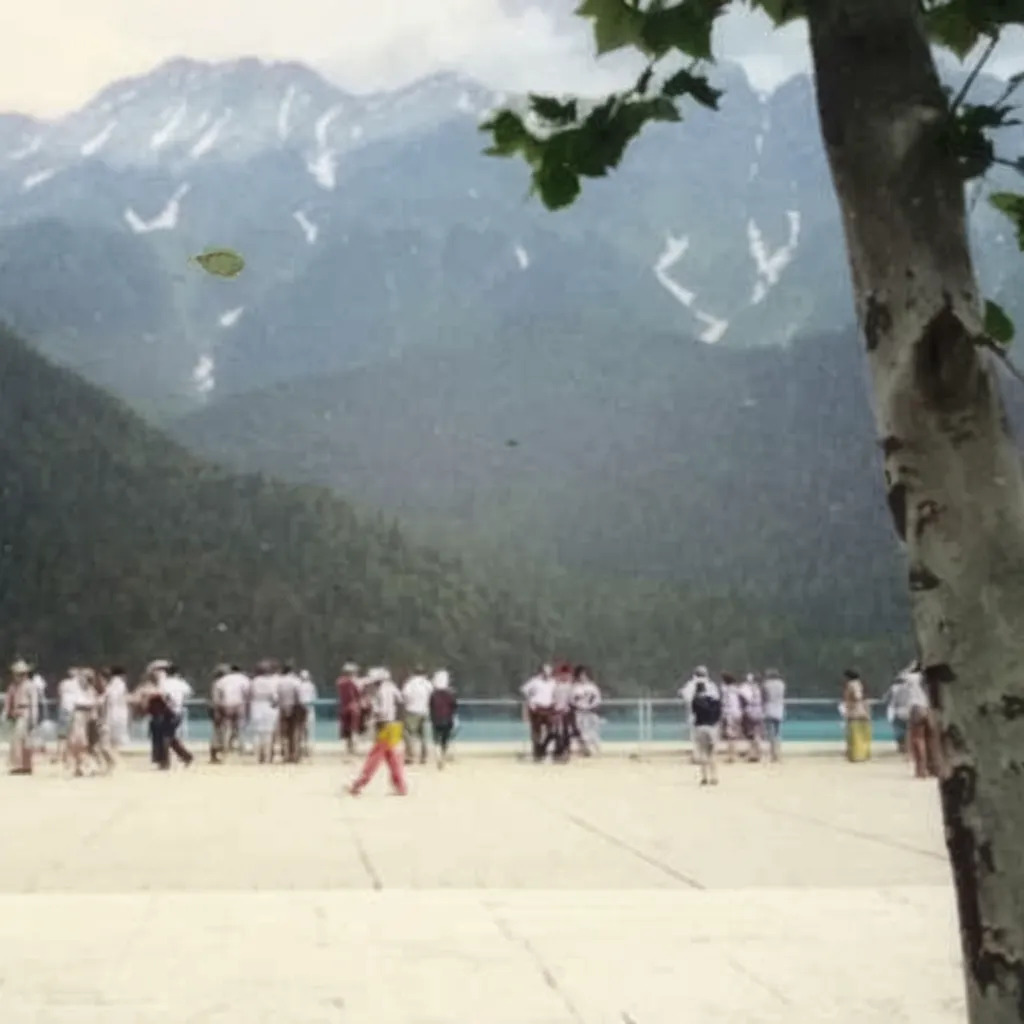}
  \end{minipage}\hfill
  \begin{minipage}[t]{0.17\textwidth}
    \centering
    \model{\gpt}\\[2mm]
    \includegraphics[width=\textwidth]{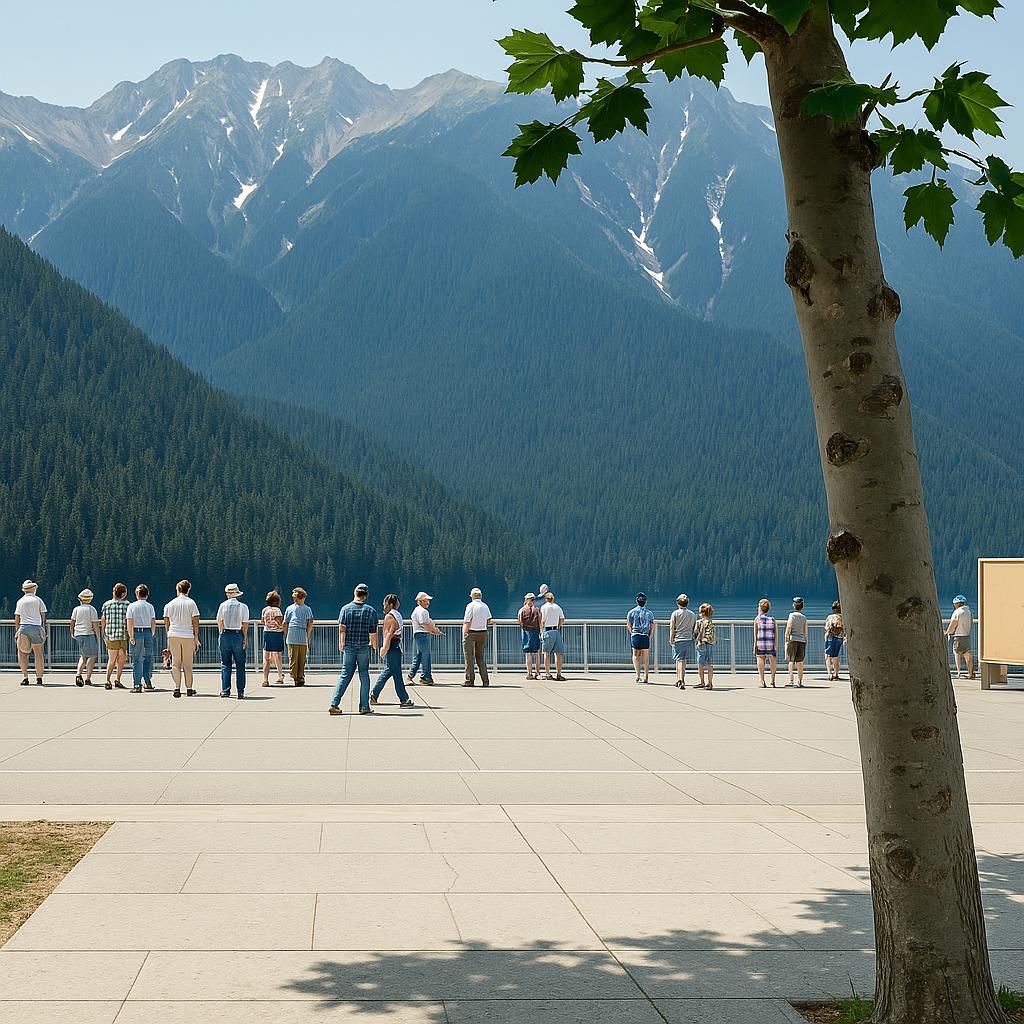}
  \end{minipage}\hfill  
  
  \caption{Models succeed at recoloring and restoring images. User request: \textit{Perform color correction and remove spots from the slide scan.}}
\end{figure*}

\begin{figure*}[htbp]
  \centering
  \begin{minipage}[t]{0.15\textwidth}
    \centering
    \textbf{Original}\\[2mm]
    \includegraphics[width=\textwidth]{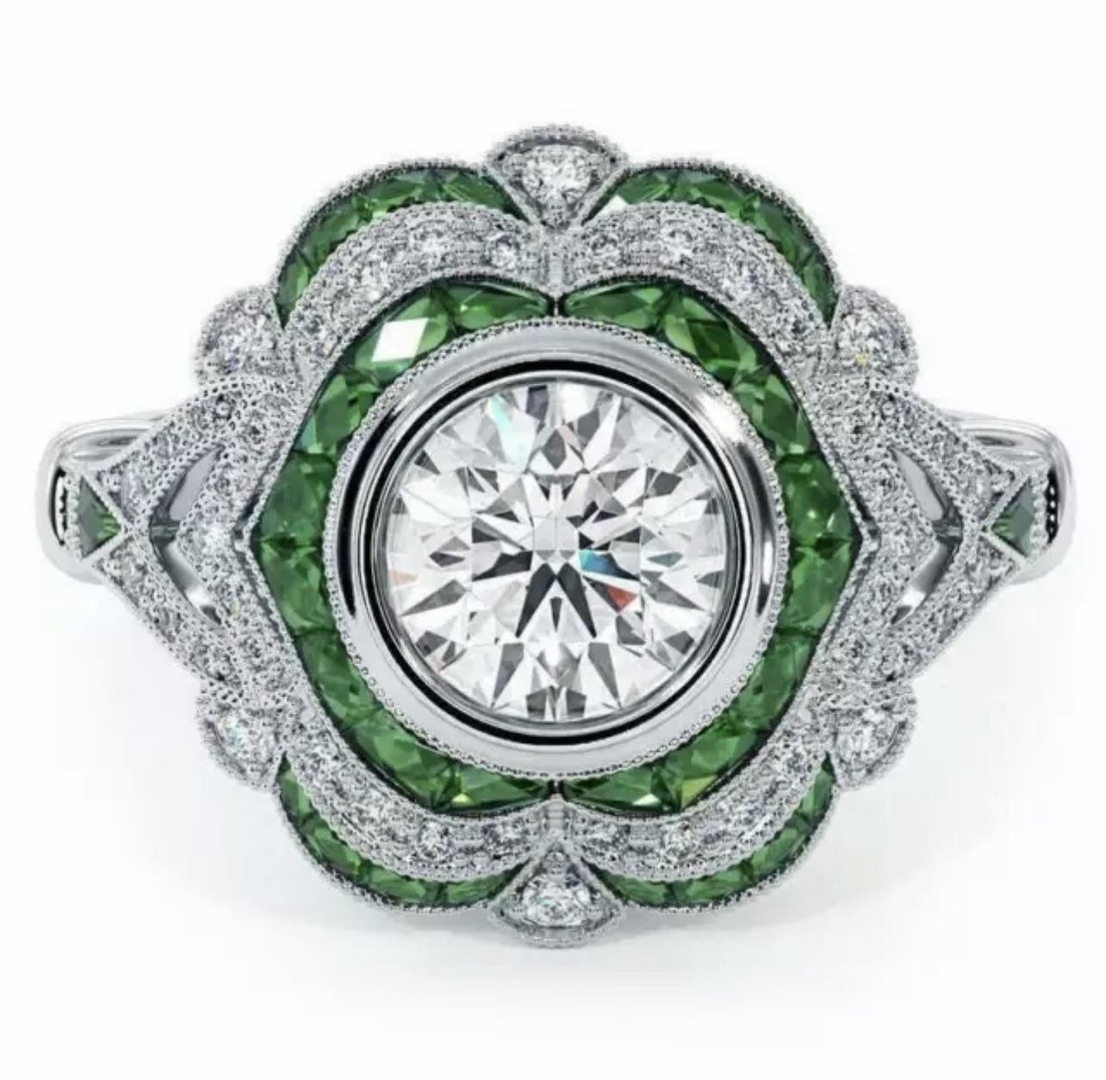}
  \end{minipage}\hfill
  \begin{minipage}[t]{0.15\textwidth}
    \centering
    \textbf{Human Edit}\\[2mm]
    \includegraphics[width=\textwidth]{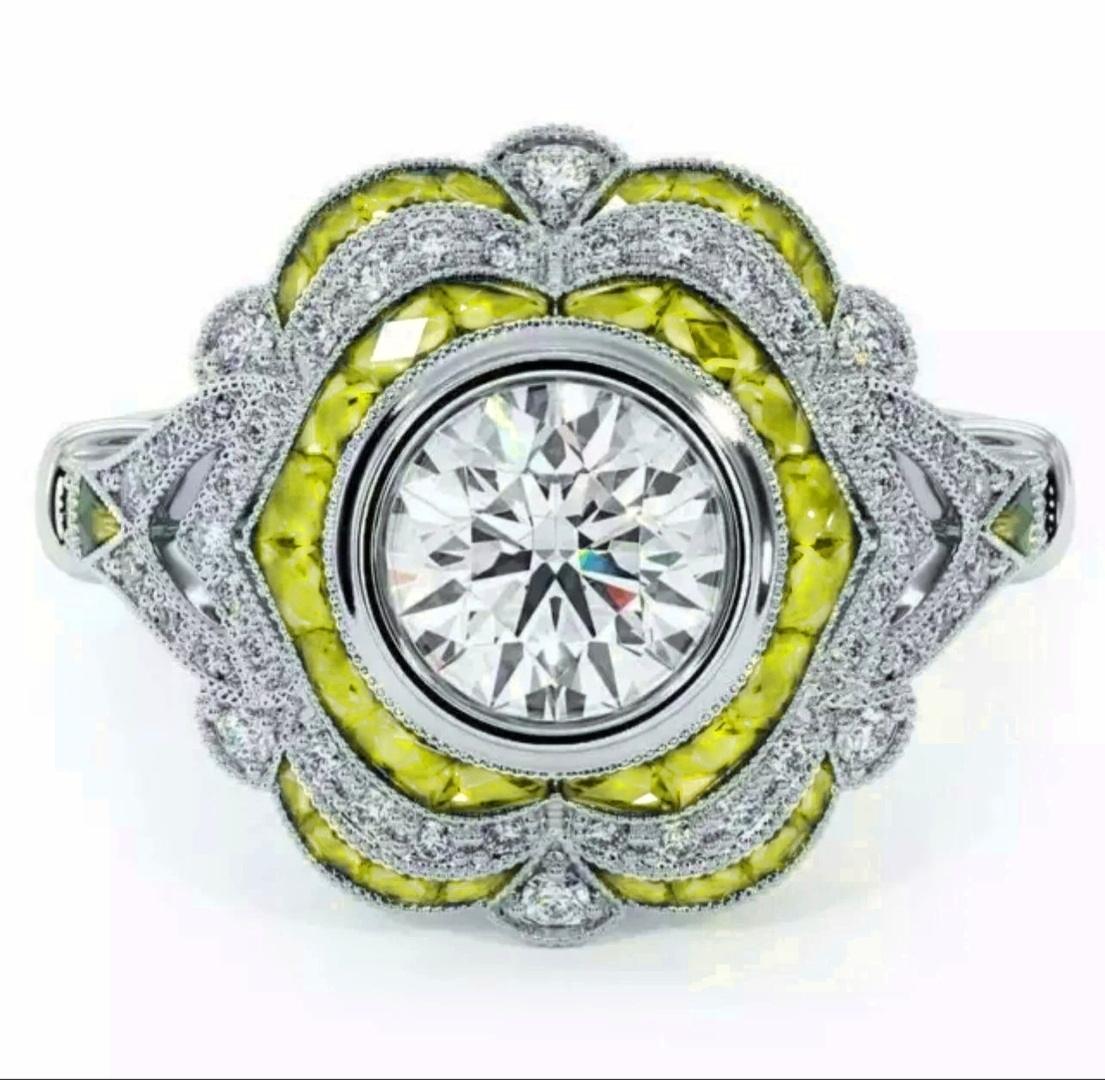}
  \end{minipage}\hfill
  \begin{minipage}[t]{0.15\textwidth}
    \centering
    \model{\SeedEdit}\\[2mm]
    \includegraphics[width=\textwidth]{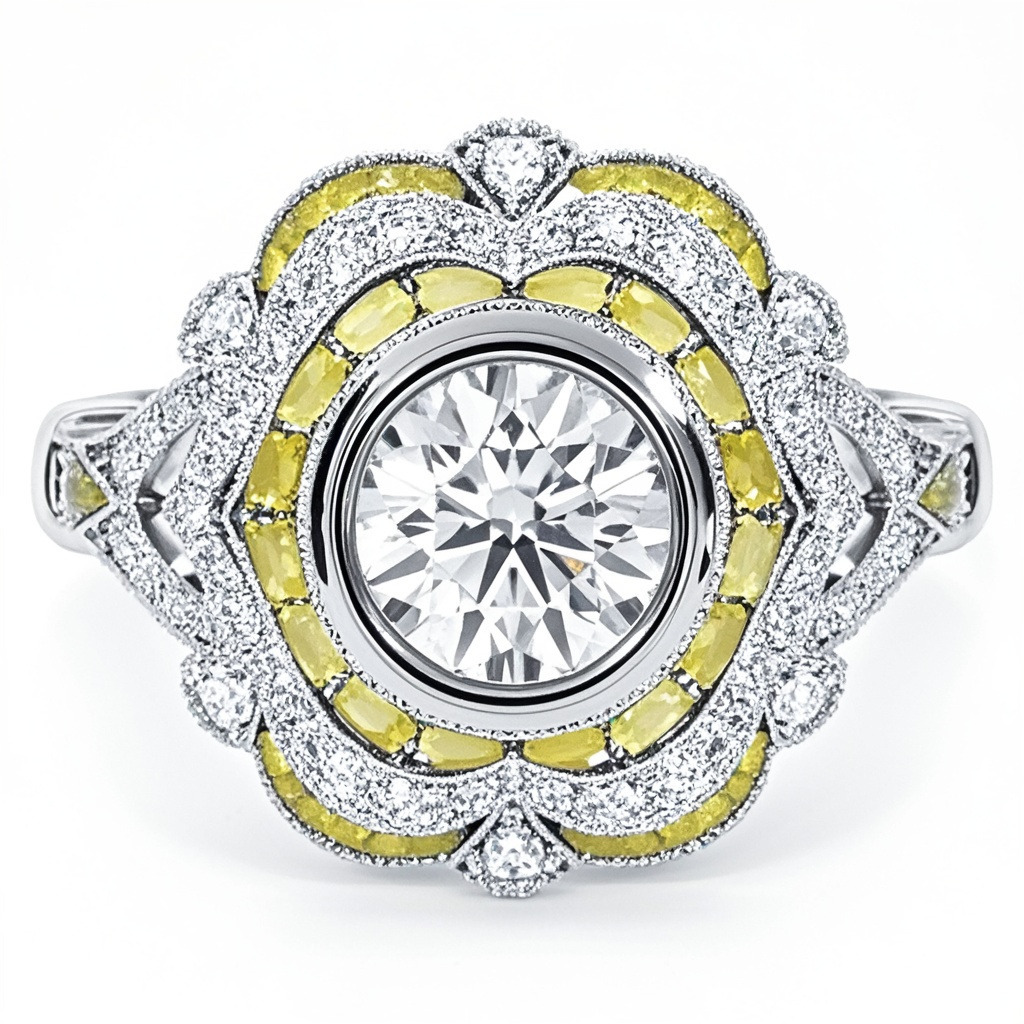}
  \end{minipage}\hfill
  \begin{minipage}[t]{0.15\textwidth}
    \centering
    \model{\InstructPix}\\[2mm]
    \includegraphics[width=\textwidth]{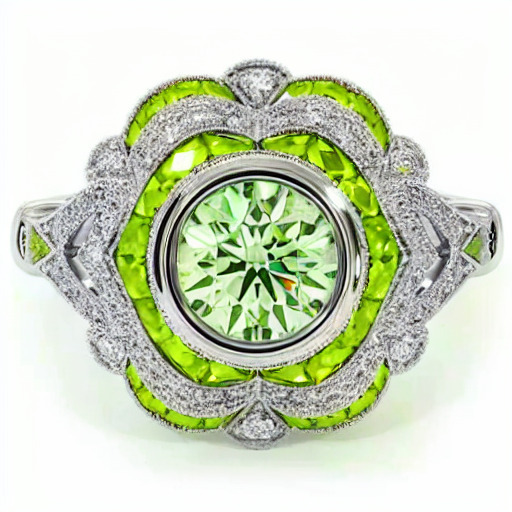}
  \end{minipage}\hfill
  \begin{minipage}[t]{0.15\textwidth}
    \centering
    {\tiny\scalebox{0.85}{\mbox{\model{\geminiflash{}}}}}\\[1mm]
    \includegraphics[width=\textwidth]{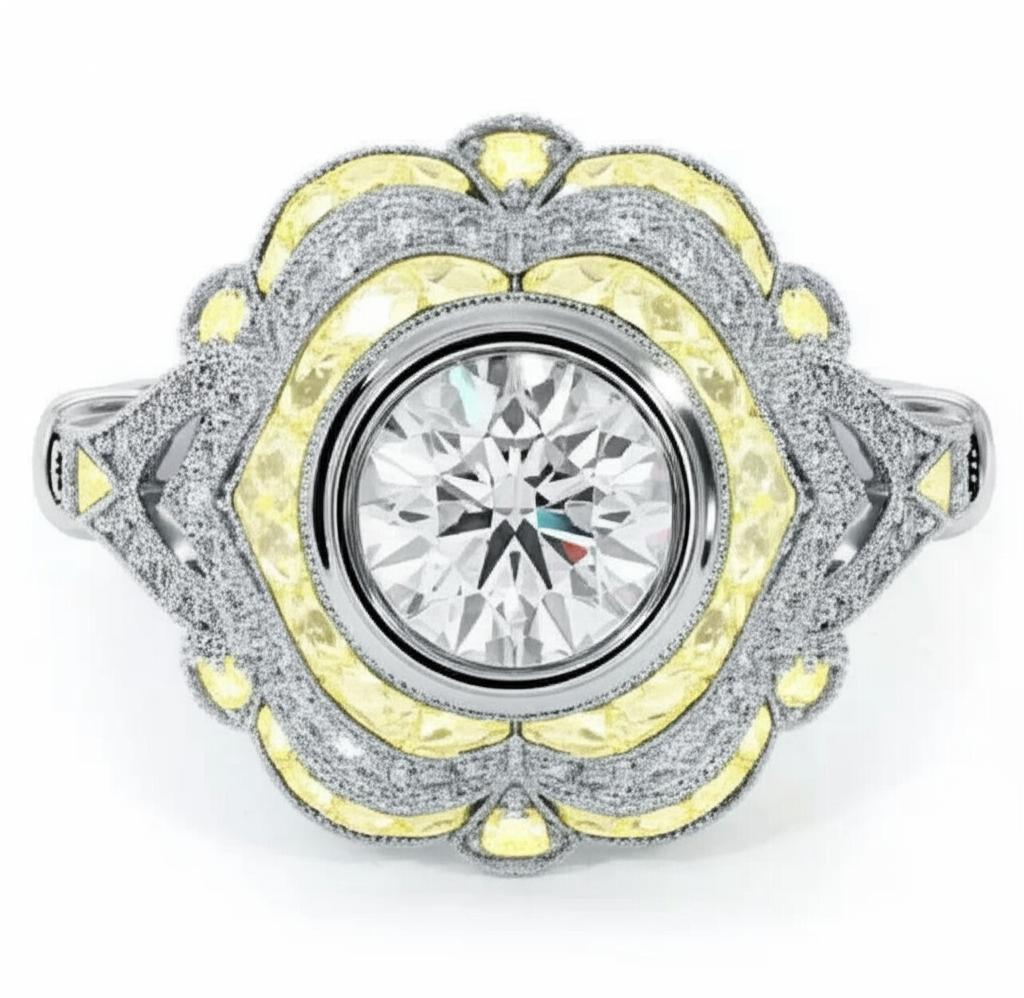}
  \end{minipage}\hfill
  \begin{minipage}[t]{0.15\textwidth}
    \centering
    \model{\gpt}\\[2mm]
    \includegraphics[width=\textwidth]{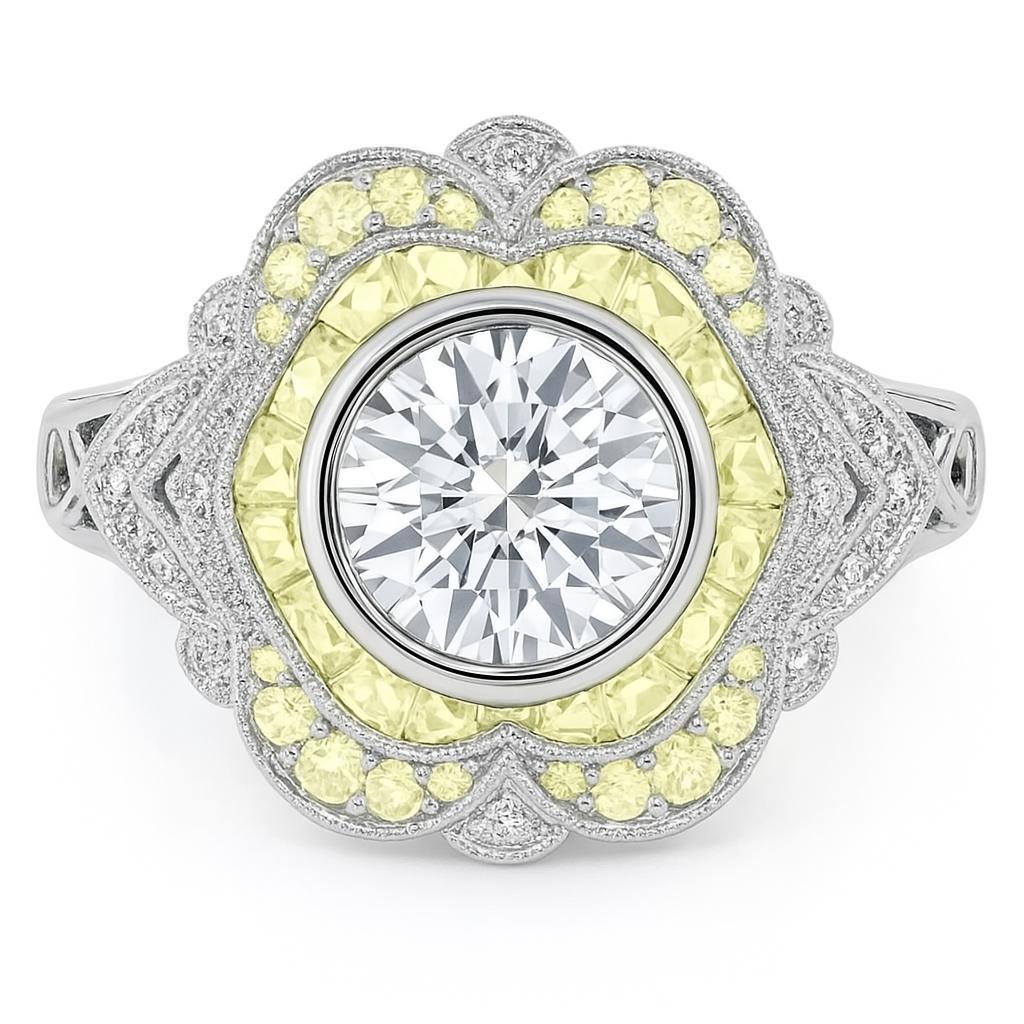}
  \end{minipage}\hfill
  
  \caption{Models can recolor specific parts of images. User request: \textit{Change the color of the green stones to pale yellow.}}
\end{figure*}

\begin{figure*}[htbp]
  \centering
  \begin{minipage}[t]{0.14\textwidth}
    \centering
    \textbf{Original}\\[2mm]
    \includegraphics[width=\textwidth]{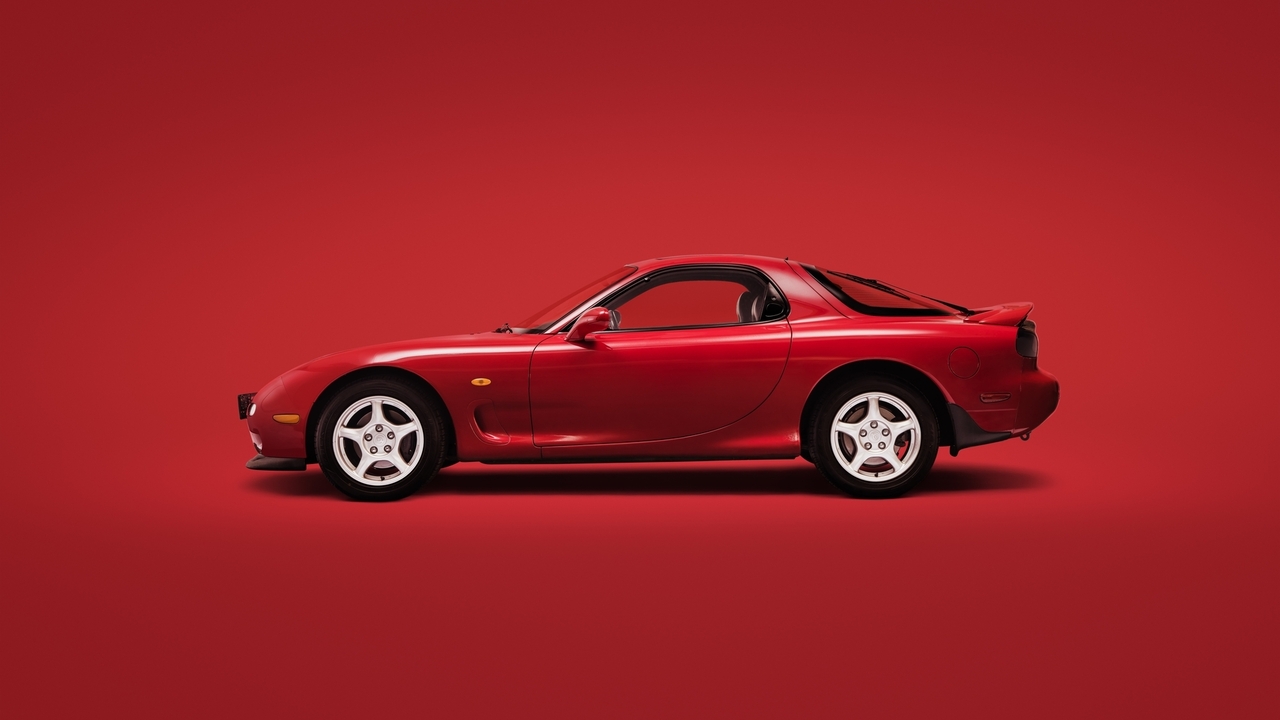}
  \end{minipage}\hfill
  \begin{minipage}[t]{0.14\textwidth}
    \centering
    \textbf{Human Edit}\\[2mm]
    \includegraphics[width=\textwidth]{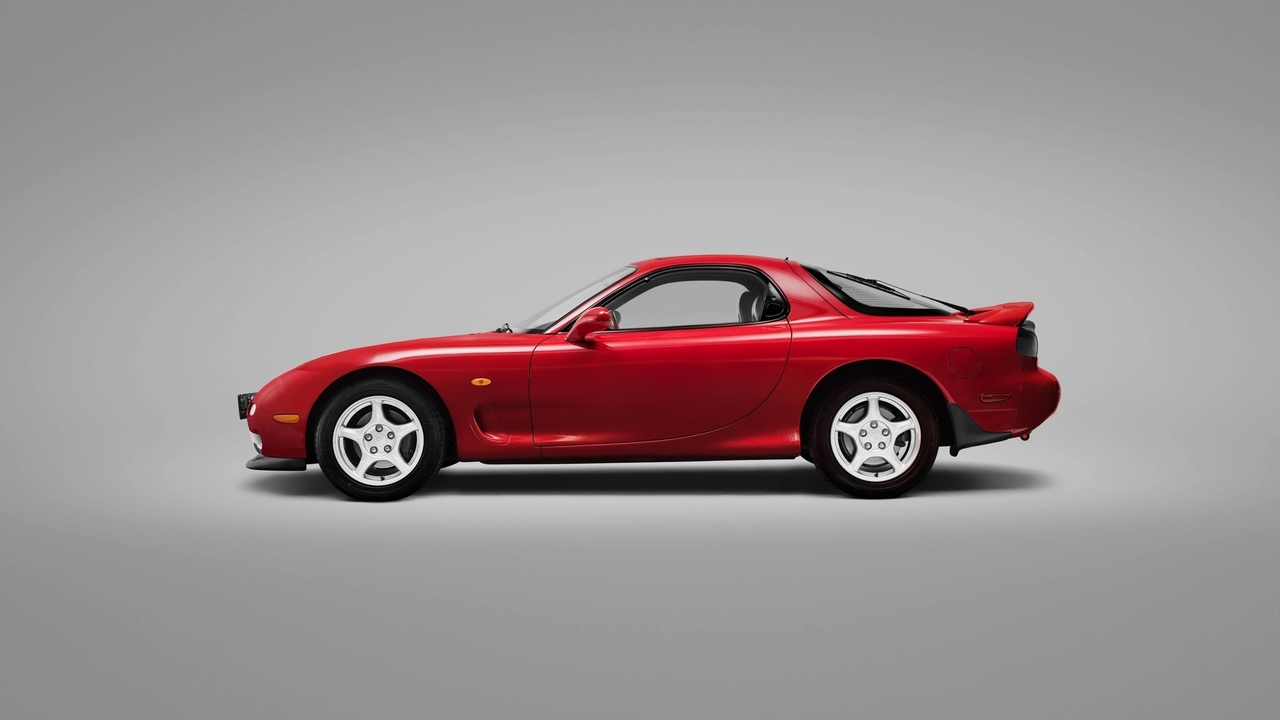}
  \end{minipage}\hfill
  \begin{minipage}[t]{0.17\textwidth}
    \centering
    \model{\ReplaceAny}\\[2mm]
    \includegraphics[width=\textwidth]{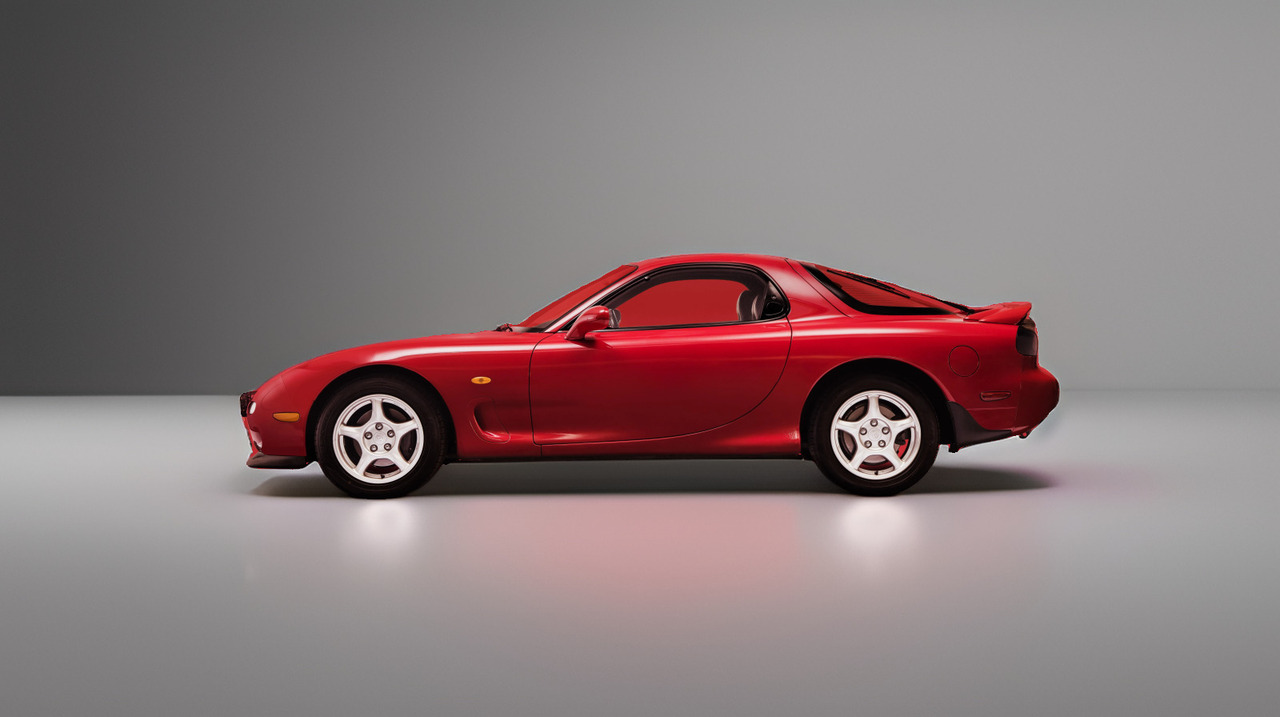}
  \end{minipage}\hfill
  \begin{minipage}[t]{0.16\textwidth}
    \centering
    \model{\InstructPix}\\[2mm]
    \includegraphics[width=\textwidth]{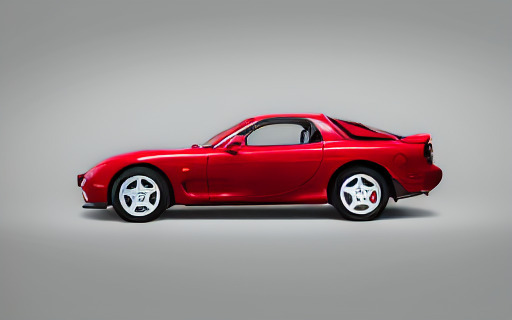}
  \end{minipage}\hfill
  \begin{minipage}[t]{0.15\textwidth}
    \centering
    {\tiny\scalebox{0.85}{\mbox{\model{\geminiflash{}}}}}\\[2mm]
    \includegraphics[width=\textwidth]{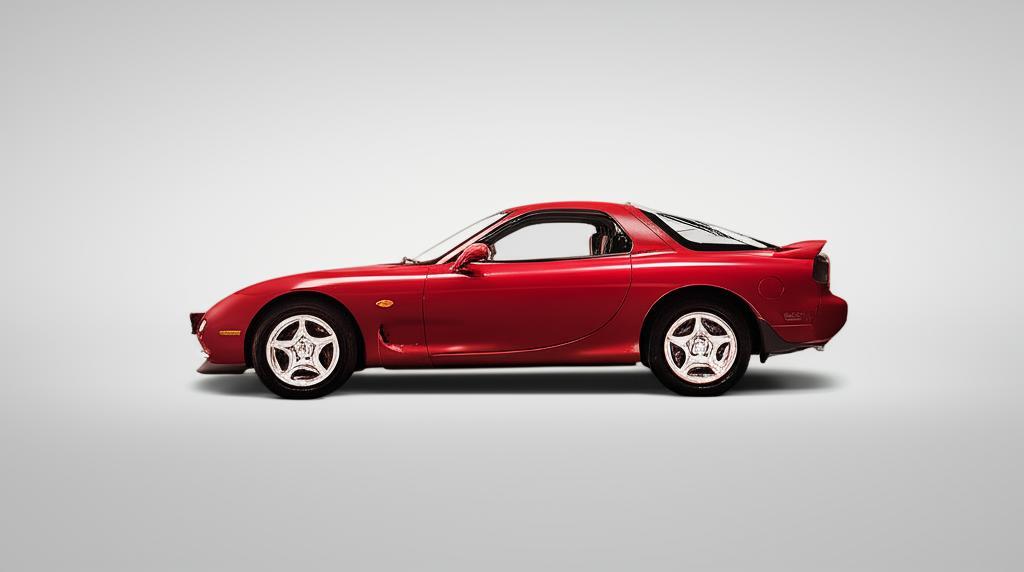}
  \end{minipage}\hfill
  \begin{minipage}[t]{0.15\textwidth}
    \centering
    \model{\gpt}\\[2mm]
    \includegraphics[width=\textwidth]{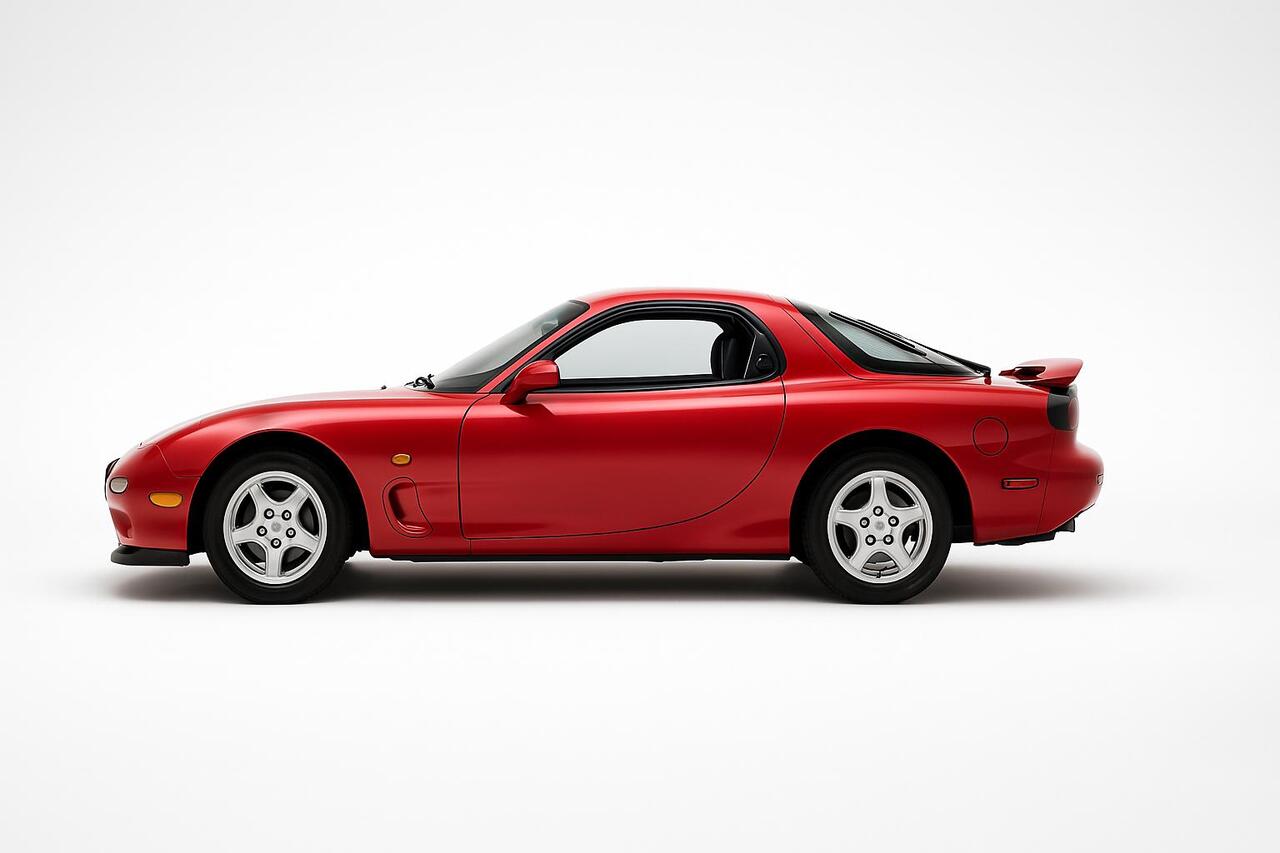}
  \end{minipage}\hfill

  \caption{Models can usually change the background properly. User request: \textit{Change the background to white or light grey to simulate a studio setting.}}
\end{figure*}

\begin{figure*}[htbp]
  \centering
  \begin{minipage}[t]{0.16\textwidth}
    \centering
    \textbf{Original}\\[1mm]
    \includegraphics[width=\textwidth]{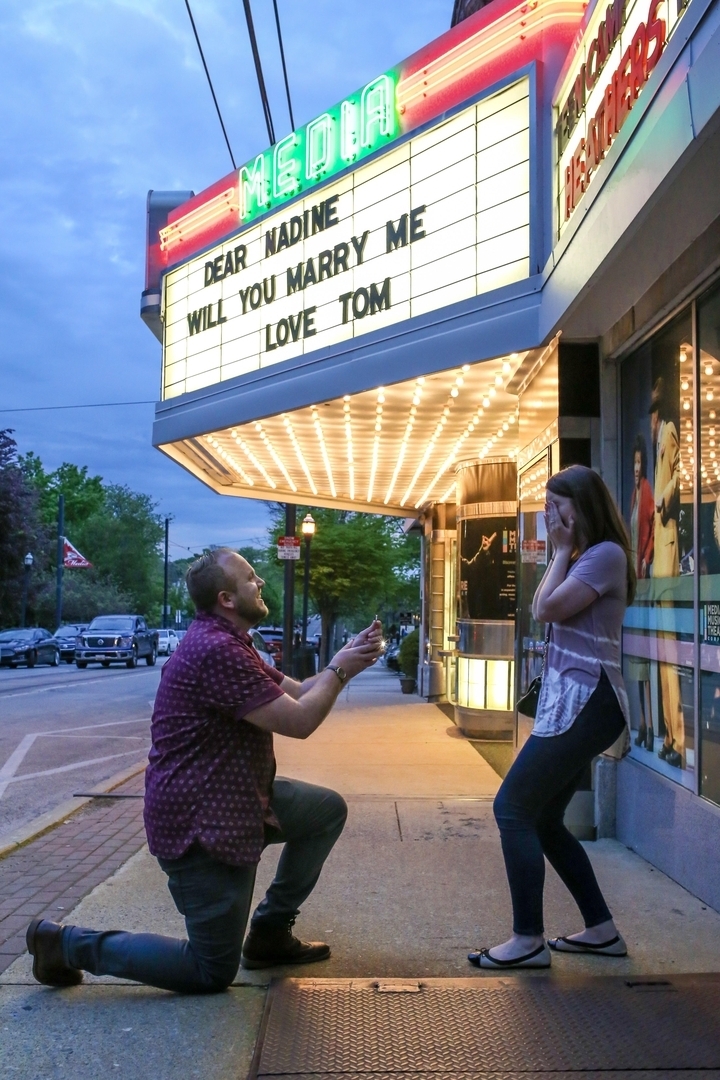}
  \end{minipage}\hfill
  \begin{minipage}[t]{0.16\textwidth}
    \centering
    \textbf{Human Edit}\\[2mm]
    \includegraphics[width=\textwidth]{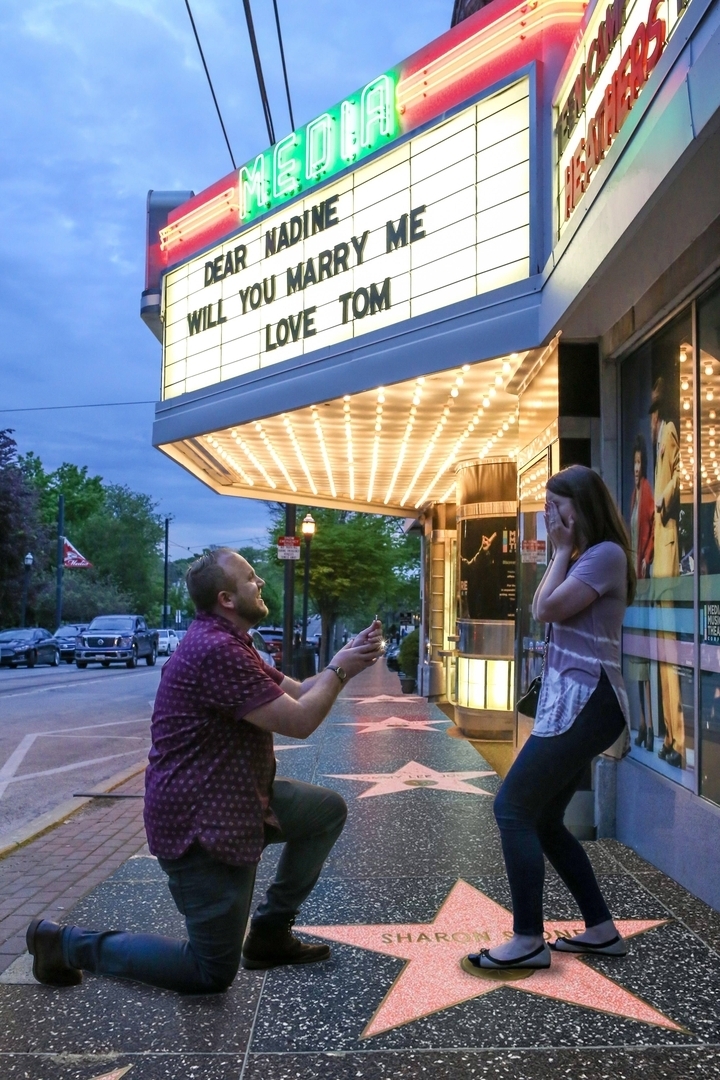}
  \end{minipage}\hfill
  \begin{minipage}[t]{0.16\textwidth}
    \centering
    \model{\SeedEdit}\\[2mm]
    \includegraphics[width=\textwidth]{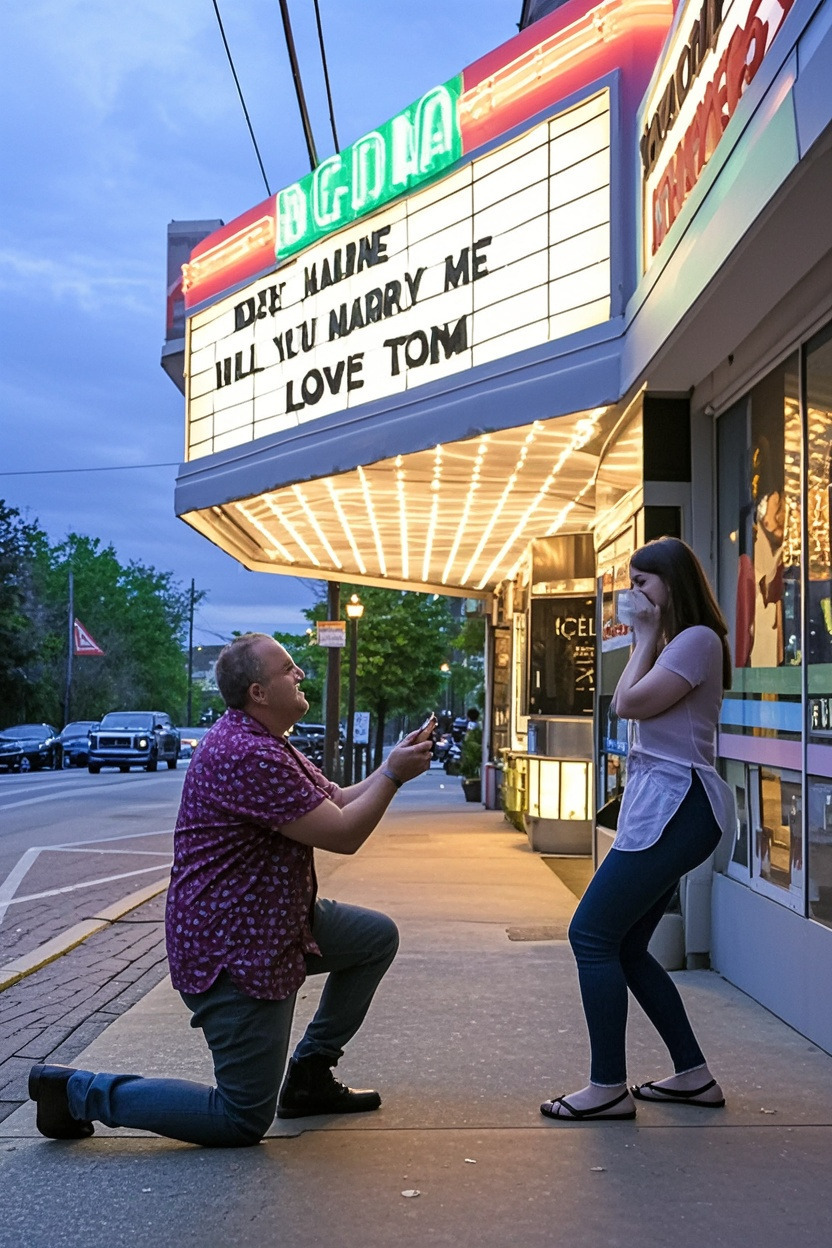}
  \end{minipage}\hfill
  \begin{minipage}[t]{0.16\textwidth}
    \centering
    \model{\Magic}\\[2mm]
    \includegraphics[width=\textwidth]{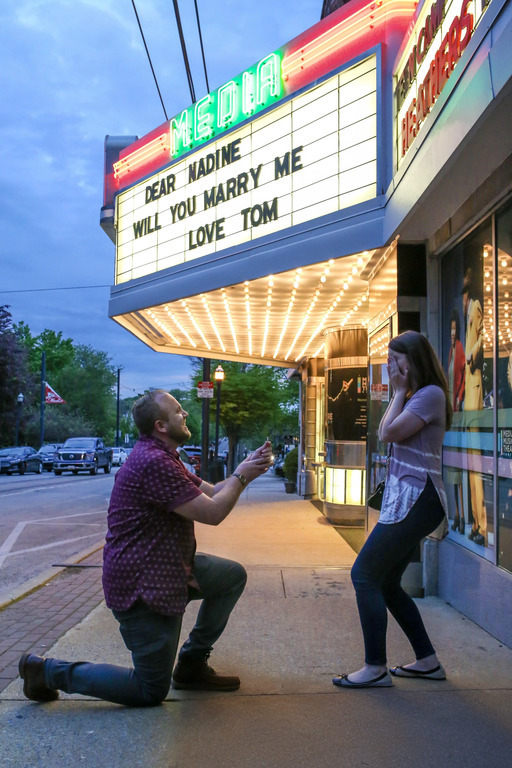}
  \end{minipage}\hfill
  \begin{minipage}[t]{0.16\textwidth}
    \centering
    {\tiny\scalebox{0.85}{\mbox{\model{\geminiflash{}}}}}\\[2mm]
    \includegraphics[width=\textwidth]{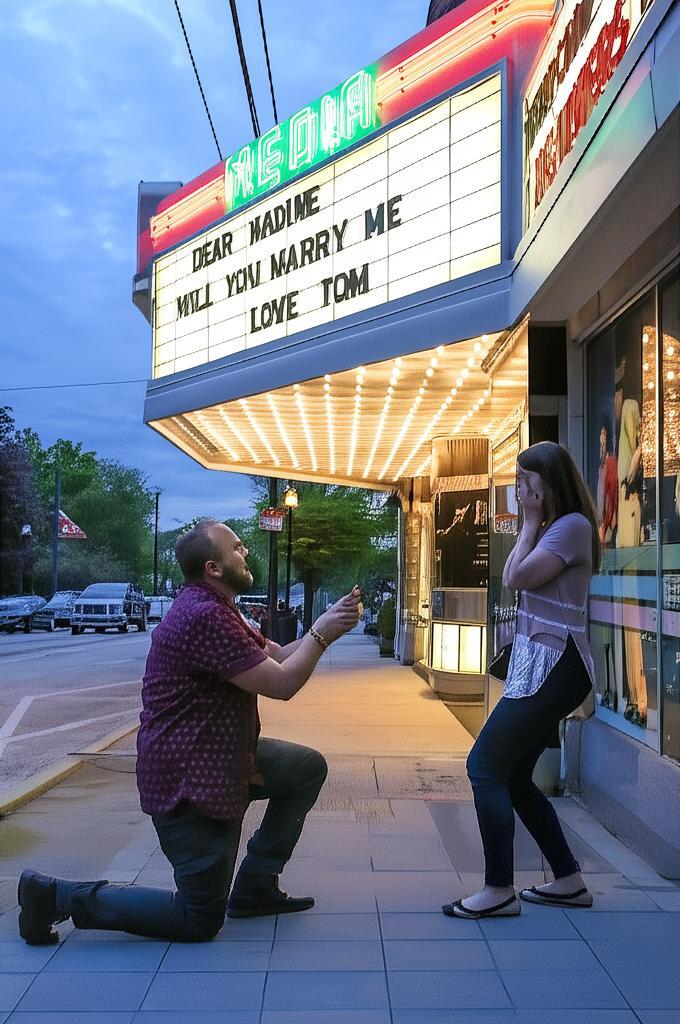}
  \end{minipage}\hfill
  \begin{minipage}[t]{0.16\textwidth}
    \centering
    \model{\gpt}\\[2mm]
    \includegraphics[width=\textwidth]{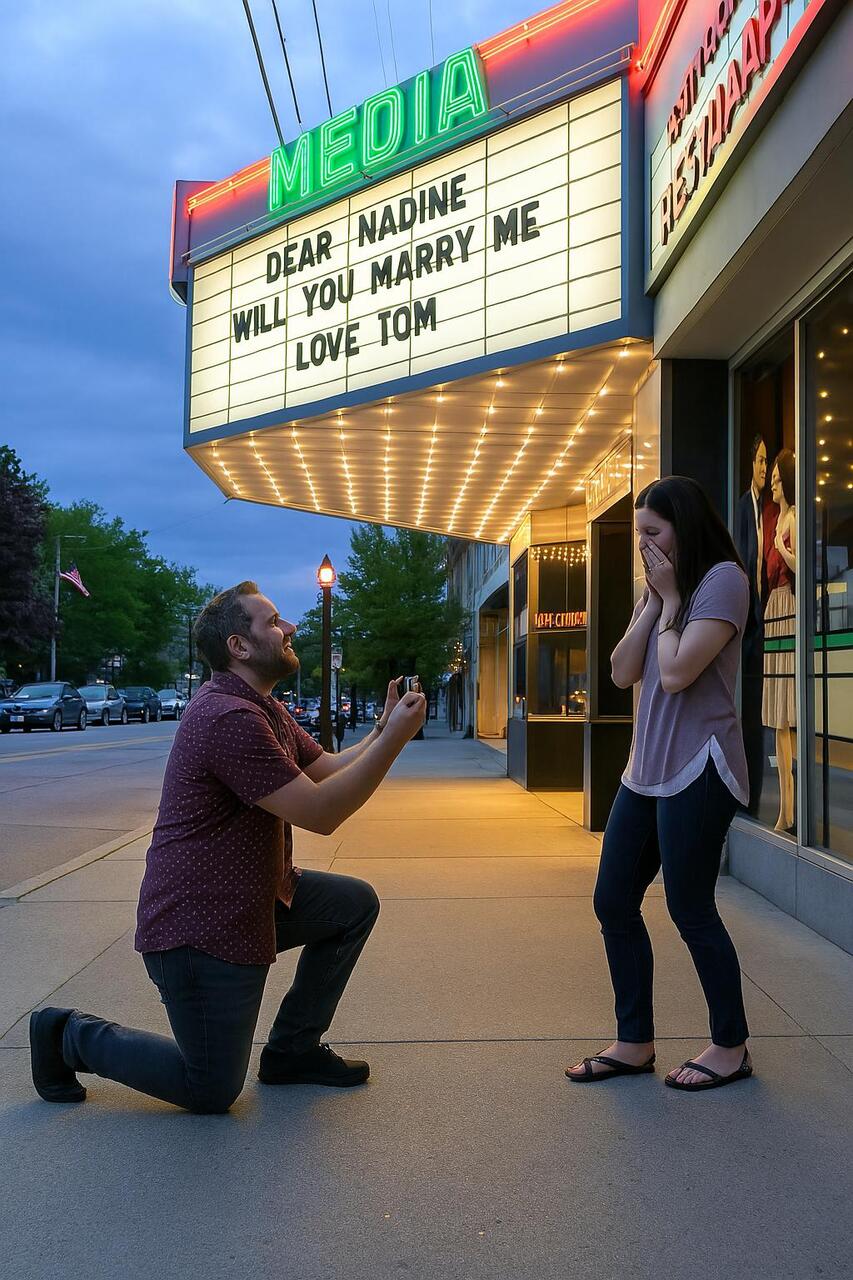}
  \end{minipage}\hfill

  \caption{Models can handle requests to replace objects. User request: \textit{Replace the metal crate at the bottom of the image with additional sidewalk.}}
\end{figure*}

\begin{figure*}[htbp]
  \centering
  \begin{minipage}[t]{0.18\textwidth}
    \centering
    \textbf{Original}\\[2mm]
    \includegraphics[width=\textwidth]{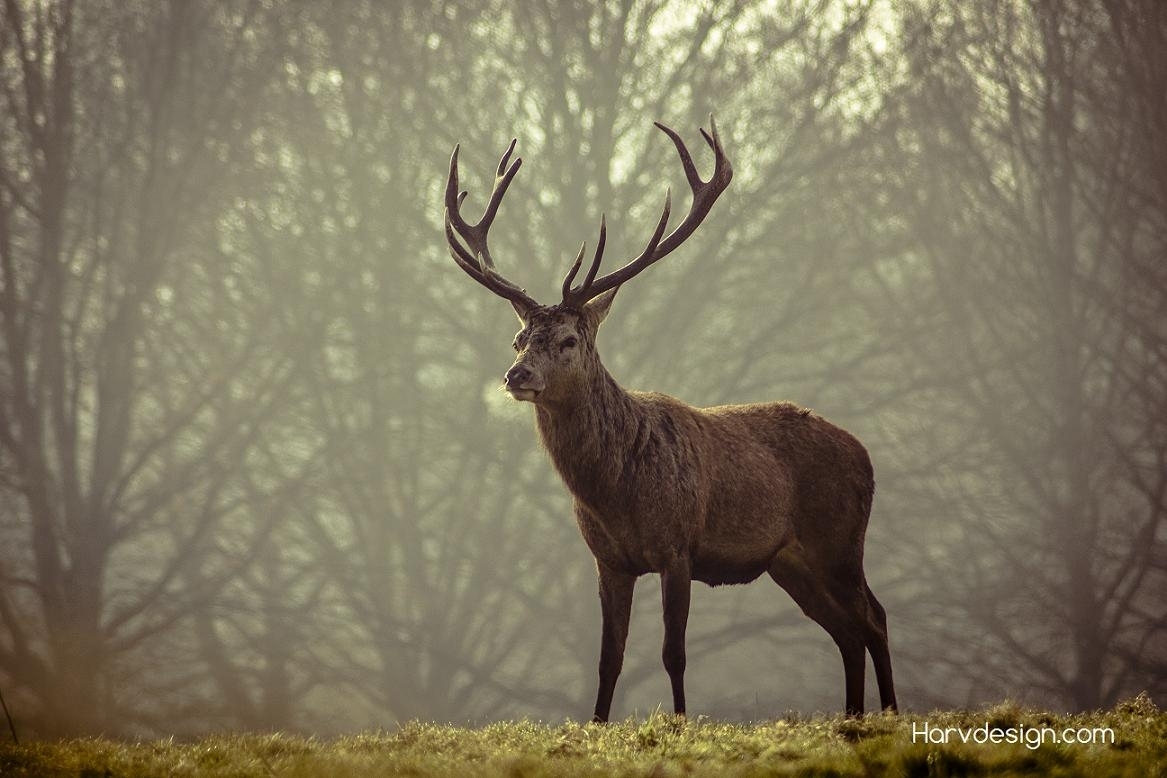}
  \end{minipage}\hfill
  \begin{minipage}[t]{0.18\textwidth}
    \centering
    \textbf{Human Edit}\\[2mm]
    \includegraphics[width=\textwidth]{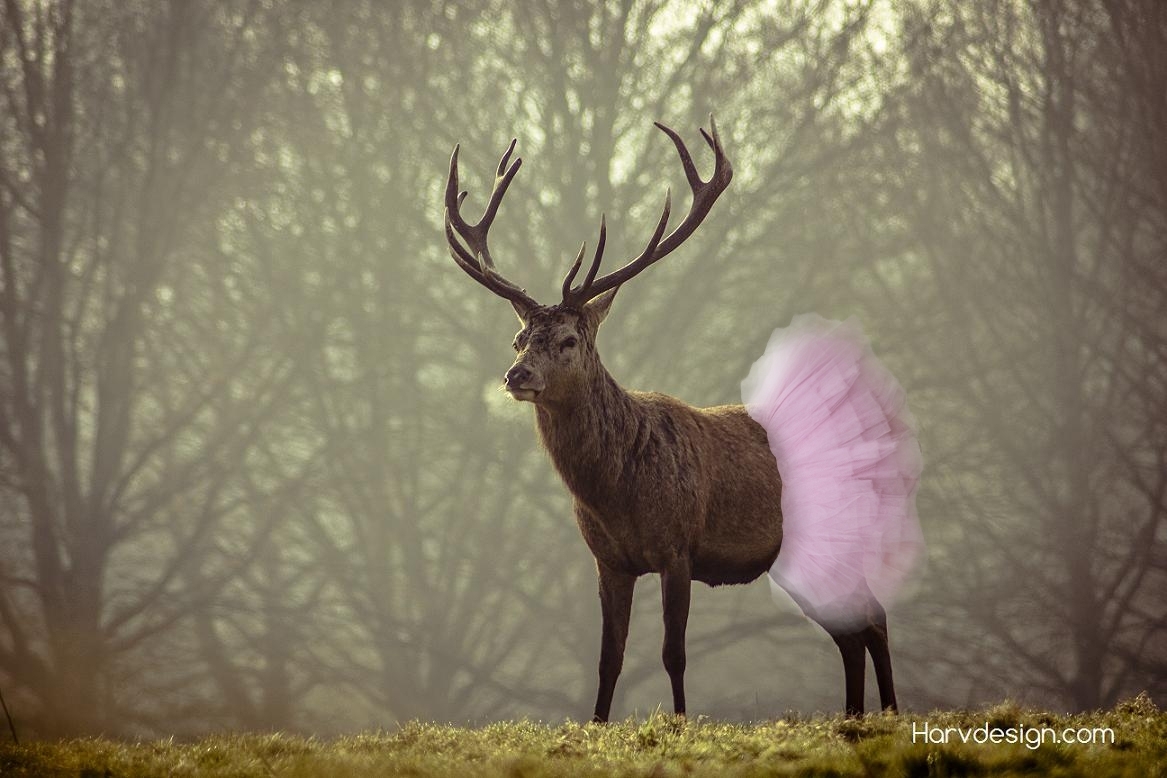}
  \end{minipage}\hfill
  \begin{minipage}[t]{0.18\textwidth}
    \centering
    \model{\SeedEdit}\\[2mm]
    \includegraphics[width=\textwidth]{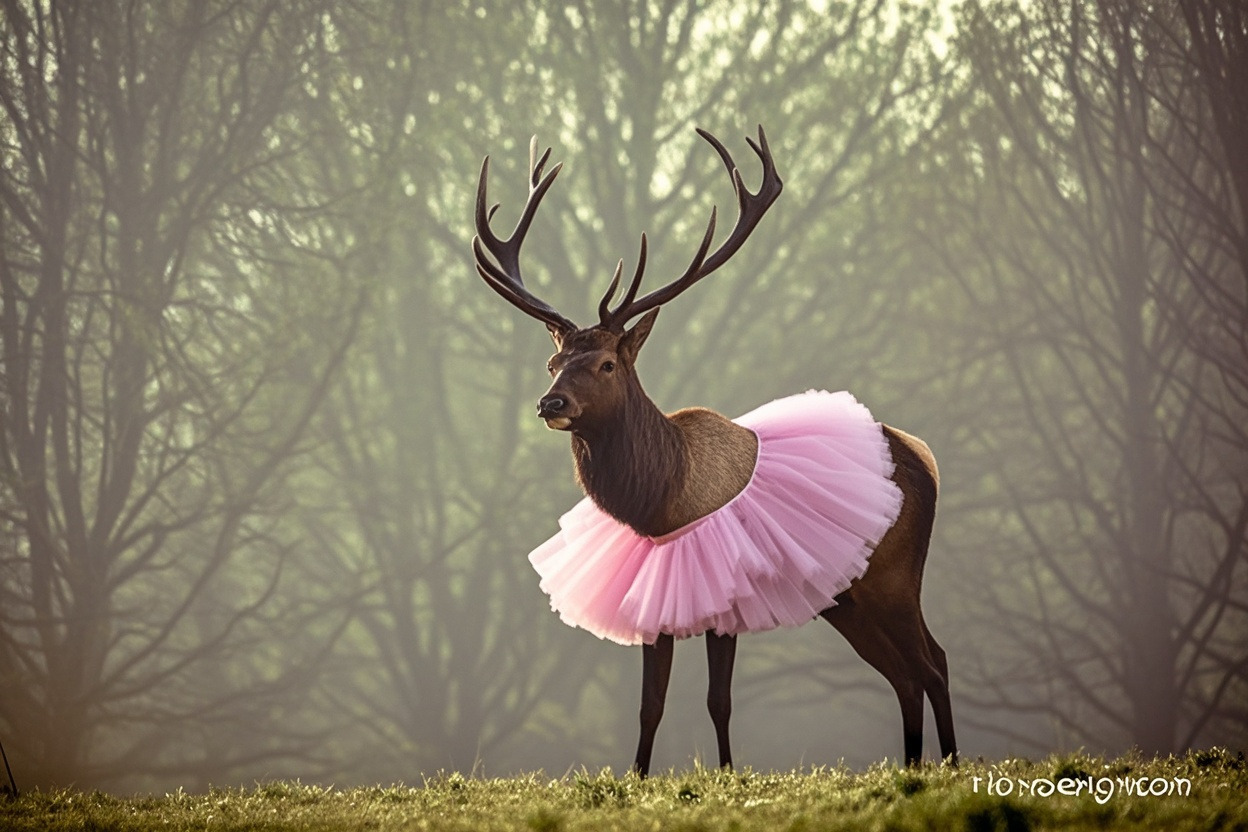}
  \end{minipage}\hfill
  \begin{minipage}[t]{0.18\textwidth}
    \centering
    \model{\CosXL}\\[2mm]
    \includegraphics[width=\textwidth]{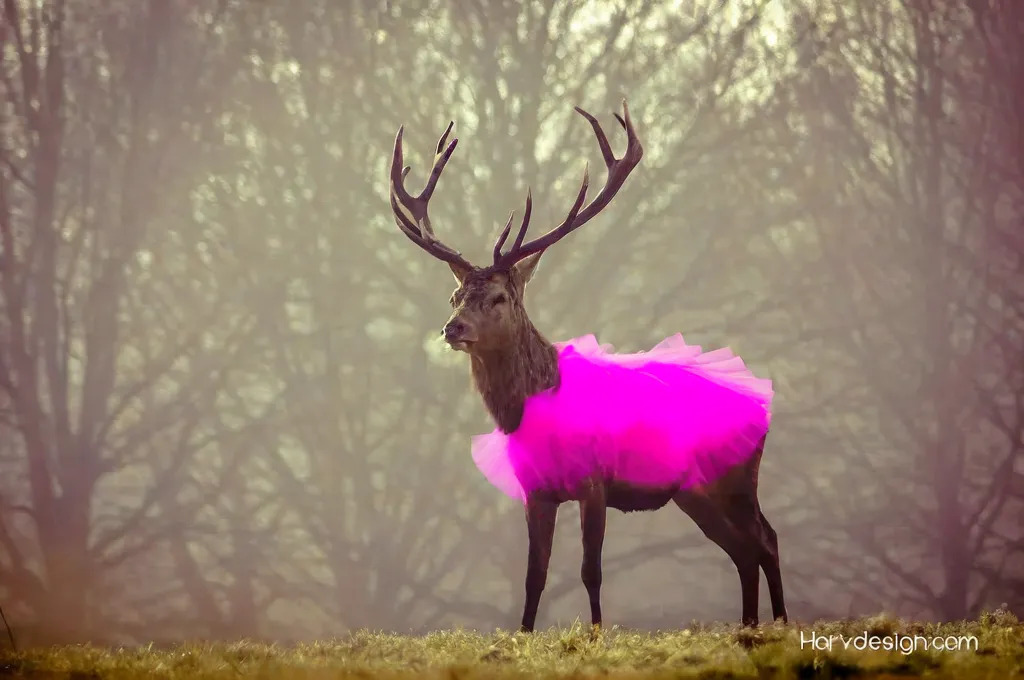}
  \end{minipage}\hfill
  \begin{minipage}[t]{0.18\textwidth}
    \centering
    \model{\geminiflash}\\[2mm]
    \includegraphics[width=\textwidth]{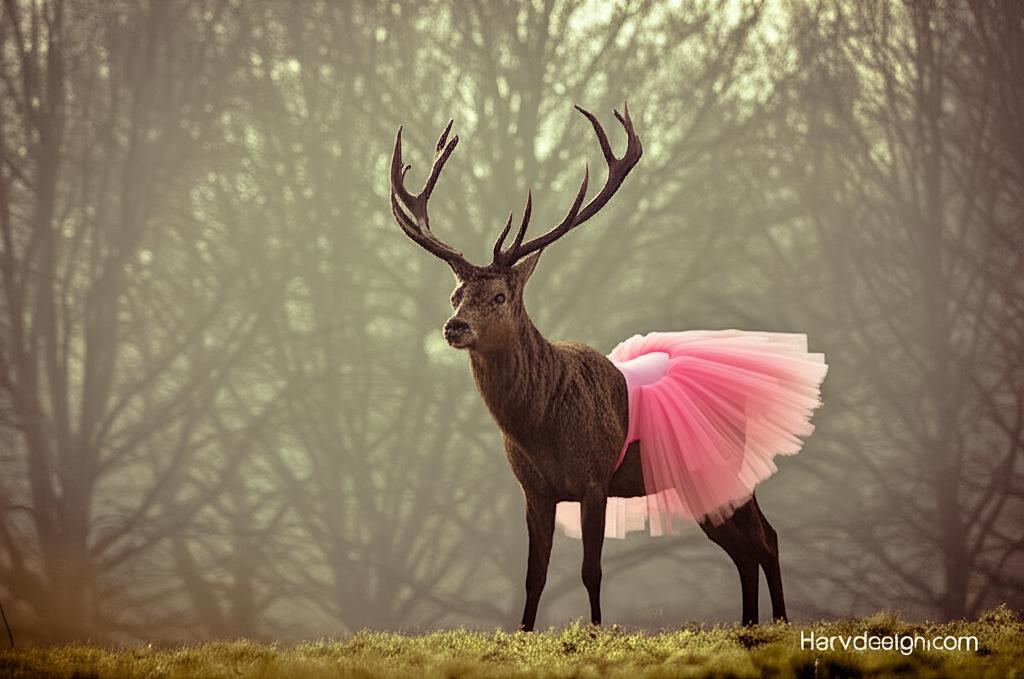}
  \end{minipage}\hfill

  \caption{Models can add objects to images, including as part of another object. User request: \textit{Add a pink tutu dress to the elk in the image.}}
\end{figure*}

\begin{figure*}[htbp]
  \centering
  \begin{minipage}[t]{0.23\textwidth}
    \centering
    \textbf{Original}\\[2mm]
    \includegraphics[width=\textwidth]{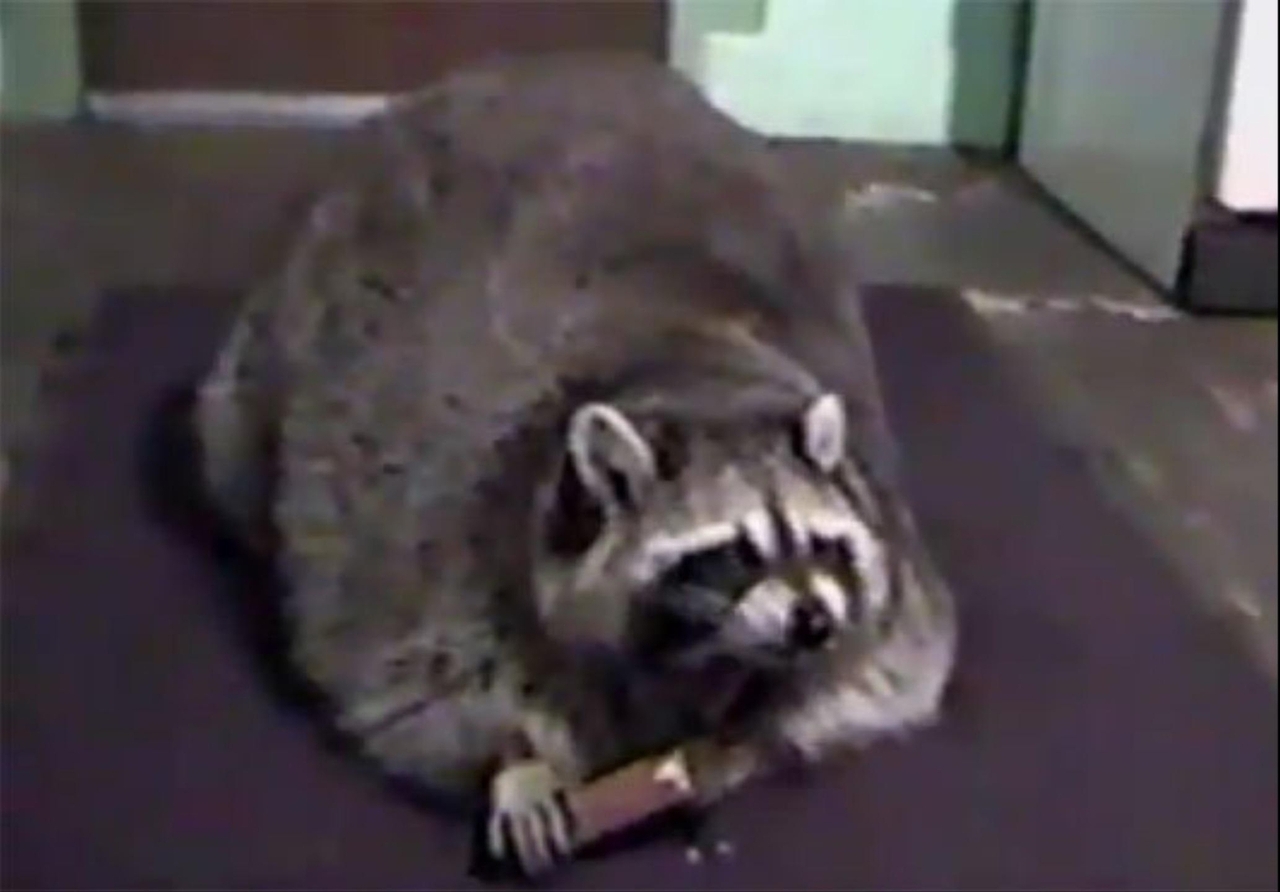}
  \end{minipage}\hfill
  \begin{minipage}[t]{0.23\textwidth}
    \centering
    \textbf{Human Edit}\\[2mm]
    \includegraphics[width=\textwidth]{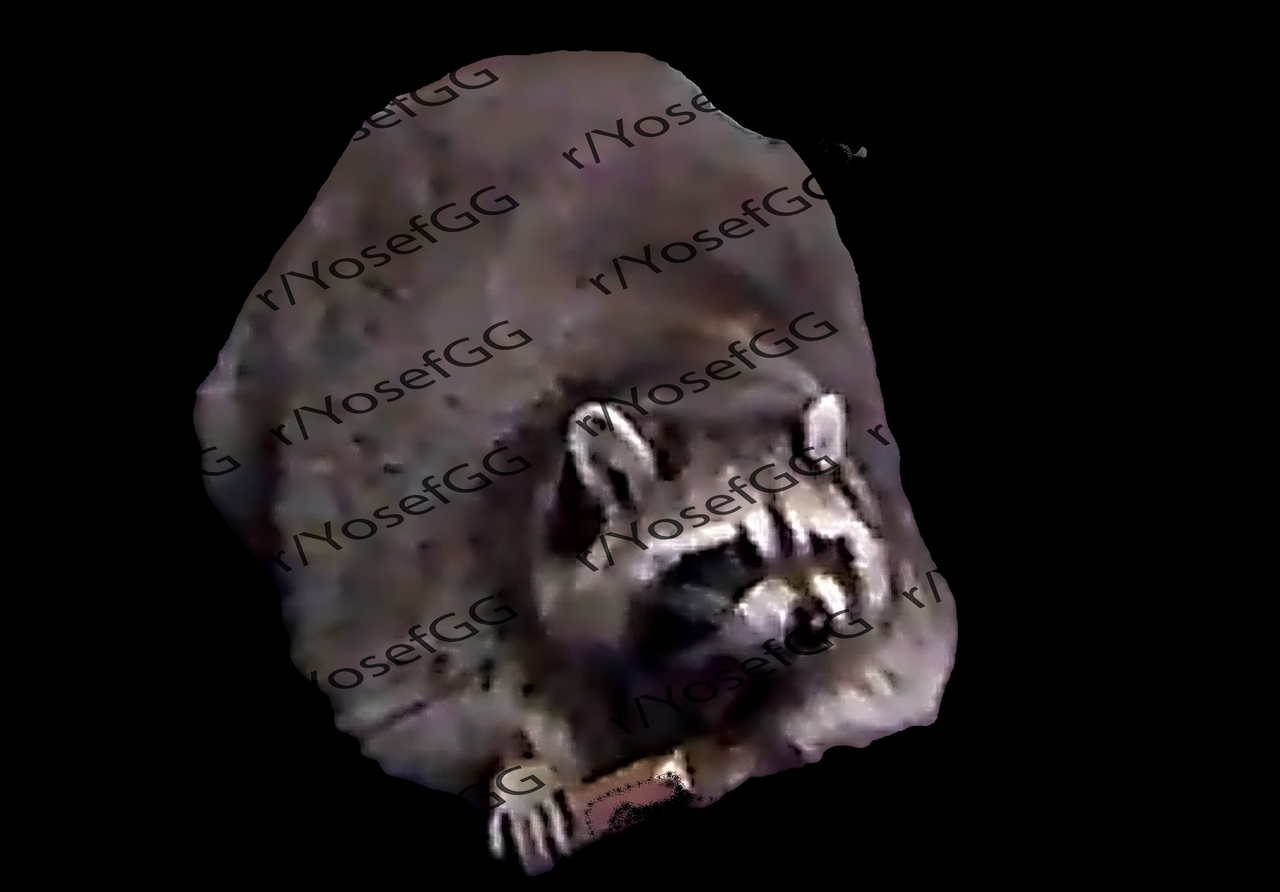}
  \end{minipage}\hfill
  \begin{minipage}[t]{0.23\textwidth}
    \centering
    \model{\TextCut}\\[2mm]
    \includegraphics[width=\textwidth]{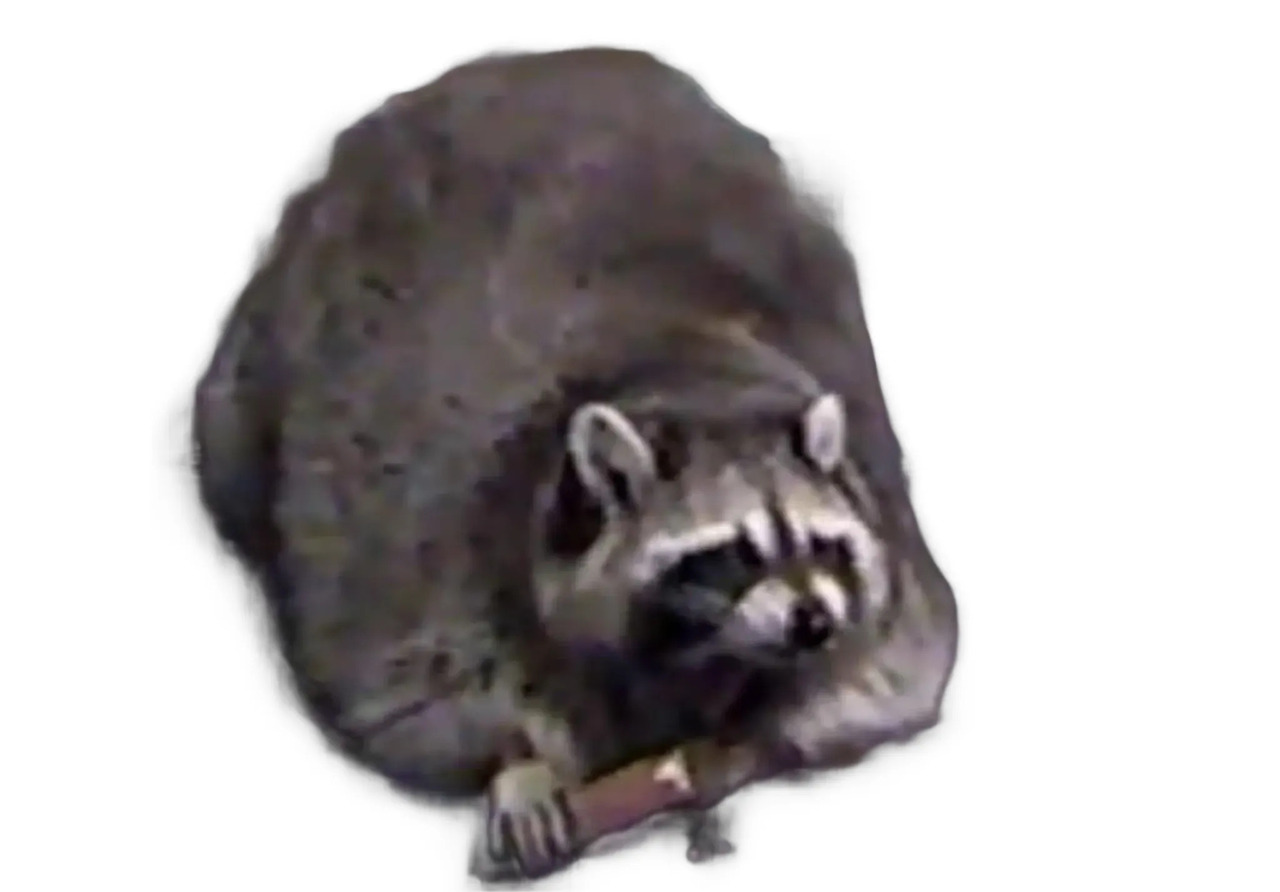}
  \end{minipage}\hfill
  \begin{minipage}[t]{0.23\textwidth}
    \centering
    {\tiny\scalebox{0.85}{\mbox{\model{\FineCutter}}}}\\[2mm]
    \includegraphics[width=\textwidth]{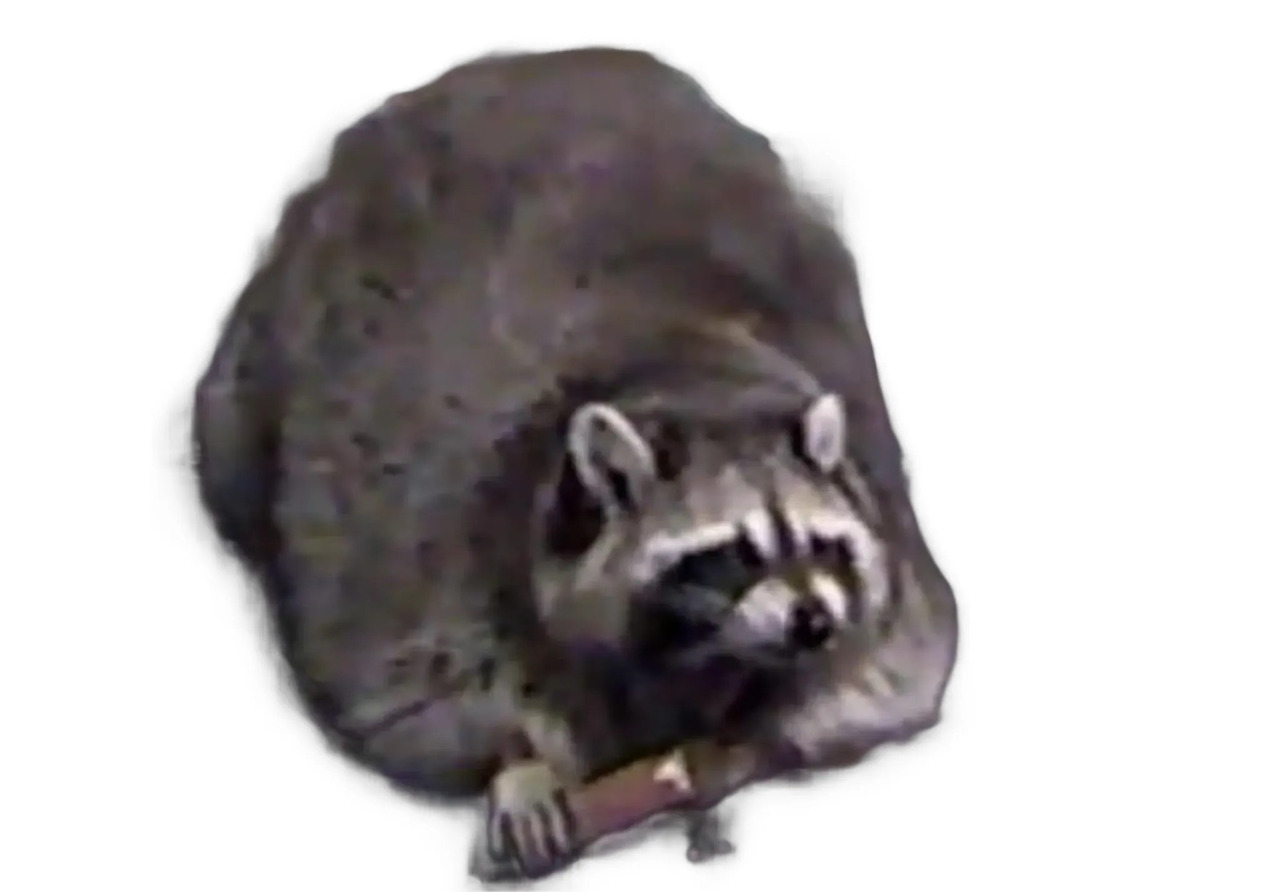}
  \end{minipage}\hfill

  \caption{Models excel at removing the backgrounds of images. User request: \textit{Create a cutout of the raccoon and enhance the image quality to HD.}}
\end{figure*}

\begin{figure*}[htbp]
  \centering
  \begin{minipage}[t]{0.18\textwidth}
    \centering
    \textbf{Original}\\[2mm]
    \includegraphics[width=\textwidth]{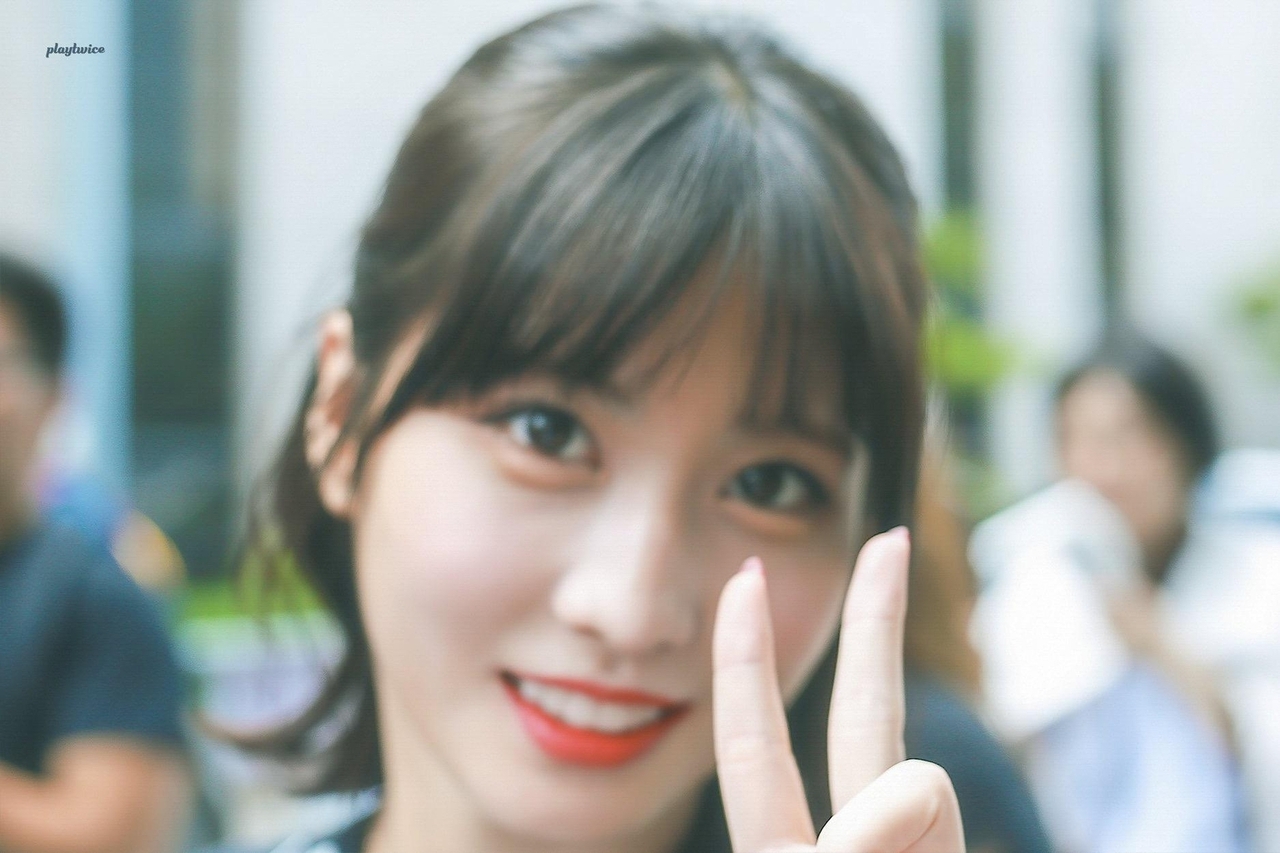}
  \end{minipage}\hfill
  \begin{minipage}[t]{0.18\textwidth}
    \centering
    \textbf{Human Edit}\\[2mm]
    \includegraphics[width=\textwidth]{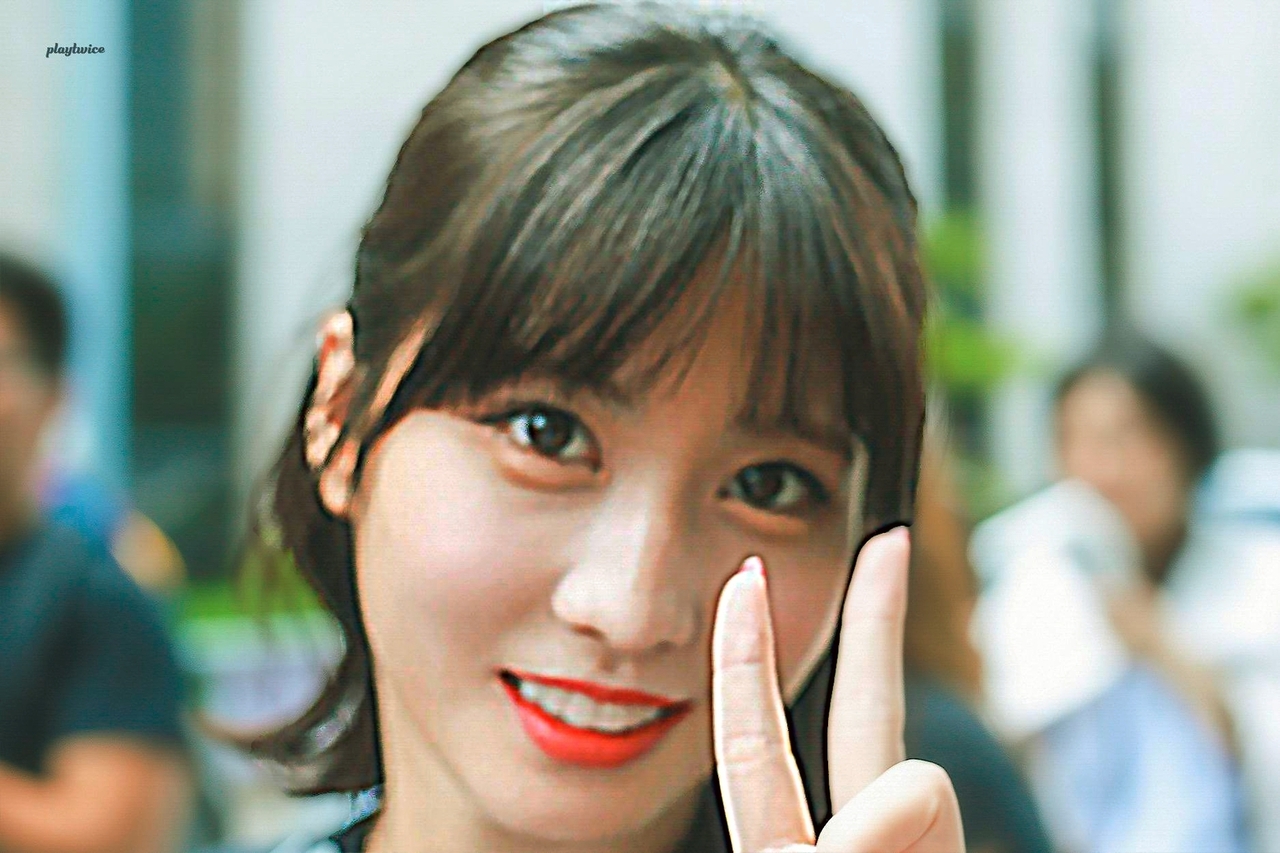}
  \end{minipage}\hfill
  \begin{minipage}[t]{0.18\textwidth}
    \centering
    \model{\SeedEdit}\\[2mm]
    \includegraphics[width=\textwidth]{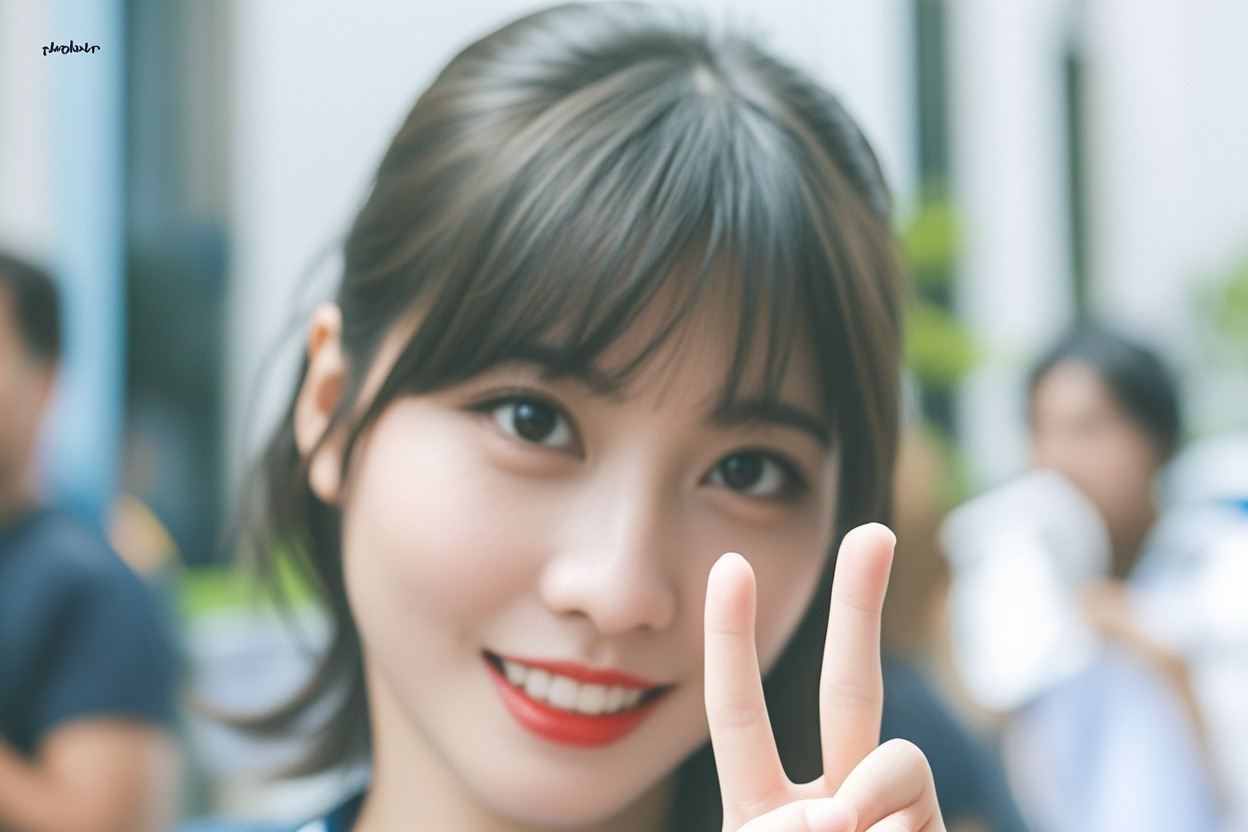}
  \end{minipage}\hfill
  \begin{minipage}[t]{0.18\textwidth}
    \centering
    \model{\CodeFormer}\\[2mm]
    \includegraphics[width=\textwidth]{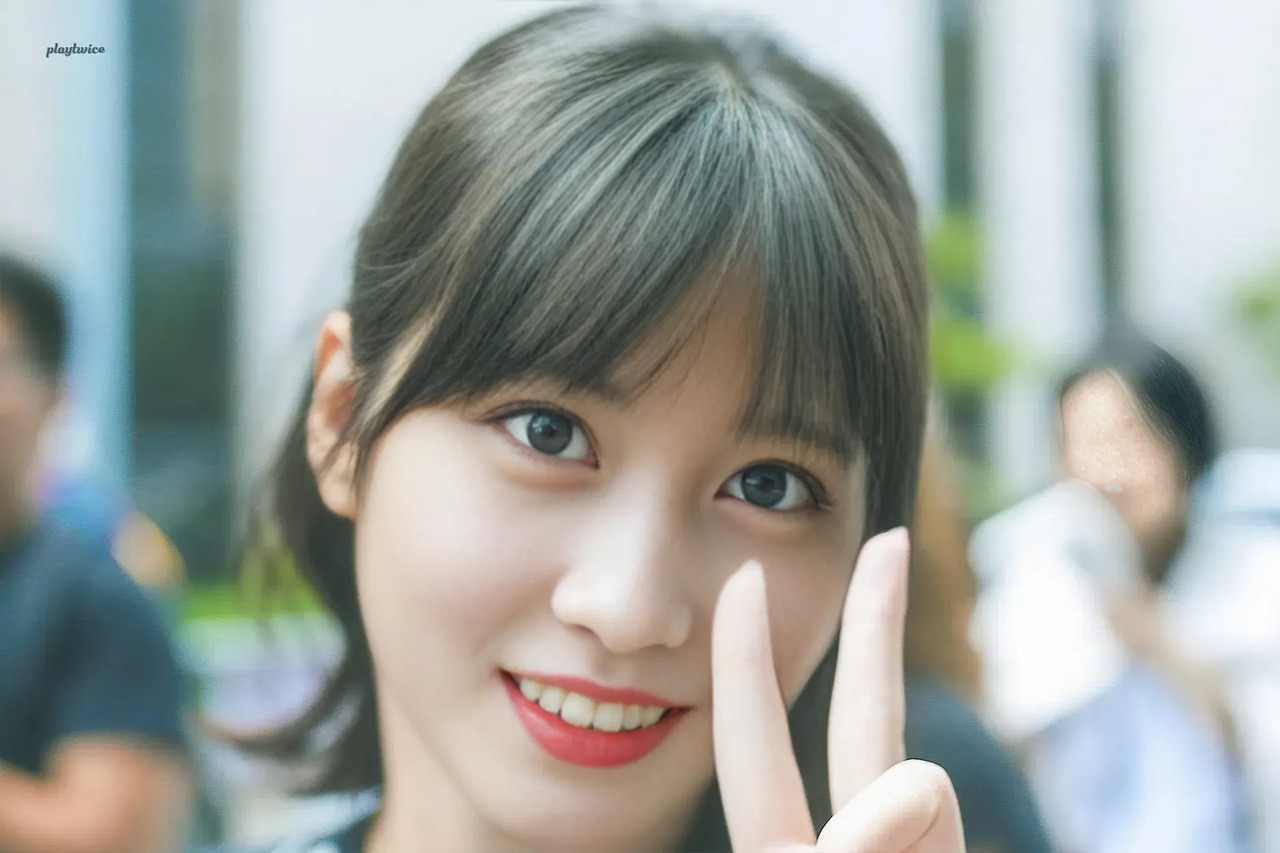}
  \end{minipage}\hfill
  \begin{minipage}[t]{0.18\textwidth}
    \centering
    \model{\gpt}\\[2mm]
    \includegraphics[width=\textwidth]{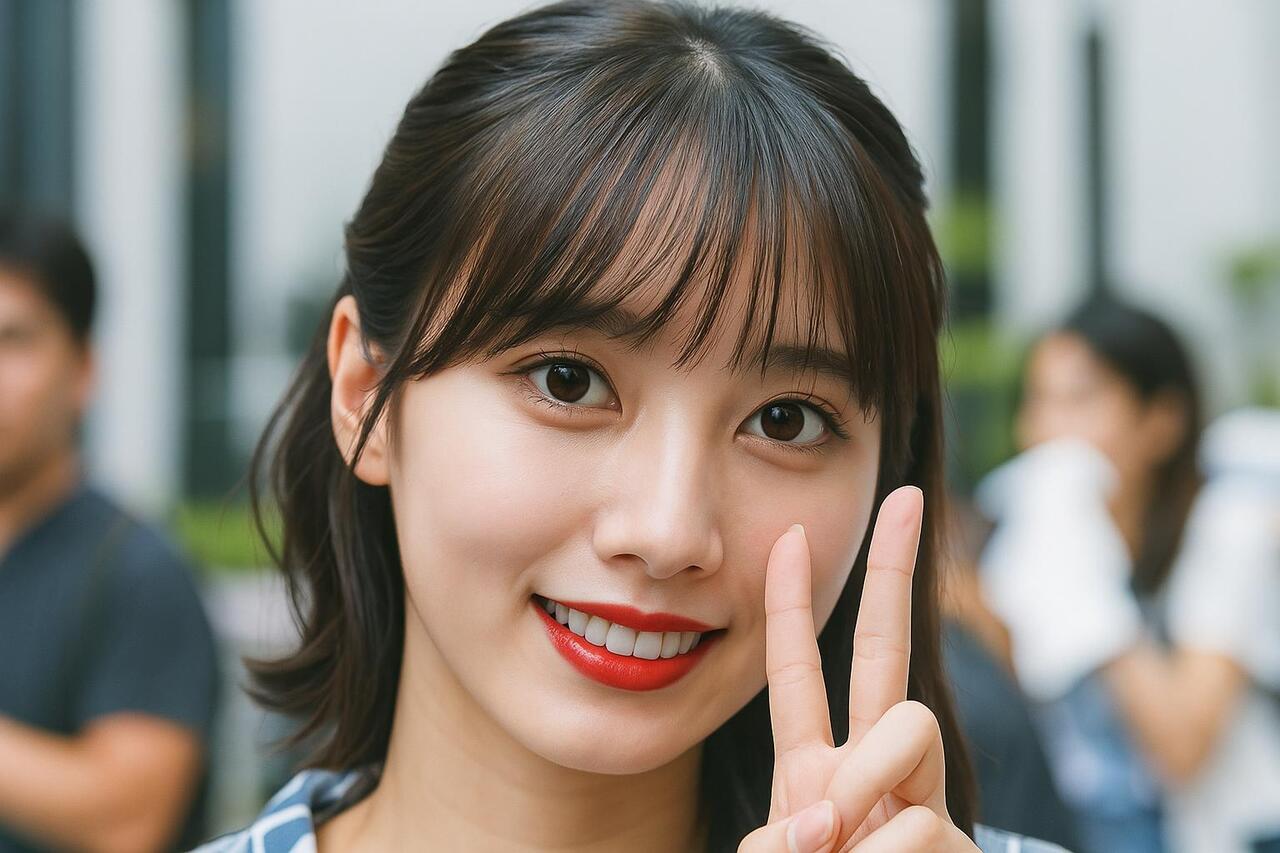}
  \end{minipage}\hfill

  \caption{Models can enhance images by removing blur. User request: \textit{Remove blur from the subject while keeping the background intact.}}
\end{figure*}

\begin{figure*}[htbp]
  \centering
  \begin{minipage}[t]{0.18\textwidth}
    \centering
    \textbf{Original}\\[2mm]
    \includegraphics[width=\textwidth]{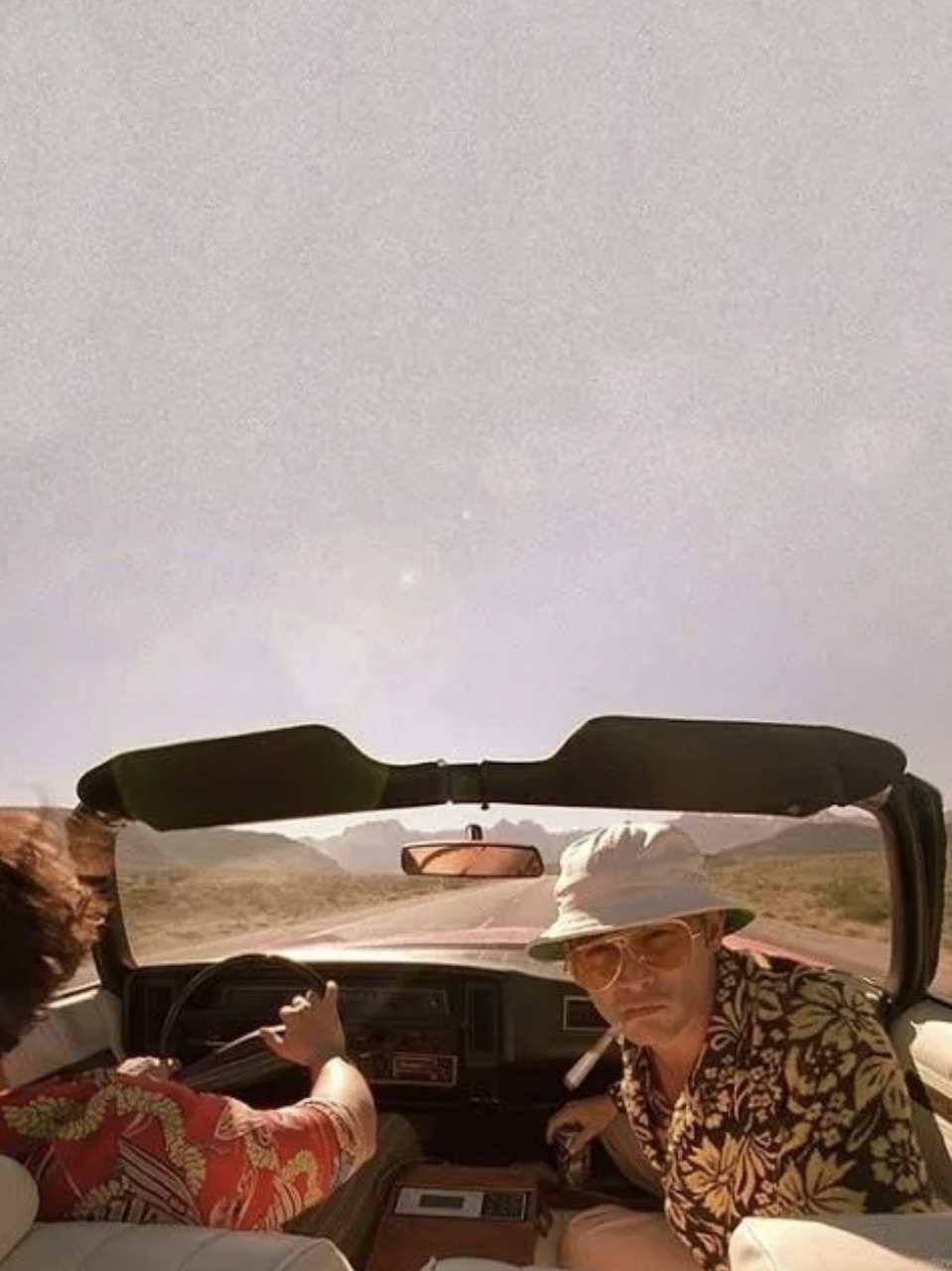}
  \end{minipage}\hfill
  \begin{minipage}[t]{0.18\textwidth}
    \centering
    \textbf{Human Edit}\\[2mm]
    \includegraphics[width=\textwidth]{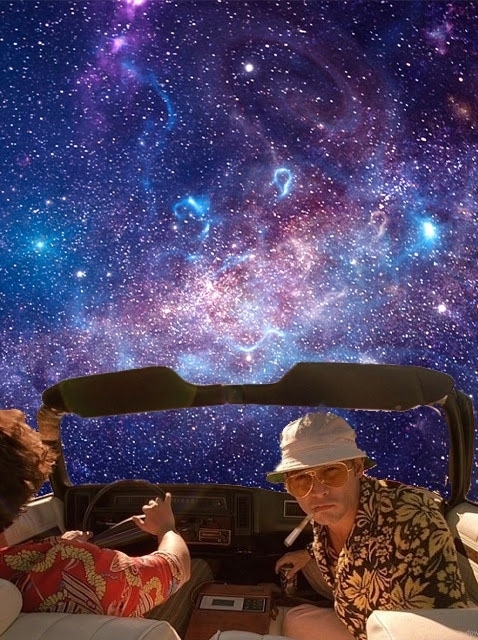}
  \end{minipage}\hfill
  \begin{minipage}[t]{0.18\textwidth}
    \centering
    \model{\Magic}\\[2mm]
    \includegraphics[width=\textwidth]{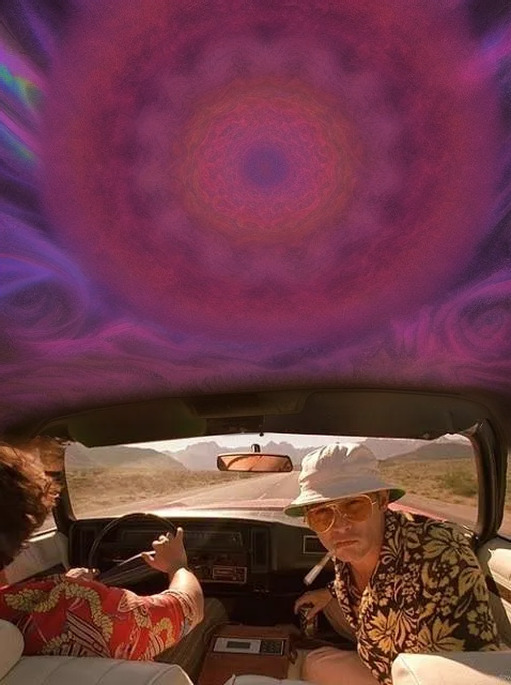}
  \end{minipage}
  \begin{minipage}[t]{0.18\textwidth}
    \centering
    \model{\InstructPix}\\[2mm]
    \includegraphics[width=\textwidth]{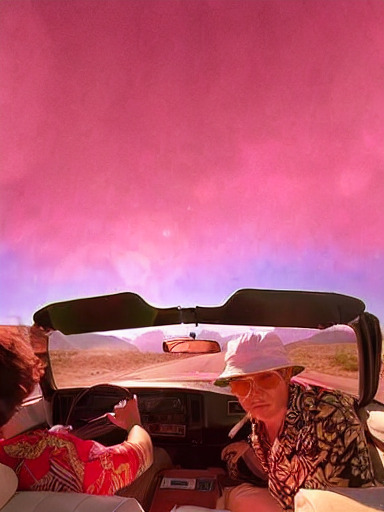}
  \end{minipage}\hfill
  \begin{minipage}[t]{0.18\textwidth}
    \centering
    \model{\geminiflash{}}\\[2mm]
    \includegraphics[width=\textwidth]{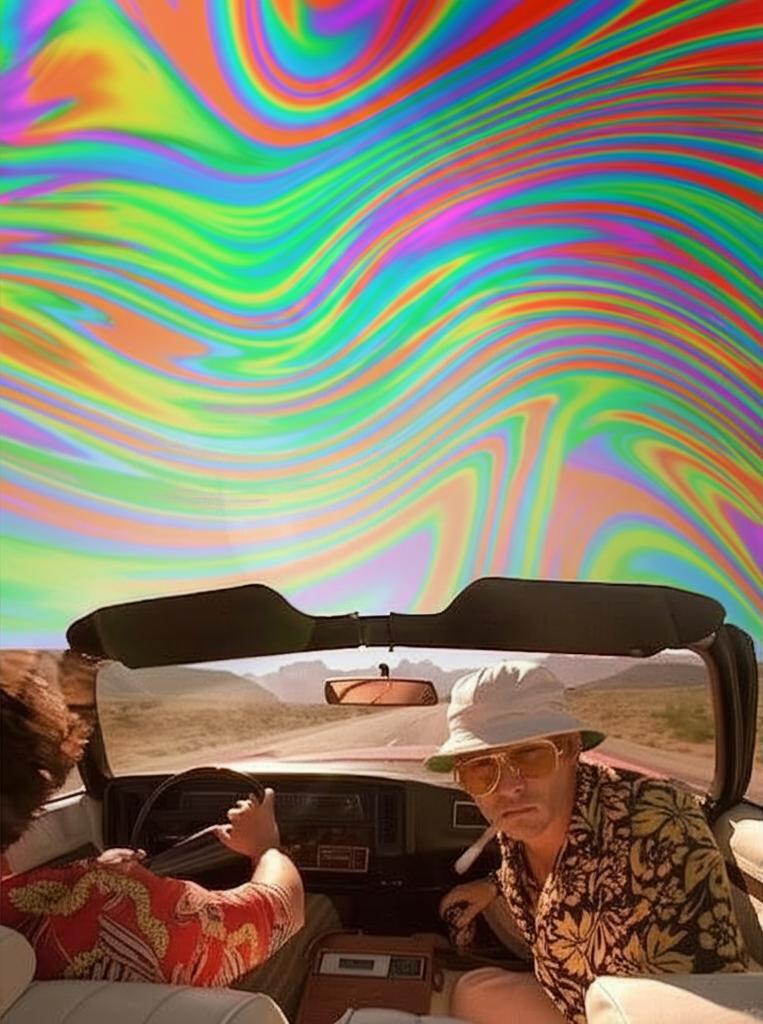}
  \end{minipage}\hfill
  
  \caption{Models succeed at edits that do not require high amounts of realism. \textit{User request: Transform the sky to appear psychedelic.}}
  \label{fig:abstract_sample}
\end{figure*}

\begin{figure*}[htbp]
  \centering
  \begin{minipage}[t]{0.16\textwidth}
    \centering
    \textbf{Original}\\[2mm]
    \includegraphics[width=\textwidth]{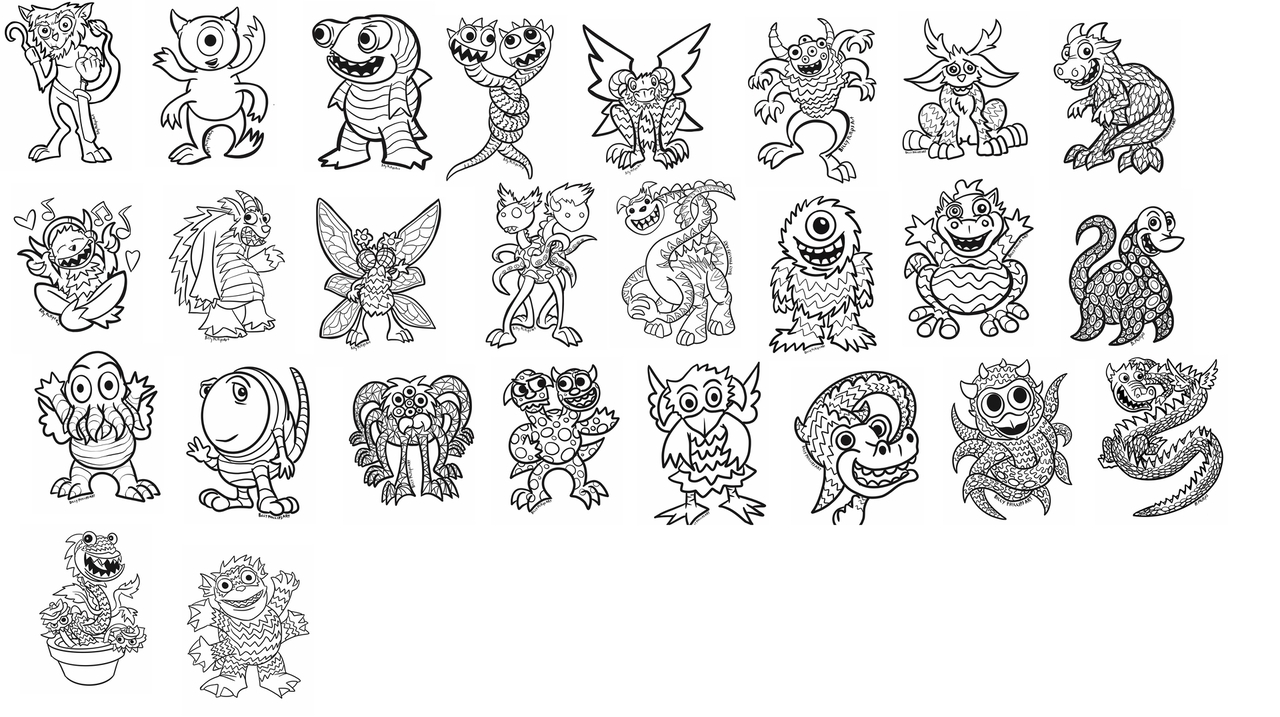}
  \end{minipage}\hfill
  \begin{minipage}[t]{0.16\textwidth}
    \centering
    \textbf{Human Edit}\\[2mm]
    \includegraphics[width=\textwidth]{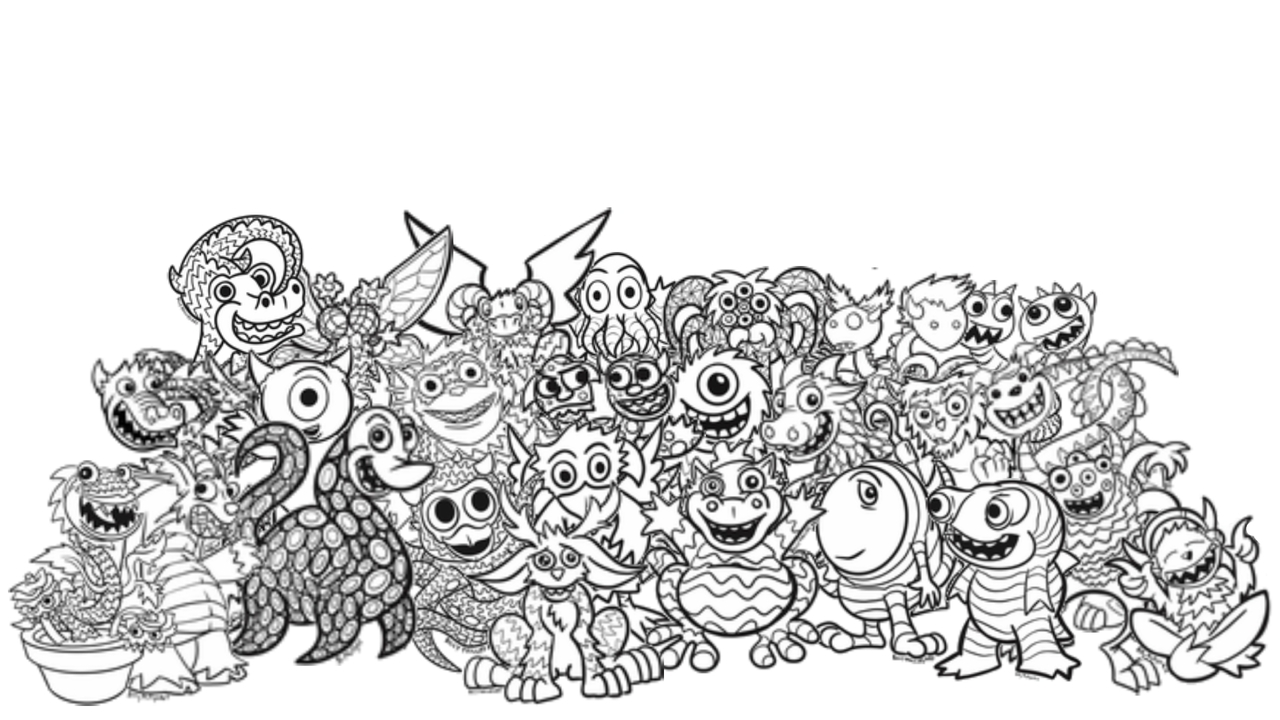}
  \end{minipage}\hfill
  \begin{minipage}[t]{0.16\textwidth}
    \centering
    \model{\CosXL}\\[2mm]
    \includegraphics[width=\textwidth]{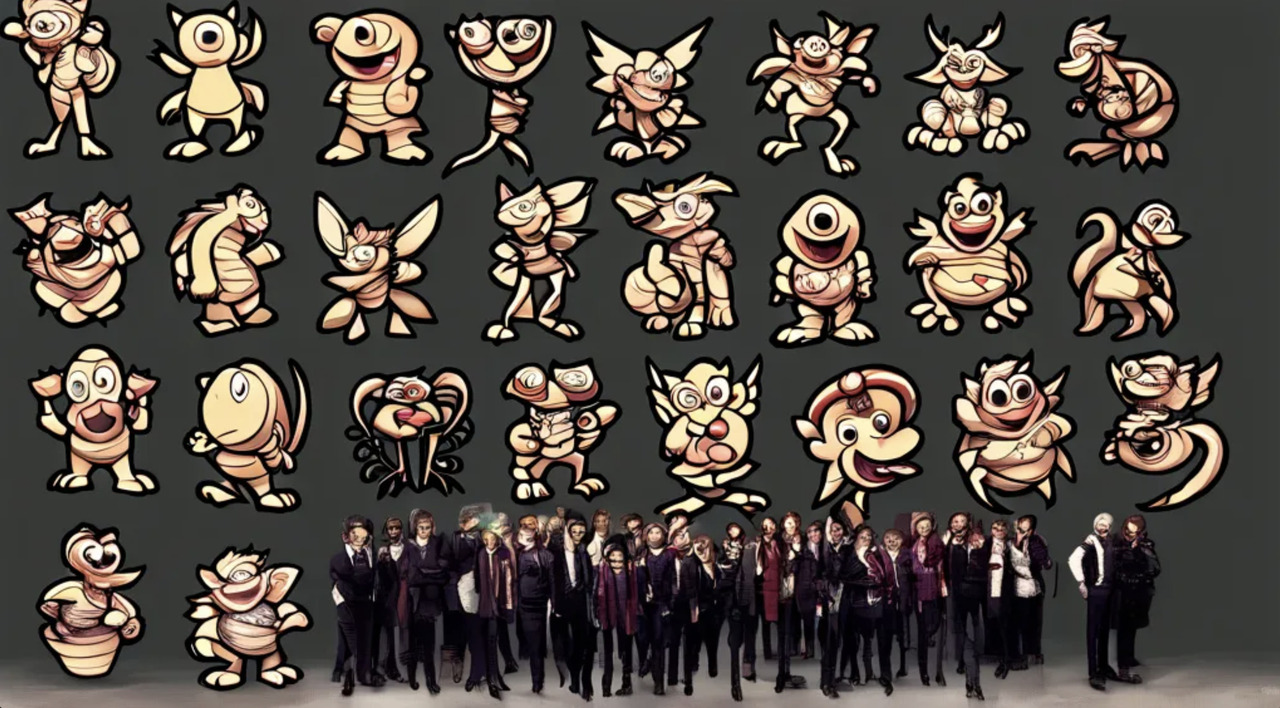}
  \end{minipage}\hfill
  \begin{minipage}[t]{0.16\textwidth}
    \centering
    \model{\InstructPix}\\[2mm]
    \includegraphics[width=\textwidth]{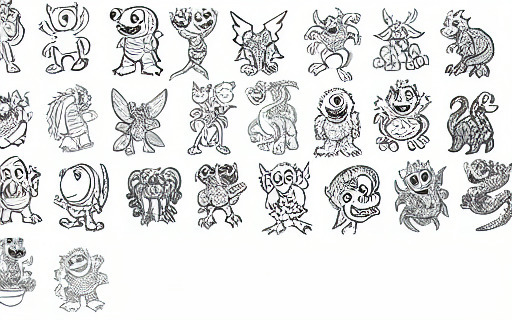}
  \end{minipage}
  \begin{minipage}[t]{0.16\textwidth}
    \centering
    {\tiny\scalebox{0.85}{\mbox{\model{\geminiflash{}}}}}\\[2mm]
    \includegraphics[width=\textwidth]{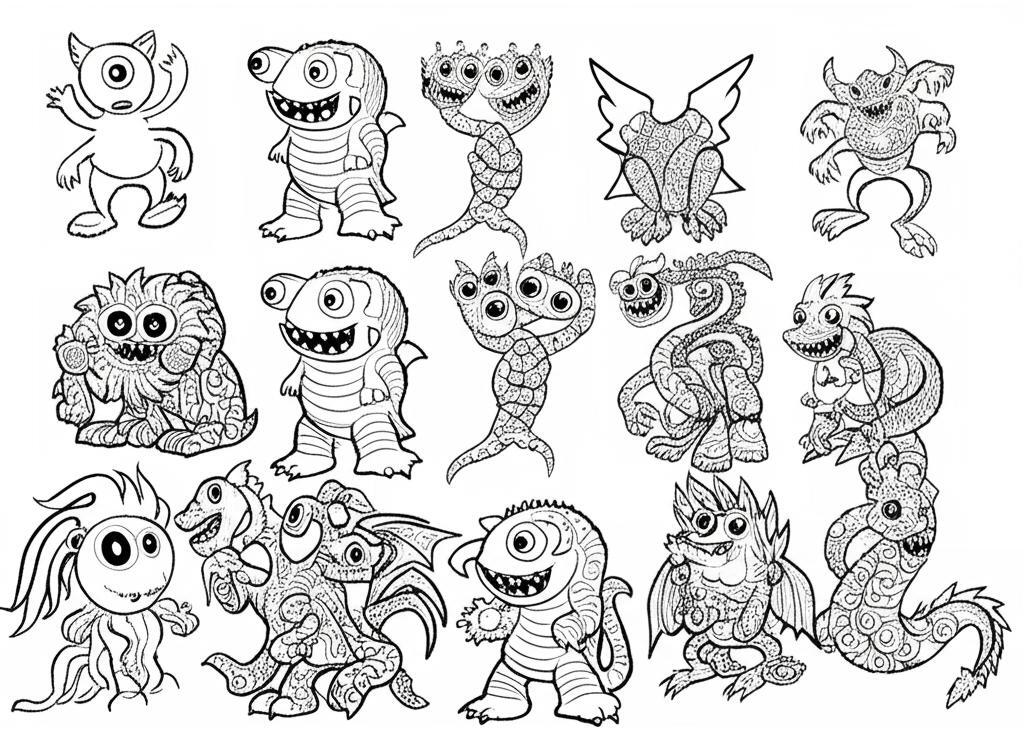}
  \end{minipage}
  \begin{minipage}[t]{0.16\textwidth}
    \centering
    \model{\gpt}\\[2mm]
    \includegraphics[width=\textwidth]{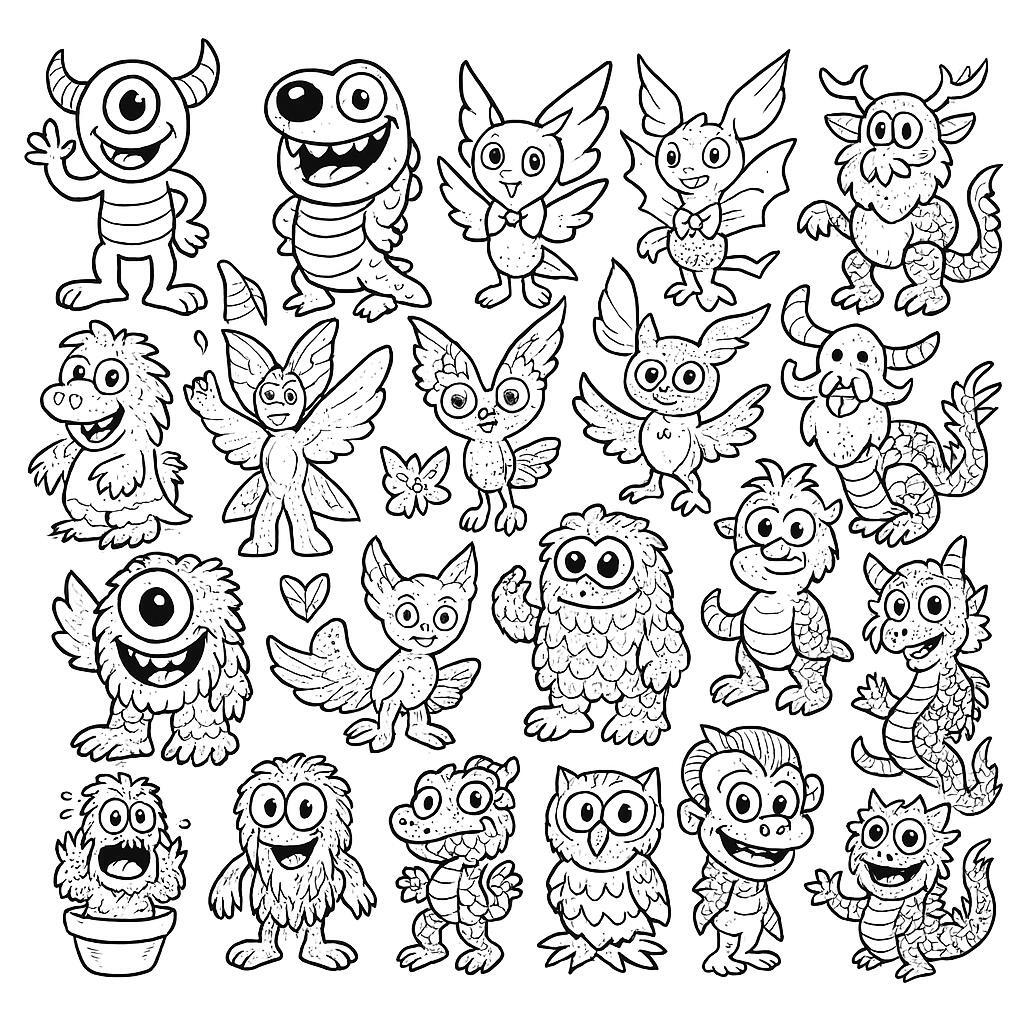}
  \end{minipage}
  
  \caption{Models fail at edits that require manipulating several objects within the same image. \textit{User request: Arrange the characters into a group picture, resizing as necessary and omitting a few for aesthetics.}}
  \label{fig:multi_obj_sample}
\end{figure*}

\begin{figure*}[htbp]
  \centering
  \begin{minipage}[t]{0.18\textwidth}
    \centering
    \textbf{Original}\\[2mm]
    \includegraphics[width=\textwidth]{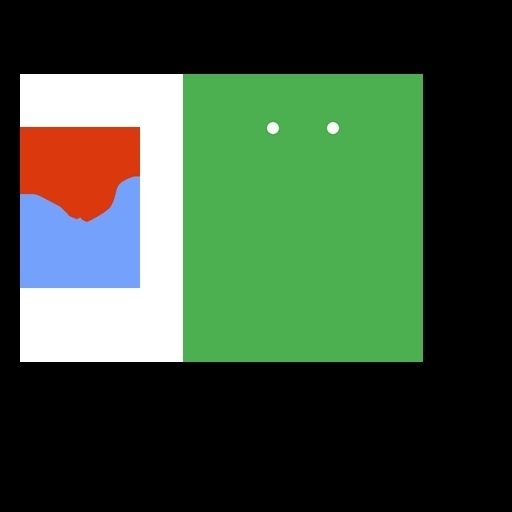}
  \end{minipage}\hfill
  \begin{minipage}[t]{0.18\textwidth}
    \centering
    \textbf{Human Edit}\\[2mm]
    \includegraphics[width=\textwidth]{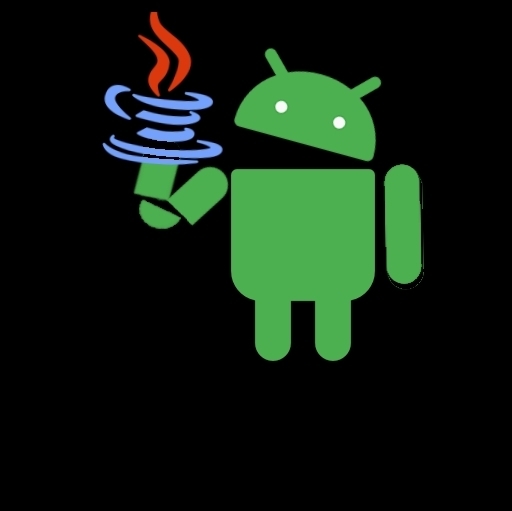}
  \end{minipage}\hfill
  \begin{minipage}[t]{0.24\textwidth}
    \centering
    \model{\CosXL}\\[2mm]
    \includegraphics[width=\textwidth]{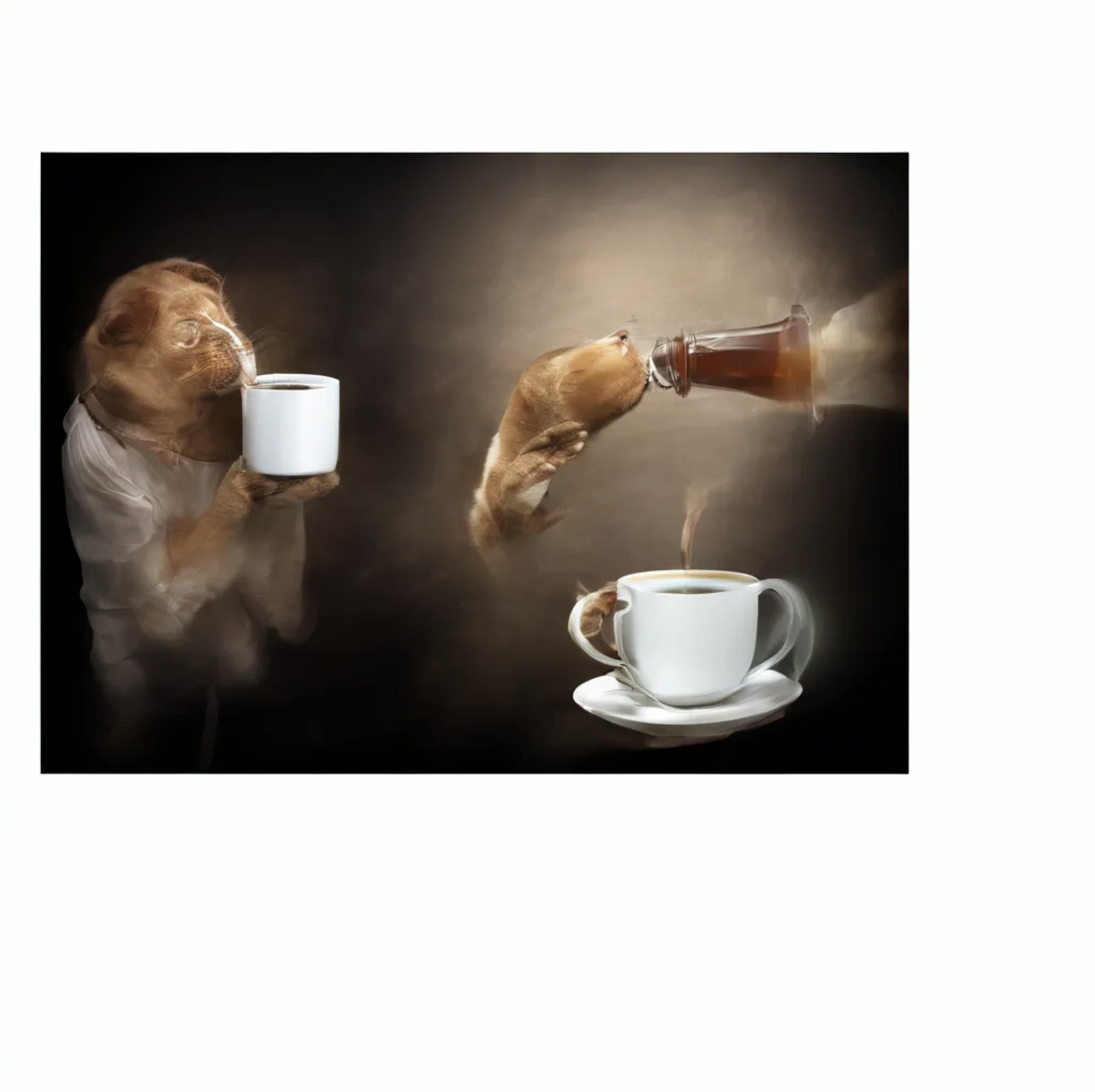}
  \end{minipage}\hfill
  \begin{minipage}[t]{0.18\textwidth}
    \centering
    \model{\InstructPix}\\[2mm]
    \includegraphics[width=\textwidth]{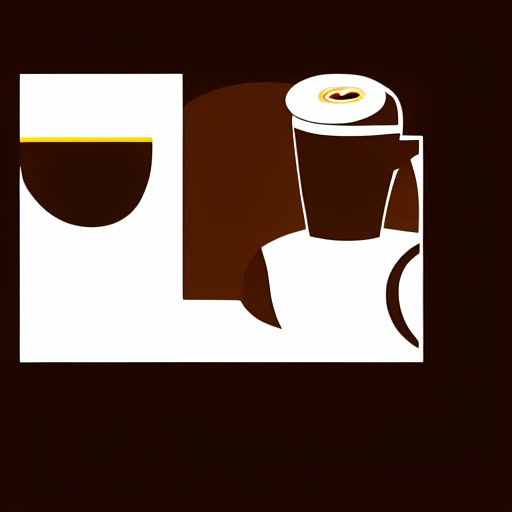}
  \end{minipage}
  \begin{minipage}[t]{0.18\textwidth}
    \centering
    \model{\gpt}\\[2mm]
    \includegraphics[width=\textwidth]{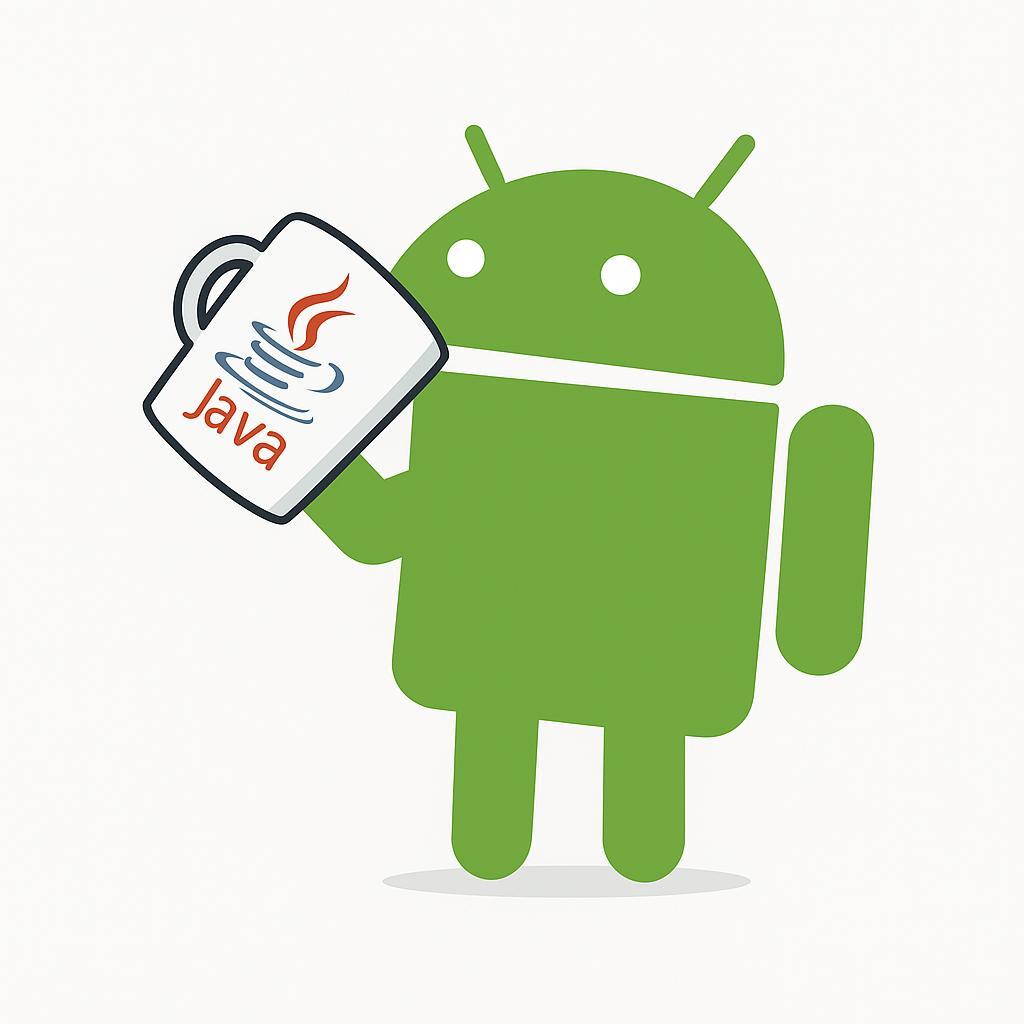}
  \end{minipage}
  
  \caption{Human users often provide rough reference images as a guide for the type of image they want to create. Models typically struggle to understand these reference images when they are not exact representations of what the final output should appear like.  \textit{User request: Edit the android image to appear as if it is drinking from a Java cup.}}
  \label{fig:reference_guides_1}
\end{figure*}

\begin{figure*}[htbp]
  \centering
  \begin{minipage}[t]{0.18\textwidth}
    \centering
    \textbf{Original}\\[2mm]
    \includegraphics[width=\textwidth]{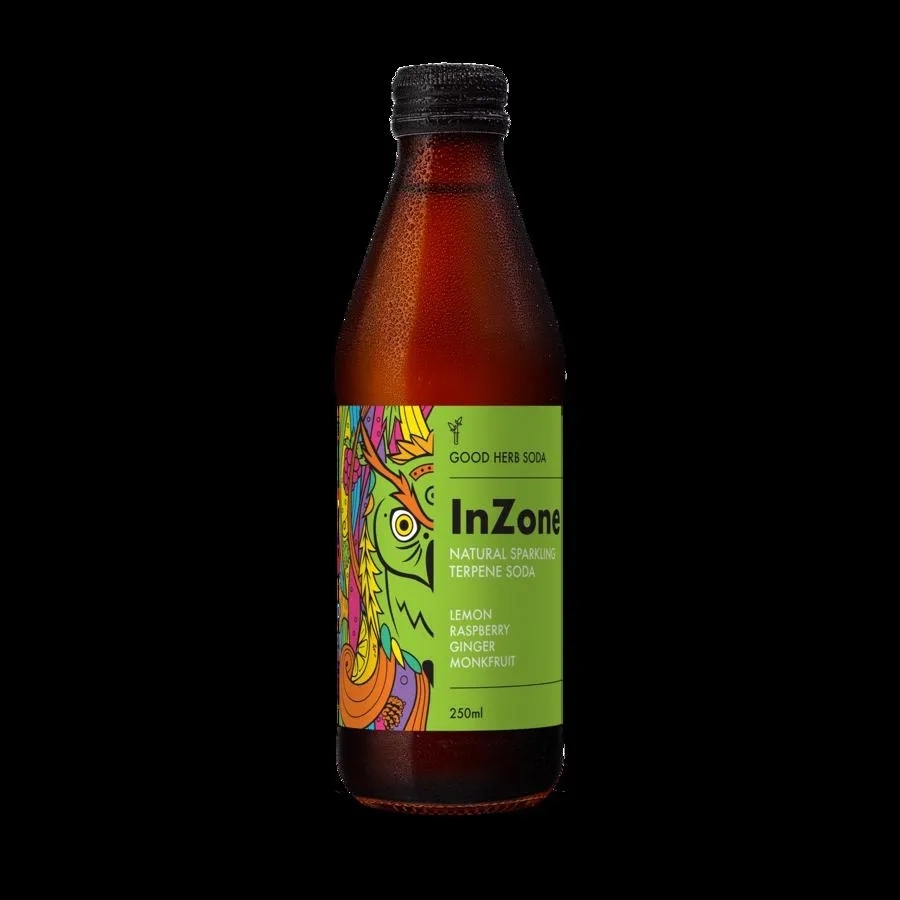}
  \end{minipage}\hfill
  \begin{minipage}[t]{0.18\textwidth}
    \centering
    \textbf{Human Edit}\\[2mm]
    \includegraphics[width=\textwidth]{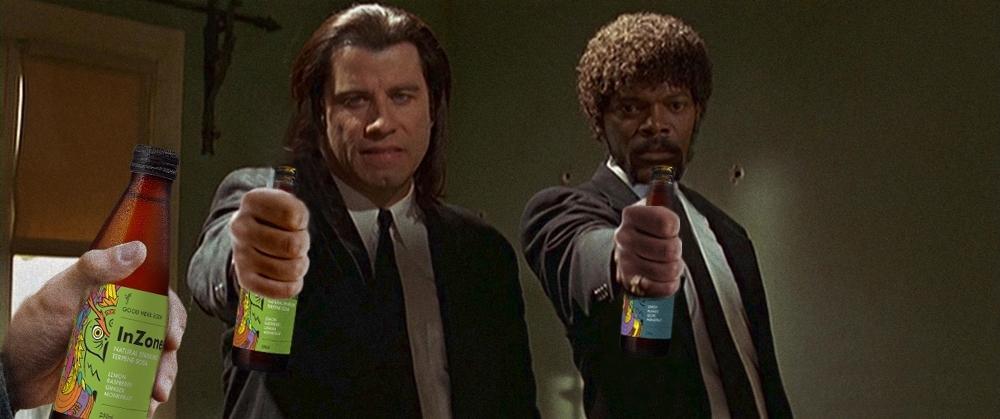}
  \end{minipage}\hfill
  \begin{minipage}[t]{0.18\textwidth}
    \centering
    \model{\SeedEdit}\\[2mm]
    \includegraphics[width=\textwidth]{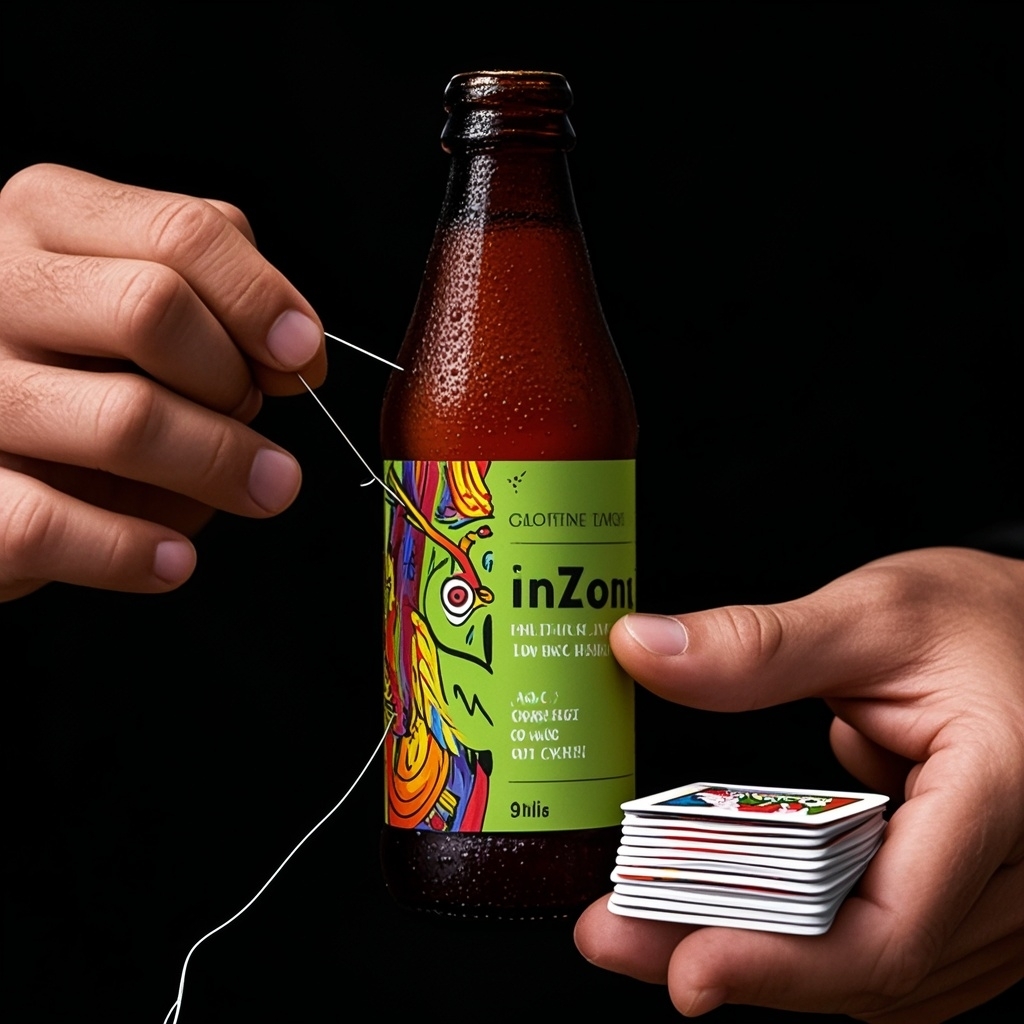}
  \end{minipage}\hfill
  \begin{minipage}[t]{0.18\textwidth}
    \centering
    \model{\geminiflash{}}\\[2mm]
    \includegraphics[width=\textwidth]{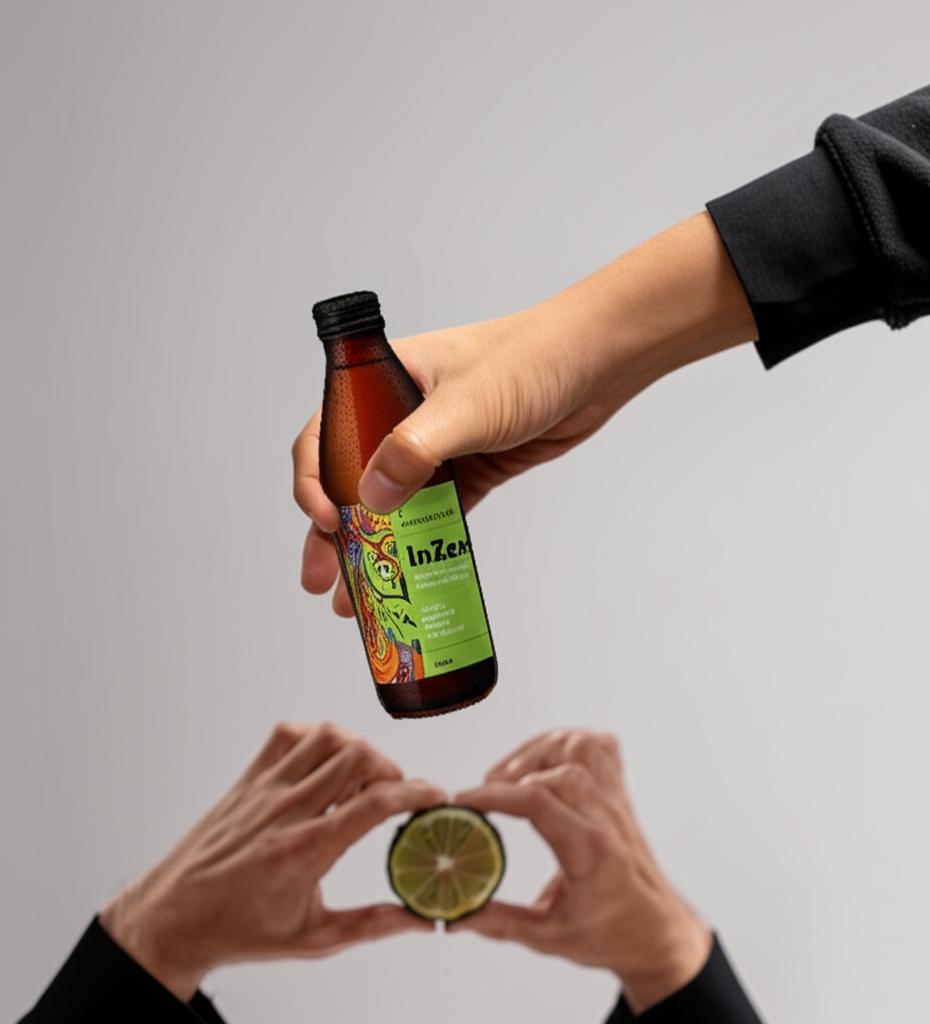}
  \end{minipage}
  \begin{minipage}[t]{0.18\textwidth}
    \centering
    \model{\gpt}\\[2mm]
    \includegraphics[width=\textwidth]{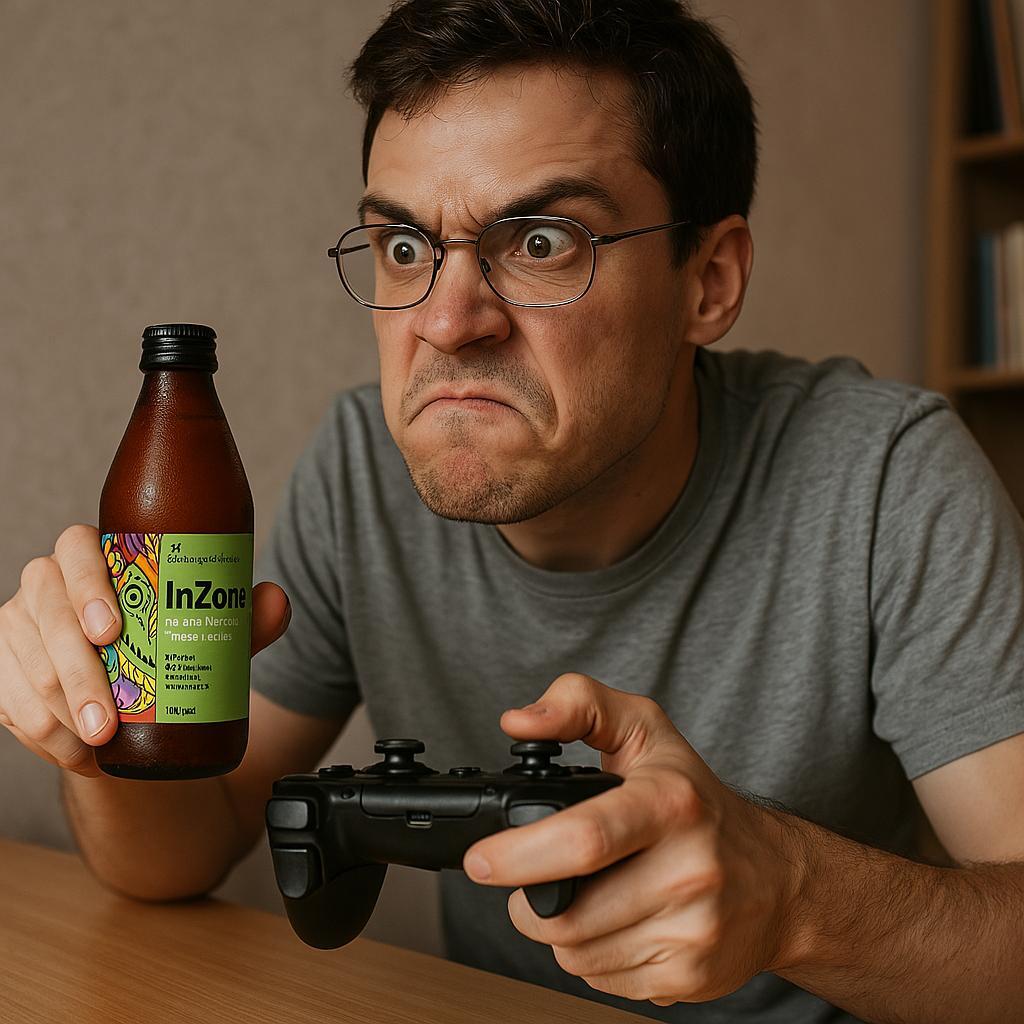}
  \end{minipage}
  
  \caption{Models struggle with requests requiring humor. User request: \textit{Photoshop the soda into humorous scenarios where people's hands are doing something focused or with extreme concentration.}}
  \label{fig:humor_sample}
\end{figure*}

\clearpage
\section{Generative AI models fail to preserve identity in image editing}
\label{sec:appendix-face_diff}

We generate an initial image of a person wearing a white T-shirt across various ages and genders using \model{FLUX-Pro}~\cite{flux2024}. These images are then processed using \gpt{} and \geminiflash{}{} to apply different shirt colors. At each step, we send the image from the previous step along with a new color name, repeating this process for eight steps. Finally, we instruct the model to revert the shirt color back to white (the original color). 
At each step, we compute the $L_2$ distance between the \model{DINOv2\_ViT-B/14}~\cite{oquab2023dinov2} feature representations of the modified images and the original image (white shirt) to quantify feature deviations.

\begin{figure}[h]
    \begin{subfigure}{\linewidth}
        \includegraphics[width=\linewidth]{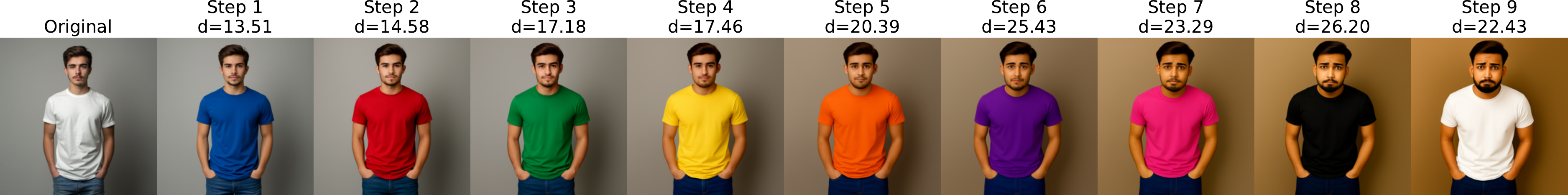}
    \end{subfigure}
    
    \vspace{-\baselineskip} 
    \begin{subfigure}{\linewidth}
        \includegraphics[width=\linewidth]{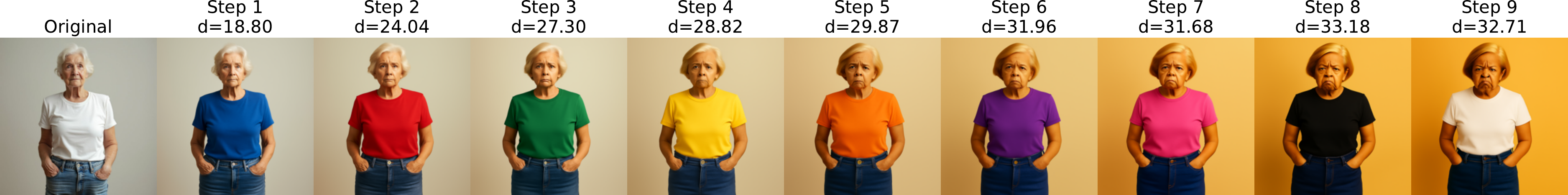}
    \end{subfigure}
    
    \vspace{-\baselineskip}
    \begin{subfigure}{\linewidth}
        \includegraphics[width=\linewidth]{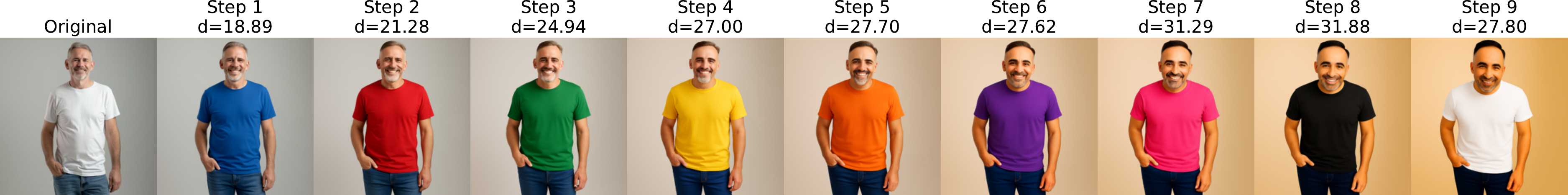}
    \end{subfigure}
    
    \vspace{-\baselineskip}
    \begin{subfigure}{\linewidth}
        \includegraphics[width=\linewidth]{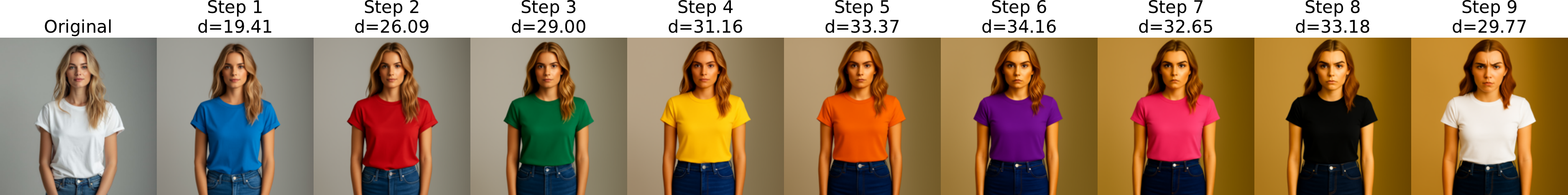}
    \end{subfigure}
    
    \vspace{-\baselineskip}
    \begin{subfigure}{\linewidth}
        \includegraphics[width=\linewidth]{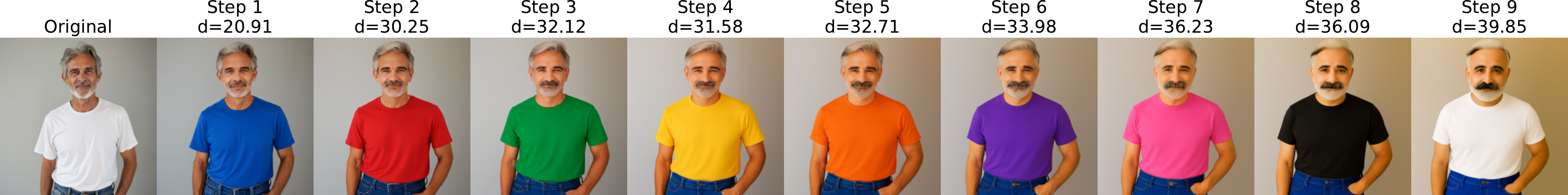}
    \end{subfigure}
    {\small
    \flushleft \hspace{0.3cm}\texttt{color}: \hspace{0.5cm} blue
    \hspace{0.9cm} red
    \hspace{0.8cm} green
    \hspace{0.6cm} yellow
    \hspace{0.6cm} orange
    \hspace{0.6cm} purple
    \hspace{0.7cm} pink
    \hspace{0.7cm} black
    \hspace{0.8cm} white
    }
    {\small
    \flushleft
    \hspace{3cm} Prompt: \emph{``Make the shirt} \texttt{\{color\}}''
    }
    \caption{
    \gpt \gptlogo fails to preserve the identity of individuals in image editing tasks. When tasked with changing the color of a shirt in a sequence, the identity of the person changes—after a few iterations, the person loses their likeness to the original image. The \textit{d} value indicates the L2 distance between the  \model{DINOv2\_ViT-B/14} embedding of the image and the original one.
    }
    \label{fig:supp-gpt_identity}
\end{figure}

\begin{figure}[h]
    \centering
    \begin{subfigure}{\linewidth}
        \includegraphics[width=\linewidth]{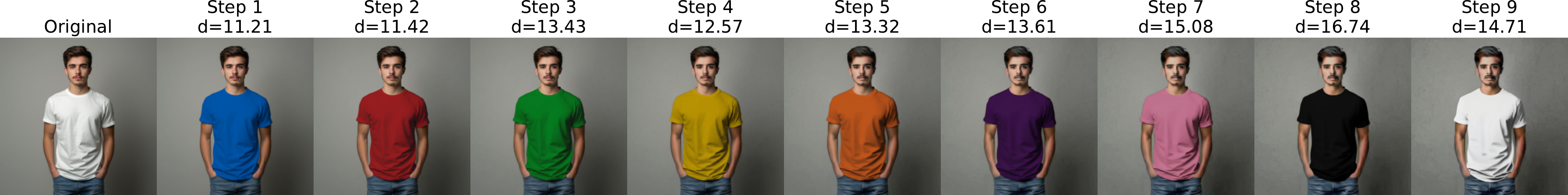}
    \end{subfigure}
    
    \vspace{-\baselineskip} 
    \begin{subfigure}{\linewidth}
        \includegraphics[width=\linewidth]{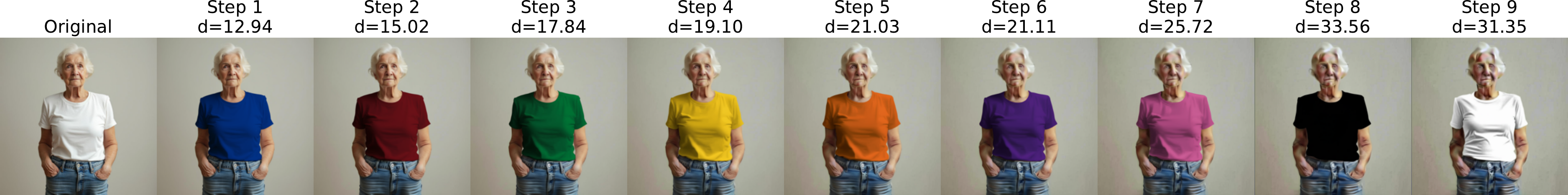}
    \end{subfigure}
    
    \vspace{-\baselineskip}
    \begin{subfigure}{\linewidth}
        \includegraphics[width=\linewidth]{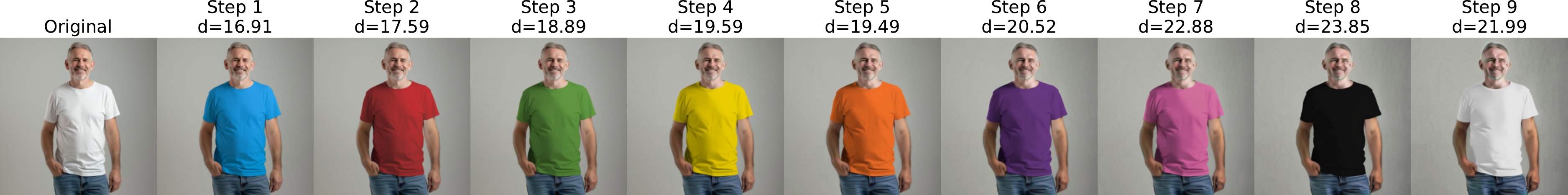}
    \end{subfigure}
    
    \vspace{-\baselineskip}
    \begin{subfigure}{\linewidth}
        \includegraphics[width=\linewidth]{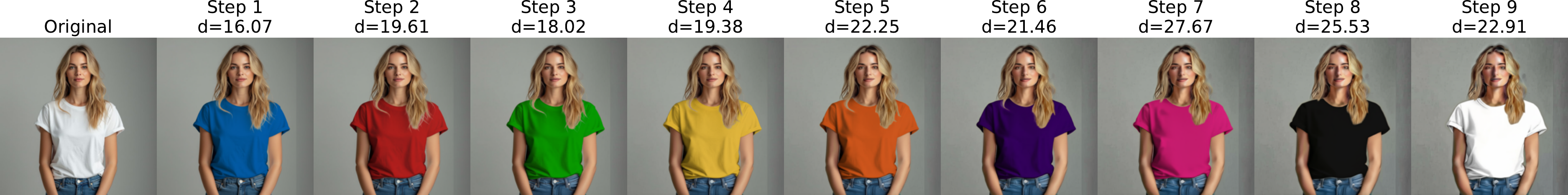}
    \end{subfigure}
    
    \vspace{-\baselineskip}
    \begin{subfigure}{\linewidth}
        \includegraphics[width=\linewidth]{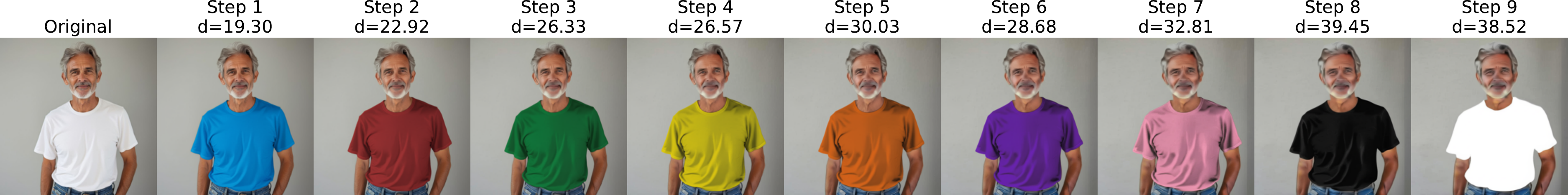}
    \end{subfigure}
    {\small
    \flushleft \hspace{0.3cm}\texttt{color}: \hspace{0.3cm} blue
    \hspace{0.9cm} red
    \hspace{0.8cm} green
    \hspace{0.6cm} yellow
    \hspace{0.6cm} orange
    \hspace{0.6cm} purple
    \hspace{0.7cm} pink
    \hspace{0.7cm} black
    \hspace{0.8cm} white
    }
    {\small
    \flushleft
    \hspace{3cm} Prompt: \emph{``Make the shirt} \texttt{\{color\}}''
    }
    \caption{
    \geminiflash{} \geminilogo fails to preserve the identity of individuals in image editing tasks. When tasked with changing the color of a shirt in a sequence, the identity of the person changes—after a few iterations, the person loses their likeness to the original image. The \textit{d} value indicates the L2 distance between the \model{DINOv2\_ViT-B/14} embedding of the image and the original one.
    }
    \label{fig:supp-gemini_identity}
\end{figure}

\begin{figure}
    \centering
    \includegraphics[width=\linewidth]{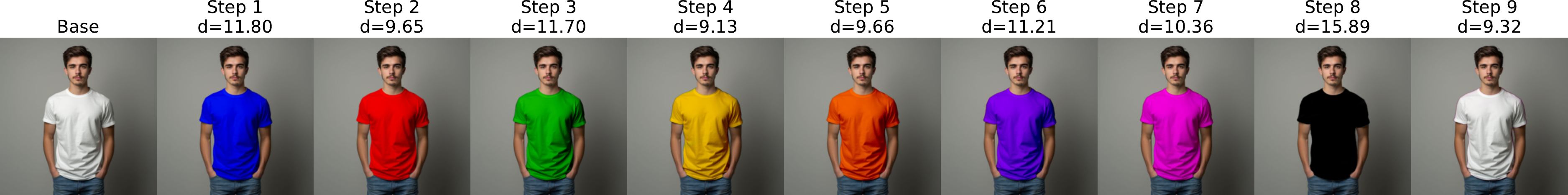}
    \caption{Although changing the shirt color while preserving every other detail in the image affects the  \model{DINO} embeddings, the magnitude of the change is much smaller than that caused by any edits made by AI models such as \gpt{} and \geminiflash{}.
    In this figure, we change the shirt color using an image editing tool while keeping all other elements unchanged. We report the variation in the \model{DINOv2\_ViT-B/14} embedding compared to the original image.}
    \label{fig:supp_gpt_gemini_identity_baseline}
\end{figure}

\begin{figure}
    \centering
    \includegraphics[width=\linewidth]{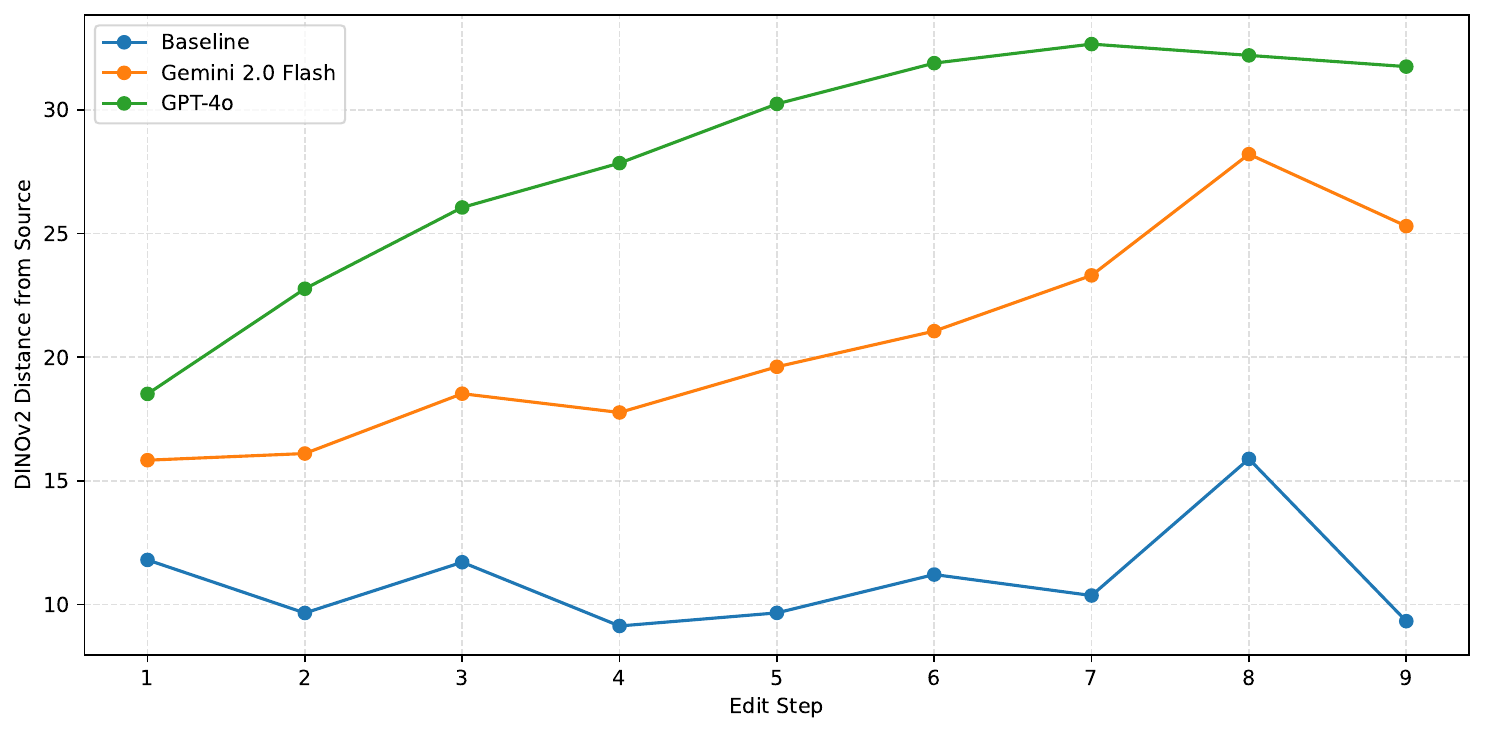}
    \caption{In a sequential image editing request, where a model is subsequently asked to change a shared color, the distance between the edited image at step $n$ and the original image increases as the requests continue.
    The value for each data point step indicates the L2 distance between the \model{DINOv2\_ViT-B/14} embedding of the image and that of the original image.
    }
    \label{fig:supp_gpt_gemini_identity_chart}
\end{figure}

\clearpage
\section{Dataset License}
\label{sec:appendix-dataset_license}

All data collected in this study comes from publicly available Reddit posts, which fall under Reddit’s user agreement policy.

\gpt{} image edits are produced using the ChatGPT website.

\geminiflash{} image edits are produced using the standard API.

\SeedEdit{} images edits are generated by the SeedEdit team.

Other AI generated edits were created through the Hugging Face (\huggingface{}) website.


\clearpage
\begin{thebibliography}{58}
\providecommand{\natexlab}[1]{#1}

\bibitem[{Basu et~al.(2023)Basu, Saberi, Bhardwaj, Chegini, Massiceti, Sanjabi, Hu, and Feizi}]{basu2023editval}
Samyadeep Basu, Mehrdad Saberi, Shweta Bhardwaj, Atoosa~Malemir Chegini, Daniela Massiceti, Maziar Sanjabi, Shell~Xu Hu, and Soheil Feizi. 2023.
\newblock Editval: Benchmarking diffusion based text-guided image editing methods.
\newblock \emph{arXiv preprint arXiv:2310.02426}.

\bibitem[{Baumgartner et~al.(2020)Baumgartner, Zannettou, Keegan, Squire, and Blackburn}]{baumgartner2020pushshift}
Jason Baumgartner, Savvas Zannettou, Brian Keegan, Megan Squire, and Jeremy Blackburn. 2020.
\newblock The pushshift reddit dataset.
\newblock In \emph{Proceedings of the international AAAI conference on web and social media}, volume~14, pages 830--839.

\bibitem[{Brixey et~al.(2018)Brixey, Manuvinakurike, Le, Lai, Chang, and Bui}]{brixey2018system}
Jacqueline Brixey, Ramesh Manuvinakurike, Nham Le, Tuan Lai, Walter Chang, and Trung Bui. 2018.
\newblock A system for automated image editing from natural language commands.
\newblock \emph{arXiv preprint arXiv:1812.01083}.

\bibitem[{Brooks et~al.(2023)Brooks, Holynski, and Efros}]{brooks2023instructpix2pix}
Tim Brooks, Aleksander Holynski, and Alexei~A Efros. 2023.
\newblock Instructpix2pix: Learning to follow image editing instructions.
\newblock In \emph{Proceedings of the IEEE/CVF Conference on Computer Vision and Pattern Recognition}, pages 18392--18402.

\bibitem[{Bychkovsky et~al.(2011)Bychkovsky, Paris, Chan, and Durand}]{fivek}
Vladimir Bychkovsky, Sylvain Paris, Eric Chan, and Fr{\'e}do Durand. 2011.
\newblock Learning photographic global tonal adjustment with a database of input / output image pairs.
\newblock In \emph{The Twenty-Fourth IEEE Conference on Computer Vision and Pattern Recognition}.

\bibitem[{Chen et~al.(2024{\natexlab{a}})Chen, Chen, Zhang, Wang, Liu, Zhou, Zhang, Wan, Zhou, and Sun}]{chen2024mllm}
Dongping Chen, Ruoxi Chen, Shilin Zhang, Yaochen Wang, Yinuo Liu, Huichi Zhou, Qihui Zhang, Yao Wan, Pan Zhou, and Lichao Sun. 2024{\natexlab{a}}.
\newblock Mllm-as-a-judge: Assessing multimodal llm-as-a-judge with vision-language benchmark.
\newblock In \emph{Forty-first International Conference on Machine Learning}.

\bibitem[{Chen et~al.(2024{\natexlab{b}})Chen, Wang, Cao, Liu, Gao, Cui, Zhu, Ye, Tian, Liu et~al.}]{chen2024expanding}
Zhe Chen, Weiyun Wang, Yue Cao, Yangzhou Liu, Zhangwei Gao, Erfei Cui, Jinguo Zhu, Shenglong Ye, Hao Tian, Zhaoyang Liu, and 1 others. 2024{\natexlab{b}}.
\newblock Expanding performance boundaries of open-source multimodal models with model, data, and test-time scaling.
\newblock \emph{arXiv preprint arXiv:2412.05271}.

\bibitem[{Chen et~al.(2024{\natexlab{c}})Chen, Hu, Niu, Chen, Li, Shan, and Wang}]{chen2024iqagpt}
Zhihao Chen, Bin Hu, Chuang Niu, Tao Chen, Yuxin Li, Hongming Shan, and Ge~Wang. 2024{\natexlab{c}}.
\newblock \href {https://doi.org/10.3389/fradi.2024.1310076} {Iqagpt: computed tomography image quality assessment with vision-language and chatgpt models}.
\newblock \emph{Frontiers in Radiology}, 4:131.

\bibitem[{D.(2025)}]{medium_photoshoprequest_2025}
Marx D. 2025.
\newblock \href {https://readmedium.com/if-i-was-good-at-photoshop-or-graphic-design-id-do-this-side-hustle-b8897429e79f} {If i was good at photoshop or graphic design, i'd do this side hustle}.
\newblock Reports r/PhotoshopRequest receives an average of 226 posts per day, based on Subredditstats and Social Rise analytics as of April 2025.

\bibitem[{Dai et~al.(2023)Dai, Hou, Ma, Tsai, Wang, Wang, Zhang, Vandenhende, Wang, Dubey et~al.}]{dai2023emu}
Xiaoliang Dai, Ji~Hou, Chih-Yao Ma, Sam Tsai, Jialiang Wang, Rui Wang, Peizhao Zhang, Simon Vandenhende, Xiaofang Wang, Abhimanyu Dubey, and 1 others. 2023.
\newblock Emu: Enhancing image generation models using photogenic needles in a haystack.
\newblock \emph{arXiv preprint arXiv:2309.15807}.

\bibitem[{Feedspot(2025)}]{feedspotPhotoshopForums}
Feedspot. 2025.
\newblock {T}op 15 {P}hotoshop {F}orums in 2025 --- forums.feedspot.com.
\newblock \url{https://forums.feedspot.com/photoshop_forums/}.
\newblock [Accessed 14-05-2025].

\bibitem[{Fellbaum(2010)}]{fellbaum2010wordnet}
Christiane Fellbaum. 2010.
\newblock Wordnet.
\newblock In \emph{Theory and applications of ontology: computer applications}, pages 231--243. Springer.

\bibitem[{Ge et~al.(2024)Ge, Zhao, Li, Ge, and Shan}]{ge2024seed}
Yuying Ge, Sijie Zhao, Chen Li, Yixiao Ge, and Ying Shan. 2024.
\newblock Seed-data-edit technical report: A hybrid dataset for instructional image editing.
\newblock \emph{arXiv preprint arXiv:2405.04007}.

\bibitem[{{Google DeepMind}(2024)}]{GoogleDeepMind2024Gemini2}
{Google DeepMind}. 2024.
\newblock Gemini 2.0 flash thinking.
\newblock \url{https://deepmind.google/technologies/gemini/}.
\newblock Experimental AI model.

\bibitem[{Hentschel et~al.(2022)Hentschel, Kobs, and Hotho}]{hentschel2022clip}
Simon Hentschel, Konstantin Kobs, and Andreas Hotho. 2022.
\newblock Clip knows image aesthetics.
\newblock \emph{Frontiers in Artificial Intelligence}, 5:976235.

\bibitem[{Hertz et~al.(2023)Hertz, Mokady, Tenenbaum, Aberman, Pritch, and Cohen-or}]{hertz2023prompttoprompt}
Amir Hertz, Ron Mokady, Jay Tenenbaum, Kfir Aberman, Yael Pritch, and Daniel Cohen-or. 2023.
\newblock \href {https://openreview.net/forum?id=_CDixzkzeyb} {Prompt-to-prompt image editing with cross-attention control}.
\newblock In \emph{The Eleventh International Conference on Learning Representations}.

\bibitem[{Huang et~al.(2024)Huang, Huang, Liu, Yan, Lv, Liu, Xiong, Zhang, Chen, and Cao}]{huang2024diffusion}
Yi~Huang, Jiancheng Huang, Yifan Liu, Mingfu Yan, Jiaxi Lv, Jianzhuang Liu, Wei Xiong, He~Zhang, Shifeng Chen, and Liangliang Cao. 2024.
\newblock Diffusion model-based image editing: A survey.
\newblock \emph{arXiv preprint arXiv:2402.17525}.

\bibitem[{{Hugging Face}(2023)}]{huggingface_spaces}
{Hugging Face}. 2023.
\newblock \href {https://huggingface.co/spaces} {{Hugging Face Spaces}}.
\newblock \url{https://huggingface.co/spaces}.
\newblock Accessed on March 02, 2025.

\bibitem[{Insights(2025)}]{marketshare}
Fortune~Business Insights. 2025.
\newblock \href {https://www.fortunebusinessinsights.com/ai-image-generator-market-108604} {Ai image generator market size, share \& industry growth 2030}.
\newblock [Online; accessed 2025-03-06].

\bibitem[{Jiang et~al.(2025)Jiang, Ku, Li, Ni, Sun, Fan, and Chen}]{jiang2025genai}
Dongfu Jiang, Max Ku, Tianle Li, Yuansheng Ni, Shizhuo Sun, Rongqi Fan, and Wenhu Chen. 2025.
\newblock Genai arena: An open evaluation platform for generative models.
\newblock \emph{Advances in Neural Information Processing Systems}, 37:79889--79908.

\bibitem[{Kampf and Brichtova(2025)}]{geminiImage}
Kat Kampf and Nicole Brichtova. 2025.
\newblock Experiment with gemini 2.0 flash native image generation.
\newblock \url{https://developers.googleblog.com/en/experiment-with-gemini-20-flash-native-image-generation/}.
\newblock Accessed: 2025-03-21.

\bibitem[{Kawar et~al.(2023)Kawar, Zada, Lang, Tov, Chang, Dekel, Mosseri, and Irani}]{kawar2023imagic}
Bahjat Kawar, Shiran Zada, Oran Lang, Omer Tov, Huiwen Chang, Tali Dekel, Inbar Mosseri, and Michal Irani. 2023.
\newblock Imagic: Text-based real image editing with diffusion models.
\newblock In \emph{Proceedings of the IEEE/CVF conference on computer vision and pattern recognition}, pages 6007--6017.

\bibitem[{Kim et~al.(2024)Kim, Ryu, Jung, Lee, Kim, Yang, Hwang, and Yang}]{kim2024augmentation}
Yoonjeon Kim, Soohyun Ryu, Yeonsung Jung, Hyunkoo Lee, Joowon Kim, June~Yong Yang, Jaeryong Hwang, and Eunho Yang. 2024.
\newblock Augmentation-driven metric for balancing preservation and modification in text-guided image editing.
\newblock \emph{arXiv preprint arXiv:2410.11374}.

\bibitem[{Korhonen and You(2012)}]{korhonen2012peak}
Jari Korhonen and Junyong You. 2012.
\newblock Peak signal-to-noise ratio revisited: Is simple beautiful?
\newblock In \emph{2012 Fourth international workshop on quality of multimedia experience}, pages 37--38. IEEE.

\bibitem[{Krojer et~al.(2025)Krojer, Vattikonda, Lara, Jampani, Portelance, Pal, and Reddy}]{krojer2025learning}
Benno Krojer, Dheeraj Vattikonda, Luis Lara, Varun Jampani, Eva Portelance, Chris Pal, and Siva Reddy. 2025.
\newblock Learning action and reasoning-centric image editing from videos and simulation.
\newblock \emph{Advances in Neural Information Processing Systems}, 37:38035--38078.

\bibitem[{Labs(2024)}]{flux2024}
Black~Forest Labs. 2024.
\newblock Flux.
\newblock \url{https://github.com/black-forest-labs/flux}.

\bibitem[{Laput et~al.(2013)Laput, Dontcheva, Wilensky, Chang, Agarwala, Linder, and Adar}]{laput2013pixeltone}
Gierad~P Laput, Mira Dontcheva, Gregg Wilensky, Walter Chang, Aseem Agarwala, Jason Linder, and Eytan Adar. 2013.
\newblock Pixeltone: A multimodal interface for image editing.
\newblock In \emph{Proceedings of the SIGCHI Conference on Human Factors in Computing Systems}, pages 2185--2194.

\bibitem[{Lee et~al.(2024)Lee, Kim, Park, Kim, and Seo}]{lee2024prometheus}
Seongyun Lee, Seungone Kim, Sue Park, Geewook Kim, and Minjoon Seo. 2024.
\newblock Prometheus-vision: Vision-language model as a judge for fine-grained evaluation.
\newblock In \emph{Findings of the Association for Computational Linguistics ACL 2024}, pages 11286--11315.

\bibitem[{Li et~al.(2022)Li, Li, Xiong, and Hoi}]{li2022blip}
Junnan Li, Dongxu Li, Caiming Xiong, and Steven Hoi. 2022.
\newblock Blip: Bootstrapping language-image pre-training for unified vision-language understanding and generation.
\newblock In \emph{International conference on machine learning}, pages 12888--12900. PMLR.

\bibitem[{Ma et~al.(2024)Ma, Ji, Ye, Lin, Wang, Zheng, Zhou, Sun, and Ji}]{ma2024i2ebench}
Yiwei Ma, Jiayi Ji, Ke~Ye, Weihuang Lin, Zhibin Wang, Yonghan Zheng, Qiang Zhou, Xiaoshuai Sun, and Rongrong Ji. 2024.
\newblock \href {https://proceedings.neurips.cc/paper_files/paper/2024/file/48fecef47b19fe501d27d338b6d52582-Paper-Conference.pdf} {I2ebench: A comprehensive benchmark for instruction-based image editing}.
\newblock In \emph{Advances in Neural Information Processing Systems (NeurIPS)}.

\bibitem[{Manuvinakurike et~al.(2018)Manuvinakurike, Brixey, Bui, Chang, Kim, Artstein, and Georgila}]{manuvinakurike2018edit}
Ramesh Manuvinakurike, Jacqueline Brixey, Trung Bui, Walter Chang, Doo~Soon Kim, Ron Artstein, and Kallirroi Georgila. 2018.
\newblock Edit me: A corpus and a framework for understanding natural language image editing.
\newblock In \emph{Proceedings of the Eleventh International Conference on Language Resources and Evaluation (LREC 2018)}.

\bibitem[{{OpenAI}(2024)}]{openai2024gpt4o}
{OpenAI}. 2024.
\newblock \href {https://openai.com/index/hello-gpt-4o/} {Hello {GPT-4o}}.
\newblock \url{https://openai.com/index/hello-gpt-4o/}.
\newblock Accessed: March 2, 2025.

\bibitem[{OpenAI(2024)}]{openai_o1}
OpenAI. 2024.
\newblock \href {https://openai.com/o1/} {Openai o1 model series}.
\newblock Accessed 2025-03-02.

\bibitem[{{OpenAI}(2025)}]{openai2025gpt4omini}
{OpenAI}. 2025.
\newblock \href {https://openai.com/index/gpt-4o-mini-advancing-cost-efficient-intelligence/} {Gpt-4o mini: advancing cost-efficient intelligence}.
\newblock Accessed on February 13, 2025.

\bibitem[{OpenAI(2025{\natexlab{a}})}]{openai2025_4oimage}
OpenAI. 2025{\natexlab{a}}.
\newblock \href {https://openai.com/index/introducing-4o-image-generation/} {Introducing 4o image generation}.
\newblock Accessed: 2025-05-10.

\bibitem[{OpenAI(2025{\natexlab{b}})}]{openai20254o}
OpenAI. 2025{\natexlab{b}}.
\newblock {I}ntroducing 4o {I}mage {G}eneration --- openai.com.
\newblock \url{https://openai.com/index/introducing-4o-image-generation/}.
\newblock [Accessed 14-05-2025].

\bibitem[{{OpenAI}(2025)}]{openai2025o1pro}
{OpenAI}. 2025.
\newblock Introducing {ChatGPT} pro.
\newblock \url{https://openai.com/index/introducing-chatgpt-pro/}.
\newblock Accessed: February 28, 2025.

\bibitem[{Oquab et~al.(2023)Oquab, Darcet, Moutakanni, Vo, Szafraniec, Khalidov, Fernandez, Haziza, Massa, El-Nouby et~al.}]{oquab2023dinov2}
Maxime Oquab, Timoth{\'e}e Darcet, Th{\'e}o Moutakanni, Huy Vo, Marc Szafraniec, Vasil Khalidov, Pierre Fernandez, Daniel Haziza, Francisco Massa, Alaaeldin El-Nouby, and 1 others. 2023.
\newblock Dinov2: Learning robust visual features without supervision.
\newblock \emph{arXiv preprint arXiv:2304.07193}.

\bibitem[{Radford et~al.(2021)Radford, Kim, Hallacy, Ramesh, Goh, Agarwal, Sastry, Askell, Mishkin, Clark et~al.}]{radford2021learning}
Alec Radford, Jong~Wook Kim, Chris Hallacy, Aditya Ramesh, Gabriel Goh, Sandhini Agarwal, Girish Sastry, Amanda Askell, Pamela Mishkin, Jack Clark, and 1 others. 2021.
\newblock Learning transferable visual models from natural language supervision.
\newblock In \emph{International conference on machine learning}, pages 8748--8763. PmLR.

\bibitem[{Rahmanzadehgervi et~al.(2024)Rahmanzadehgervi, Bolton, Taesiri, and Nguyen}]{Rahmanzadehgervi_2024_ACCV}
Pooyan Rahmanzadehgervi, Logan Bolton, Mohammad~Reza Taesiri, and Anh~Totti Nguyen. 2024.
\newblock Vision language models are blind.
\newblock In \emph{Proceedings of the Asian Conference on Computer Vision (ACCV)}, pages 18--34.

\bibitem[{Schuhmann(2023)}]{schuhmann2023improved}
Christoph Schuhmann. 2023.
\newblock Improved aesthetic predictor.
\newblock \url{https://github.com/christophschuhmann/improved-aesthetic-predictor}.

\bibitem[{Sheynin et~al.(2024)Sheynin, Polyak, Singer, Kirstain, Zohar, Ashual, Parikh, and Taigman}]{sheynin2024emu}
Shelly Sheynin, Adam Polyak, Uriel Singer, Yuval Kirstain, Amit Zohar, Oron Ashual, Devi Parikh, and Yaniv Taigman. 2024.
\newblock Emu edit: Precise image editing via recognition and generation tasks.
\newblock In \emph{Proceedings of the IEEE/CVF Conference on Computer Vision and Pattern Recognition}, pages 8871--8879.

\bibitem[{Shi et~al.(2020)Shi, Xu, Bui, Dernoncourt, Wen, and Xu}]{shi2020benchmark}
Jing Shi, Ning Xu, Trung Bui, Franck Dernoncourt, Zheng Wen, and Chenliang Xu. 2020.
\newblock A benchmark and baseline for language-driven image editing.
\newblock In \emph{Proceedings of the Asian Conference on Computer Vision}.

\bibitem[{Shi et~al.(2021)Shi, Xu, Xu, Bui, Dernoncourt, and Xu}]{shi2021learning}
Jing Shi, Ning Xu, Yihang Xu, Trung Bui, Franck Dernoncourt, and Chenliang Xu. 2021.
\newblock Learning by planning: Language-guided global image editing.
\newblock In \emph{Proceedings of the IEEE/CVF Conference on Computer Vision and Pattern Recognition}, pages 13590--13599.

\bibitem[{Shi et~al.(2024)Shi, Wang, and Huang}]{shi2024seededit}
Yichun Shi, Peng Wang, and Weilin Huang. 2024.
\newblock Seededit: Align image re-generation to image editing.
\newblock \emph{arXiv preprint arXiv:2411.06686}.

\bibitem[{SubredditStats.com(2025)}]{subredditstatsRPhotoshopRequestSubreddit}
SubredditStats.com. 2025.
\newblock r/{P}hotoshop{R}equest {S}ubreddit {S}tats ({P}hotoshop {R}equest) --- subredditstats.com.
\newblock \url{https://subredditstats.com/r/PhotoshopRequest}.
\newblock [Accessed 14-05-2025].

\bibitem[{Sushko et~al.(2025)Sushko, Bharadwaj, Lim, Ilin, Caffee, Chen, Salehi, Hsieh, and Krishna}]{sushko2025realedit}
Peter Sushko, Ayana Bharadwaj, Zhi~Yang Lim, Vasily Ilin, Ben Caffee, Dongping Chen, Mohammadreza Salehi, Cheng-Yu Hsieh, and Ranjay Krishna. 2025.
\newblock \href {https://cvpr.thecvf.com/virtual/2025/poster/34596} {Realedit: Reddit edits as a large-scale empirical dataset for image transformations}.
\newblock In \emph{Proceedings of the IEEE/CVF Conference on Computer Vision and Pattern Recognition (CVPR)}.
\newblock To appear.

\bibitem[{Tan et~al.(2019)Tan, Dernoncourt, Lin, Bui, and Bansal}]{tan2019expressing}
Hao Tan, Franck Dernoncourt, Zhe Lin, Trung Bui, and Mohit Bansal. 2019.
\newblock \href {https://doi.org/10.18653/v1/P19-1182} {Expressing visual relationships via language}.
\newblock In \emph{Proceedings of the 57th Annual Meeting of the Association for Computational Linguistics}, pages 1873--1883, Florence, Italy. Association for Computational Linguistics.

\bibitem[{Wang et~al.(2023{\natexlab{a}})Wang, Chan, and Loy}]{wang2023exploring}
Jianyi Wang, Kelvin~CK Chan, and Chen~Change Loy. 2023{\natexlab{a}}.
\newblock Exploring clip for assessing the look and feel of images.
\newblock In \emph{Proceedings of the AAAI conference on artificial intelligence}, volume~37, pages 2555--2563.

\bibitem[{Wang et~al.(2023{\natexlab{b}})Wang, Saharia, Montgomery, Pont-Tuset, Noy, Pellegrini, Onoe, Laszlo, Fleet, Soricut et~al.}]{wang2023imagen}
Su~Wang, Chitwan Saharia, Ceslee Montgomery, Jordi Pont-Tuset, Shai Noy, Stefano Pellegrini, Yasumasa Onoe, Sarah Laszlo, David~J Fleet, Radu Soricut, and 1 others. 2023{\natexlab{b}}.
\newblock Imagen editor and editbench: Advancing and evaluating text-guided image inpainting.
\newblock In \emph{Proceedings of the IEEE/CVF conference on computer vision and pattern recognition}, pages 18359--18369.

\bibitem[{Wang et~al.(2004)Wang, Bovik, Sheikh, and Simoncelli}]{wang2004image}
Zhou Wang, Alan~C Bovik, Hamid~R Sheikh, and Eero~P Simoncelli. 2004.
\newblock Image quality assessment: from error visibility to structural similarity.
\newblock \emph{IEEE transactions on image processing}, 13(4):600--612.

\bibitem[{Wu et~al.(2024)Wu, Zhang, Zhang, Chen, Liao, Li, Gao, Wang, Zhang, Sun, Yan, Min, Zhai, and Lin}]{wu2024q}
Haoning Wu, Zicheng Zhang, Weixia Zhang, Chaofeng Chen, Liang Liao, Chunyi Li, Yixuan Gao, Annan Wang, Erli Zhang, Wenxiu Sun, Qiong Yan, Xiongkuo Min, Guangtao Zhai, and Weisi Lin. 2024.
\newblock Q-align: teaching lmms for visual scoring via discrete text-defined levels.
\newblock In \emph{Proceedings of the 41st International Conference on Machine Learning}, ICML'24. JMLR.org.

\bibitem[{Xiong et~al.(2024)Xiong, Wang, Guo, Ye, Fan, Gu, Huang, and Li}]{xiong2024llava}
Tianyi Xiong, Xiyao Wang, Dong Guo, Qinghao Ye, Haoqi Fan, Quanquan Gu, Heng Huang, and Chunyuan Li. 2024.
\newblock Llava-critic: Learning to evaluate multimodal models.
\newblock \emph{arXiv preprint arXiv:2410.02712}.

\bibitem[{Zhang et~al.(2023)Zhang, Mo, Chen, Sun, and Su}]{zhang2023magicbrush}
Kai Zhang, Lingbo Mo, Wenhu Chen, Huan Sun, and Yu~Su. 2023.
\newblock Magicbrush: A manually annotated dataset for instruction-guided image editing.
\newblock \emph{Advances in Neural Information Processing Systems}, 36:31428--31449.

\bibitem[{Zhang et~al.(2018)Zhang, Isola, Efros, Shechtman, and Wang}]{zhang2018unreasonable}
Richard Zhang, Phillip Isola, Alexei~A Efros, Eli Shechtman, and Oliver Wang. 2018.
\newblock The unreasonable effectiveness of deep features as a perceptual metric.
\newblock In \emph{Proceedings of the IEEE conference on computer vision and pattern recognition}, pages 586--595.

\bibitem[{Zhang et~al.(2024)Zhang, Yang, Feng, Qin, Chen, Yu, Chen, Wang, Savarese, Ermon et~al.}]{zhang2024hive}
Shu Zhang, Xinyi Yang, Yihao Feng, Can Qin, Chia-Chih Chen, Ning Yu, Zeyuan Chen, Huan Wang, Silvio Savarese, Stefano Ermon, and 1 others. 2024.
\newblock Hive: Harnessing human feedback for instructional visual editing.
\newblock In \emph{Proceedings of the IEEE/CVF Conference on Computer Vision and Pattern Recognition}, pages 9026--9036.

\bibitem[{Zhao et~al.(2024)Zhao, Ma, Chen, Si, Wu, An, Yu, Zhang, Li, and Chang}]{zhao2024ultraedit}
Haozhe Zhao, Xiaojian~Shawn Ma, Liang Chen, Shuzheng Si, Rujie Wu, Kaikai An, Peiyu Yu, Minjia Zhang, Qing Li, and Baobao Chang. 2024.
\newblock Ultraedit: Instruction-based fine-grained image editing at scale.
\newblock \emph{Advances in Neural Information Processing Systems}, 37:3058--3093.

\bibitem[{Zhao et~al.(2025)Zhao, Ma, Chen, Si, Wu, An, Yu, Zhang, Li, and Chang}]{zhao2025ultraedit}
Haozhe Zhao, Xiaojian~Shawn Ma, Liang Chen, Shuzheng Si, Rujie Wu, Kaikai An, Peiyu Yu, Minjia Zhang, Qing Li, and Baobao Chang. 2025.
\newblock Ultraedit: Instruction-based fine-grained image editing at scale.
\newblock \emph{Advances in Neural Information Processing Systems}, 37:3058--3093.

\end{thebibliography}
\end{document}